\documentclass[hidelinks]{thesisUnal}

\usepackage[table,xcdraw]{xcolor}
\usepackage{amsmath}
\usepackage{algorithm}
\usepackage{algorithmicx}
\usepackage{algpseudocode}
\usepackage{todonotes}
\usepackage{multirow}

\usepackage{xspace}
\newcommand{\haea}{\textsc{HaEa}\xspace}
\usepackage{amssymb}
\usepackage{pdflscape}

\begin{document}

\logouniversity[45mm]{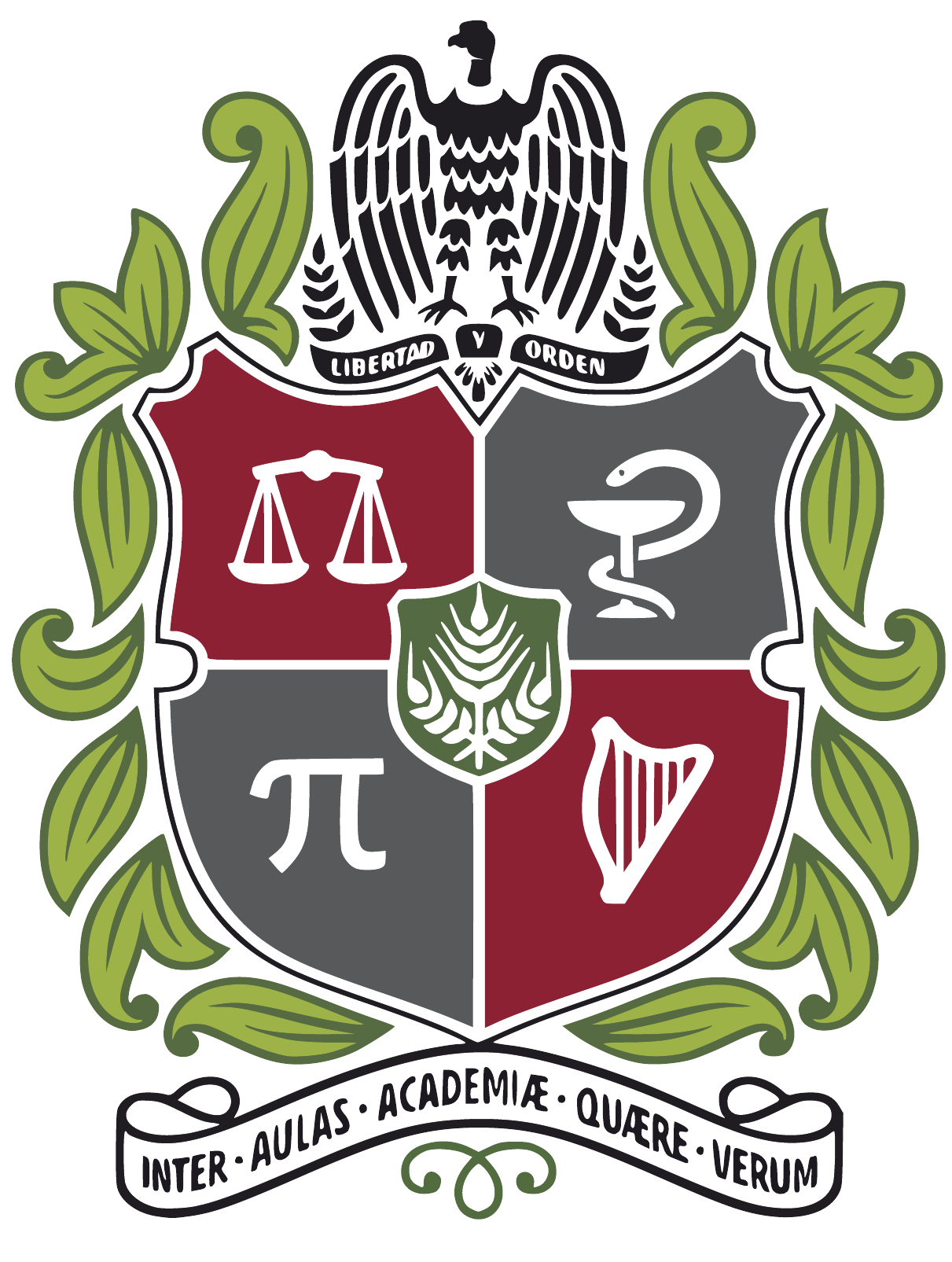}

\infothesis[
    author = {Lifeth \'Alvarez Camacho},
    degreeauthor = {Software Engineer},
    code = {1047410852},
    advisor = {Jonatan G\'omez Perdomo, Ph.D.},
    degreeadvisor = {Computer scientist},
    title = {Modeling Epigenetic Evolutionary Algorithms: An approach based on the Epigenetic Regulation process},
    titledegree = {Dissertation to apply for the title of},
    degree = {Master in Computer Systems Engineering},
    researcharea = {Evolutionary Computing, Evolutionary Algorithms, Epigenetics},
    researchline = {Artificial Life, Optimization},
    researchgroup = {Research Group in Artificial Life, ALife},
    university = {Universidad Nacional de Colombia},
    faculty = {Engineering School},
    department = {Computer Systems Engineering},
    city = {Bogot\'a, D.C.},
    date = {December 14, 2020},
]


\abstractthesis[
    titleenglish = {Modeling Epigenetic Evolutionary Algorithms: An approach based on the Epigenetic Regulation process.},
    abstractenglish = {Many biological processes have been the source of inspiration for heuristic methods that generate high-quality solutions to solve optimization and search problems. This thesis presents an epigenetic technique for Evolutionary Algorithms, inspired by the epigenetic regulation process, a mechanism to better understand the ability of individuals to adapt and learn from the environment. Epigenetic regulation comprises biological mechanisms by which small molecules, also known as epigenetic tags, are attached to or removed from a particular gene, affecting the phenotype. Five fundamental elements form the basis of the designed technique: first, a metaphorical representation of {\em Epigenetic Tags} as binary strings; second, a layer on chromosome top structure used to bind the tags (the {\em Epigenotype layer}); third, a {\em Marking Function} to add, remove, and modify tags; fourth, an {\em Epigenetic Growing Function} that acts like an interpreter, or decoder of the tags located over the alleles, in such a way that the phenotypic variations can be reflected when evaluating the individuals; and fifth, a tags inheritance mechanism. A set of experiments are performed for determining the applicability of the proposed approach.
    },
    keywordenglish = {evolutionary algorithms, evolution, epigenetics, gene regulation.}
    ]
 
\acceptationnote[
    note = {Approved},
    jury = {German Hernandez P\'erez},
    jury = {Jorge Ortiz Triviño},
    advisor = {Jonatan G\'omez Perdomo},
    date = {Bogot\'a, D.C., December 14, 2020}
]

\frontmatter
    \chapter{Dedicatory}
\begin{quote}
\centering
\large{\textit{To Abba...}}
\end{quote}

    \chapter{Acknowledgments}

I want to thank my advisor, Professor Jonatan Gomez, for his guidance and academic support during this research. I value his insights, remarks, motivation, and expertise that notably contributed to this thesis.

I want to thank Professor Luis Eugenio Andrade, Biologist, for his support at the beginning of this work, his help with Epigenetics understanding, and the comments and suggestions during academic meetings.

I express my gratitude to Aimer Alonso Gutierrez, Biologist and passionate about Epigenetics, for all the comments and suggestions. His knowledge reinforces and refines different aspects of this thesis.

My recognition goes out to the Artificial Life (ALife) Research Group, for all the comments and suggestions during formal and informal meetings.

Finally, I would like to acknowledge my family for their support and constant encouragement to finish this research.




    \tableofcontents
    \listoftables
    \listoffigures
\mainmatter
    \chapter{Introduction}

Optimization is a common task in people's lives. Investors use passive investment strategies that avoid excessive risk while obtaining great benefits. A conventional application of calculus is calculating a function minimum or maximum value. Manufacturers strive for the maximum efficiency of their production procedures. Companies lessen production costs or maximize revenue, for example, by reducing the amount of material used to pack a product with a particular size without detriment of quality. Software Engineers design applications to improve the management of companies' processes. The school bus route that picks a group of students up from their homes to the school and vice-versa, every school day, must take into account distances between homes and time. Optimization is an essential process, is present in many activities, contributes to decision science, and is relevant to the analysis of physical systems.

Making use of the optimization process requires identifying some objective, a quantitative measure of the system's performance under consideration. The objective can be profit, time, energy, or any resource numerically represented; the objective depends on problem characteristics, named as variables or unknowns. The purpose is to obtain variables values that optimize the objective; variables may present constraints or restrictions, for example, quantities such as the distance between two points and the interest rate on loan must be positive. The process of identifying objectives, variables, and constraints for a problem is known as modeling. The first step in the optimization process is to build an appropriate model. Once the model is formulated, an optimizer (a problem-solving strategy for solving optimization problems, such as equations, analytic solvers, algorithms, among others) can be implemented to find a satisfactory solution. There is no unique optimization solver but a set of optimizers, each of which is related to a particular optimization problem type. Picking a suitable optimizer for a specific problem is fundamental, it may determine whether the problem is tractable or not and whether it finds the solution \cite{CAVAZZUTI, KOZIEL, NOCEDAL}.

After a problem-solving strategy is applied to the model, the next step is to recognize if it succeeds in finding a solution. Some mathematical expressions known as optimal conditions or assumptions help to validate variable sets in order to know if they satisfy assumptions. If assumptions are not satisfied, variables may facilitate the current estimate of the solution to be adjusted. The model may be enhanced by implementing techniques such as a one-at-a-time {\em sensitivity analysis} technique that analyzes the influence of one parameter on the cost function at a time and exposes the solution susceptibility to changes in the model and data.  Interpretation in terms of the applicability of obtained solutions may recommend ways in which the model can be refined or corrected. The optimization process is repeated if changes are introduced to the model \cite{CAVAZZUTI, KOZIEL, NOCEDAL}.

An optimization algorithm is a method that iteratively executes and compares several solutions until it finds an optimum or satisfactory solution. Two optimization algorithms types widely used today are deterministic and stochastic. Deterministic algorithms do not involve randomness; these algorithms require well-defined rules and steps to find a solution. In contrast, stochastic algorithms comprise in their nature probabilistic translation rules \cite{CAVAZZUTI}. The use of randomness might enable the method to escape from local optima and subsequently reach a global optimum. Indeed, this principle of randomization is an effective way to design algorithms that perform consistently well across multiple data sets, for many problem types \cite{CAVAZZUTI, KOZIEL}. Evolutionary algorithms are a kind of stochastic optimization methods.

Evolutionary Algorithms (EAs) are a subset of population-based, metaheuristic optimization algorithms of the Evolutionary Computation field, which describes mechanisms inspired by natural evolution, the process that drives biological evolution. There are many types of evolutionary algorithms, the most widely known: Genetic Algorithm (GA), Genetic Programming (GP), Evolutionary Strategies (ES), and Hybrid Evolutionary Algorithms (HEAs). The common underlying idea behind all EAs is the same, an initial population of individuals, a parent selection process that considerate the aptitude of each individual, and a transformation process that allows the creation of new individuals through crossing and mutation. Candidate solutions act like the population's individuals for an optimization problem, and the fitness function determines the quality of solutions. The population's evolution then occurs after the repeated application of the above mechanisms \cite{CAVAZZUTI, EIBEN, HOLLAND, KOZIEL}. 

Many computer scientists have been interested in understanding the phenomenon of adaptation as it occurs in nature and developing ways in which natural adaptation mechanisms might be brought into computational methods. Current evolutionary algorithms are suitable for some of the most important computational problems in many areas, for example, linear programming problems (manufacturing and transportation), convex problems (communications and networks), complementarity problems (economics and engineering), and combinatorial problems (mathematics) such as the traveling salesman problem (TSP), the minimum spanning tree problem (MST), and the knapsack problem \cite{EIBEN, MITCHELL}. However, some computational problems involve searching through a large number of solution possibilities and require strategies that facilitate the adaptability of individuals in order to perform well in a changing environment \cite{MITCHELL}. In recent years, several authors have worked on hybrid strategies to improve the efficiency of population-based global search methods, so the adaptive behavior of populations can be rapidly manifested under selective pressure. Nevertheless, the possibility of finding good solutions is parameter dependant or long term search dependant. 


Developing problem solvers (algorithms) is a common task in Computer Science. There are different motives behind the design of algorithms. One is that engineers analyze processes in nature to mimic ``natural problem solvers'' which may help to achieve approximate solutions to the optimal one \cite{BEDAU, EIBEN}. Another motivation from a technical perspective is the growing demand for algorithms to solve hard or intractable problems (with time and space requirements). The above implies a requirement for robust algorithms with satisfying performance; consequently, there is a need for algorithms that apply to an extended number of problems, algorithms with less customization for specific optimization problems, and produce suitable (not necessarily optimal) solutions within a reasonable time \cite{EIBEN}. A third motivation is one that is present in every science: inquisitiveness. For example, evolutionary processes are topics of scientific investigation where the primary purpose is to understand evolution. From this viewpoint, evolutionary computing serves as a tool to conduct experiments to emulate processes from traditional biology. Evolutionary Algorithms may provide an answer to the challenge of using automated solution methods for a broader set of problems \cite{BEDAU, EIBEN}.

It is challenging to find simpler ways to improve any kind of search process. However, a future full of possibilities may start by designing new models that capture the essence of vital mechanisms present in living systems. An example of this is Epigenetics, which studies the epigenome, the epigenome-influencing factors present in the environment, epigenetic changes, and the inheritable epigenetic changes in gene expression. The DNA consists of two long chains of nucleotides, on which thousands of genes are encoded. The complete set of genes in an organism is known as its genome. The DNA is spooled around proteins called histones. Both the DNA and histones are marked with chemical tags, also known as epigenetic tags. Some regulatory proteins, histones, and epigenetic tags form a second structural layer called the epigenome \cite{GHR, NIGMS, UTAH}. All chemical tags that are adhered to the entire DNA and histones regulate the activity and expression of all genes within the genome. The biological mechanisms that involve attaching epigenetic tags to or removing them from the genome are known as epigenetic changes or epigenetic mechanisms. Two examples of epigenetic changes are DNA Methylation and Histone Acetylation. When epigenetic tags bind to DNA and alter its function, they have `` marked'' the genome; such marks do not modify the DNA sequence. Instead, they modify the way DNA's instructions are used by cells \cite{GHR, NHGRI, NIGMS}.

Epigenetics also brings up the concept of environment. Some changes occur during individuals' lifespan, and environmental factors can originate those changes. Epigenetic changes remain as cells divide and, in some cases, might be inherited through generations. Environmental factors, such as an individual's diet and exposure to pollutants, may also impact the epigenome \cite{GHR, NIGMS, UTAH}. Epigenetics has a potential role in our understanding of inheritance, natural selection, and evolution. Therefore, a change in gene expression in the adult \emph{$P_0$} generation caused by the environment might also be carried over into the \emph{$P_1$} generation or beyond, leading to a kind of long term memory \cite{UTAH}. 

Epigenetics drives modern technological advances, also challenges and reconsiders conventional paradigms of evolution and biology. Due to recent epigenetic discoveries, early findings on genetics are being explored in different directions. Both genetics and epigenetics help to better understand functions and relations that DNA, RNA, proteins, and the environment produce regarding heritage and health conditions. Epigenetics will not only help to understand complex processes related to embryology, imprinting, cellular differentiation, aging, gene regulation, and diseases but also study therapeutic methods. The incorporation of epigenetic elements in EAs allows robustness, a beneficial feature in changing environments where learning and adaptation are required along the evolutionary process \cite{RICALDE, GHR, NIGMS, UTAH}. Adaptation is a crucial evolutionary process that adjusts the fitness of traits and species to become better suited for survival in specific environments \cite{JEFFERY}. 

Epigenetic mechanisms like DNA Methylation and Histone Modification are vital memory process regulators. Their ability to dynamically control gene transcription in response to environmental factors promotes prolonged memory formation. The consistent and self-propagating nature of these mechanisms, especially DNA methylation, implies a molecular mechanism for memory preservation. Learning and memory are seen as permanent changes of behavior generated in response to a temporary environmental input \cite{ZOVKIC}. Organisms' ability to permanently adapt their behavior in response to environmental stimulus relies on functional phenotypic plasticity \cite{DEANS}. Epigenetic mechanisms intervene in biological processes such as phenotype plasticity, memory formation between generations, and epigenetic modification of behavior influenced by the environment. The previous fact leads researchers to improve evolutionary algorithms' performance in solving hard mathematical problems or real-world problems with continuous environmental changes by contemplating the usage of epigenetic mechanisms \cite{RICALDE}.

\section{Contributions of the Thesis}


This approach is inspired by the epigenetic regulation process, a biological mechanism widely studied by the Epigenetic field. This thesis aims to present a technique that models the adaptive and learning principles of epigenetics. The dynamics of DNA Methylation and Histone Modification has been summarized into five fundamental elements: first, a metaphorical representation of {\em Epigenetic tags}; second, a structural layer above the chromosome structure used to bind tags ({\em Epigenotype}); third, a marker ({\em Marking Function}) that comprises three actions: add, modify, and remove tags located on alleles; fourth, a tags interpreter or decoder ({\em Epigenetic Growing function}); and fifth, an inheritance process ({\em Crossover Operator}) to pass such tags on to the offspring. So that, the technique may reflect the adaptability of individuals during evolution, the ability of individuals to learn from the environment, the possibility of reducing time in the search process, and the discovery of better solutions in a given time.




The epigenetic mechanisms DNA Methylation and Histone Modification have been characterized to abstract principles and behaviors (from the epigenotype, tags, marking, reading, and inheritance biological elements) for the metaphorical designing of epigenetic components. In this thesis, the {\em Epigenotype} structure represents individuals' second layer, where tags are attached. The designed technique takes advantage of such a layer to influence the direction of the search process. The {\em Epigenotype} is made up of tags, {\em Epigenetic Tags} are represented with a binary string sequence of 0's and 1's, a tag implies a rule to interpret segments (alleles) of an individual's genome. The tags contain two sections; the first section denotes an operation, that is, a binary operation that operates on an individual's chromosome; the second section of the tags contains the gene size. The gene size indicates the number of alleles on which operates a binary operation. Tags determine individuals' gene expression; in other words, how alleles will be expressed, whether 1 or 0, depending on the operation applied to them.

The {\em Marking Function} involves writing, modifying, and erasing tags based on a metaphorical representation from writers, erasers, and maintenance enzymes. These actions act with a distributed probability of being applied to a single allele of a chromosome. Also, these epigenetic changes are framed into marking periods; such periods represent the environment, an abstract element that has been a point of reference to assess results of this technique. This mechanism allows performing the marking process in defined periods during the evolution process. The {\em Epigenetic Growing Function} represents the behavior of reader enzymes to interpret the epigenetic code (tags) on genotypes and then build phenotypes. The {\em Epigenetic Growing Function} plays the role of tags decoder or interpreter. After scanning the original individual's genome with its equivalent epigenotype (tags) and applying the operations to a copy of the chromosome, the {\em Epigenetic Growing Function} generates a resulting bit string to build phenotypes that are evaluated and scored. The {\em Crossover Operator} keeps its essence, but now, it includes tags as part of the exchange of genetic and epigenetic material between two chromosomes to create progeny. These epigenetic components are part of the proposed technique for the evolutionary algorithms' framework.



The epigenetic technique is implemented in classic Genetic Algorithms (GAs) and standard versions of \haea (Hybrid Adaptive Evolutionary Algorithm, designed to adapt genetic operators rates while solving a problem \cite{GOMEZa, GOMEZb}). The epigenetic components described previously are embedded in the algorithms' logic. The epigenetic implementations are named ReGen GA (GA with regulated genes) and ReGen \haea (\haea with regulated genes). Finally, the validation of the technique is made by comparing GA and \haea classic versions versus epigenetic implementations of GA and \haea, through a set of experiments/benchmarks to determine the proposed approach applicability. The comparison includes a set of experiments with binary and real encoding problems to identify the behavior of individuals under marking pressure. After comparing results between EAs standard versions and epigenetic EAs, it can be noticed that the current technique finds similar good solutions to standard EAs and better ones in subsequent iterations. The optimal solution (global optimum) is not always reached, but still, this model performs better in most cases.

\section{Overview}

Chapter~\ref{chapter2} presents state of the art. An overview of optimization, evolutionary algorithms with an emphasis on genetic algorithms, and hybrid evolutionary algorithms. The chapter also describes the biological basis of this research and a brief review of related work in the literature.

Chapter~\ref{chapter3} explains the proposed approach in detail. The chapter includes {\em Tags} encoding, selected operations, gene sizes, {\em Epigenotype} representation, the {\em Marking Function}, the {\em Epigenetic Growing Function}, the {\em Crossover Operator}, and a generic evolutionary algorithm pseudocode with the epigenetic components of the proposed technique.

Chapter~\ref{chapter4} introduces the implementation of the epigenetic technique on Genetic Algorithms. This chapter reports results on selected experimental functions with binary and real encoding for determining the model's applicability. Additionally, the chapter presents the analysis and discussion of results.

Chapter~\ref{chapter5} presents the implementation of the epigenetic technique on \haea. The \haea variations use two genetic operators: single point Crossover (enhanced to include tags) and single bit Mutation. Experimental functions with binary and real encoding are represented along with the analysis and discussion of results.

Chapter~\ref{chapter6} brings this book to a close with a short recapitalization and future research directions of this thesis.

Appendix~\ref{append} exhibits an example of individuals representation. The appendix includes individuals with a marked genotype in binary representation and its phenotypic interpretation. In Appendix~\ref{appendB} standard and ReGen EAs Samples for statistical analysis are included.
    \chapter{State of the Art}\label{chapter2}

Optimization is a significant paradigm with extensive applications. In many engineering and industrial tasks, some processes require optimization to minimize cost and energy consumption or maximize profit, production, performance, and process efficiency. The effective use of available resources requires a transformation in scientific thinking. The fact that most real-world tasks have much more complicated circumstances and variables to change systems' behavior may help in switching current reasoning. Optimization is much more meaningful in practice since resources and time are always limited. Three components of an optimization process are modeling, the use of specific problem-solving strategies (optimizer), and a simulator \cite{CAVAZZUTI, KOZIEL, NOCEDAL}.

A problem can be represented by using mathematical equations that can be transformed into a numerical model and be numerically solved. This phase must ensure that the right numerical schemes for discrete or continuous optimization are used. Another fundamental step is to implement the right algorithm or optimizer to find an optimal combination of design variables. A vital optimization capability is to produce or search for new solutions from previous solutions, which leads to the search process convergence. The final aim of a search process is to discover solutions that converge at the global optimum, even though this can be hard to achieve. In terms of computing time and cost, the most crucial step is using an efficient evaluator or simulator. In most cases, an optimization process often involves evaluating an objective function, which will verify if a proposed solution is feasible  \cite{CAVAZZUTI, KOZIEL, NOCEDAL}. 

Optimization includes problem-solving strategies in which randomness is present in the search procedure (Stochastic) or mechanical rules without any random nature (Deterministic). Deterministic algorithms work in a mechanically and predetermined manner without any arbitrary context. For such an algorithm, it reaches the same final solution if it starts with the same state. Oppositely, if there is some randomness in the algorithm, it usually reaches a different output every time the algorithm runs, even though it starts with the same input. Evolutionary Algorithms are examples of stochastic strategies \cite{CAVAZZUTI, KOZIEL}. Stochastic optimization methods seem to be the most innovative and advanced strategies for optimization. Compared to deterministic optimization methods, they have both advantages and disadvantages: they are less mathematically complex, use randomness in the search scheme, and have a slower convergence approaching the optimum solution \cite{CAVAZZUTI}. Optimization attempts to find the best possible solution among all available ones; optimization models a problem in terms of some evaluation function and then employ a problem solver strategy to minimize or maximize the objective function. The evaluation function represents the quality of the given solutions. Some methodologies aim to optimize some functions, but most of the problems are so large that it can be impossible to guarantee whether the obtained solution is optimal or not \cite{BURKE}.

Developing problem solvers (algorithms) is a common task in Computer Science. Engineers have always looked at nature's solutions to mimic ``natural problem solvers'' which may help to achieve approximate solutions to the optimal one \cite{EIBEN}. When complex natural phenomena are analyzed in the context of computational processes, our understanding of nature changes, leading to the design of powerful bio-inspired techniques to solve hard problems. Life has shown outstanding growth in complexity over time; life exhibits complex adaptive behavior in many natural elements, starting from individual cells to any living being, and even to evolving systems. Artificial Life (ALife), at first, focuses on understanding the essential properties of any life form, then uses synthetic methods ({\em soft, hard, and wet}) to represent such systems \cite{BEDAU, MITPRESS}. A characteristic of computing inspired by nature is the metaphorical use of concepts, principles, and biological mechanisms. ALife concentrates on complex systems that involve life, adaptation, and learning. By creating new types of life-like phenomena, artificial life continually challenges researchers to review and think over what it is to be alive, adaptive, intelligent, creative, and whether it is possible to represent such phenomena. Besides, ALife aims to capture the simple essence of vital processes, abstracting away as many details of living systems or biological mechanisms as possible \cite{BEDAU}. 

An example of this is the evolution process by natural selection, a central idea in biology. Biological evolution is the change in acquired traits over succeeding generations of organisms. The alteration of traits arises when variations are introduced into a population by gene mutation, genetic recombination, or erased by natural selection or genetic drifts. Adaptation is a crucial evolutionary process in which traits and species' fitness adjust for being better suited for survival in specific environments. The environment acts to promote evolutionary change through shifts in development \cite{JEFFERY}. The evolution of artificial systems is an essential element of ALife, facilitating valuable modeling tools and automated design methods \cite{MITCHELL3}. Evolutionary Algorithms are used as tools to solve real problems and as scientific models of the evolutionary process. They have been applied to a large variety of optimization tasks, including transportation problems, manufacturing, networks, as well as numerical optimization \cite{EIBEN, MITCHELL3}. However, the search for optimal solutions to some problem is not the only application of EAs; their nature as flexible and adaptive methods allow them to be applied in diverse areas from economic modeling and simulation to the study of diverse biological processes during adaptation \cite{EIBEN}.

\section{Evolutionary Algorithms}

Evolutionary Algorithms (EAs) are a subset of population-based, metaheuristic optimization algorithms of the Evolutionary Computation field, which uses mechanisms inspired by natural evolution, the process that drives biological evolution. There are many sorts of evolutionary algorithms, the most widely known: Genetic Algorithm (GA), Genetic Programming (GP), Evolutionary Strategies (ES), and Hybrid Evolutionary Algorithms (HEA). The general fundamental idea behind all these EAs is the same; given a population of individuals within some environment with limited resources, only the fittest survive. To define a particular EA, there are some components, or operators that need to be specified. The most important are: the representation of individuals, an evaluation (fitness) function, an initial population of individuals, a parent selection process that considerate the aptitude of each individual, a transformation process that allows the creation of new individuals through crossing and mutation, and a survivor selection mechanism (replacement) \cite{CAVAZZUTI, EIBEN, HOLLAND, KOZIEL}.

\subsection{Genetic Algorithms}

GAs are adaptive heuristic search computational methods based on genetics and the process that drives biological evolution, which is natural selection \cite{EIBEN}. Holland \cite{HOLLAND} presented the GA as the biological evolution process abstraction and formulated a theory about adaptation. Holland intended to understand adaptation and discover alternatives in which natural adaptation mechanisms might be brought into computer methods. The most used EA to solve constrained and unconstrained optimization problems is the traditional GA \cite{CAVAZZUTI, HOLLAND, KOZIEL}, also today, the most prominent and widely evolution models used in artificial life systems. They have been implemented as tools for solving scientific models of evolutionary processes and real problems \cite{MITCHELL3}.

A Genetic Algorithm explores through a space of chromosomes, and each chromosome denotes a candidate solution to a particular problem. Bit strings usually represent chromosomes in a GA population; each bit position (locus) in the chromosome has one out of two possible values (alleles), 0 and 1. These concepts are analogically brought from biology, but GAs use a simpler abstraction of those biological elements \cite{MITCHELL, MITCHELL2}. The most important elements in defining a GA are the encoding scheme (hugely depends on the problem), an initial population, a parent selection mechanism, variation operators such as recombination, mutation, and a replacement mechanism \cite{EIBEN, MITCHELL2}. The GA often requires a fitness objective function that assigns a score to each chromosome in the current population \cite{MITCHELL, MITCHELL2}. Once an optimization problem has been set up, the search process takes place by evaluating the population of individuals during several iterations. In the course of the evolution process, the chromosomes change from one population to another. An individual's performance depends on how well it satisfies the specified problem with its current schema of strings (the most common is binary alphabet \{0,1\}). The genetic algorithms obey to a population evolution model where the fittest survive \cite{HOLLAND}.

\subsubsection{Binary Codification}

The binary encoding uses the binary digit, or a bit, as the fundamental unit of information, there are only two possibilities $0$ or $1$. The genotype simply consists of a string or vector of ones and zeroes. For a particular problem, it is important to decide the string's length and how it will be interpreted to produce a phenotype. When deciding the genotype to phenotype mapping for a problem, it is essential to ensure the encoding allows all possible bit strings to express a valid solution to a given problem \cite{EIBEN}.

\subsubsection{Real Codification}

Real numbers represent any quantity along a number line. Because reals lie on a number line, their size is comparable. One real can be greater or less than another and used on arithmetic operations. Real numbers ($\mathbb{R}$) include the rational numbers ($\mathbb{Q}$), which include the integers ($\mathbb{Z}$), which include the natural numbers ($\mathbb{N}$). Examples: $3.44, -56.1, 2, 3/6, -1$. When values come from a continuous rather than a discrete distribution, usually, the most sensible way to represent a problem's candidate solution is through real values. For example, they may represent physical quantities such as the dimension of an object. The genotype for a solution with {\em k} genes is a vector $(x_1,...,x_k)$ with $x_i$ $\in \mathbb{R}$ \cite{EIBEN}.

\subsection{Hybrid Evolutionary Algorithms}

Hybridization of evolutionary algorithms is growing in the EA community due to their capabilities in handling several real-world problems, including complexity, changing environments, imprecision, uncertainty, and ambiguity. For diverse problems, a standard evolutionary algorithm might be efficient in finding solutions. As stated in the literature, standard evolutionary algorithms may fail to obtain optimal solutions for many types of problems. The above exposes the need for creating hybrid EAs, mixed with other heuristics. Some of the possible motives for hybridization include performance improvement of evolutionary algorithms (example: speed of convergence), quality enhancement of solutions obtained by evolutionary algorithms, and to include evolutionary algorithms as part of a larger system \cite{EIBEN, GROSAN}.

There are many ways of mixing techniques or strategies from population initialization to offspring generation. Populations may be initialized by consolidating previous solutions, using heuristics, or local search, among others. Local search methods may be included within initial population members or among the offspring. EAs Hybridation may involve operators from other algorithms, penalty-reward mechanisms, or domain-specific knowledge to the search process. The exploitation and exploration relationship produced during the execution determine the evolutionary algorithm behavior. Adaptive evolutionary algorithms produce exploitation/exploration relations that may avoid premature convergence and improve final results. Merging problem-specific knowledge for particular problems can also enhance Evolutionary Algorithms' performance \cite{EIBEN, GROSAN}.

As described in the literature, various techniques, heuristics, or metaheuristics are used to improve the evolutionary algorithms' general efficiency. Common hybrid strategies are compiled as follows: Hybridization between two EAs, Neural network-assisted EAs, Fuzzy logic assisted EA, Particle swarm optimization (PSO) assisted EA, Ant colony optimization (ACO) assisted EA, Bacterial foraging optimization assisted EA, and Hybridization between EAs and other heuristics (such as local search, tabu search, simulated annealing, hill climbing, dynamic programming, greedy random adaptive search procedure, among others.) \cite{GROSAN}.

\section{Overview of Epigenetics}

How living beings (particularly humans) respond to their environment is determined by inheritance, and the different experiences lived during development. Inheritance is traditionally viewed as the transfer of variations in DNA (Deoxyribonucleic Acid) sequence from parent to child. However, another possibility to consider in the gene-environment interaction is the trans-generational response. This response requires a mechanism to transmit environmental exposure information that alters the gene expression of the next generations \cite{PEMBREY}.

Two examples of trans-generational effects were found in Överkalix, a remote town in northern Sweden, and the Netherlands. The study conducted in Överkalix with three generations born in 1890, 1905, and 1920 revealed that the high or low availability of food for paternal grandfathers and fathers (during childhood or their slow growth period) influenced the risk of cardiovascular disease and diabetes mellitus mortality in their male children and grandchildren \cite{KAATIa, KAATIb, PEMBREY}. On the other hand, the study carried out on a group of people in gestation and childhood during the period of famine experienced between the winter of 1944 and 1945 in the Netherlands, evidenced that people with low birth weight, developed with higher probability, health problems such as diabetes, hypertension, obesity or cardiovascular disease during their adult life. The research concludes that famine during gestation and childhood has life-long effects on health. Such effects vary depending on the timing of exposure and the evolution of the recovery period \cite{KYLE}.


In this sense, gene expression can be affected in such a way that it reflects habits that shape an individual's lifestyle, even the ``experiences" of a generation might be passed down to progeny that have not necessarily lived in similar conditions to their parents. It is in this context that epigenetics, area that studies the modifications that affect gene expression, offers many answers regarding epigenetic regulation (this includes gene activation and recruitment of enzymes that add or remove epigenetic tags) and the inheritance of genetic conditions (susceptibility to disease) onto offspring \cite{GHR, UTAH}. As an organism grows and develops, chemical markers silence genome sections at strategic times and specific locations. Epigenetics is the study of those chemical reactions and the factors that influence them \cite{UTAH}. Certain factors, conditioned by habits and environment, are capable of interacting with genes and modifying their function without altering their molecular composition (DNA base sequence) \cite{GHR, LOBO}.

In 1942, Waddington described epigenetics as ``the branch of biology which studies the causal interactions between genes and their products, which bring the phenotype into being'' \cite{WADDINGTON}. However, the term epigenetics has been approached more broadly, recognizing its practical and theoretical importance in biology, without leaving aside Waddington's conception and the different concepts that have emerged to refer to the subject \cite{TAMAYO}. Such as the recognition of alternative development pathways, the existence of complex networks in development processes, phenotypic stability, plasticity, and the influence of the environment on organisms throughout their development \cite{JABLONKA, TAMAYO}. Today, Waddington's views on epigenetics are closely associated with phenotypic plasticity, which is a gene's ability to produce multiple phenotypes. Identifying regulatory interactions gene to gene and gene to protein explain the gene expression changes that Waddington named epigenetics \cite{DEANS}.

Holliday (1994) offered two definitions of epigenetics. His first definition describes epigenetics as ``the study of the changes in gene expression, which occur in organisms with differentiated cells, and the mitotic inheritance of given patterns of gene expression'' \cite{HOLLIDAY}. A second definition states that epigenetics is ``nuclear inheritance, which is not based on differences in DNA sequence'' \cite{HOLLIDAY}. Holliday redefined epigenetics in a way that was more precise and considerately focused on the inheritance of expression states \cite{DEANS}. For these two definitions, literature refers to gene regulation (Waddington definition) and epigenetic inheritance (Holliday definition) as intragenerational and transgenerational epigenetics. Epigenetics study involves both intragenerational and transgenerational epigenetics. The former refers to gene expression modifications through epigenetic marks (e.g., DNA methylation and Histone modification) that result in a modified phenotype within an individual's lifespan. The latter refers to the inheritance of a modified phenotype from parental generations with no DNA sequence changes. The same epigenetic markers mentioned above may be responsible; however, this category focuses on the act of inheritance \cite{BURGGREN}.

On the NIH Roadmap Initiative side, the epigenetic definition includes: ``Epigenetics is an emerging frontier of science that involves the study of changes in gene activity regulation and expression that are not dependent on gene sequence. Epigenetics refers to both heritable changes in gene activity and expression (in the progeny of cells or individuals) and also stable, long-term alterations in the transcriptional potential of a cell that are not necessarily heritable'' \cite{NIEHS}. Both Waddington's and Holliday's definitions seem to be part of the contemporary epigenetic description.

\begin{figure*}
\centering
\includegraphics[width=6.5in]
{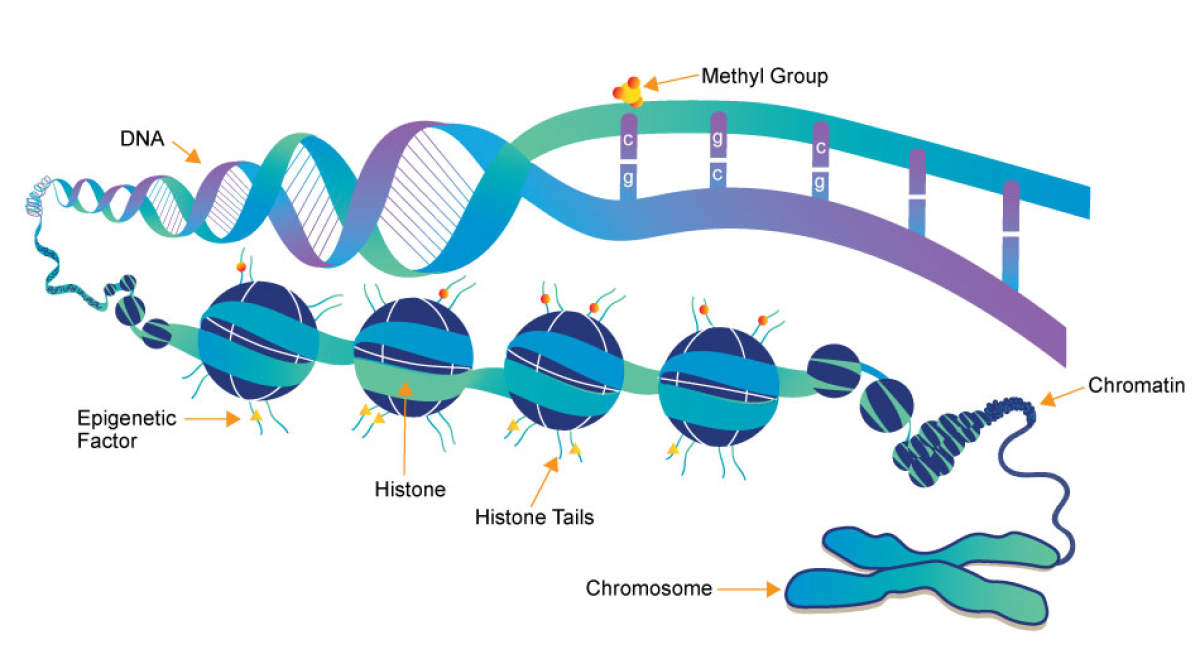}

\includegraphics[width=6.5in]
{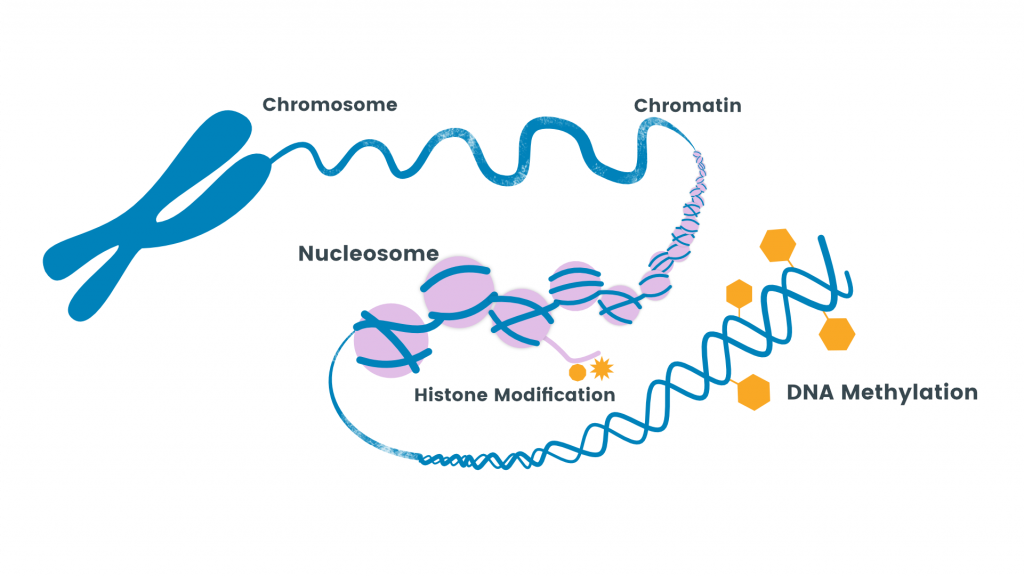}
\caption{Epigenetic Mechanisms.}
\label{c2fig1}
\end{figure*}

Today, a wide variety of behaviors and health indicators in people are linked to epigenetic mechanisms \cite{WEINHOLD}. Epigenetic processes are natural and essential for many organism functions; however, if epigenetic misregulation occurs (due to unfavorable environmental conditions, for example), significant adverse effects on health and behavior can happen. Some types of epigenetic modifications include methylation, acetylation, phosphorylation, ubiquitylation, and sumoylation. Among the best known and widely studied epigenetic mechanisms are DNA methylation and Histone modification (see Fig. \ref{c2fig1}). Other epigenetic modifications and considerations may emerge as research in the area of epigenetics progress \cite{GHR, NIGMS, WEINHOLD}. In fact, any mechanism that allocates regulatory data on genes without altering the nucleotide sequence is considered ``epi'', ``on top of'' or ``in addition to'' genetics. Examples of how epigenetic mechanisms affect gene expression are seen in processes like gene imprinting, cellular differentiation, and gene regulation during lifetime \cite{GHR, NIGMS, UTAH}.


\subsection{Epigenetic Mechanisms}

Humans have 23 pairs of chromosomes in each body cell; each pair has one chromosome from the mother and another from the father. A chromosome is composed of DNA and proteins. The DNA consists of two long chains made up of nucleotides, on which thousands of genes are encoded. The complete set of genes in an organism is known as its genome. The DNA is spooled around proteins called histones. Both the DNA and the Histones are marked with chemical tags, also known as epigenetic tags. The histones and the epigenetic tags form a second structural layer that is called the epigenome. The epigenome (epigenotype) comprises all chemical tags adhered to the entire DNA and Histones as a way to regulate the genes' activity (gene expression) within the genome. The biological mechanisms that involve attaching epigenetic tags to or removing them from the genome are known as epigenetic changes or epigenetic mechanisms \cite{GHR, NIGMS, UTAH}.

\subsubsection{DNA Methylation}
DNA methylation mechanism conducts the addition or elimination of methyl groups ({\em CH3}), predominantly where cytosine bases consecutively occur \cite{WEINHOLD}. In other words, chemical markers called methyl groups are bound to cytosines at {\em CpG sites} in DNA. Methyl groups silence genes by disrupting the interactions between DNA and the proteins that regulate it \cite{NHGRI}. Genome regions that have a high density of {\em CpGs} are known as {\em CpG islands}, and DNA methylation of these islands leads to transcriptional repression \cite{AARON}.

Methylation is sparsely found but globally spread in indefinite {\em CpG sequences} throughout the entire genome, except for {\em CpG islands}, or specific stretches (approximately one kilobase in length) where high {\em CpG contents} are found. These methylated sequences can drive to improper gene silencing, such as tumor suppressor genes' silencing in cancer cells. Studies confirm that methylation close to gene promoters differs considerably among cell types, with methylated promoters associated with low or no transcription \cite{PHILLIPS}.

DNA methylation represents the best-characterized and best-known epigenetic mechanism. DNA methylation is bound to the genome by {\em Dnmt3a} and {\em Dnmt3b} methyltransferases. These enzymes catalyze a methyl group's attachment to the cytosine DNA base on the fifth carbon ({\em C5}). DNA methylation maintenance is preserved when cells divide, and it is carried out by {\em Dnmt1} enzyme. Together, the mentioned enzymes guarantee that DNA methylation markers are fixed and passed onto succeeding cellular generations. In this way, DNA methylation is a cellular memory mechanism that transmits essential gene expression programming data along with it \cite{PAIGE}.

\subsubsection{Histone Modification}
Histone modification is a covalent posttranslational change to histone proteins, which includes methylation, acetylation, phosphorylation, ubiquitylation, and sumoylation. All these changes influence the DNA transcription process. Histone Acetyltransferases, for exmple, are responsible for Histone Acetylation; these enzymes attach acetyl groups to lysine residues on Histone tails. In contrast, Histone Deacetylases ({\em HDACs}) remove acetyl groups from acetylated lysines. Usually, the presence of acetylated lysine on Histone tails leads to an accessible chromatin state that promotes transcriptional activation of selected genes; oppositely, lysine residues deacetylation conducts to restricted chromatin and transcriptional inactivation \cite{GILBERT}.

The DNA is indirectly affected, DNA in cells is wounded around proteins called Histones, which form reel-like structures, allowing DNA molecules to stay ordered in the form of chromosomes within the cell's nucleus as depicted in Fig~\ref{c2fig1}. When Histones have chemical labels, other proteins in cells detect these markers and determine if the DNA region is accessible or ignored in a particular cell \cite{NHGRI}.

\subsection{Gene Regulatory Proteins}

Epigenetic regulation comprises the mechanisms by which epigenetic changes such as methylation, acetylation, and others can impact phenotype. Regulatory proteins conduct the epigenetic regulation process. These proteins have two main functions; the first involves switching specific genes on or off (gene activation); the second is related to the recruitment of enzymes that add, read or remove epigenetic tags from genes \cite{UTAH}.

\subsubsection{Enzymes: Writers, Readers, and Erasers}

Gene regulatory proteins recruit enzymes to add, read, and remove epigenetic tags; these processes are performed on the DNA, the Histones, or both, as explained previously \cite{UTAH}. These enzymes are seen as epigenetic tools, a family of epigenetic proteins known as readers, writers, and erasers \cite{RAO}. Epigenetics involves a highly complex and dynamically reversible set of structural modifications to DNA and histone proteins at a molecular level; these modifications evidence a second layer on the chromatin structure. 
The progress of epigenetic research has allowed the identification of crucial players performing these changes. These chemical alterations are catalyzed by enzymes referred to as epigenetic modifiers of the chromatin. The different enzymes that catalyze these modifications portray the epigenetic space's diversity and the complexity of gene expression regulation \cite{RAO}. 

These epigenetic modifiers are classified as ``writers'', ``readers'', and ``erasers''. {\em Writers} add to DNA or Histones chemical units ranging from a single methyl group to ubiquitin proteins. For example, DNA methyltransferases ({\em DNMTs}) are responsible for introducing the {\em C5-methylation} on {\em CpG} dinucleotide sequences. Such molecular structures not only influence the relation between DNA and histone proteins but also recruit non-coding RNAs ({\em ncRNAs}) and chromatin remodellers. On the other hand, the specialized domain-containing proteins that recognize and interpret those modifications are {\em Readers}; the binding interactions recognize through so-called reader modules specific modification codes or marks within the chemically modified nucleic acids and proteins and then perform conformational changes in chromatins and provide signals to control chromatin dynamics. Finally, {\em Erasers}, a dedicated type of enzyme expert in removing chemical markers, guarantee a reversible process. In order to achieve that, a group of eraser enzymes catalyzes the removal of the written information, ensuring a balanced and dynamic process \cite{RAO, PAIGE, UTAH}.

\subsubsection{Gene Silencing and Repression}

As explained above, epigenetics means  ``upon'', ``above'' or ``over'' genetics. Epigenetics describes a type of chemical reaction resulting from epigenetic modifications that alter DNA's physical structure without altering its nucleotide sequence. These epigenetic modifications cause genes to be more or less accessible during the transcription process. In general, environmental conditions influence the interactions and chemical reactions of the epigenotype, which can mark genes with specific chemical labels that direct actions such as gene silencing, activation or repression of a gene (activity), which translates into a modification in its function \cite{GHR, NIGMS, UTAH}.

Epigenetic mechanisms, in particular, Histone and DNA modifications, go beyond the idea of switching genes off and on. Gene silencing refers to a mechanism where large regions of a chromosome become transcriptionally inactive due to the compact wrapping of histone proteins that restricts the activity of DNA polymerases, which situate nucleotide units into a new chain of nucleic acid \cite{SANTER}. DNA regions that are highly packed are known to be part of the heterochromatin structure. In contrast, DNA relatively broadened form what is known as euchromatin. For a long time, it was assumed that heterochromatin is transcriptionally deedless compared to euchromatin. Nevertheless, many recent studies have questioned this conception of transcriptionally silent heterochromatin \cite{SHAFA}. Those studies indicate that the concept of equivalence between open chromatin with active transcription and compact chromatin with inactive transcription is not always applicable to all genes. Active genes have been located in tight chromatin regions and inactive genes in open chromatin regions \cite{CLARK}.
 
Gene repression is the inactivation of individual genes whose products are necessary to preserve cell functions, such as producing essential enzymes or cofactors \cite{SANTER}. The prevention of basal transcription machinery formation is considered the first mechanism through which gene expression down-regulation occurs. This type of transcriptional repression is obtained by directly altering the functional interactions of a transcription factor. Another gene repression mechanism is the inactivation of the transcriptional function of an activator. In this case, through different mechanisms (for example, the protein-protein interaction, covalent modification, and degradation), the repressor can affect an activator's capacity. The repression mechanisms require repressors binding to components of the basal transcriptional machinery or transcriptional activators. Epigenetic modifications that affect the chromatin structure close to the target genes may trigger these repression mechanisms \cite{CESARO}.

The recruitment of Histone acetyltransferase enzymes ({\em HATs}) allows the {\em H3} and {\em H4} histone tails acetylation. This mechanism promotes interactions between DNA and histones. The result is a relaxed structure surrounding the core promoter that is available to the general transcription process. Activator proteins interact with the general transcription factors to intensify DNA binding and initiation of transcription. The earlier means that activator proteins' recruitment helps raise the transcription rate, leading to gene activation \cite{CLARK}. Methylation is related to both gene activation and repression; and each mechanism depends on the degree of methylation. Inactive genes or silent chromosome regions are highly methylated in their {\em CpG islands} compared with the same gene on the active chromosome \cite{FESTEN, SHAFA}.

There are other important considerations around the expression of genes. Gene regulation mechanisms (silencing, repression, activation) depend mostly on the cell's epigenetic condition, which controls the gene expression timing and degree at a specific time \cite{SHAFA}. Silencing gene expression is not just about switching chromatin areas entirely off, or gene repression fully suppressing a gene function. The dynamics of these mechanisms also involve decreasing the level of transcription by gradually reducing gene expression, depending on tags bind location or regions, and how many tag groups are attached. So, it is possible to evidence sections of the chromosome where the gene expression is not totally inactivated but strongly reduced. In the same way, active genes and regions with expression levels moderated. The binding of proteins to particular DNA elements or regulatory regions to control transcription and mechanisms that modulate translation of {\em mRNA} may also be moderated \cite{CESARO, CLARK, FESTEN, SHAFA}.

\subsection{Epigenetic Memory and Adaptation}

Today epigenetic modifications such as DNA methylation and histone tail modifications are known as essential regulators in the consolidation and propagation of memory. These mechanisms' ability to dynamically control gene transcription in response to environmental factors establishes the consolidation of long-term memory formation. Additionally, these mechanisms, particularly DNA methylation, has a persistent and self-propagating nature that suggests a molecular mechanism for memory maintenance \cite{ZOVKIC}.

Learning and memory are seen as permanent alterations of behavior produced as a result of a changing environment \cite{ZOVKIC}. For a temporal stimulus to promote long-term behavior changes, cells must experience several cellular stimuli and molecular modifications that establish a lasting memory. The molecular foundation for memory preservation is notable when short-lived environmental stimuli induce self-perpetuating biochemical reactions required to maintain molecular changes. Holliday \cite{HOLLIDAYb} proposed that epigenetic mechanisms, particularly DNA methylation, possess the biochemical properties required to transmit memories throughout life. DNA methylation is recognized as a solid and self-perpetuating regulator of cellular identity through the establishment and spread of persistent heritable changes in gene expression through cell divisions \cite{BIRD}. This earlier suggests that epigenetic mechanisms are able to provide a suitable molecular baseline for memory consolidation and maintenance.

In this case, epigenetic tags help to long-term memorize how genes should be expressed; changes in gene expression can lead living beings to adapt to their environment. Epigenetic markers represent a type of cellular memory, a cell epigenetic profile, a collection of tags that describe expression states of genes, and also the totality of the signals received during an individual's existence \cite{UTAH}. Adaptation is vital in organisms' development process. The fitness of traits and species is continuously adjusted, so individuals are better suited to survive in particular environments and qualified to face different conditions \cite{JEFFERY}. The environment continually acts to promote transformation through changes in development. The organism's ability to permanently adapt its behavior to environmental changes depends entirely on functional phenotypic plasticity and the genome's capability to produce multiple phenotypes. Propagation of expression states and cell memory is part of the heritable memory conception, an explicit property of epigenetic gene regulation \cite{DEANS, UTAH}.

\subsection{Epigenetics and Evolution}

Nowadays, epigenetics is known not only because of its relevance for medicine, farming, and species preservation, but also because studies have revealed the importance of epigenetic mechanisms in inheritance and evolution. Particularly, evidencing epigenetic inheritance in systems where non-DNA alterations are transmitted in cells. Also, the organism diversity broadens the heredity theory and defy the current gene-centered neo-Darwinian version of Darwinism \cite{JABLONKA}. Epigenetics as science does not intend to oppose early ideas of evolutionary theory. In fact, some authors suggest considering modern epigenetics as neo-Lamarckian \cite{PENNY} or close to the original argument proposed by Baldwin (known as Baldwin effect) \cite{EGAS}. Early authors were undergoing studies that are expanding the knowledge about inheritance and evolution. Currently, the epigenetics research community continues learning through epigenetics studies \cite{PENNY}, even when the idea of epigenetic inheritance and its influence on evolution is still controversial \cite{BURGGREN}. This set of theories, along with others like Mendelian principles and Hardy-Weinberg law, try explaining inheritance and living organisms diversity as stated by the tenets of genetic traits heredity from parent organisms to their children  \cite{BURGGREN, JABLONKA, PENNY}.  Those theories are based on some factors or conditions, name them statistical, environmental, needs, survival, among others. Despite this, some researchers think it is not helpful attributing modern ideas to early researchers, since, it can be misleading \cite{PENNY}.

A fundamental principle of evolution is that natural selection alters organisms behavior over long periods by shaping populations traits. Natural selection has no particular inclination, this process acts on organisms with poor or improved fitness, which derives from mutations accumulation; these mutations can enhance resulting phenotypic modifications. However, phenotypic changes at the population level and beyond generally occur over many thousands of generations when a genotype with a modified phenotype of higher fitness slowly places in the general population or a genotype with lower fitness is eliminated from the population \cite {BURGGREN}. Epigenetic inheritance changes the evolutionary perspective, as mentioned previously, the genome slowly transforms through the processes of random variation and natural selection; it takes a large number of generations for a genetic trait to prevail in a population. The epigenome, otherwise, changes rapidly as a consequence of being affected by signals from the environment. Epigenetic changes may occur in many organisms at one time; through epigenetic heritage, some parents experiences may pass on to the next generations; and the epigenome remains flexible as the environment changes. Epigenetic inheritance allows an organism to continuously adjust its gene expression to suit its environment without affecting its DNA code \cite {UTAH}.

The increment of individuals' fitness in a population may derive from epigenetic or genetic changes over thousands and thousands of generations. However, the epigenetic inheritance impact might not only be potentially evidenced in posterity generations but also be perceived immediately in a population. The inheritance of epigenetically shaped phenotypes may result from the continuous inheritance of epigenetic tags over generations. An epigenetically-inherited phenotype does not need to be fixed on the genotype to have a prominent influence on the evolution of traits. Instead, what directs the genotype variation in a population is the individuals' capability to survive despite unevenly-distributed epigenetic tags that produce suitable or unsuitable phenotypes subject to natural selection. Intragenerational and transgenerational epigenetics, therefore, are not mutually exclusive. It is evident that an alteration in gene expression in the adult generation phenotype {\ em P0} by DNA methylation and Histone Acetylation, for example, might also be passed onto {\ em P1} generation or beyond \cite{BURGGREN}.

\section{Related Work}

There is a predominant focus in the literature on the in-depth study of epigenetic mechanisms, especially those that may be associated with the diagnosis, prevention, and treatment of diseases with apparently less emphasis on what mechanisms do at the phenotypic level of an individual, particularly between generations. Models/Strategies focused on epigenetic changes occurring during one generation's life span or transmitted through generations, or at an individual level have been the target to identify the most recent achievements around this topic in the evolutionary algorithms community. Those models have been developed with different approaches. Several authors have worked on hybrid strategies to improve the solution capacity of population-based global search methods, so the adaptive behavior of populations can be rapidly manifested under selective pressure. Such strategies aim to address a wider variety of computational problems by mimicking biological mechanisms or social changes. Below, some approaches are briefly described; they entail adaptation and learning behaviors, two characteristics that this thesis is studying.

Dipankar Dasgupta et al. (1993) \cite{DASGUPTA} introduce the structured Genetic Algorithm (sGA). Though this strategy does not mention epigenetic mechanisms, it involves gene activation, an essential mechanism in gene regulation to control genes states: repression (different from silencing) and expression. This genetic model includes redundant genetic material and a gene activation mechanism that utilizes a multi-layered structure (hierarchical) for the chromosome. Each gene in higher levels acts as a switchable pointer that activates or deactivates sets of lower-level genes. At the evaluation stage, only the active genes of an individual are translated into phenotypic functionality. It also includes a long-term distributed memory within the population enabling adaptation in non-stationary environments. Its main disadvantage is the use of a multi-level representation with optional search spaces that could be activated at the same time, leading to express a bit string that may be too long for the problem solution.

Tanev and Yuta (2008) \cite{TENEV} describe the first model mentioning Epigenetics in the EA community. In this model, they focus on an improved predator-prey pursuit problem. They present individuals with double cell, comprising somatic cell and germ cell, both with their respective chromatin granules. In the simulation, they use the Modification of Histones to evidence the role this mechanism plays in regulating gene expression and memory (epigenetic learning, EL). The Genetic Programming Algorithm defines a set of stimulus-response rules to model the reactive behavior of predator agents. The beneficial effect of EL on GP's performance characteristics is verified on the evolution of predator agents' social behavior. They report that EL helps to double improve the computational performance of the implemented GP. Additionally, the simulation evidences the phenotypic variety of genotypically similar individuals and their preservation from the negative effects of genetic recombination. Phenotypic preservation is achieved by silencing particular genotypic regions and turning them on when the probability of expressing beneficial phenotypic traits is high.

Satish Periyasamy et al. (2008) present the Epigenetic Algorithm (EGA), based on the intragenerational adaptation of biological cells, optimization strategy that bio-molecules use for their functional processes. They adapt concepts of bio-molecular degradation and autocatalysis, which are ubiquitous cellular processes and pivotal for the adaptive dynamics and evolution of an intelligent cellular organization. The algorithm is used to achieve optimization for organizations' internal structures, with a focus on the autopoietic behavior of the systems. Additionally, the authors present a simulation with agent-based cell modeling. This artificial model is called SwarmCell; the model is built as an autopoietic system that represents a minimal biological cell. The authors state that their epigenetic algorithm can demonstrate to be a fundamental extension to existing evolutionary systems and swarm intelligence models. They discuss improving problem-solving capabilities by implementing epigenetic strategies in their model \cite{PERIYA}.

Epigenetic Tracking by Alessandro Fontana (2009) is a mathematical model of biological cells \cite{FONTANA}. The computer simulation generates complex 3-dimensional cellular structures with the potential to reproduce the complexity typical of living beings. The simulation shows the homogeneous distribution of stem cells, which are dynamically and continuously created during development from non-stem cells. The model uses an integer number genetic encoding scheme controlled by a regulatory network with epigenetic regulatory states (on and off) to represent signals in distinct development phases. A two-dimensional cellular grid and a GA operating on the genome allow generating arbitrary 2-or-3-dimensional shapes.

The EpiAL model by Sousa and Costa (2010) \cite{SOUSAa, SOUSAb}, is based on two main entities: agents and the environment, for which, epigenetics is considered as the ability for an agent to modify its phenotypic expression due to environmental conditions. An agent has regulatory structures that, given inputs from the environment, can act upon the genotype, regulating its expression. They also consider the epigenetic marks to be inherited between generations, through the transmission of partial or full markers (methylation values off/on), allowing the existence of acquired traits (methyl marks) to be transmitted through generations of agents. The environment models a two-dimensional grid with transposable locations or separated ones by a wall. Each location has different characteristics, temperature, light, and food that can change gradually; the agents intend to survive and procreate. Agents behavior is encoded on binary strings. Methylation marks regulate the activation of genes. An EA controls the survival and reproduction of different organisms. Non-epigenetically modified populations find it difficult to survive in changing environments, while epigenetically modified populations are capable to regulate themselves under changing conditions.

Chikumbo et al. (2012) \cite{CHIKUMBOa} propose a Multi-Objective Evolutionary Algorithm with epigenetic silencing for the land use management problem. The algorithm intention is to decrease the ecological footprint while ensuring sustainability in land use management through asserted decision making. The chromosome encodes each possible paddock use, and the system emulates gene regulation with epigenetic silencing based on histone modification and RNA editing mechanisms. A visualization tool of pareto frontier is used, composing fourteen objectives into three general criteria: a set of time-series, farm management strategies, and their related spatial arrangements of land uses. However, improvements in epigenetic variations are not estimated as the approach is not compared to any other standard Multi-Objective EA. In 2015, the authors introduced an improvement by using a similar epigenetics-based modification, such improvent is described in Triple Bottomline Many-Objective-Based Decision Making for a Land Use Management Problem \cite{CHIKUMBOb}. The change involves the use of Hyper Radial Visualization (HRV), 3D Modeling, and Virtual Reality to diminish the functions of fourteen objectives and visualize solutions in a simpler representation to be interpreted by an expert group. The triple bottom line is represented by the economic, environmental, and social complex (stakeholder preferences) factors.

Arnold C and et al. (2013) \cite{ARNOLD} propose a theoretical mechanism to explain how a cell can reconstruct its epigenetic state after the replication process. The replication process may be responsible for epigenetic information loss, such information is accumulated over a cell lifetime. They hypothesize that the different combinations of reader and writer units in histone-modifying enzymes use local rewriting rules capable of computing the acquired patterns. To demonstrate this, they use a flexible stochastic simulation system to study histone modification states' dynamics. The implementation is a flexible stochastic simulation system based on the Gillespie algorithm, which models the master equation of a detailed chemical model. An evolutionary algorithm is then implemented to find enzyme mixtures for stable inheritance and patterns across multiple cell divisions with high precision. Once such patterns are found, chromatin is rebuilt.

Turner et al. (2013) formally describe the Artificial Epigenetic Regulatory Network (AERN), an extended version of their previous artificial gene regulation (AGN) model. AERN uses an analog of DNA methylation combined with chromatin modifications as its epigenetic elements, giving the network the ability to change its epigenetic information during the evolution and execution of epigenetic frames. Epigenetic control enables the network to evolve. In the model, subsets of genes are more likely to perform a given objective, when present an active state. The inclusion of epigenetic data gives the network the capacity to designate different genes to diverse tasks, completely directing gene expression as stated by its operating environment. The goal is to follow specific trajectories governed by evolution rules and represented in chaotic dynamics (Chirikov's standard map). The net evolves by making use of a GA. Because of the ability to deactivate genes, the network increases its efficiency. Consequently, with objectives including deactivated genes,  a minimum computational effort is required to achieve at least the first iteration of the network simulation. The epigenetic mechanism improves the performance of the model based on the authors' report \cite{TURNER}.

Przybilla and colleagues (2014) \cite{PRZYBILLA} introduce a computational model to understand epigenetic changes in aging stem cells in a population of cells where each contains an artificial genome. The transcription of the genes encoded in the genome is controlled by DNA methylation, histone modification, and the action of a CIS-regulatory network. The dynamic of the model is determined by the molecular crosstalk between the different epigenetic mechanisms. The epigenetic states of genes are subject to different types of fluctuations. The model provides a mechanistic understanding of age‐related epigenetic drifts. The researchers aim at linking epigenetic mechanisms to phenotypic changes of cells to derive hypotheses on the emergence of age‐related phenotypes (ARPs) on a population level. They combine their model of transcriptional regulation with an individual cell‐based model of stem cell populations, which allows them to simulate aging on the molecular, cellular, and population level.  The authors hypothesize that ARPs are a consequence of epigenetic drifts, which originate the limited cellular capability to inherit epigenetic information.

La Cava and colleagues (2014) \cite{LACAVAa} describe a method to solve the symbolic regression problem using Developmental Linear Genetic Programming (DLGP) with an epigenetic hill climber (EHC). The EHC helps to optimize the epigenetic properties of linear genotypes that represent equations. In addition to having a genotype composed of a list of instructions, the Epigenetic Hill Climber (EHC) creates a binary array of equivalent length in each individual, referred to as an Epiline. During genotype to phenotype mapping, only instructions from the list with an active state in the corresponding Epiline are executed. Their implementation is based on two main characteristics: first, inheritability by coevolution of Epilines with their respective genotype; and second, the use of EHC, which mimics the acquisition of lifetime learning by epigenetic adaptation. The EHC implementation evaluates epigenetic changes to determine whether individuals should be updated. Epigenetic modifications to an individual are kept only if the fitness is improved or not changed, based on the active genes (instructions). The same method is implemented to solve program synthesis problems in Inheritable Epigenetics in GP (2015) \cite{LACAVAb} and GP with Epigenetic Local Search (2015) \cite{LACAVAc}. La Cava reports that the addition of epigenetics results in faster convergence, less bloat, and an improved ability to find exact solutions on several symbolic regression problems.

The epigenetic algorithm (EGA) by Birogul (2016) \cite{BIROGUL} adapts epigenetic concepts to the classical GA structure. The EGA includes epicrossover operator, epimutation operator, and epigenetic factors. Also, his technique explains how epigenetic inheritance is achieved across populations. The designed EGA is applied to the Mobile Network Frequency Planning that is a constrained optimization problem. He uses data from real base stations BCCH (Broadcast Control Channel) in a GSM network (Global System for Mobile Communications) to test his approach. He states that EGA obtained better results in a shorter time and less iteration than classical GAs when implementing both algorithms in order to solve the mentioned constrained optimization problem.

The epiGenetic algorithm (epiGA) by Daniel Stolfi and Enrique Alba (2018) \cite{STOLFIa, STOLFIb}, consists of four epigenetic operators. The first operator is the Individual Initialization that creates individuals made up of cells. Second, the Nucleosome Generation operator that creates the nucleosome structure where the DNA is collapsed and made inaccessible during reproduction. Third operator, the Nucleosome Based Reproduction, where the most promising cells combine following epigenetic rules. The last operator, called Epigenetic Mechanisms, is where some rules are applied according to DNA methylation and the surrounding environment. Each individual in the population contains M cells, which can represent different solutions to the problem. Four binary vectors of the same size of the chromosome with the problem representation integrate each cell. One vector contains the encoded solution; two vectors comprise the chromosomes of the cell's parents and another vector where the binary mask (nucleosome mask) representing the nucleosome structure is stored. The foundation of epiGA is epigenesis. The authors focused on how the DNA and histones are collapsed to form nucleosomes, how this affects the gene replication during reproduction, and how the epigenetic mechanisms modify the gene expression through methylation, contributing to building the bio-inspired operators of the algorithm. The epiGA is used to solve the Multidimensional Knapsack Problem (MKP). They report that the algorithm does perform similarly or better than published results in the literature.

Esteban Ricalde (2019) proposes an approach that describes a new mechanism for Genetic Programming inspired by epigenetics. The mechanism activates and deactivates chromosomal regions triggered by environmental changes. The epigenetic approach is based on the DNA methylation mechanism and implements a tree-based scheme for evolving executable instructions. Only conditional nodes are affected by the mechanism. Ricalde also introduces an adaptive factor strategy to assess the environment local variation. The mechanism takes into account changes in the environment to conduct epigenetic mutations. The author reports GP performance improvements when solving problems with changing environmental conditions, such environments promote individuals to adapt easier. This strategy aims to present an innovative method for the traffic signal control problem. The method defines the evolution process of actuated traffic controllers by the use of GP. The adaptive factor strategy is focused on traffic signals optimization context \cite{RICALDE}.

The Memetic Algorithm (MA), is a cultural-based strategy (1989) \cite{MOSCATO}, inspired by the description of memes in Dawkins {\em The Selfish Gene} book. A `meme' denotes the idea of a unit of imitation in cultural evolution, which in some aspects, is analogous to the gene in GAs. Examples of memes are tunes, ideas, catch-phrases, clothes fashions, ways of making pots, food, music, or ways of building arches \cite{DAWKINS}. The MAs extend the notion of memes to cover conceptual entities of knowledge-enhanced procedures or representations. The MA combines the population-based global search and the local search heuristic made by each individual, capable of performing local refinements without genetic representation constraints. The earlier may represent a high computational cost due to the separated individual learning process or local improvement for a problem search. Moscato coined the name `Memetic Algorithm' to cover a wide range of techniques where the evolutionary search is extended by adding one or more phases of local search, or the use of problem-specific information.

According to the analysis of state of the art, it can be noticed that the different approaches (except the Memetic Algorithm) focus on common elements abstracted from the dynamics of the epigenetic mechanisms. These elements involve: 

\begin{enumerate}
    \item The activation and deactivation (gene activation) of individuals' chromosomes through epigenetic mutations triggered by environmental changes. These mutations modify the markers (with off and on states) during the lifespan of the individual.
    
    \item The use of active genes to evaluate individuals' ability to adapt and survive (fitness). 
    
    \item The learning behavior through the notion of memory across generations by propagating the best active genes (the ones that make the individual fittest). 
    
    \item Moreover, the particular effects the environment can produce during the development of individuals within a generation and their progeny. 
    
\end{enumerate}
 
Despite the different usage of Epigenetic, previous approaches have evidenced that the incorporation of epigenetic components in EAs facilitates robustness. Robustness \cite{FELIX} is essential to ensure the permanence of phenotypic attributes potentially subjected to genetic and non-genetic modification; robustness also permits genetic and non-genetic changes to increase. Such variations will possibly introduce evolutionary alterations to a population and make individuals adapt.

    \chapter{Evolutionary Algorithms with Regulated Genes: ReGen EAs}\label{chapter3}

The previous chapter described optimization and evolutionary processes as inspiration to design problem solvers; also, an epigenetics overview, the relation between epigenetics and evolution, memory and adaptation from epigenetics point of view, and the different approaches that implemented Epigenetics into Evolutionary Algorithms.  

State of the art shows that epigenetic mechanisms play a fundamental role in biological processes. Some of such processes are phenotype plasticity, memory consolidation within generations, and environmentally influenced epigenetic modifications. The earlier leads researchers to think about applying epigenetic mechanisms to enhance evolutionary algorithms performance in solving hard mathematical problems or real-world problems with continuous environmental changes \cite{RICALDE}.

This approach is not supported on the main idea of switching genes off and on (gene activation mechanism), or silencing chromosome sections like most of the approaches previously described. Epigenetic mechanisms, in particular, Histone and DNA modifications, go beyond the idea of activating and deactivating genes. As mentioned in state of the art, these mechanisms also involve decreasing or promoting the level of transcription by gradually reducing or increasing expression, depending on tags bind location or regions, and how many tag groups are attached \cite{CESARO, CLARK, FESTEN, SHAFA}. Methylation, for example, is sparsely found, but globally spread in indefinite {\em CpG sequences} throughout the entire genome, except for {\em CpG islands}, or specific stretches where high {\em CpG contents} are found \cite{PHILLIPS}.

Based on the preceding, this thesis assumes individuals' chromosomes to be entirely active; that is to say, epigenetic states on/off do not restrict gene/allele expression. Individuals' genotype is regulated by designed epigenetic tags that encode different meanings from on and off states. Tags encode rules with specific operations to be applied to the chromosome during the reading process (decoding) to generate the respective phenotype at a subsequent time and then evaluate it. 

Another important consideration is that the operations described in this thesis to perform the decoding process are not brought from biology. These operations are not present in the Epigenetic mechanisms neither but are plausible to solve computational problems. They are part of the elements involved in the epigenetic components of this approach and do not represent any biological operation. Biochemical processes Methylation and Histone Modification are regulated by an ``epigenetic code'' written as modifications in DNA and Histones and read by molecular units that specifically identify single or combined modifications, in this approach, an operation plus gene size represent a modification.

Epigenetics is the set of self-referential interactions between a group of genes, not necessarily from a gene towards itself, but a gene effect on another gene. Writers, readers, and erasers, for example, direct some interactions.  These kinds of interactions are referred to as epigenetic mutations in the literature and are reversible \cite{RAO}. Hence, they are not classical mutations (on nucleotide sequence), neither in the algorithmic context, but they are transient mutations, highly reversible, without any restriction to reversion. They can be seen as mutations on interactions level, not on the interacting objects (genes). Traditional evolutionary algorithms assume a finite number of genes, and to obtain novelty, they require not only mutations in their chromosome but also new genes. Epigenetics satisfies that need. Epigenetics becomes a problem-solver; it optimizes the number of genes and reduces classic mutation dependence. Epigenetics accelerates reversible mutation (environment may help on this) and reduces the cost of deleterious mutation (that reduce individual's fitness) or unresponsive mutations. In this approach, the given interactions during the marking (writing, erasing, and modifying actions) and reading processes may be seen as ``mutations''. However, these ``mutations'' are reversible, which is not the case with biological mutations on nucleotide sequences.

This approach aims to introduce a technique for Evolutionary Algorithms based on the adaptive and learning dynamics of the Epigenetic Regulation process. So that, the technique may reflect the adaptability of individuals during evolution, the ability of individuals to learn from the environment, the possibility of reducing time in the search process, and the discovery of better solutions in a given time. Based on the preceding, the dynamics of DNA Methylation and Histone Modification have been summarized into five fundamental elements that form the basis of this approach. First, a metaphorical representation of {\em Epigenetic tags} that are not off/on states, instead they represent reading rules to interpret sections (alleles) of an individual's genome. Second, a structural layer above the chromosome structure used to bind tags ({\em Epigenotype}). Third, a marker ({\em Marking Function}) to add, remove and modify tags, this process is performed between defined marking periods, simulating periods where individuals' genetic codes are affected by external factors (as seen in study cases of Överkalix and Dutch famine). Fourth, a tags interpreter or decoder ({\em Epigenetic Growing Function}) to generate individuals' phenotypes from their epigenotype-genotype structures. The marker and decoder are based on different enzymes' behaviors and principles (writers, readers, erasers, and the ones that work on markers maintenance). Finally, an inheritance process ({\em Crossover Operator}) to pass such tags onto the offspring (transgenerational non-genetic inheritance over subsequent generations).


The purpose of this chapter is to introduce the proposed technique to design epigenetic evolutionary algorithms with binary encoding. The epigenetic components of the model are described as follows: section \ref{c3s1} shows a detailed description of {\em Tags'} representation in the model; section \ref{c3s2} briefly characterizes the {\em Epigenotype}; section \ref{c3s3} explains the {\em Marking process}; section \ref{c3s4} describes the {\em Epigenetic Growing function}; section \ref{c3s5} illustrates adjustments on a {\em Crossover operator} to inherit tags in succeeding generations; and at the end, the Pseudocode of the epigenetic EA and a summary of this chapter are presented in sections \ref{c3s6} and \ref{c3s7} respectively.

\section{Tags and Encoding}\label{c3s1}
Epigenetic tags in the ReGen EA, are represented with a binary string sequence of 0's and 1's, and are located on alleles. Each set of tags is built with {\em 8}-bits, the first three bits represent a bit operation {\em(Circular shift, Transpose, Set to, Do nothing,   Right shift by one, Add one, Left shift by one, and Subtract one)} and the last {\em 5}-bits represent the gene size. Note that, the decimal representation of the {\em 5}-bits from {\em00001} to {\em11111} is used with no changes, but for {\em00000} the decimal value is thirty two. The first 3-bits string sequence uses a one-to-one mapping to a rule that performs a simple bit operation on chromosomes. The gene size says the number of alleles that are involved in the bit operation. Fig.~\ref{c3fig2} shows the tags' representation in the ReGen EA.

\subsection{Bit Operations}
Eight operations have been defined according to the {\em 3}-bits binary strings depicted in Fig.~\ref{c3fig2}. Each combination maps to a simple bit operation to be applied on a copy of the chromosome. The operations only impact the way alleles are read when evaluating the entire chromosome. An operation can be applied in a way that can affect a specific number of alleles/bits in later positions based on the {\em 5}-bits binary string that denotes the gene size $l$. The {\em Bit Operations} are described as follows:

\subsubsection{Circular shift}({\em 000}). Circularly shifts a specified number of bits to the right: starting at the marked bit up to $l$ bits ahead. Let $x$ be a binary string $x = (x_1, x_2, x_3,..., x_n)$, and $x_k$ be a bit marked with the shift tag and $l$ the gene size encoded by the tag. If the mark is read by the decoder, the decoded bit string $y$ will have $y_{k+i} = x_{k+i-1}$ for all $i=1,...,l-1$ and $y_{k}=x_{k+l-1}$.

\subsubsection{Transpose}({\em 001}). Transposes a specified number of bits: starting at the marked bit up to $l$ bits ahead. Let $x$ be a binary string $x = (x_1, x_2, x_3,..., x_n)$, and $x_k$ be a bit marked with the transposition tag and $l$ the gene size encoded by the tag. If the mark is read by the decoder, the decoded bit string $y$ will have $y_{k+i} = x_{k+l-1-i}$ for all $i=0,1,...,l-1$.

\subsubsection{Set to}({\em 010}). Sets a specified number of bits to a given value, the value of the marked bit: starting at the marked bit up to $l$ bits ahead. Let $x$ be a binary string $x = (x_1, x_2, x_3,..., x_n)$, and $x_k$ be a bit marked with the set-to tag and $l$ the gene size encoded by the tag. If the mark is read by the decoder, the decoded bit string $y$ will have $y_{k+i} = x_k$ for all $i=0,1,...,l-1$.

\subsubsection{Do nothing}({\em 011}). Does not apply any operation to a specified number of bits: starting at the marked bit up to $l$ bits ahead. Let $x$ be a binary string $x = (x_1, x_2, x_3,..., x_n)$, and $x_k$ be a bit marked with the do-nothing tag and $l$ the gene size encoded by the tag. If the mark is read by the decoder, the decoded bit string $y$ will have $y_{k+i} = x_{k+i}$ for all $i=0,1,...,l-1$.

\subsubsection{Right shift by one}({\em 100}). A right arithmetic shift of one position moves each bit to the right by one. This operation discards the least significant bit and fills the most significant bit with the previous bit value (now placed one position to the right). This operation shifts a specified number of bits: starting at the marked bit up to $l$ bits ahead. Let $x$ be a binary string $x = (x_1, x_2, x_3,..., x_n)$, and $x_k$ be a bit marked with the right shift by one tag and $l$ the gene size encoded by the tag. If the mark is read by the decoder, the decoded bit string $y$ will have $y_{k}=x_k$ and $y_{k+i} = x_{k+i-1}$ for all $i=1,...,l-1$.

\subsubsection{Add one}({\em 101}). Adds one to a specified number of bits: starting at the marked bit up to $l$ bits ahead. Let $x$ be a binary string $x = (x_1, x_2, x_3,..., x_n)$, and $x_k$ be a bit marked with the add one tag and $l$ the gene size encoded by the tag. If the mark is read by the decoder, the decoded bit string $y$ will have $y_{k+l-1-i} = x_{k+l-1-i} + 1 + carry$ for all $i=0,1,...,l-1$. In the case the number of bits in the result is greater than the initial addends, the decoder discards the rightmost bit in the binary number (least significant bit) in order to set the bits in the resulting chromosome.

\subsubsection{Left shift by one}({\em 110}). A left arithmetic shift of one position moves each bit to the left by one. This operation fills the least significant bit with zero and discards the most significant bit. This operation shifts a specified number of bits: starting at the marked bit up to $l$ bits ahead. Let $x$ be a binary string $x = (x_1, x_2, x_3,..., x_n)$, and $x_k$ be a bit marked with the left shift by one tag and $l$ the gene size encoded by the tag. If the mark is read by the decoder, the decoded bit string $y$ will have $y_{k+i} = x_{k+i+1}$ for all $i=0,1,...,l-2$ and $y_{k+l-1}=0$.

\subsubsection{Subtract one}({\em 111}). Subtracts one to a specified number of bits: starting at the marked bit up to $l$ bits ahead. Let $x$ be a binary string $x = (x_1, x_2, x_3,..., x_n)$, and $x_k$ be a bit marked with the subtract one tag and $l$ the gene size encoded by the tag. If the mark is read by the decoder, the decoded bit string $y$ will have $y_{k+l-1-i} = x_{k+l-1-i} + borrow - borrowed - 1$ for all $i=0,1,...,l-1$. When $1$ is subtracted from $0$, the borrow method is applied. The borrowing digit (zero) essentially obtains ten from borrowing ($borrow = 10$), and the digit that is borrowed from is reduced by one ($borrowed = 1$).

\par These operations have been selected due to their simplicity and capacity to generate, discover, and combine many possible building blocks. {\em Set to} operation, for example, it can be dominant depending on the optimization problem when maximizing or minimizing a function. The current operations combine short, high-fitness schemas resulting in high-quality building blocks of solutions after the epigenetic growing function is applied. If any allele has tags bound to it, regions of the chromosome will be read as the {\em Operation} states. In section~\ref{c3s4}, the epigenetic growing function and the application of these {\em Bit Operations} are explained in more detail.

\subsection{Gene Sizes}
As mentioned above, the last {\em 5}-bits of a tag represent the gene size. The gene size determines the number of alleles involved in the bit operation during the decoding process, Table~\ref{c3table1} briefly shows some binary strings and their respective values. These genes sizes have been proposed based on the order-$i$ schemas of the functions selected to perform experiments and the transformation of binary strings of 32 bits to real values. Fig.~\ref{c3fig2} depicts the complete structure of a tag.

\begin{table}[ht]
\centering
\caption{Gene Sizes}
\label{c3table1}
\begin{tabular}{cccccccl}
 \hline
 String & Value & String & Value & String & Value & String & Value\\\hline
   \verb"00001 "& 1 & \verb"00101" & 5 & \verb"11001" & 25 & \verb"11101" & 29\\
   \verb"00010" & 2 & \verb"00110" & 6 & \verb"11010" & 26 & \verb"11110" & 30\\
   \verb"00011" & 3 & \verb"00111" & 7 & \verb"11011" & 27 & \verb"11111" & 31\\
   \verb"00100" & 4 & \verb"01000" & 8 & \verb"11100" & 28 & \verb"00000" & 32\\
  \hline
\end{tabular}
\end{table}

\section{Epigenotype}\label{c3s2}
The {\em Epigenotype} is a structural layer on the chromosome top structure used to attach tags. This second layer represents individuals' epigenome, and it is a structure with the same size as an individual's chromosome. This epigenetic component holds a set of epigenetic changes that influence the direction of the search process. It is coded as a multidimensional vector $(m \cdot n)$, where $m$ is the tag length, and $n$ is the length of the individual's chromosome.

\section{Marking Function}\label{c3s3}
The Marking function adds tags to or removes tags from alleles of any chromosome in the solution space. Additionally, it can modify the {\em8-bits} binary string tags. The marking process works with a marking rate, that is, the probability of applying the function on every bit of a chromosome. When the probability is positive, the function generates a probability of adding a tag to one single allele, removing a tag from one single allele, or modifying a tag on any allele. These actions cannot happen simultaneously and are mutually exclusive. Tags are randomly added or removed from any allele. Also, the {\em modify} action randomly changes any of the eight positions in the binary string. The distribution of these actions is given as follows in Equation~\ref{c3eq1}. 

\begin{eqnarray}
P_{Marking} = 
\begin{cases}
\text{No Marking $0.98$}\\
\text{Add a tag (8-bits) $0.007$} \\
\text{Remove a tag (8-bits) $0.007$} \\
\text{Modify a tag (any bit of a tag) $0.006$}
\end{cases}
\label{c3eq1}
\end{eqnarray}

\begin{figure}
\centering
\includegraphics[width=2.8in]{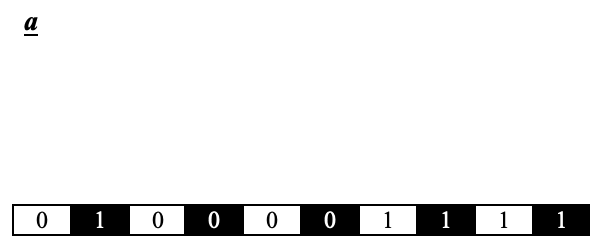}
\includegraphics[width=2.8in]{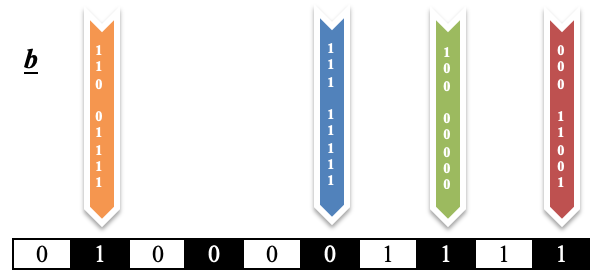}
\includegraphics[width=2.8in]{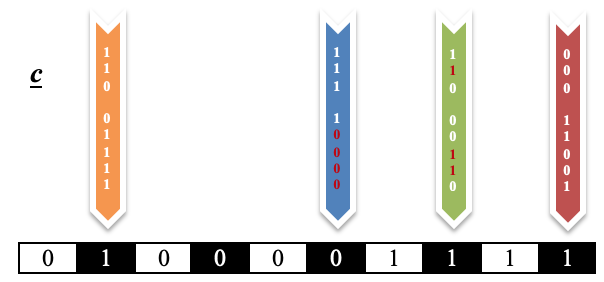}
\includegraphics[width=2.8in]{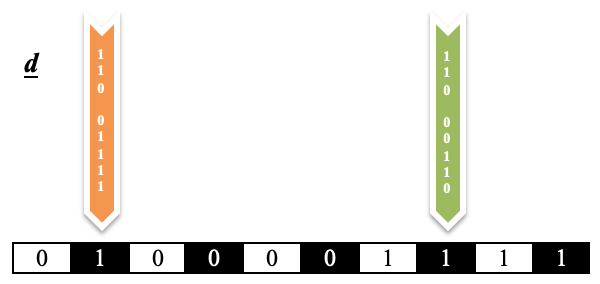}
\caption{General representation of the {\em Marking} function: a) shows a chromosome with no tags on it; b) depicts the addition of four tags to a chromosome; c) illustrates tags' bit modification in red; and d) presents a chromosome with two removed tags.}
\label{c3fig1}
\end{figure}

The probability of marking a single bit of a chromosome is defined by taking into account three factors. First, biologically epigenetics marks of the sort of Methylation, for example, are dispersed in indeterminate {\em CpG sequences} on the genome, except for {\em CpG islands}, or specific areas where high {\em CpG contents} are present. Despite that, they can affect gene expression. They are powerful because of what they encode, not for the quantity, this means for better or worse, a few tags can have the influence or potential to make individuals bits being interpreted in such a way that good or poor results could be obtained. The second factor aims to avoid chromosomes over-marking, if each bit is marked, it could cause over-processing during the decoding process. The third factor is related to the definition of a marking probability that allows the {\em Marking function} to keep the marking process balanced; a probability value that ensures tags diversity and a considerable number of marked positions. 

Based on previous considerations, experiments to define a marking probability, involve tuning the marking process with different rate values from {\em0.1} to {\em1.0}. The higher the rate, the less effective is the marking function. When the rate is reduced, the marking function reveals an equilibrium between the applied actions and the obtained solutions, after running experiments with lower rates from {\em0.01} to {\em0.09}, the rate of {\em0.02} has shown to be enough to influence the search process and help ReGen EAs in finding solutions closer to the optimum. Consequently, the probability of marking a single bit has been set to {\em\textbf{0.02}}. Following the definition of such a probability rate, the probability distribution for adding, removing, and modifying is set based on the significance of having a considerable number of tags and a variety of them. If tags are added, they should be eventually removed, or chromosomes will be over marked; for this, the approach gets rid of tags with the same probability as the {\em add} tag action. Then, the {\em modify} tag action uses a lower probability to recombine the {\em8-bits} of a tag and generate different decoding frames. 
The influence or impact of the designed actions is mostly that they altogether:
\begin{enumerate}
    \item Let individuals have a reasonable quantity of tags,
    \item Allow bit combination for tags, and
    \item Ensure discovering building blocks during tags interpretation that could generate solutions that are not neighbor to current solutions to escape from a local optimum.
\end{enumerate}

\subsection{Adding a tag}
This action writes tags on any chromosome. {\em Add} is a metaphorical representation of {\em writer} enzymes. Fig.~\ref{c3fig1} presents a chromosome (image {\em \textbf{a}}) with no tags. In image {\em \textbf{b}}, four tags are added at positions {\em2}, {\em6}, {\em8}, and {\em10}, based on the defined {\em add tag} probability of  {\em0.007}.

\subsection{Modifying a tag}
This action modifies tags on any chromosome. {\em Modify} is a metaphorical representation of {\em maintenance} enzymes. In Fig.~\ref{c3fig1}, image {\em \textbf{c}} illustrates modified tags at positions {\em6} and {\em8}, bits in red changed. This action is applied under the defined {\em modify tag} probability of {\em0.006} and then randomly changes any of the eight positions in the binary string with a rate of $1.0/l$, where $l$ is the tag's length.

\subsection{Removing a tag}
This action erases tags from any chromosome. {\em Remove} is a metaphorical representation of {\em eraser} enzymes. Fig.~\ref{c3fig1}, depicts a chromosome with removed tags. In Image {\em \textbf{d}}, two tags at positions {\em6} and {\em10} are not longer bound to the chromosome. The tag removal is performed with the defined {\em remove tag} probability of {\em0.007}.

\begin{landscape}
\begin{figure}
\centering
\includegraphics[height=4in, width=9.3in]{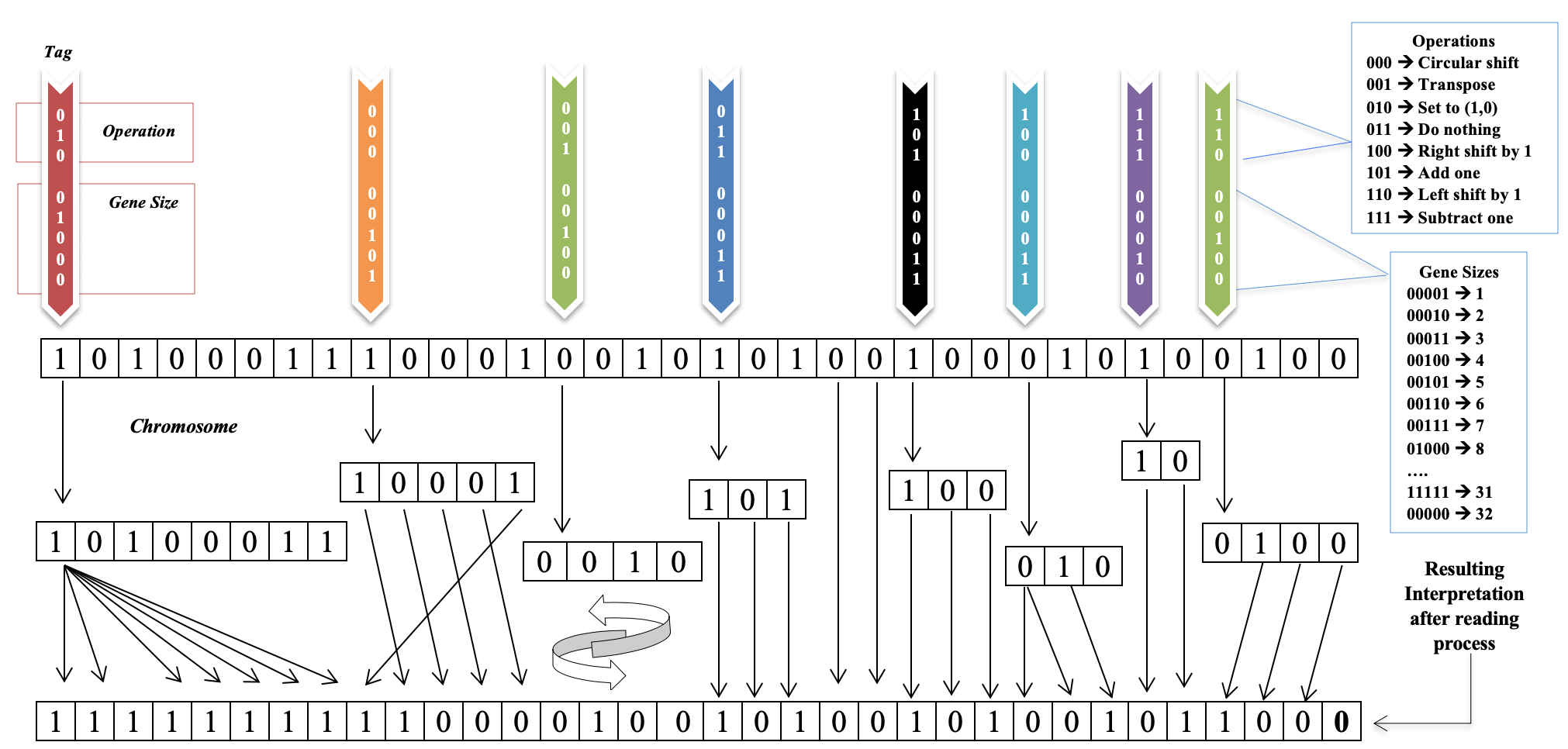}
\caption{General representation of an individual with its epigenotype. The bottom section shows the tag's interpretation process to generate a bit string used to build the individual's phenotype.}
\label{c3fig2}
\end{figure}
\end{landscape}

\section{Epigenetic Growing Function}\label{c3s4}

This function is a metaphorical representation of {\em reader} enzymes. The epigenetic growing function generates bit strings from individuals' genotypes-epigenotypes for eventual phenotypes creation. Tags allow this function to build different individuals before the quality or fitness of each solution is evaluated. The growth happens in the binary search space (coded solutions), but is reflected in the solution space (actual solutions). From a mathematical point of view, the search space is transformed and reduced when the tags' interpretation is performed; this ensures both exploration and exploitation. The first to reach different promising regions in a smaller search space and the second to search for optimal solutions within the given region. The bigger are {\em gene size} values, the less variety of building blocks will result during decoding. Tags may lead individuals to be represented as closer feasible solutions to some extreme (minimum or maximum) in the search space. When this function is applied, chromosomes grow in the direction of minimum or maximum points, depending on the problem. This process differs from mutations or hyper-mutations which modify chromosomes and maintain genetic diversity from one generation of a population of chromosomes to another on a broader search space.

The Epigenetic Growing function acts like an interpreter or decoder of tags located over a particular allele. This function scans each allele of a chromosome and the tags that directly affect it, so that, the phenotypic variations can be reflected when evaluating individuals. During the decoding process, alleles are not changed; the chromosome keeps its binary encoding fixed. This means the individual's genotype is not altered. Note that the scope of the {\em Operations} to be applied depends on the {\em gene size} indicator. If an {\em Operation} has been already applied, and there is another one to be applied, the epigenetic growing function considers the interpretation of the previous bits in order to continue its decoding process. An example of the prior process to phenotype generation is illustrated in Fig.~\ref{c3fig2}. The example shows the decoding process for each bit with or without tags.

On the top of Fig.~\ref{c3fig2}, a chromosome with a size of {\em34} and eight tags is depicted. Alleles in positions {\em1, 9, 14, 18, 23, 26, 29} and {\em31} are marked with colored tags. The decoding starts from left to right. The first position is scanned, as it has a tag bound to it, the function initiates a tag identification. The tag in red is {\em01001000}, the first three bits {\em010} indicate an operation that is {\em Set to}. It means to set a specified number of bits to the same value of the allele, which is {\em1}. The specified number of bits to be {\em Set to}, is indicated by the last five bits of the tag, which are {\em01000}, this refers to a gene size ($l$) of {\em5}-bits. Then, the resulting interpretation is to set all bits to {\em1}, starting at the marked bit up to the gene size minus one ($l-1$). After finishing the first decoding, the epigenetic growth function continues scanning at position ($l+1$) and keeps the previous result. This process is repeated until the entire chromosome is scanned; each result is concatenated to generate a final bit string; the length of the chromosome and the resulting string keep fixed. At the bottom of Fig.~\ref{c3fig2}, a final interpretation is shown, the concatenated string is the source to build the phenotype, which is evaluated and gives the score for the individual.

\section{Crossover Operator}\label{c3s5}

\begin{landscape}
\begin{figure}
 \centering
\includegraphics[height=6in, width=8in]
{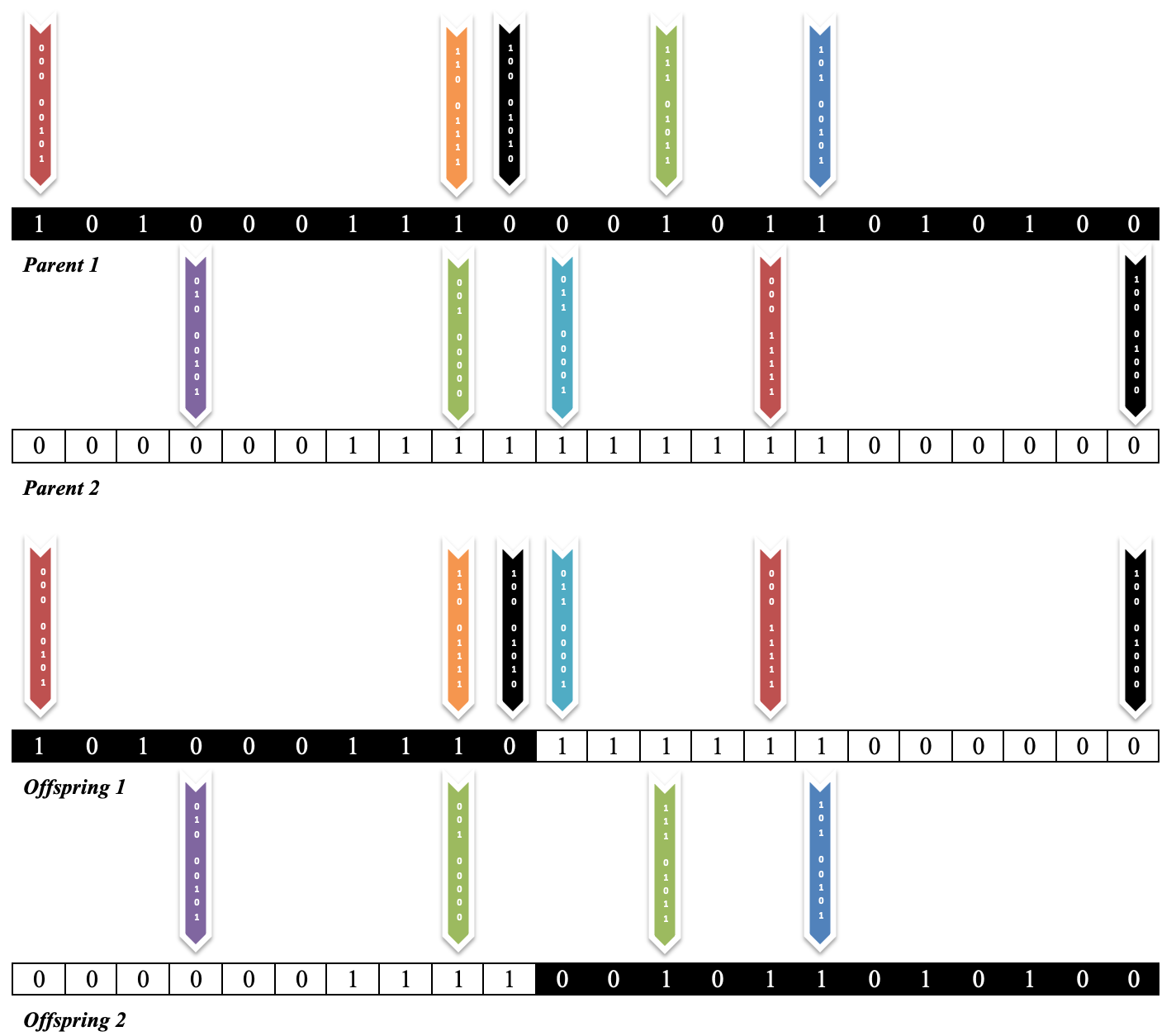}
\caption{Illustrative example of genetic and epigenetic recombination: Simple Point Crossover operation.}
\label{c3fig3}
\end{figure}
\end{landscape}

The epigenetic tags added to the chromosome are inherited by the offspring during Crossover. Transgenerational epigenetic inheritance transmits epigenetic markers from one individual to another (parent-child transmission) that affects the offspring's traits without altering the sequence of nucleotides of DNA. The idea here is to perform the recombination process based on the selected operator, as usual, the only difference is that tags located on alleles will be copied along with the genetic information. This model presents the Single Point Crossover as an illustrative example of genetic and epigenetic recombination. So, a calculated cross point {\em x} will be applied to the chromosome at {\em x} position. By doing this process, the offspring will inherit alleles with their tags. Fig.~\ref{c3fig3} shows the exchange of genetic code and epigenetic tags. A Simple Point Crossover operation is performed over given parents at cross point {\em10}. {\em Offspring 1} inherited from {\em Parent 1}, part of the genetic code plus its tags in positions {\em1, 9} and {\em10}. From {\em Parent 2}, it also inherited part of the genetic code plus some tags in positions {\em11, 15} and {\em22}. {\em Offspring 2} got part of the genetic code plus its tags in positions {\em13} and {\em16} from {\em Parent 1}. From {\em Parent 2}, it got part of the genetic code plus its tags in positions {\em4} and {\em9}.

\section{Pseudo Code}\label{c3s6}
The sequence of steps for the proposed ReGen EA is defined in Algorithm~\ref{c3algo1} and Algorithm~\ref{c3algo2}. Note that the pseudo-code includes the same elements of a generic evolutionary algorithm. The epigenetic components are embedded, as defined in Algorithm~\ref{c3algo1}. The ReGen EA's behavior is similar to EA's standard versions until defined marking periods and tags decoding processes take place. Note that the reading process is different when chromosomes are marked, phenotypes are built based on tags interpretation. This process firstly identifies the operation to be applied on a specific section of a chromosome and secondly the gene size to define the scope of the bit operation, as depicted in Fig.~\ref{c3fig2}.

The epigenetic EA incorporates a pressure function to perform the marking process during a specific period. A range of iterations determines a period. Marking periods represent the environment, an abstract element that has been a point of reference to assess the results of this model. At line $7$, the function {\em markingPeriodON} validates the beginning of defined periods, this indicates that the marking process can be performed starting from iteration $a$ to iteration $b$. Any number of marking periods can be defined. Periods could be between different ranges of iterations. Additionally, the {\em epiGrowingFunction} is embedded at line $10$. The {\em epiGrowingFunction} interprets tags and generates a bit string used to build the phenotype before initiating the fitness evaluation of individuals. The standard EA uses the individual's genotype to be evaluated; in contrast, the epigenetic technique uses the resulting phenotype from the tags decoding process. This technique is called ReGen EA, which means Evolutionary Algorithm with Regulated Genes.  

\begin{algorithm}[H]
\caption{Pseudo code of a ReGen EA}
\begin{algorithmic}[1]
\State \textbf{initialize}   {\em population with random candidate solutions}
\State \textbf{evaluate}   {\em each candidate}
\Repeat
\State \textbf{select}   {\em parents}
\State \textbf{recombine}   {\em pairs of parents}
\State \textbf{mutate}   {\em the resulting offspring}
\If {\Call{markingPeriodON}{iteration}}
 \State \Call{applyMarking}{offspring}
\EndIf
\State $phenotypes \gets$ \textsc{decode}(\Call{epiGrowingFunction}{offspring})
\State \textbf{evaluate} {\em phenotypes of the new candidates}
\State \textbf{select}   {\em individuals for the next generation}
\Until{\em Termination condition is satisfied}
\end{algorithmic}
\label{c3algo1}
\end{algorithm}

\begin{algorithm}[H]
\caption{Pseudo code of a ReGen EA}
\begin{algorithmic}[1]
\Function{markingPeriodON}{$it$}
    \State $start \gets startValue$
      \State $end \gets endValue$
      \If { $start$ $ \geq it \leq $ $end$}
         \State \Return $true$ 
     \EndIf
      \State \Return $false$
\EndFunction
\newline
\Function{applyMarking}{offspring}
   \State $mark \gets P_{Marking}$ \Comment{probability of 0.02}
   \State $notModify \gets P_{Adding} + P_{Removing}$ \Comment{same probability of 0.35 to add and remove}
   \For{\textbf{each} allele $\in$ offspring$_{i}$ chromosome}
		\If {$mark$}
            \If {$notModify$}
                \If {add}
                     \If {notMarked}
                        \State{add tag}
                      \EndIf
                \Else
                    \If {isMarked}
                        \State{remove tag}
                     \EndIf
                \EndIf
            \Else
               \If {isMarked}
                    \State{modify any tag' bit with a rate of 1.0/tag length}
                \EndIf
            \EndIf
        \EndIf
   \EndFor
\EndFunction
\newline
\Function{epiGrowingFunction}{offspring}
    \State $bitStrings \gets$ offspring
     \If {offspring isMarked}
         \State $bitStrings \gets$ \textbf{read offspring marks}
     \EndIf
      \State \Return $bitStrings$
\EndFunction
\end{algorithmic}
\label{c3algo2}
\end{algorithm}

\section{Summary}\label{c3s7}
This chapter describes the proposed epigenetic technique under the scope of this thesis. Five fundamental elements form the basis of the designed technique (ReGen EA): first, a metaphorical representation of {\em Epigenetic Tags} as binary strings; second, a layer on chromosome top structure used to bind tags ({\em Epigenotype}); third, a {\em Marking Function} to add, remove, and modify tags; fourth, a {\em Epigenetic Growing Function} that acts like an interpreter, or decoder of tags located on the {\em Epigenotype}; and fifth, tags inheritance by the offspring during {\em Crossover}. The abstraction presented in this chapter describes a way to address a large number of computational problems with binary and real encoding. This technique may find approximately optimal solutions to hard problems that are not efficiently solved with other techniques.

    \chapter{ReGen GA: Binary and Real Codification}\label{chapter4}

Genetic Algorithm with Regulated Genes (ReGen GA) is the implementation of the proposed epigenetic model on a classic GA. The general terminology of a GA includes population, chromosomes, genes, genetic operators, among others. The ReGen GA has a layer to attach tags and involves two functions named {\em Marking} and {\em Epigenetic Growing}. The first function simulates periods in which individuals' genetic codes are affected by external factors, represented by the designed tags. The second function generates bit strings from genotypes and their respective epigenotypes for phenotypes formation (see Algorithm~\ref{c4algo1}). Also, the ReGen GA uses Simple Point {\em Crossover} operator to perform recombination and transmission of epigenetic markers from one individual to its descendants. 

This chapter aims to present the application of the proposed epigenetic model. This implementation intends to address real and binary encoding problems. Experimental functions with binary and real encoding have been selected to determine the model applicability. The experiments will evidence the effect of the tags on population behavior. In section~\ref{c4s1}, experimental setups and parameters configuration used for the selected functions are described. In section~\ref{c4s2}, a set of binary experiments is presented, implementing Deceptive (orders three and four), Royal Road, and Max Ones functions. Additionally, some experimental results and their analysis are exhibited in subsections~\ref{c4s2ss2} and \ref{c4s2ss3}. In section~\ref{c4s3}, a set of real experiments is presented, implementing Rastrigin, Rosenbrock, Schwefel, and Griewank functions. Also, some experimental results and their analysis are reported in subsections~\ref{c4s3ss2} and \ref{c4s3ss3}. At the end of this chapter, a summary is given in section~\ref{c4s4}.

\begin{algorithm}[H]
\caption{Pseudo code of the ReGen GA}
\begin{algorithmic}[1]
\State \textbf{initialize}   {\em population with random candidate solutions}
\State \textbf{evaluate}   {\em each candidate}
\Repeat
\State \textbf{select}   {\em parents}
\State \textbf{recombine}   {\em pairs of parents}
\State \textbf{mutate}   {\em the resulting offspring}
\If {\Call{markingPeriodON}{iteration}}
 \State \Call{applyMarking}{offspring}
\EndIf
\State $phenotypes \gets$ \textsc{decode}(\Call{epiGrowingFunction}{offspring})
\State \textbf{evaluate} {\em phenotypes of the new candidates}
\State \textbf{select}   {\em individuals for the next generation}
\Until{\em Termination condition is satisfied}
\end{algorithmic}
\label{c4algo1}
\end{algorithm}

\section{General Configuration}\label{c4s1}

The following configuration applies to all experiments presented in this chapter. It is well known that an algorithm can be tweaked (e.g., the operators in a GA) to improve performance on specific problems, even though, this thesis intends to avoid giving too many advantages to the performed GA implementations in terms of parametrization. The classic GA and the ReGen GA parameters are tuned with some variations on only two standard operators, Single Bit Mutation and Simple Point Crossover. Also, for all experiments, three marking periods have been defined, note that the defined number of periods are just for testing purposes. Marking Periods can be appreciated in figures of reported results delineated with vertical lines. Vertical lines in blue depict the starting point of marking periods and gray lines, the end of them.

The set up for classic GAs includes: $30$ runs; $1000$ iterations; population size of $100$ individuals; a tournament of size $4$ for parents selection; generational (GGA) and steady state (SSGA, in which replacement policy is elitism) replacements to choose the fittest individuals for the new population; each bit in the chromosome has a mutation rate of $1.0/l$, where $l$ is the chromosome length, while the single point crossover rates are set from $0.6$ to $1.0$.

The set up for the ReGen GA includes: $30$ runs; $1000$ iterations; population size of $100$ individuals; a tournament of size $4$ for parents selection; generational (GGA) and steady state (SSGA, in which replacement policy is elitism) replacements to choose the fittest individuals for the new population; each bit in the chromosome has a mutation rate of $1.0/l$, where $l$ is the chromosome length, while the single point crossover rates are set from $0.6$ to $1.0$; a marking probability of $0.02$ (the probability to add a tag is $0.35$, to remove a tag is $0.35$, and to modify a tag $0.3$) and three marking periods have been defined. Such periods start at iterations 200, 500, and 800, with a duration of 150 iterations each.

It is worth mentioning that the crossover rate of $0.7$, along with the mutation rate of $1.0/l$, are considered good parameters to solve binary encoding problems \cite{TBACK, MITCHELL2}, even though, five crossover rates are used to evaluate tags inheritance impact. Table~\ref{c4table1} shows a summary of the general setup for the experiments.


\begin{table}[h]
\centering
\caption{General configuration with 5 different Crossover rates}
\label{c4table1}
\begin{tabular}{lcc}
\hline
 Factor Name & \textbf{Classic GA} & \textbf{ReGen GA}\\\hline
Mutation Operator Rate  & $1.0/l$ & $1.0/l$\\
Crossover Operator Rate  & 0.6 - 1.0 & 0.6 - 1.0 \\
Marking Rate  & {\em none} & 0.02\\
Marking Periods  & {\em none} & 3 \\
Population Size & 100 & 100 \\
Generations  & 1000 & 1000 \\
Runs & 30 & 30 \\
Parent selection & {\em Tournament} &  {\em Tournament} \\\hline
\end{tabular}
\end{table}

\section{Binary Problems}\label{c4s2}

Chromosomes encoding is one of the challenges when trying to solve problems with GAs; encoding definition depends on the given problem. Binary encoding is the most traditional and simple, essentially because earlier GA implementations used this encoding type. This section reports experiments with four different binary functions.

\subsection{Experiments}\label{c4s2ss1}

Performing experiments use binary encoding for determining the proposed technique applicability. In binary encoding, a vector with binary values encodes the problem's solution. Table~\ref{c4table2} shows a simple example of functions with a single fixed bit string length. These functions have been chosen to work on the first approximation to test this technique. Be aware that this does not mean a different bit string length is not allowed. Any length value can be set. The selected functions and fixed string length values are just for the purpose of making the experiments simpler and easier to understand.

\begin{table}[h]
\centering
\caption{Experimental Functions}
\label{c4table2}
\begin{tabular}{cccl}
\hline
 Function & Genome Length & Global Optimum\\ \hline
\verb"Deceptive 3"  & 360 & 3600 \\
\verb"Deceptive 4"  & 360 & 450 \\
\verb"Royal Road"   & 360 & 360 \\
\verb"Max Ones"     & 360 & 360 \\\hline
\end{tabular}
\end{table}

\subsubsection{Deceptive Order Three and Deceptive Order Four Trap}
The deceptive functions proposed by Goldberg in 1989 are challenging problems for conventional genetic algorithms (GAs), which mislead the search to some local optima (deceptive attractors) rather than the global optimum \cite{GOLDBERG}. An individual's fitness is defined as indicated in Table~\ref{c4table3} and Table~\ref{c4table4} for Deceptive order three and Deceptive order four trap, respectively.

\begin{table}[H]
\centering
\caption{Order Three Function}
\label{c4table3}
\begin{tabular}{ccccl}
\hline
 String & Value & String & Value \\\hline
   \verb"000 "& 28 & \verb"100" & 14\\
   \verb"001" & 26 & \verb"101" &  0\\
   \verb"010" & 22 & \verb"110" &  0\\
   \verb"011" & 0   & \verb"111" & 30\\\hline
\end{tabular}
\end{table}

\begin{table}[H]
\centering
\caption{Order Four Trap Function} 
\label{c4table4}
\begin{tabular}{ccccl}
\hline
String & Value & String & Value \\\hline
    \verb"0000" & 5 & \verb"1000" &  1\\
    \verb"0001" & 1 & \verb"1001" &  2\\
    \verb"0010" & 1 & \verb"1010" &  2\\
    \verb"0011" & 2 & \verb"1011" &  3\\
    \verb"0100" & 1 & \verb"1100" & 2\\
    \verb"0101" & 2 & \verb"1101" & 3\\
    \verb"0110" & 2 & \verb"1110" &  3\\
    \verb"0111" & 3 & \verb"1111" &  4\\\hline
\end{tabular}
\end{table}

\subsubsection{Max Ones and Royal Road}
The Max Ones' problem (or BitCounting) is a simple problem that consists of maximizing the number of $1$'s in a chain. The fitness of an individual is defined as the number of bits that are 1. Formally, this problem can be described as finding a string $x = (x_1, x_2, x_4,..., x_n)$, where $x_i \in \{0,1\}$, which maximizes the following Equation~\ref{c4eq1}:

\begin{eqnarray}
f(x)=\sum_{i=1}^{n}x_i\;
\label{c4eq1}
\end{eqnarray}

The Royal Road function developed by Forrest and Mitchell in 1993 \cite{FORREST-MITCHELL}, consists of a list of partially specified bit strings (schemas) with a sequence of 0's and 1's. A schema performs well when all bits are set to 1. For the experiments, order-8 schemas are configured. 

A simple Royal Road function, $R_{1}$ is defined by Equation~\ref{c4eq2}. $R_{1}$ consists of a list of partially specified bit strings (schemas) $s_{i}$ in which (\textquoteleft${*}$\textquoteright) denotes a wild card (i.e., allowed to be either 0 or 1). A bit string $x$ is said to be an instance of a schema $s, x \in s$, if $x$ matches $s$ in the defined (i.e., non-\textquoteleft${*}$\textquoteright) positions. The fitness $R_{1}(x)$ of a bit string $x$ is defined as follows:

\begin{eqnarray}
R_{1}(x)=\sum_{i=1}^{8}\delta_{i}(x)o(s_{i}), where \ \delta_{i}(x) = 
\begin{cases}
\text{$1$  if  $x \in s_{i}$} \\ 
\text{$0$ otherwise} 
\end{cases}
\label{c4eq2}
\end{eqnarray}

\subsection{Results}\label{c4s2ss2}
Based on the defined configuration, both classic GA and ReGen GA are compared to identify the behavior of tags during individuals' evolution. Results are tabulated from Table~\ref{c4table5} to Table~\ref{c4table8}, these tables present the binary functions: Deceptive order three (D3), Deceptive order four trap (D4), Royal Road (RR), and Max ones (MO). Both EA implementations with generational (GGA), steady state (SSGA) replacements, and five crossover rates per technique. For each rate, the best fitness based on the maximum median performance is reported, following the standard deviation of the observed value, and the iteration where the reported fitness is found. The iteration is enclosed in square brackets.

\begin{figure}[H]
\centering
\includegraphics[width=1.9in]{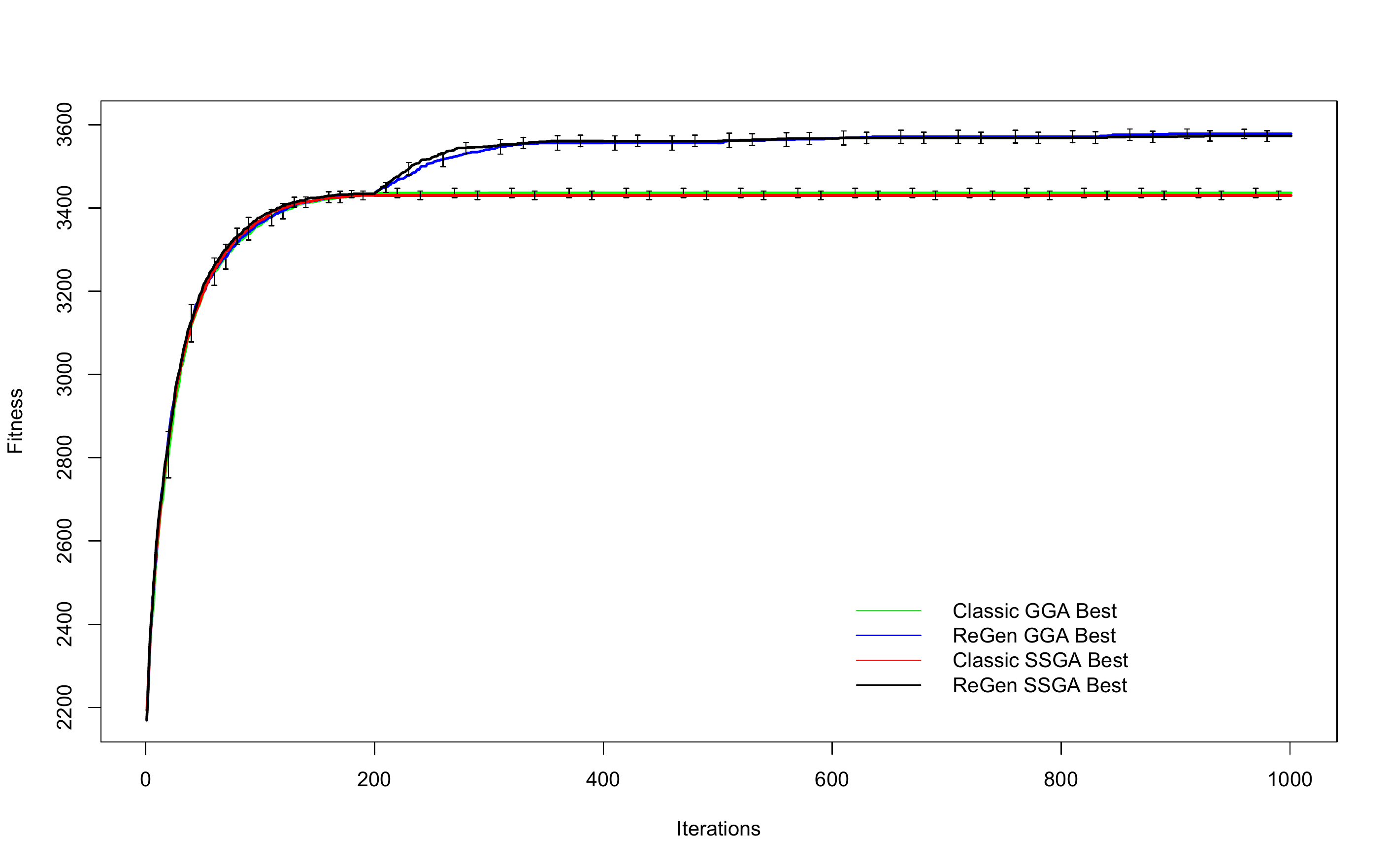}
\includegraphics[width=1.9in]{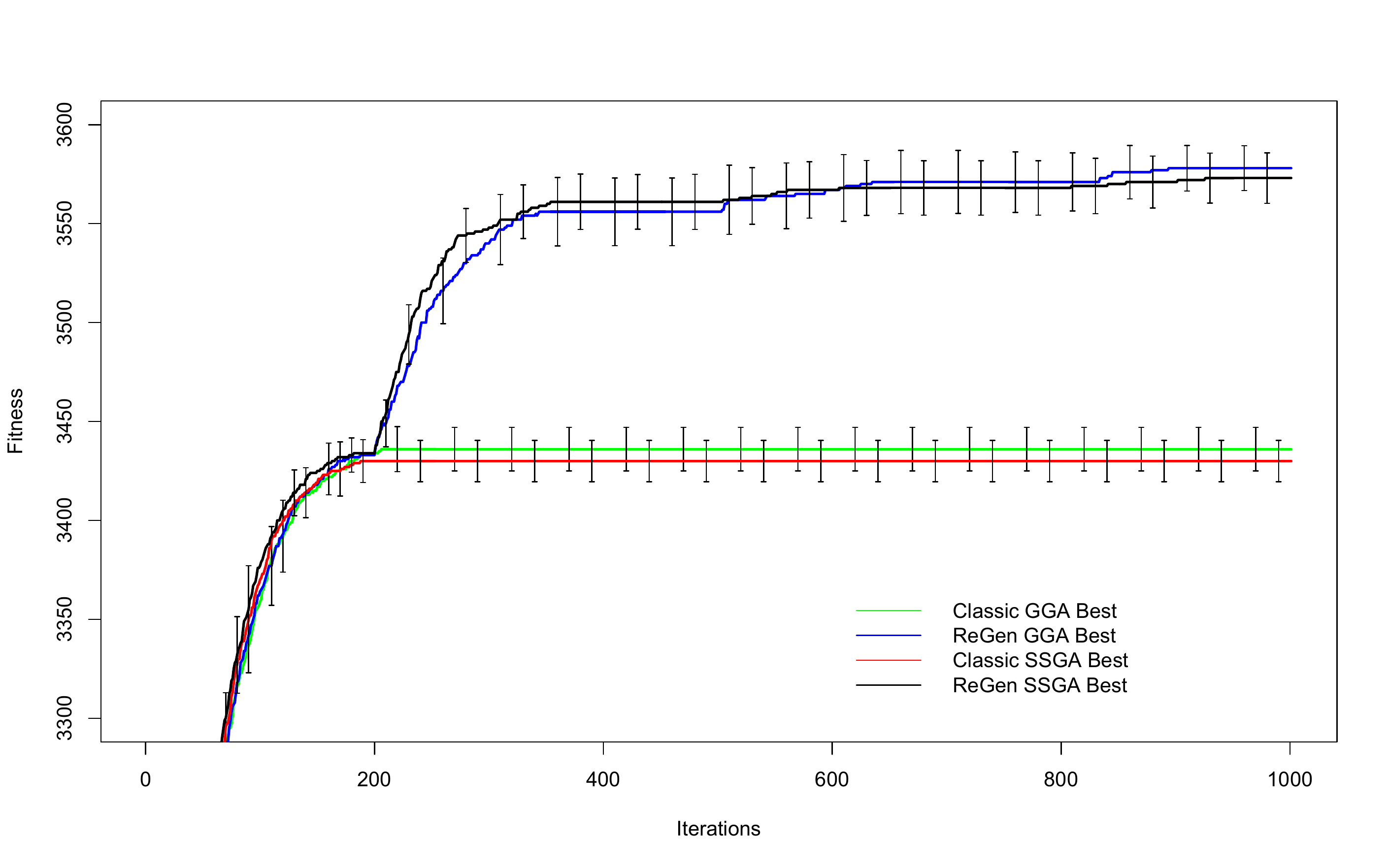}
\includegraphics[width=1.9in]{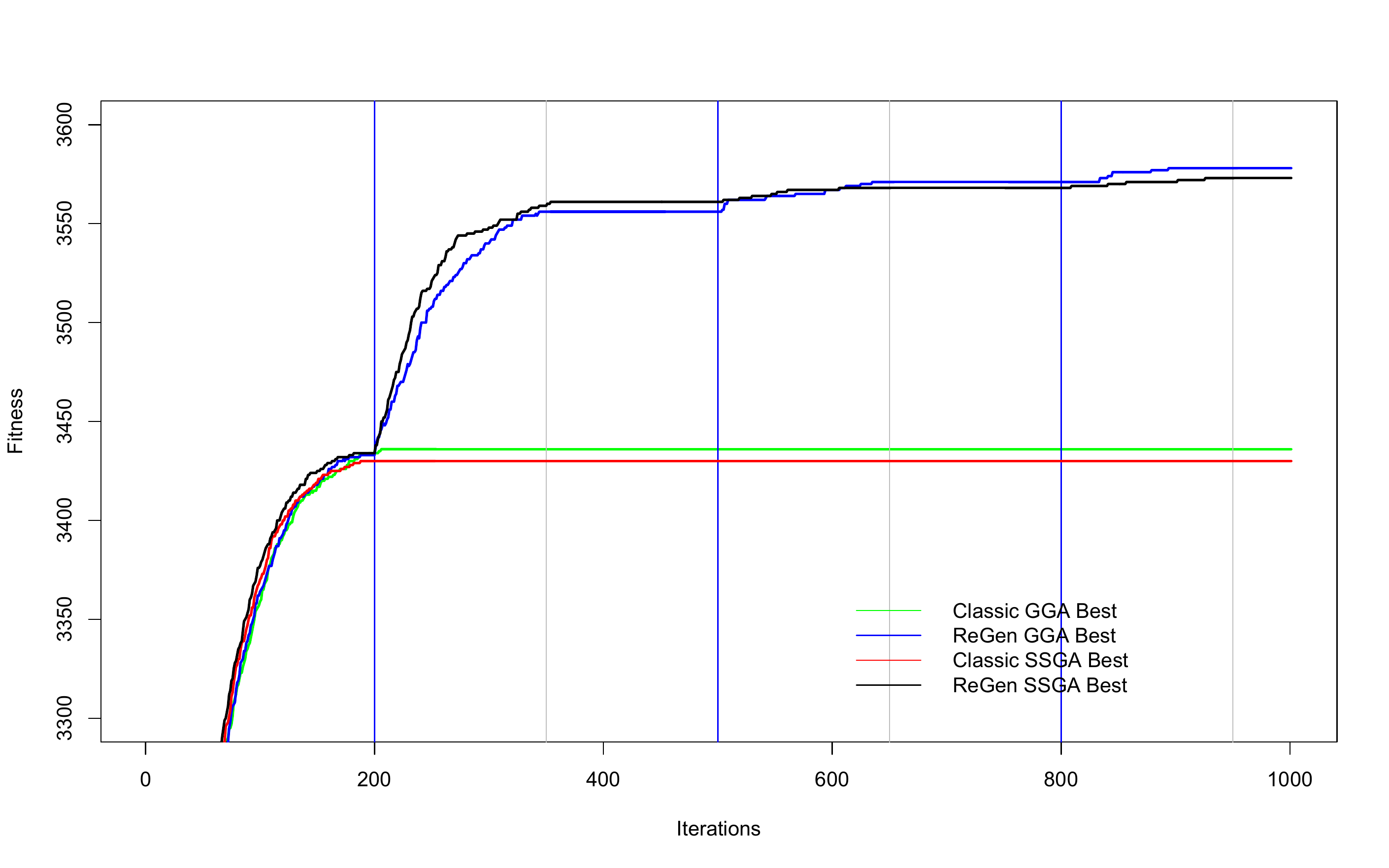}
\includegraphics[width=1.9in]{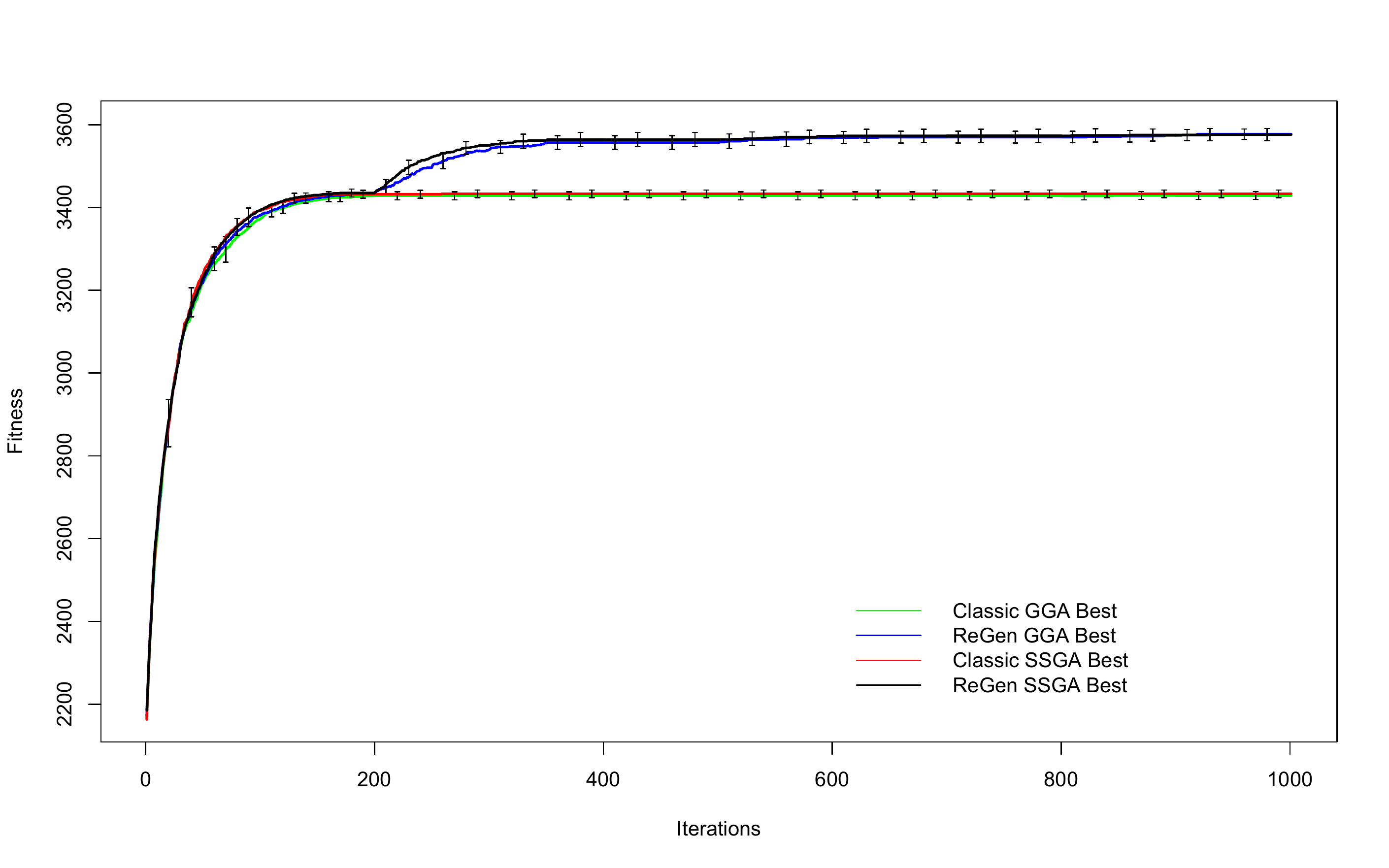}
\includegraphics[width=1.9in]{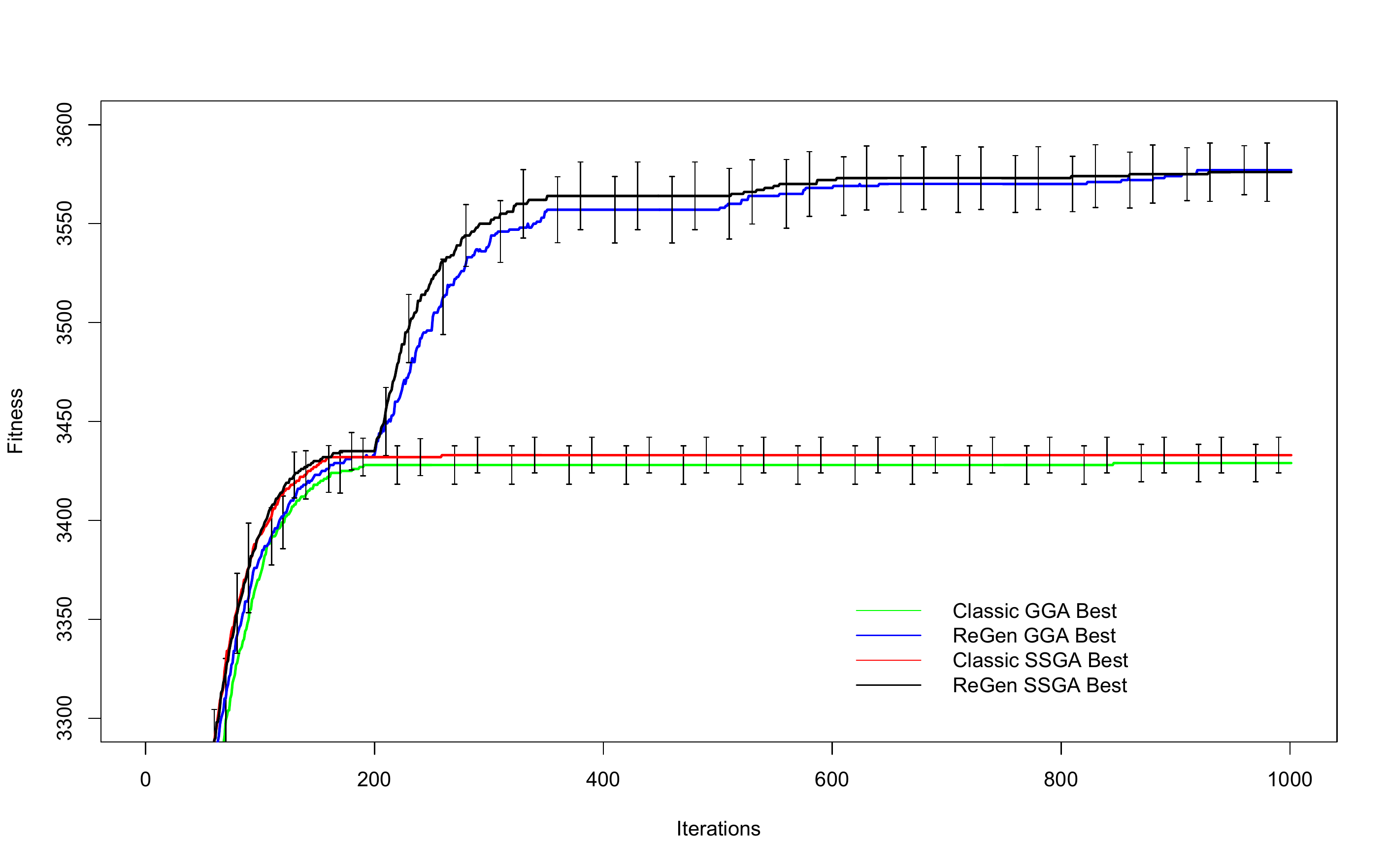}
\includegraphics[width=1.9in]{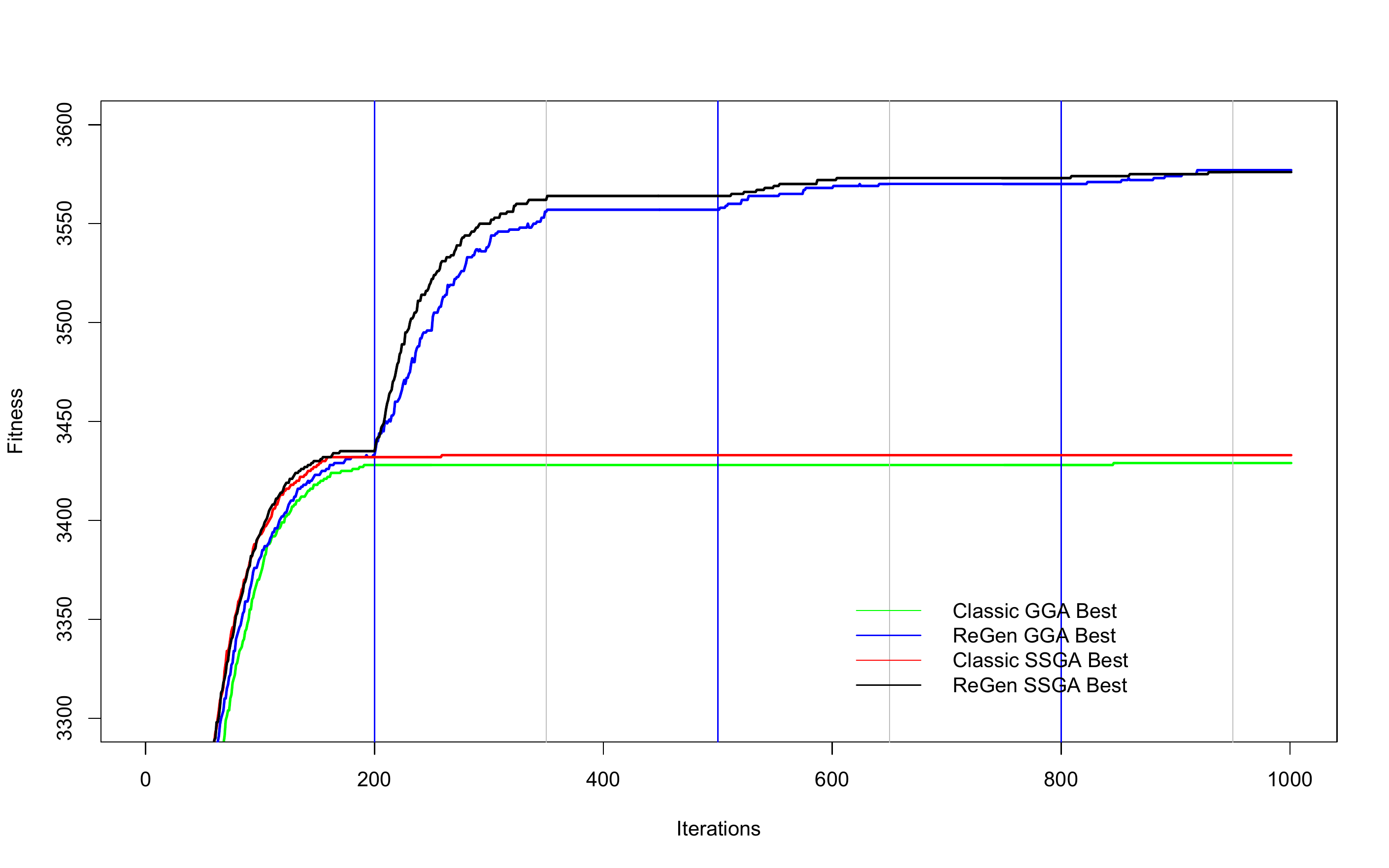}
\includegraphics[width=1.9in]{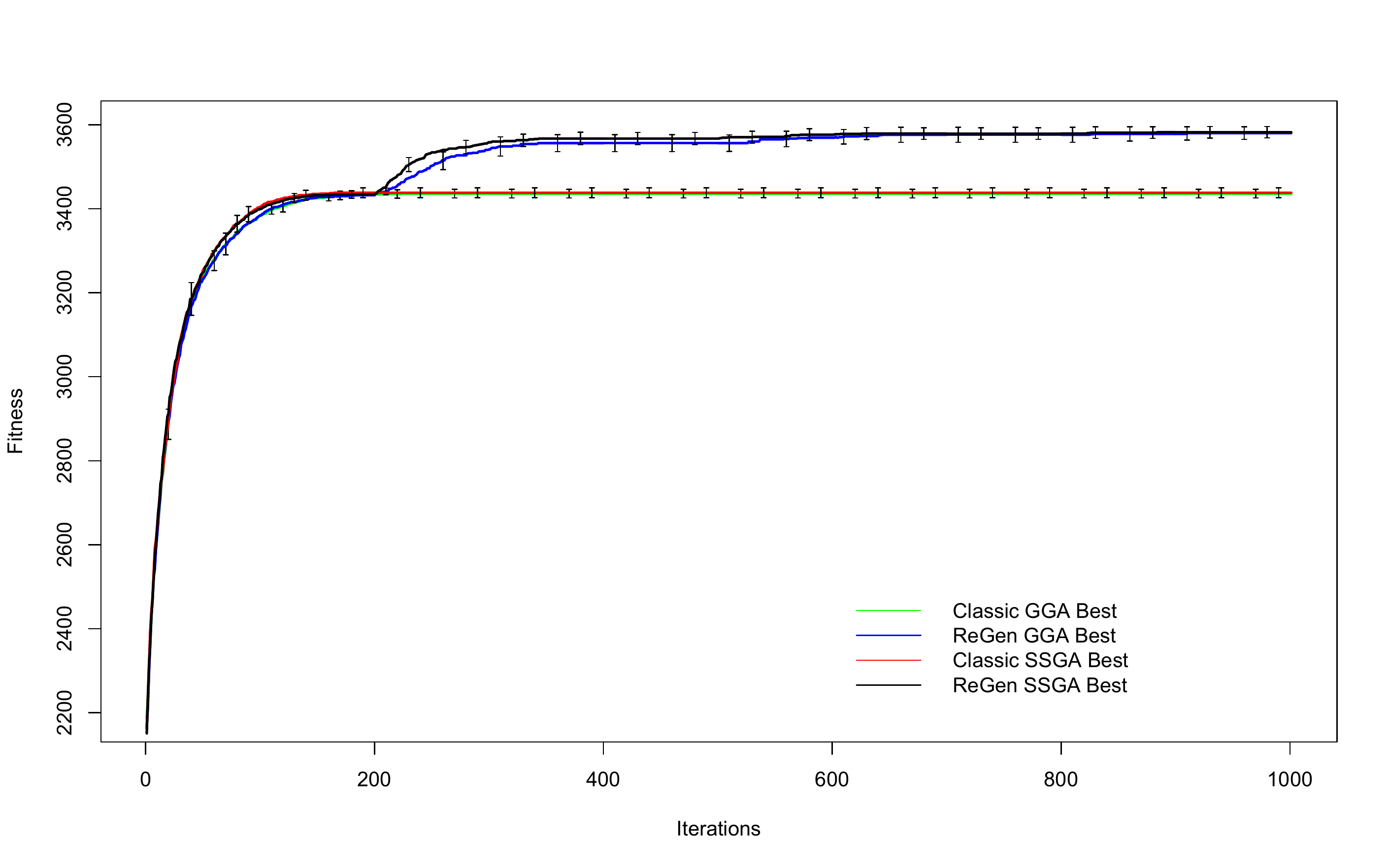}
\includegraphics[width=1.9in]{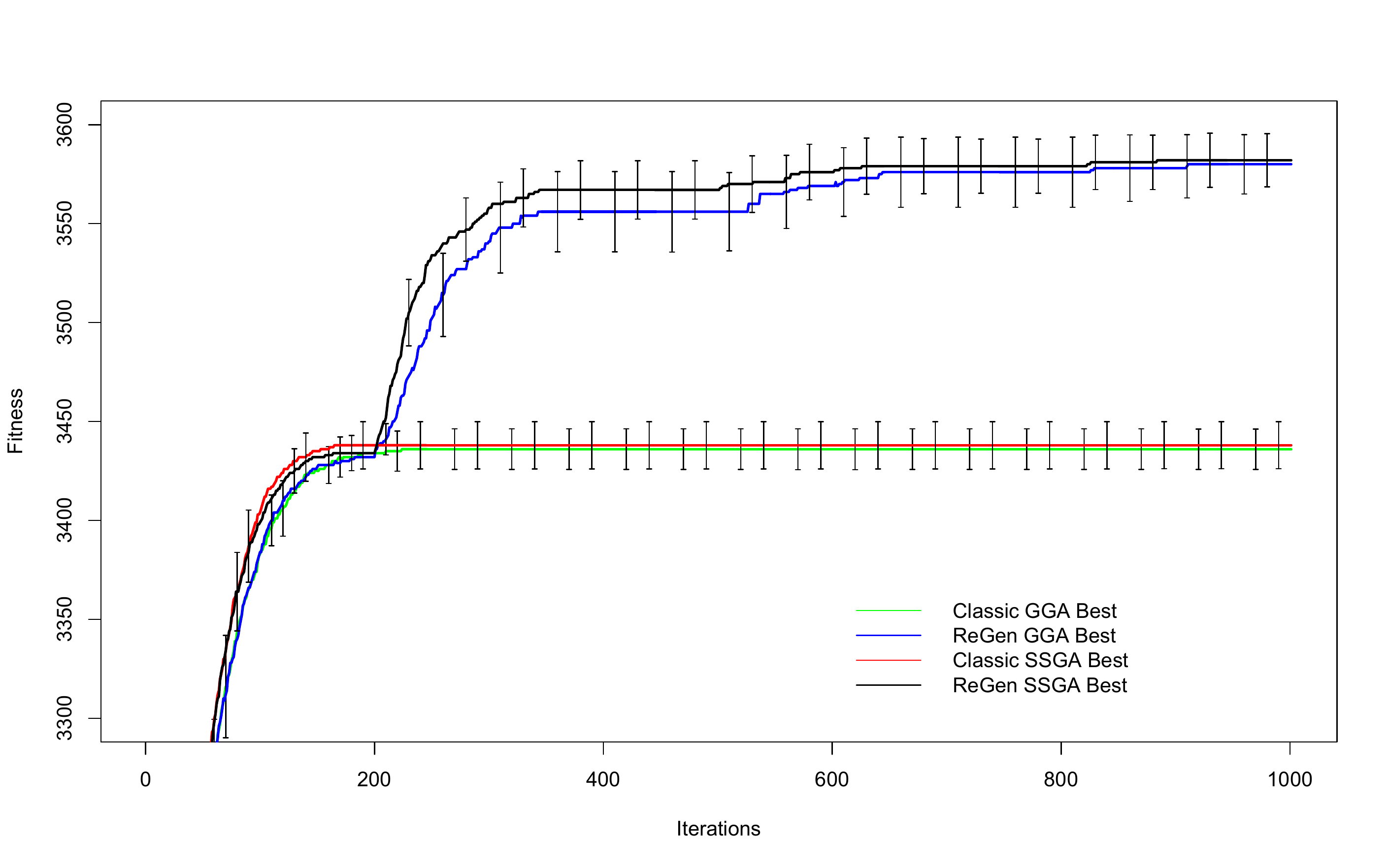}
\includegraphics[width=1.9in]{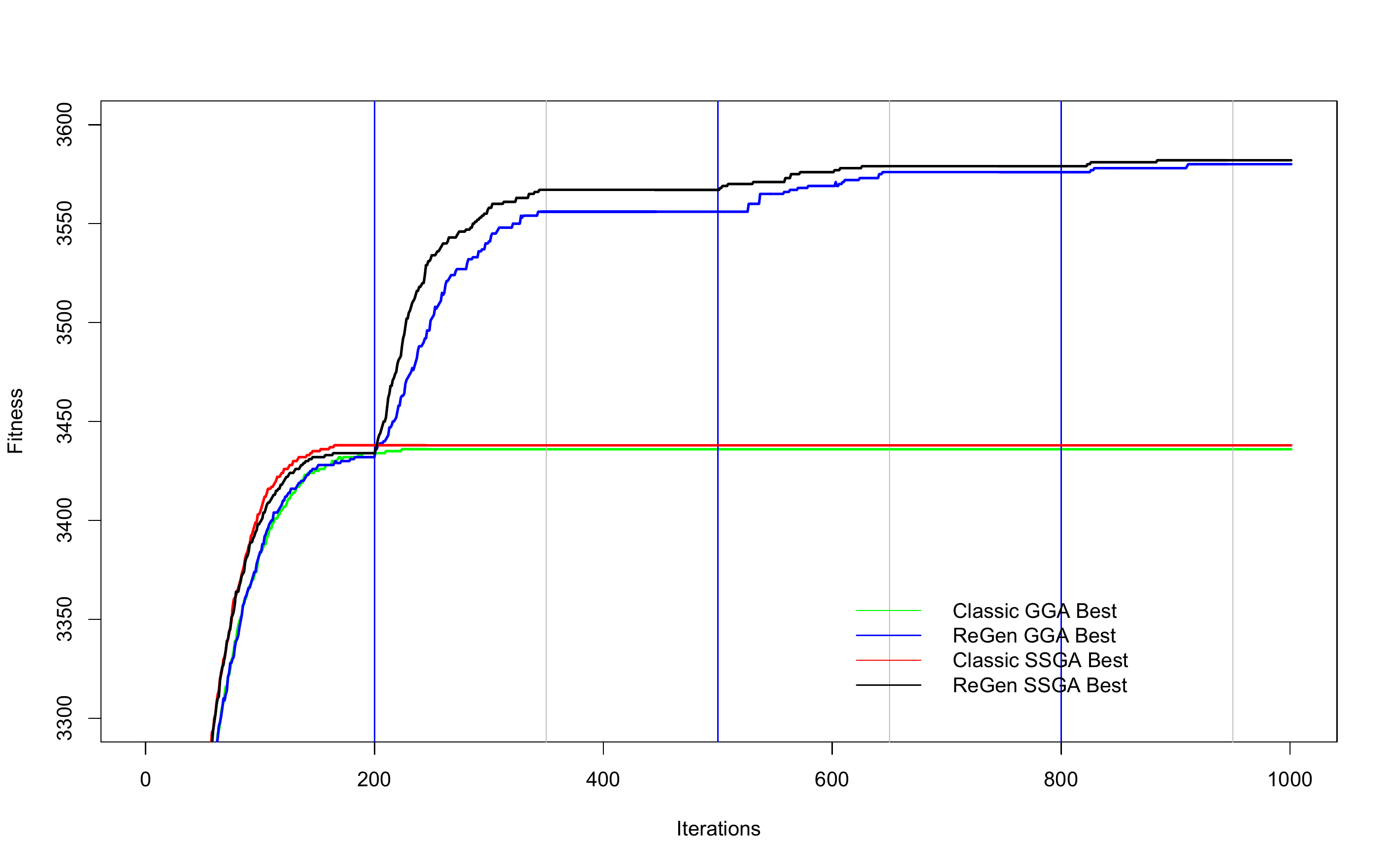}
\includegraphics[width=1.9in]{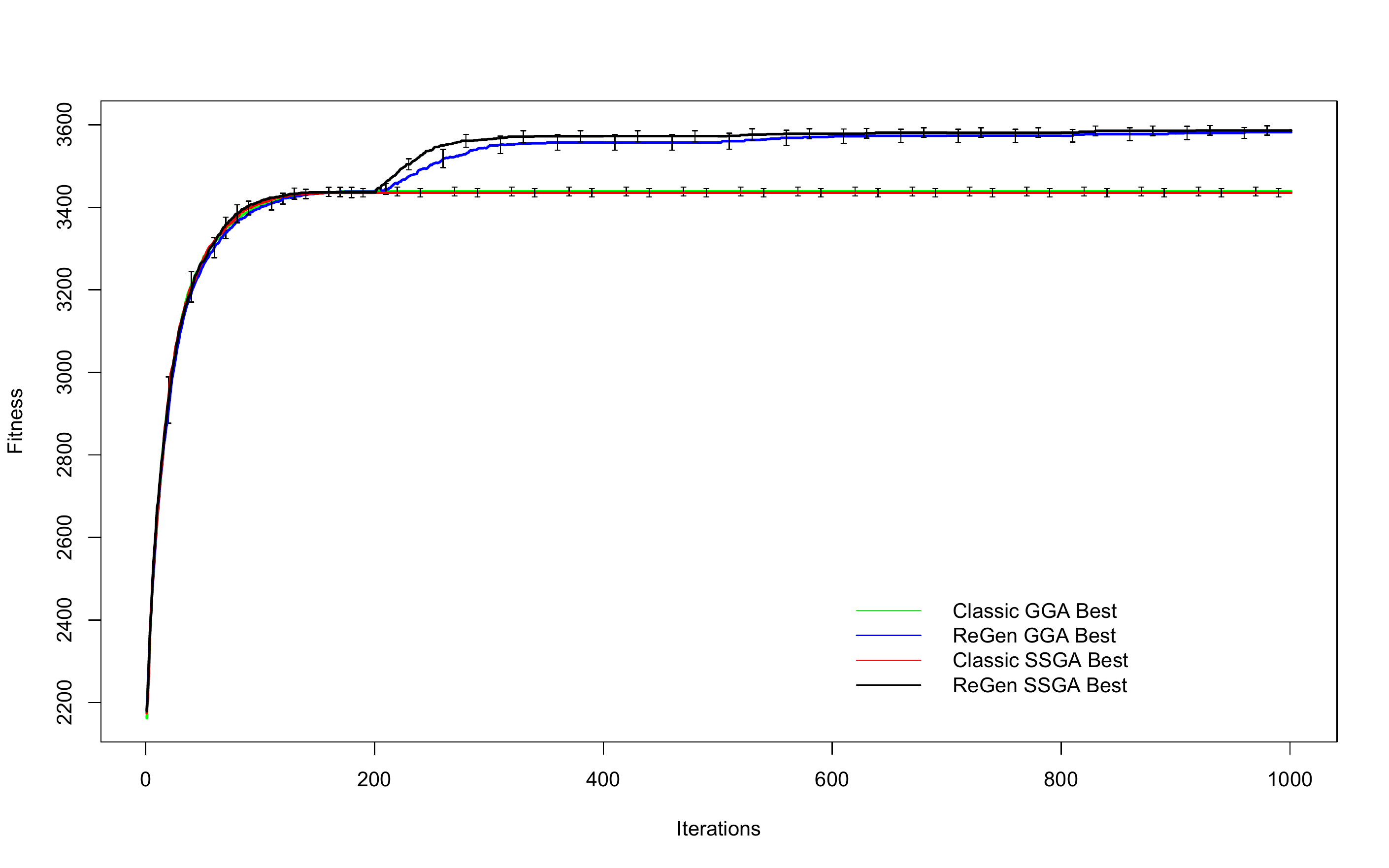}
\includegraphics[width=1.9in]{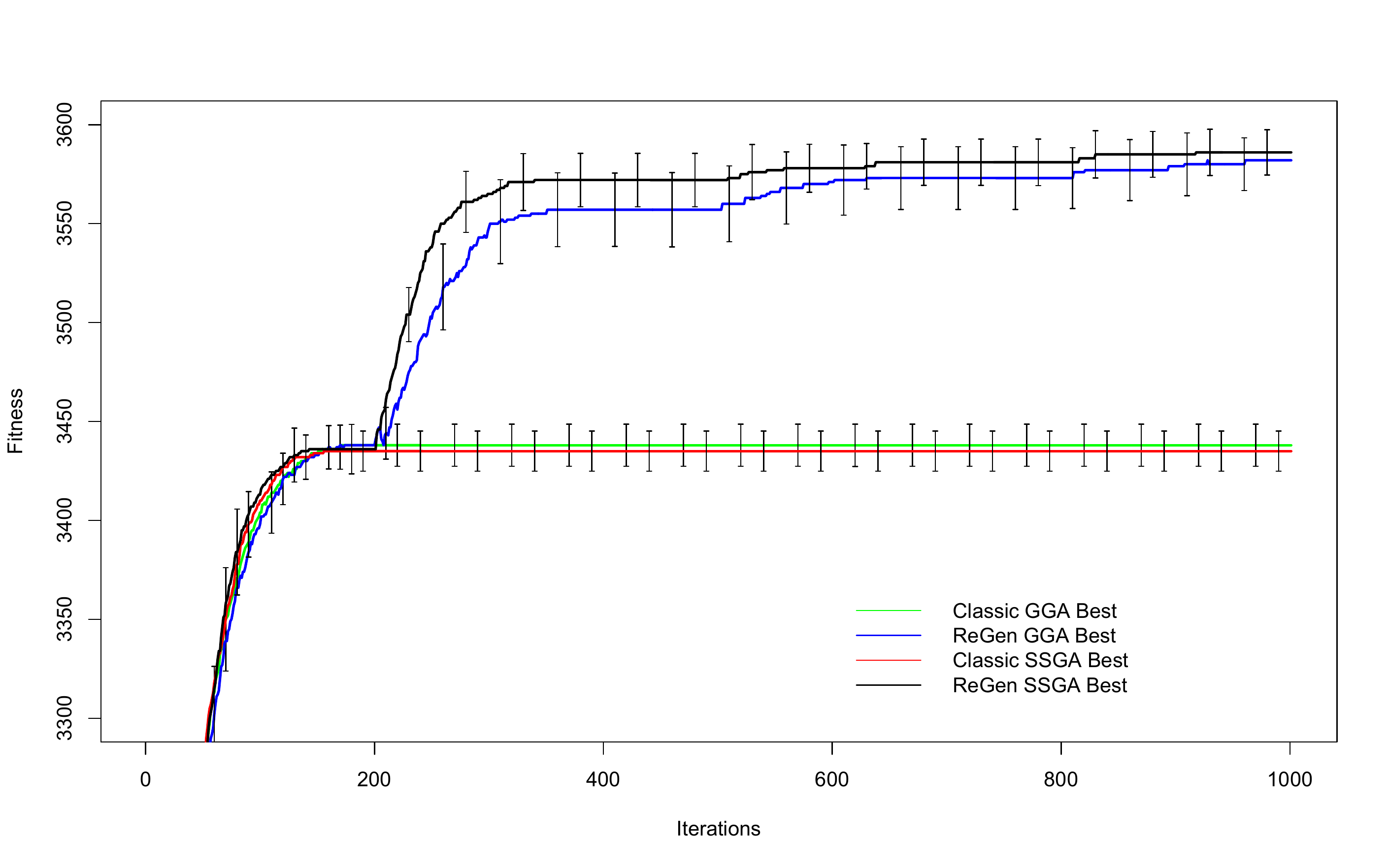}
\includegraphics[width=1.9in]{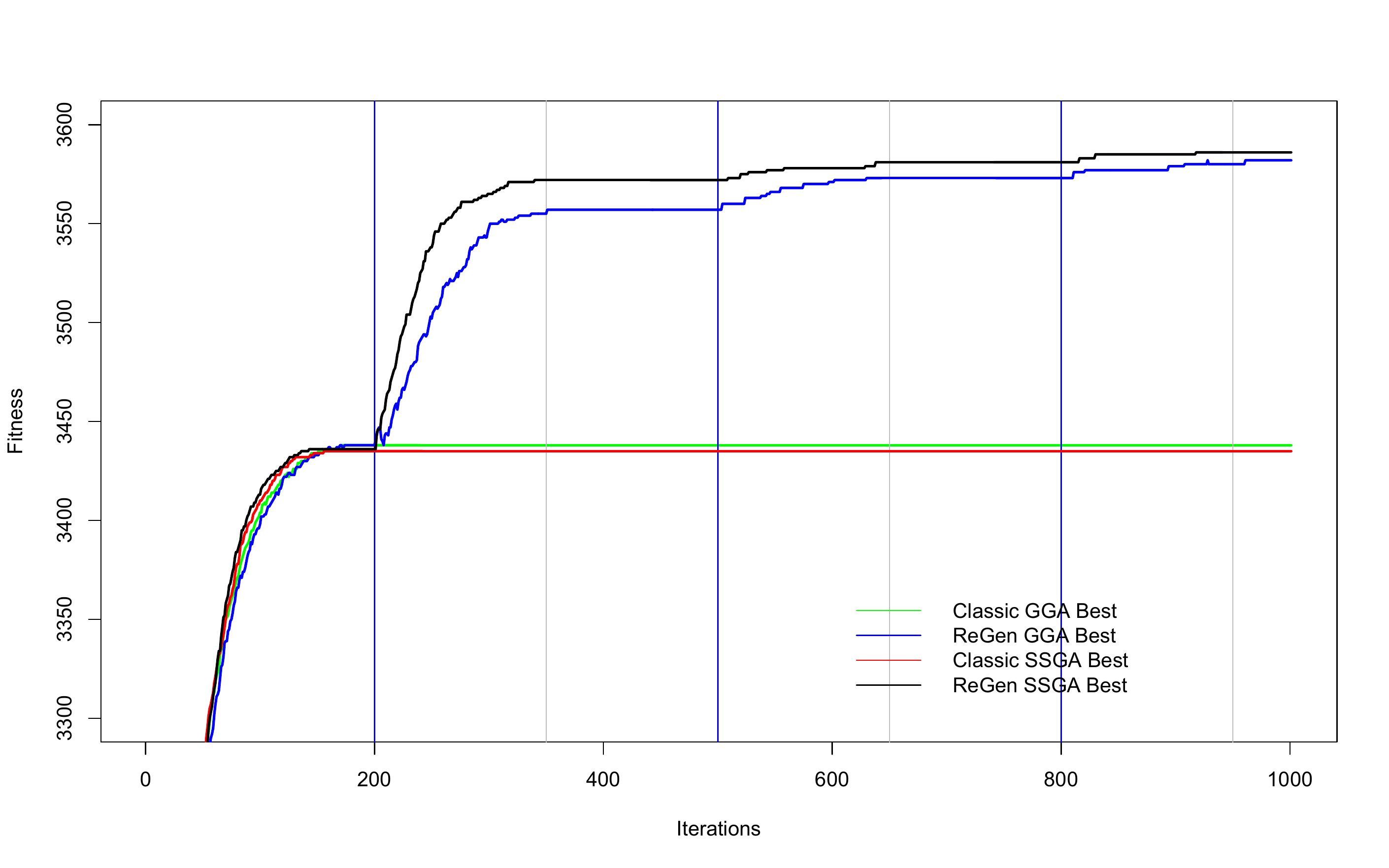}
\includegraphics[width=1.9in]{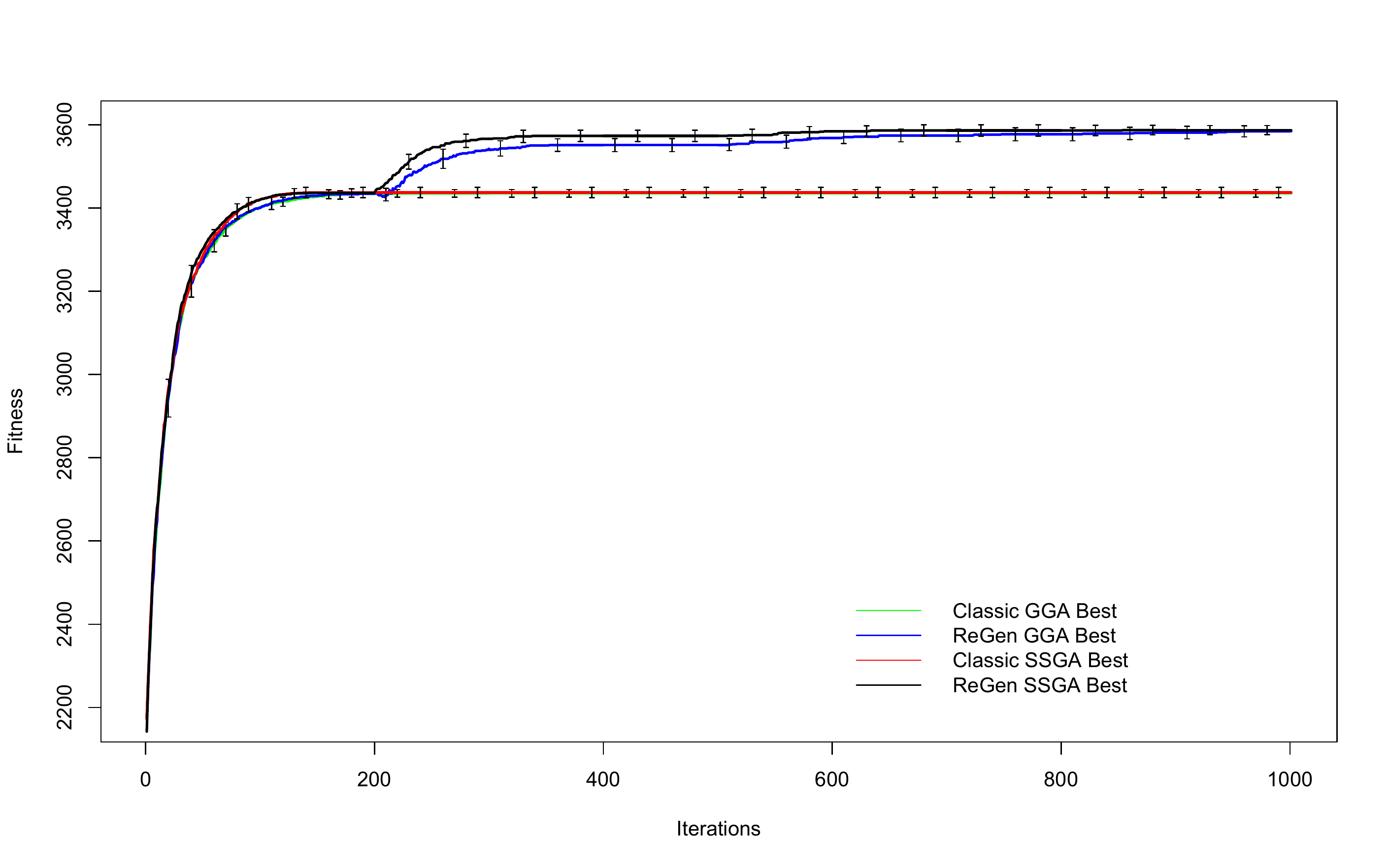}
\includegraphics[width=1.9in]{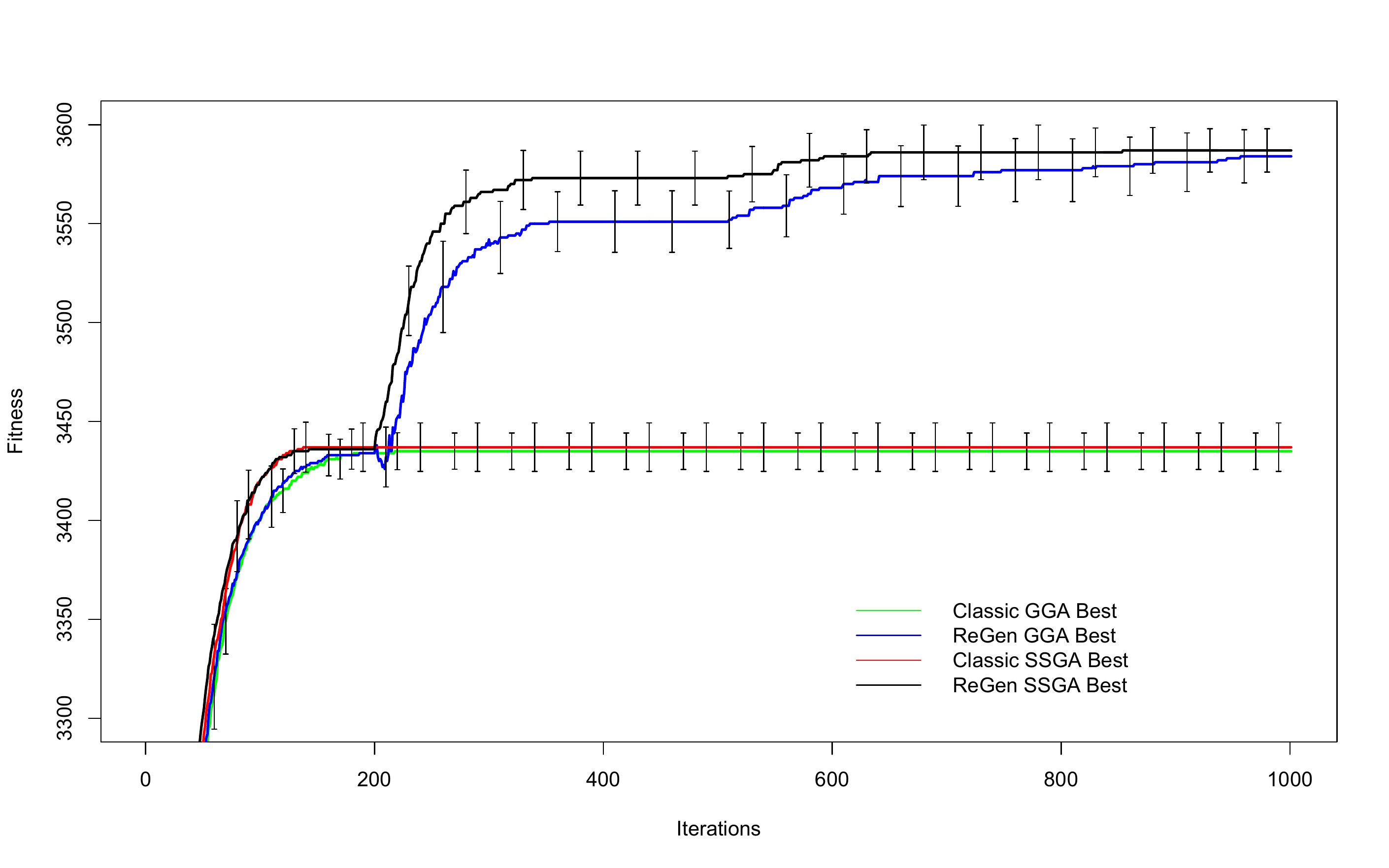}
\includegraphics[width=1.9in]{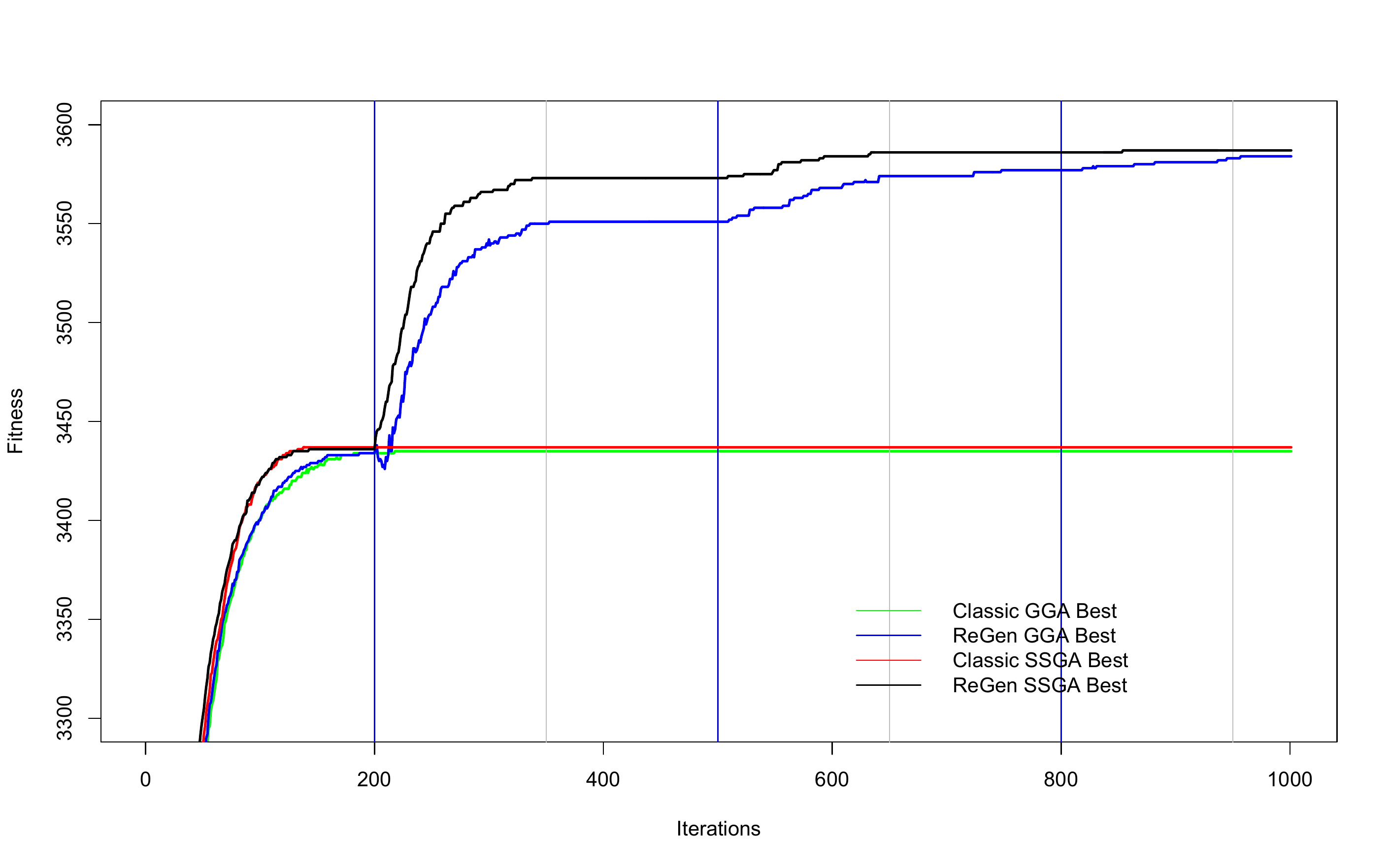}
\caption{Deceptive Order $3$. Generational replacement (GGA) and Steady State replacement (SSGA). From top to bottom, crossover rates from $0.6$ to $1.0$.}
\label{c4fig1}
\end{figure}

\begin{figure}[H]
\centering
\includegraphics[width=1.9in]{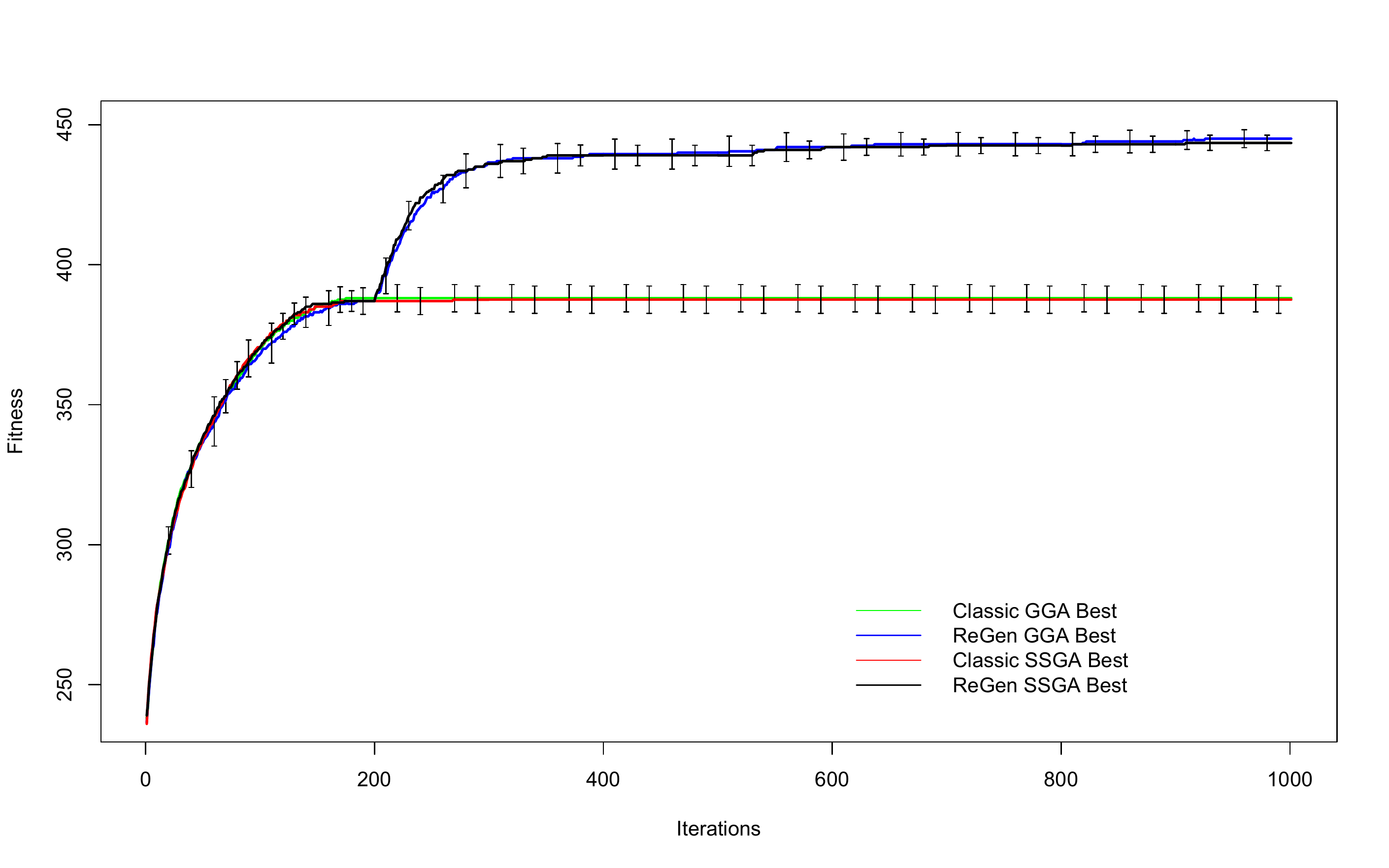}
\includegraphics[width=1.9in]{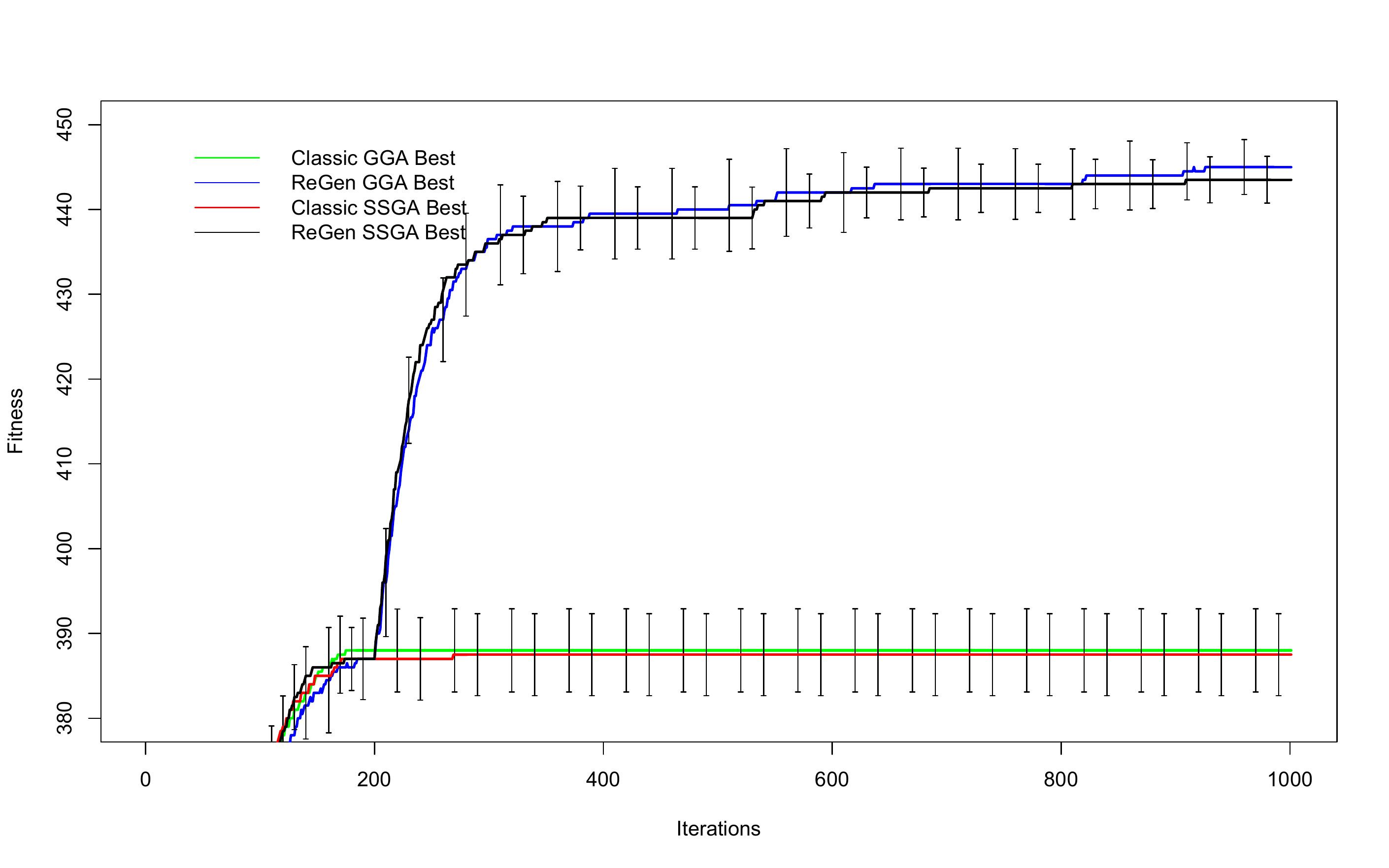}
\includegraphics[width=1.9in]{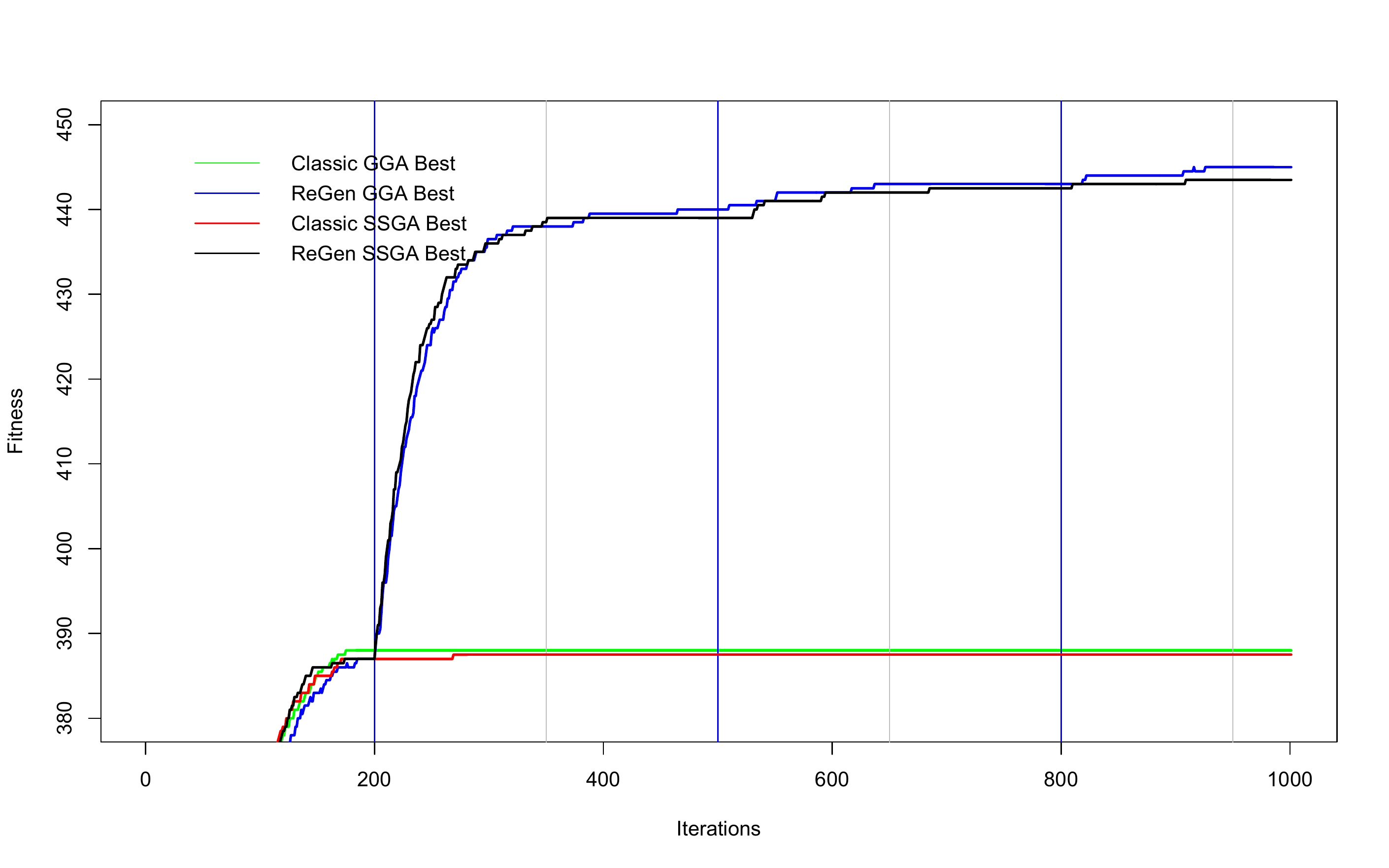}
\includegraphics[width=1.9in]{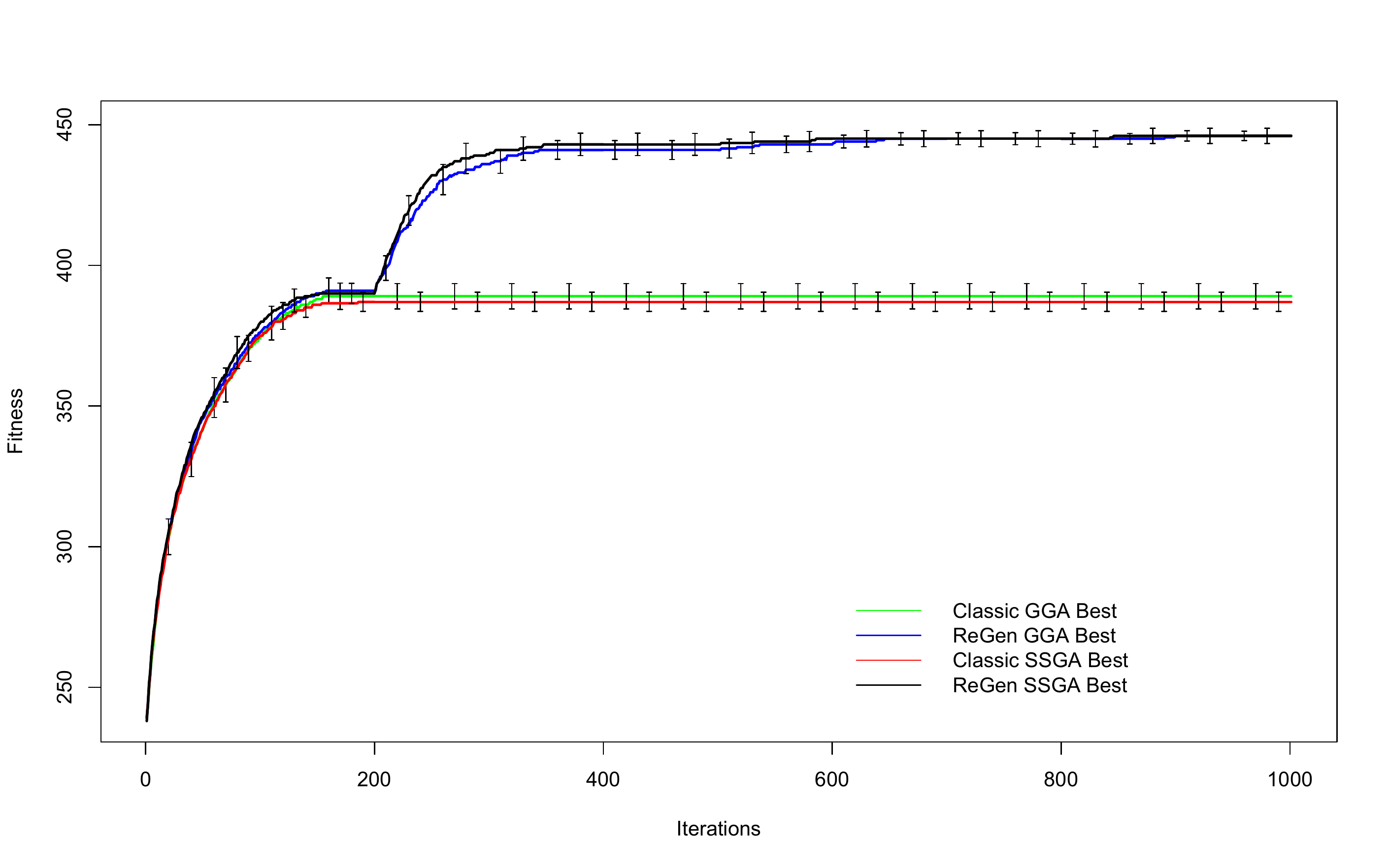}
\includegraphics[width=1.9in]{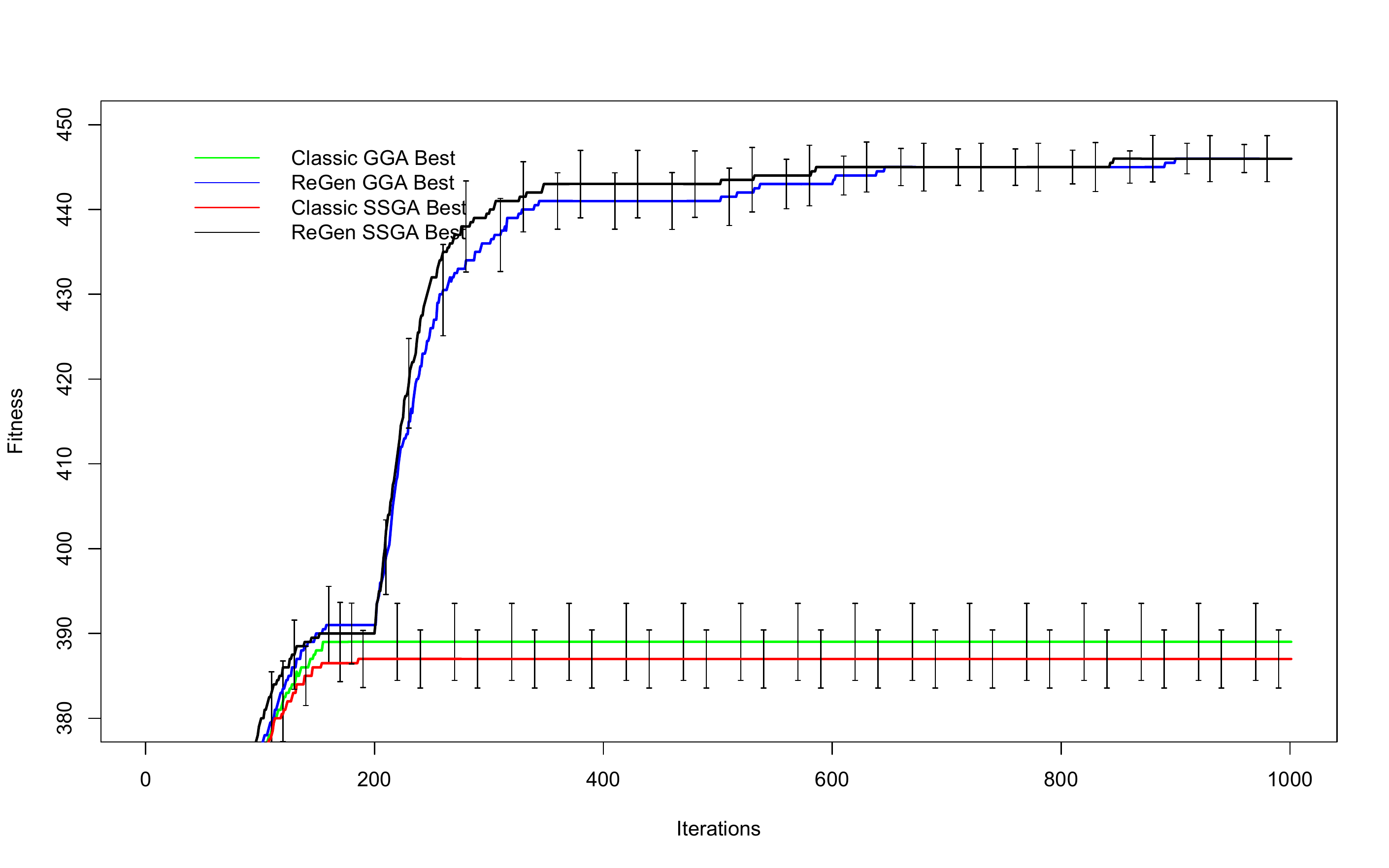}
\includegraphics[width=1.9in]{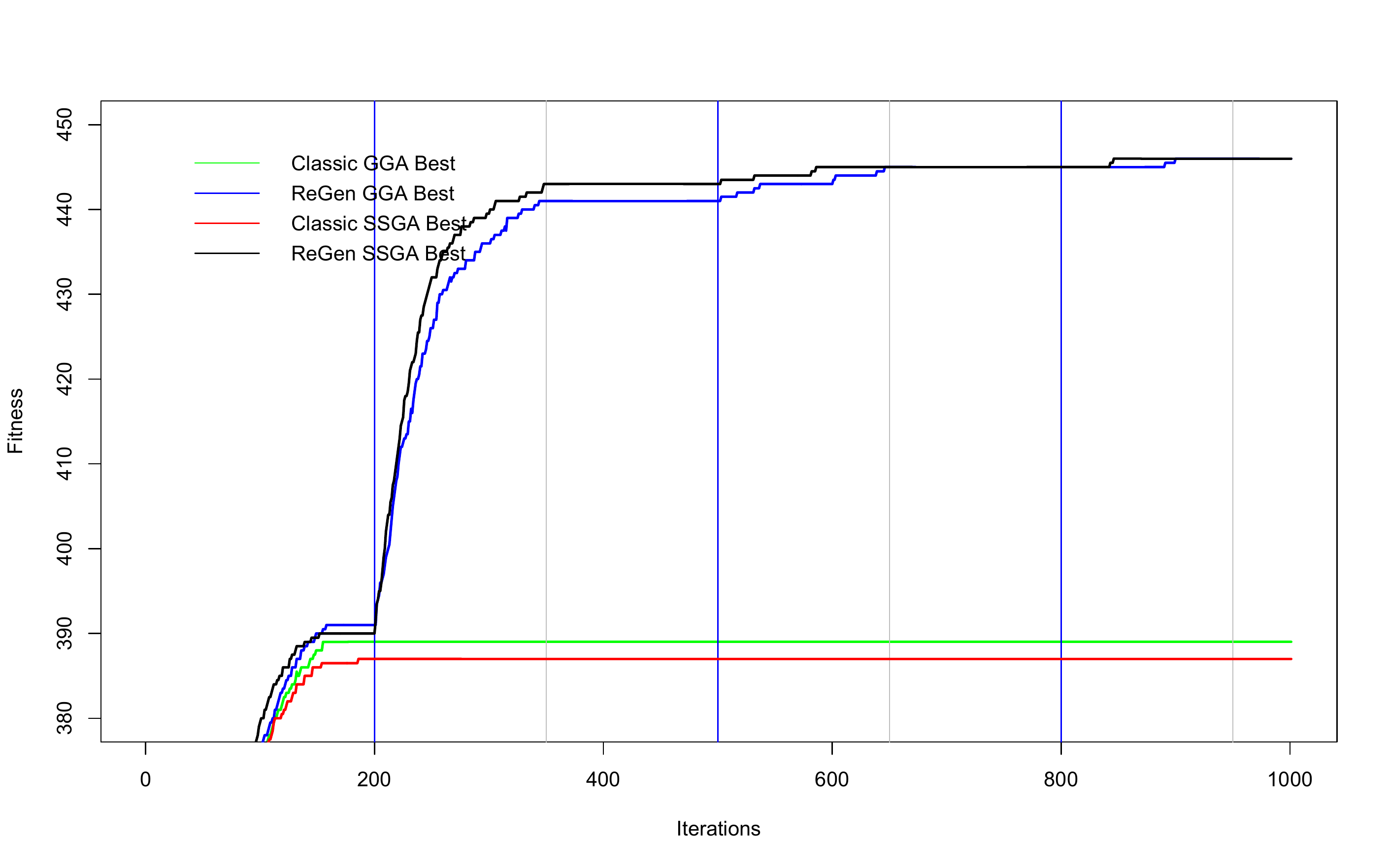}
\includegraphics[width=1.9in]{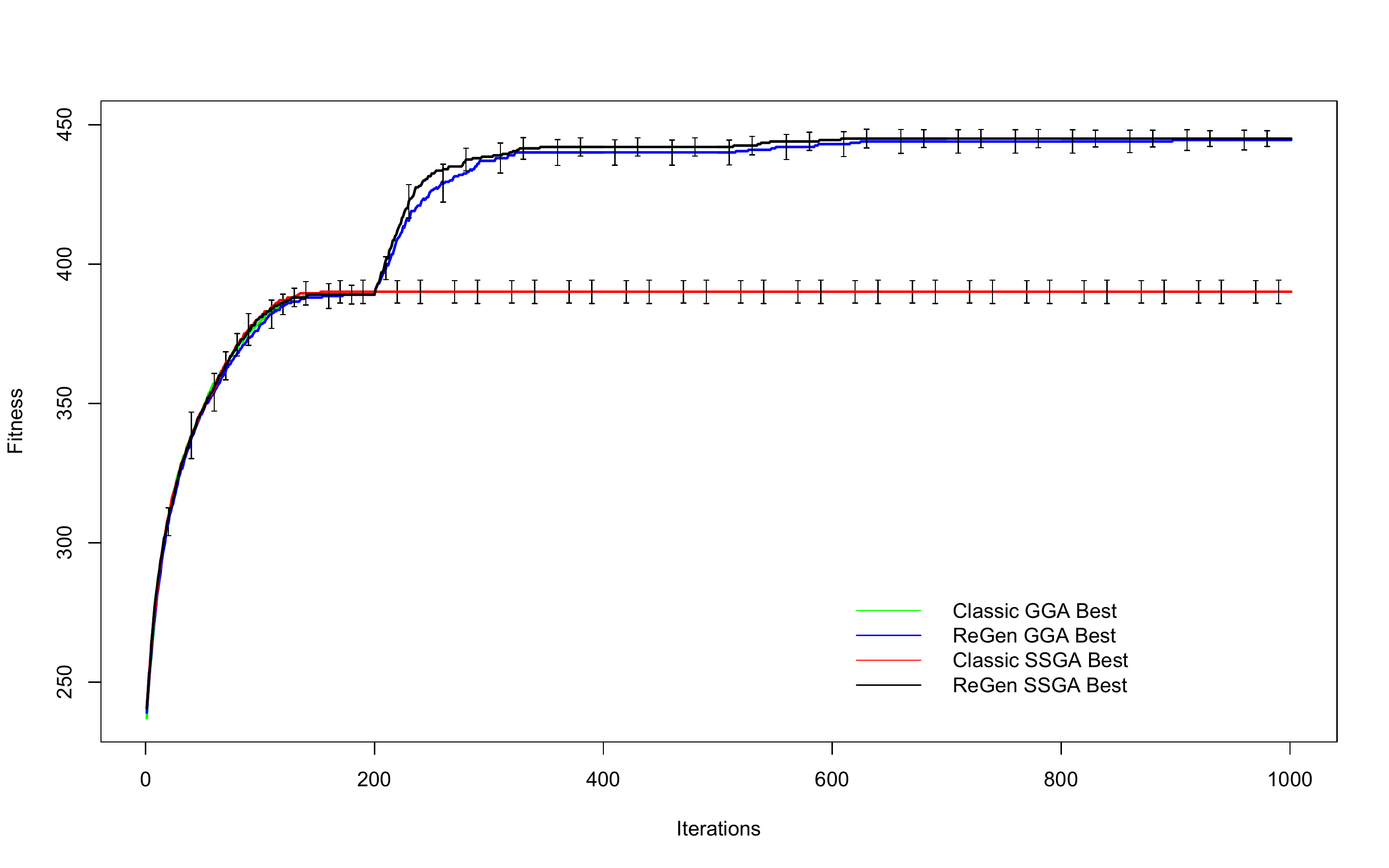}
\includegraphics[width=1.9in]{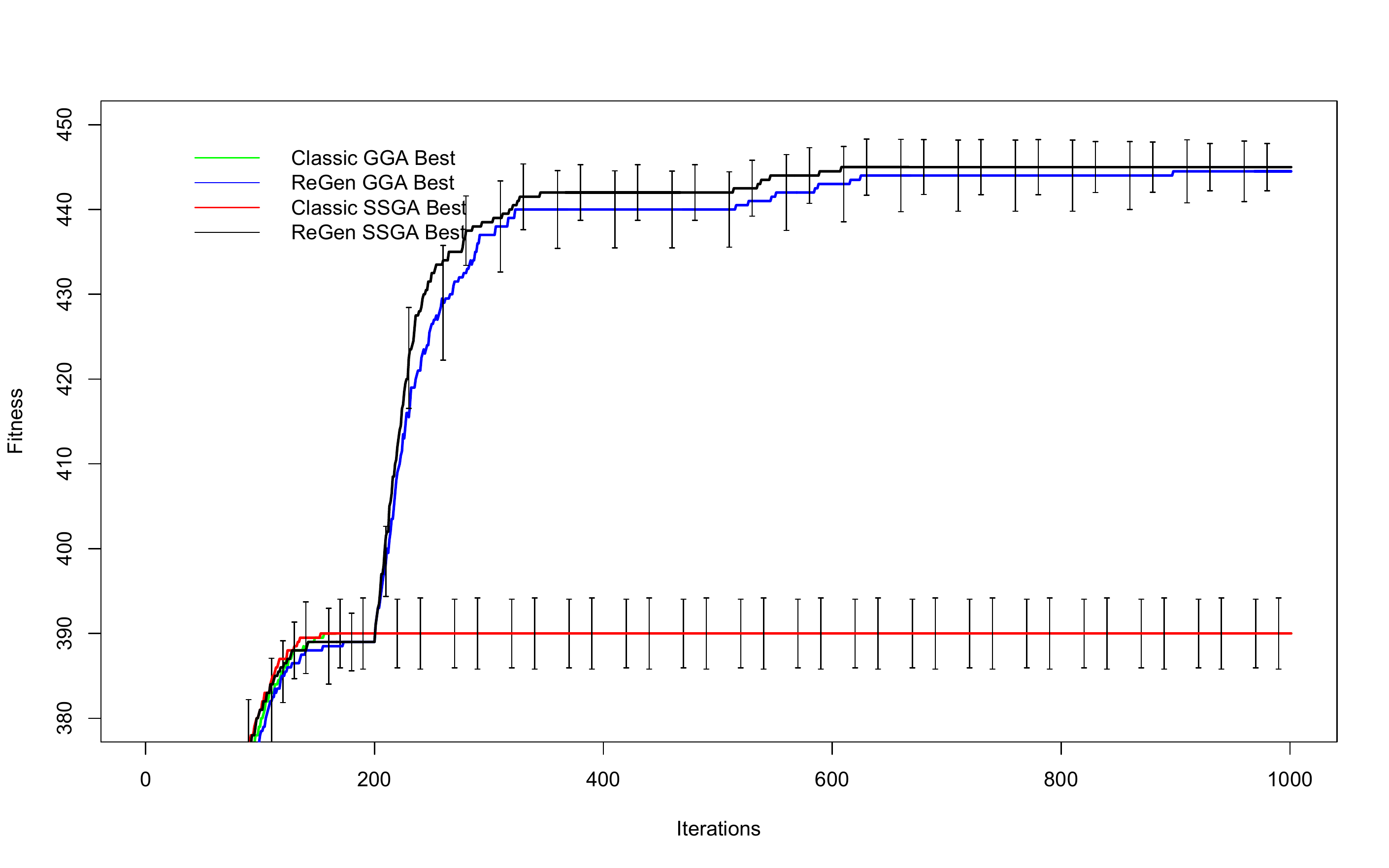}
\includegraphics[width=1.9in]{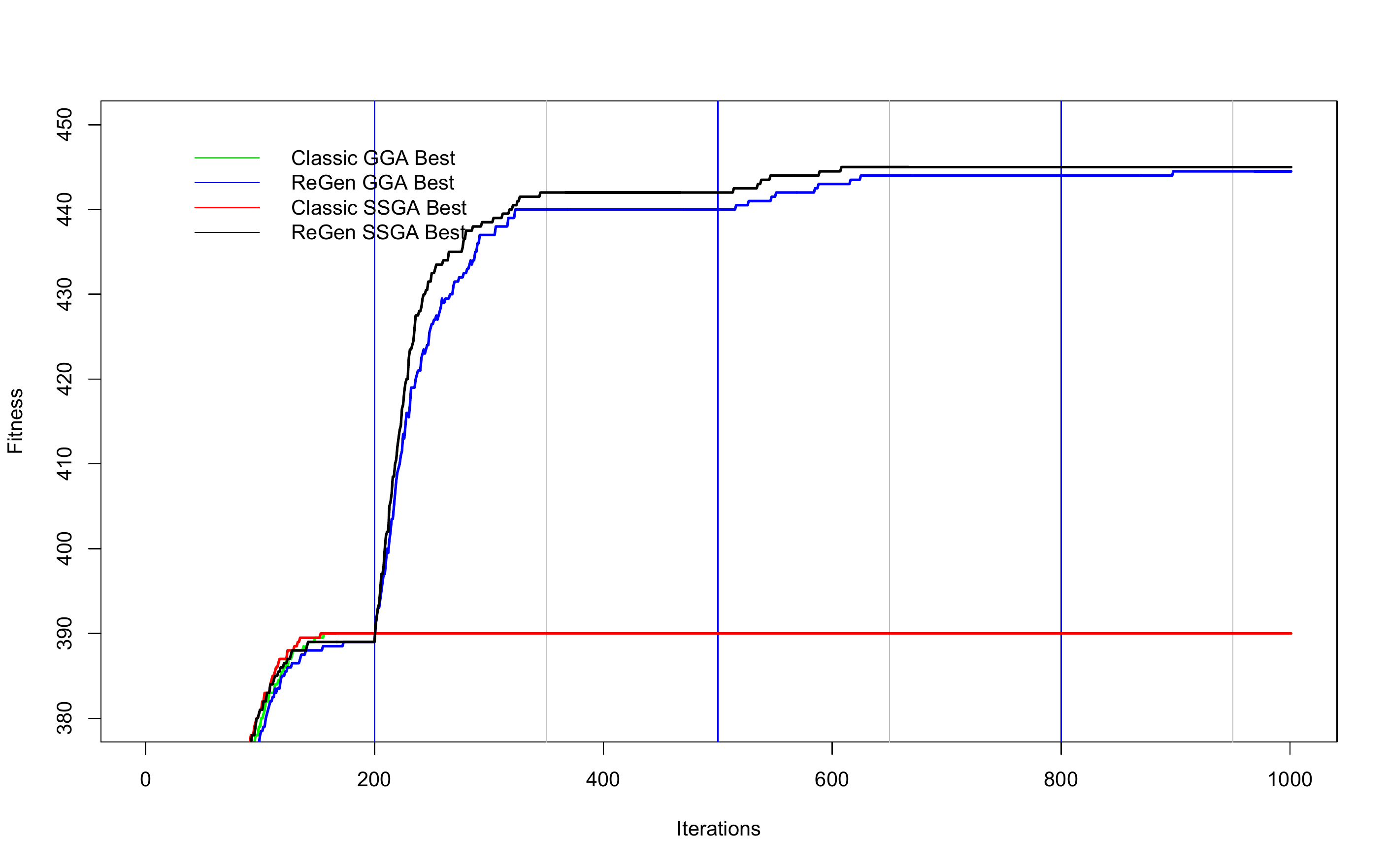}
\includegraphics[width=1.9in]{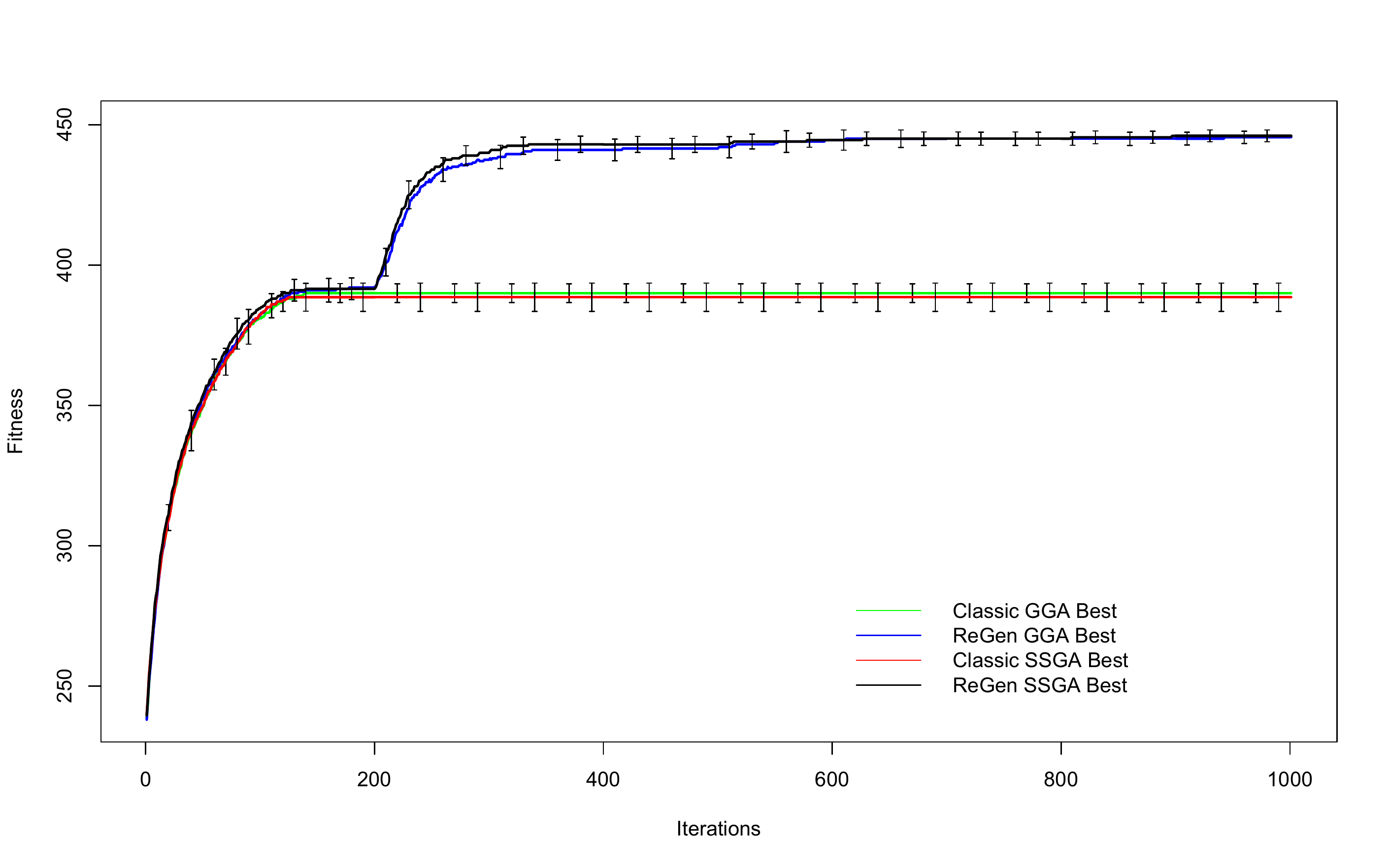}
\includegraphics[width=1.9in]{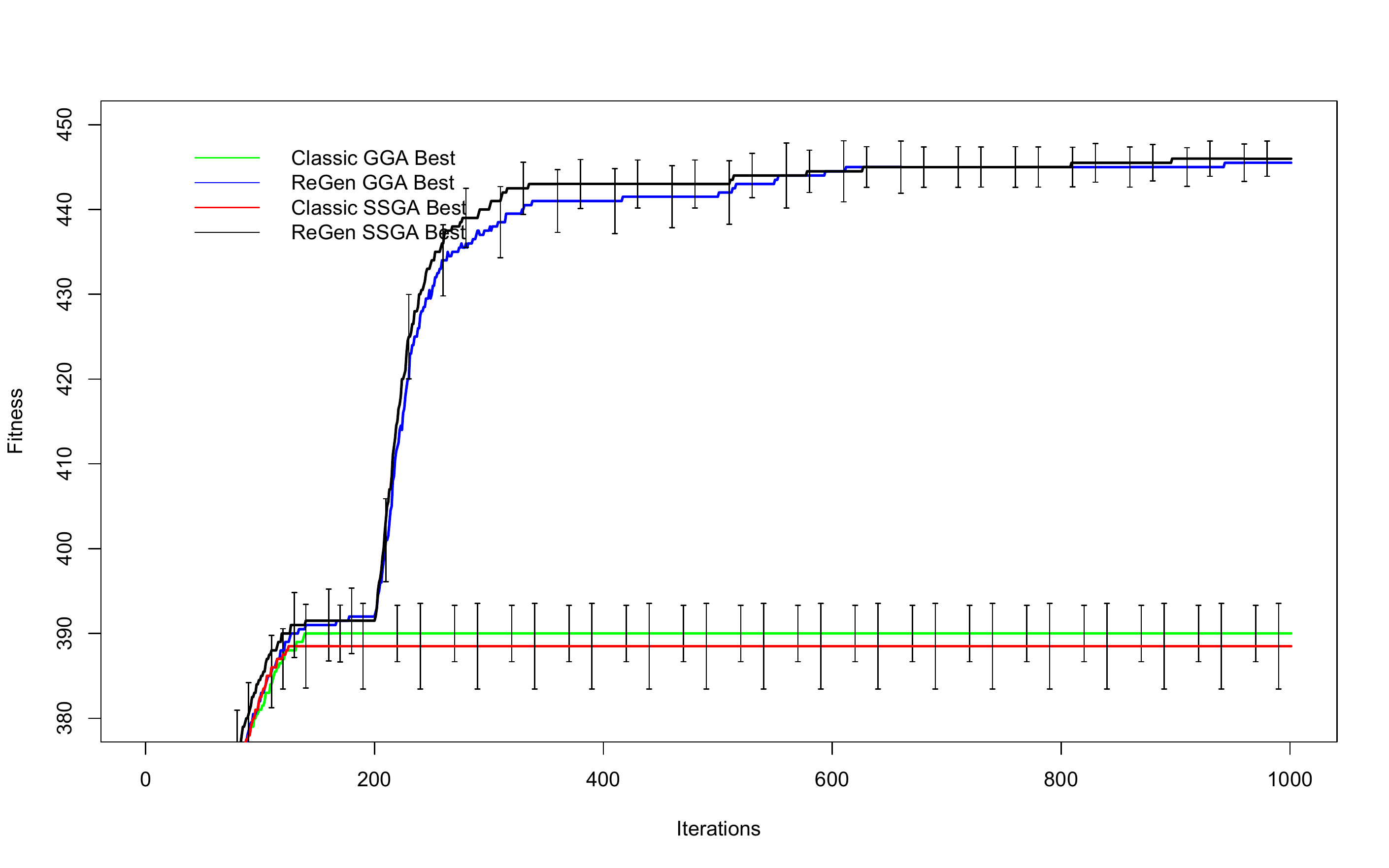}
\includegraphics[width=1.9in]{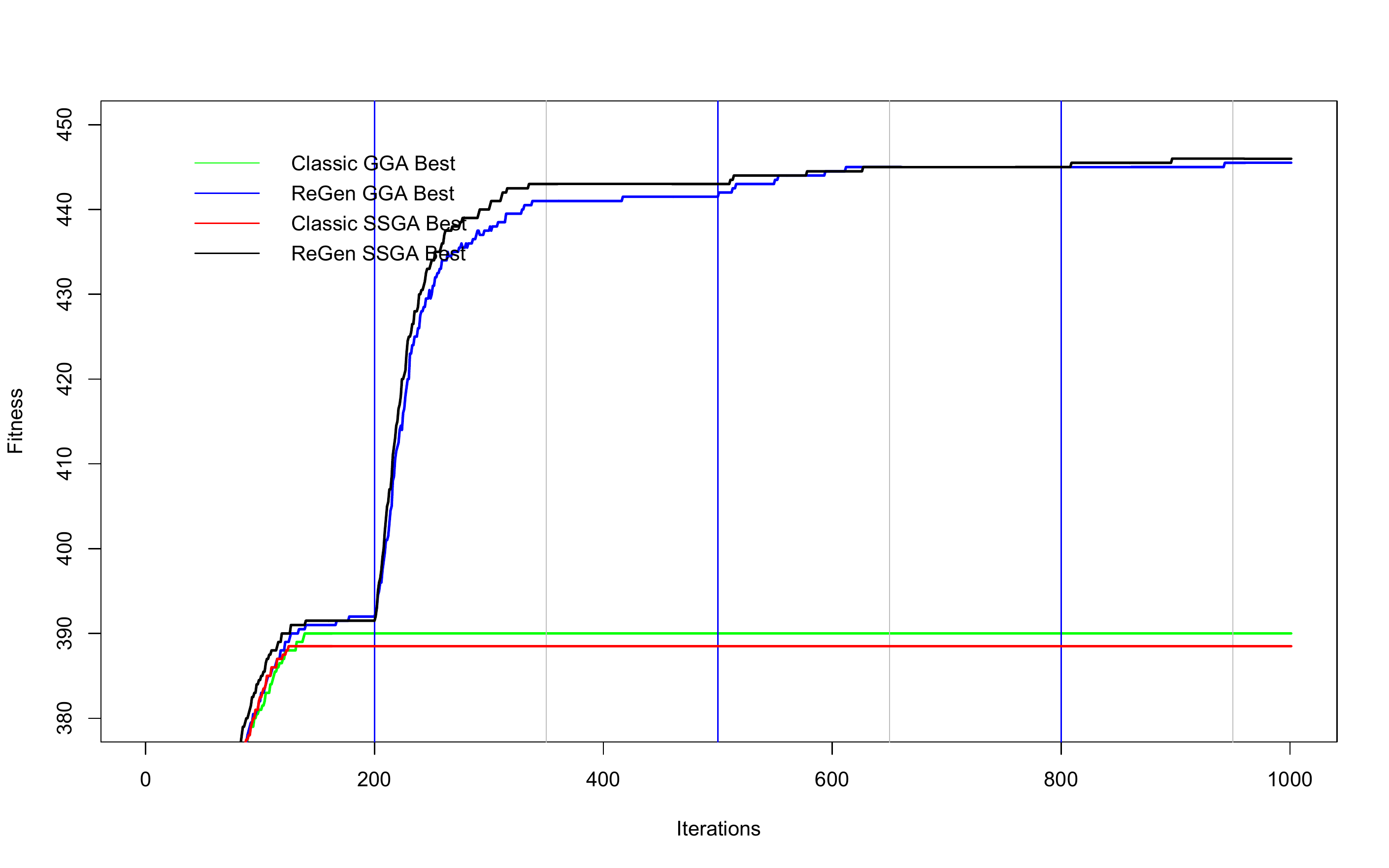}
\includegraphics[width=1.9in]{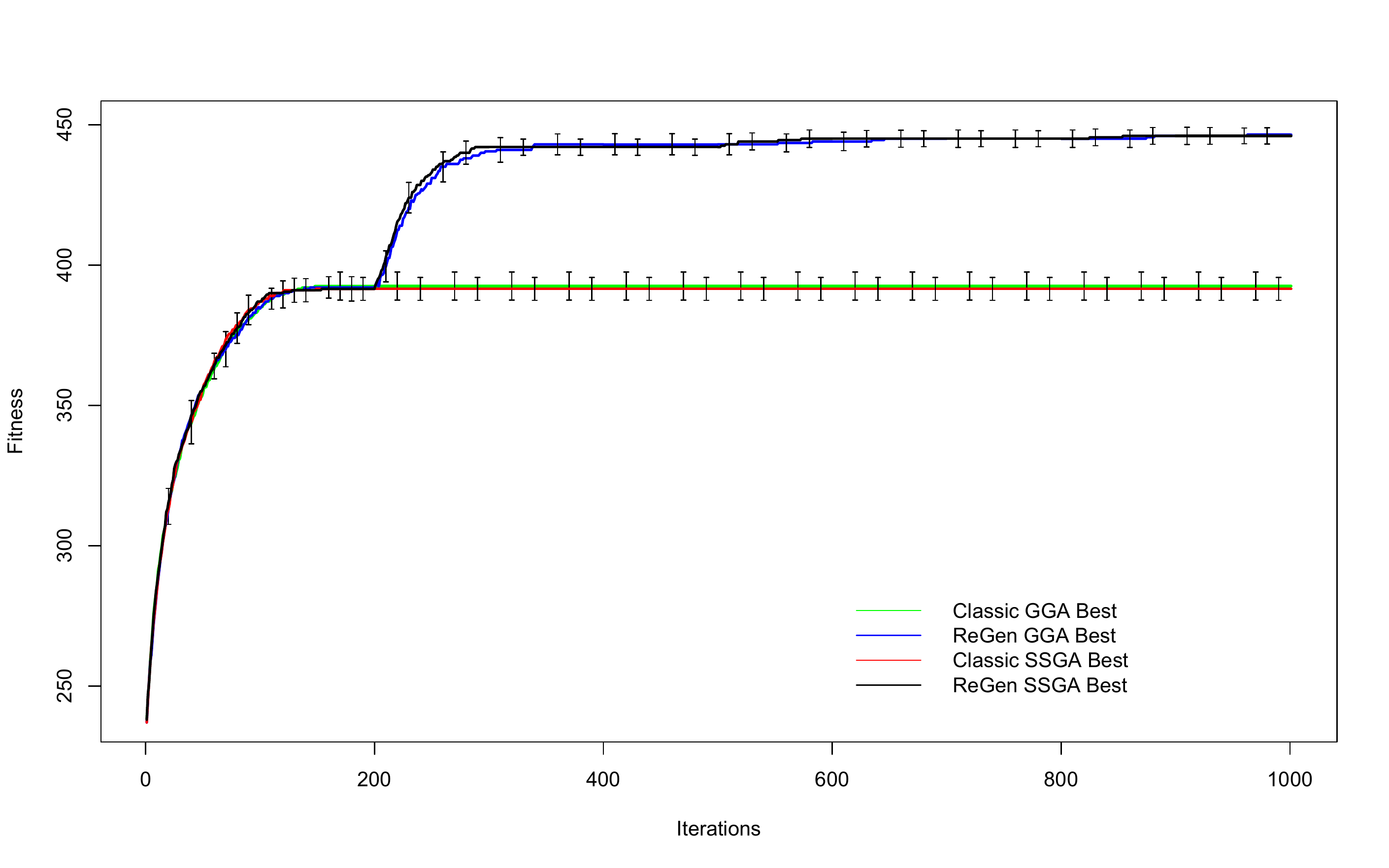}
\includegraphics[width=1.9in]{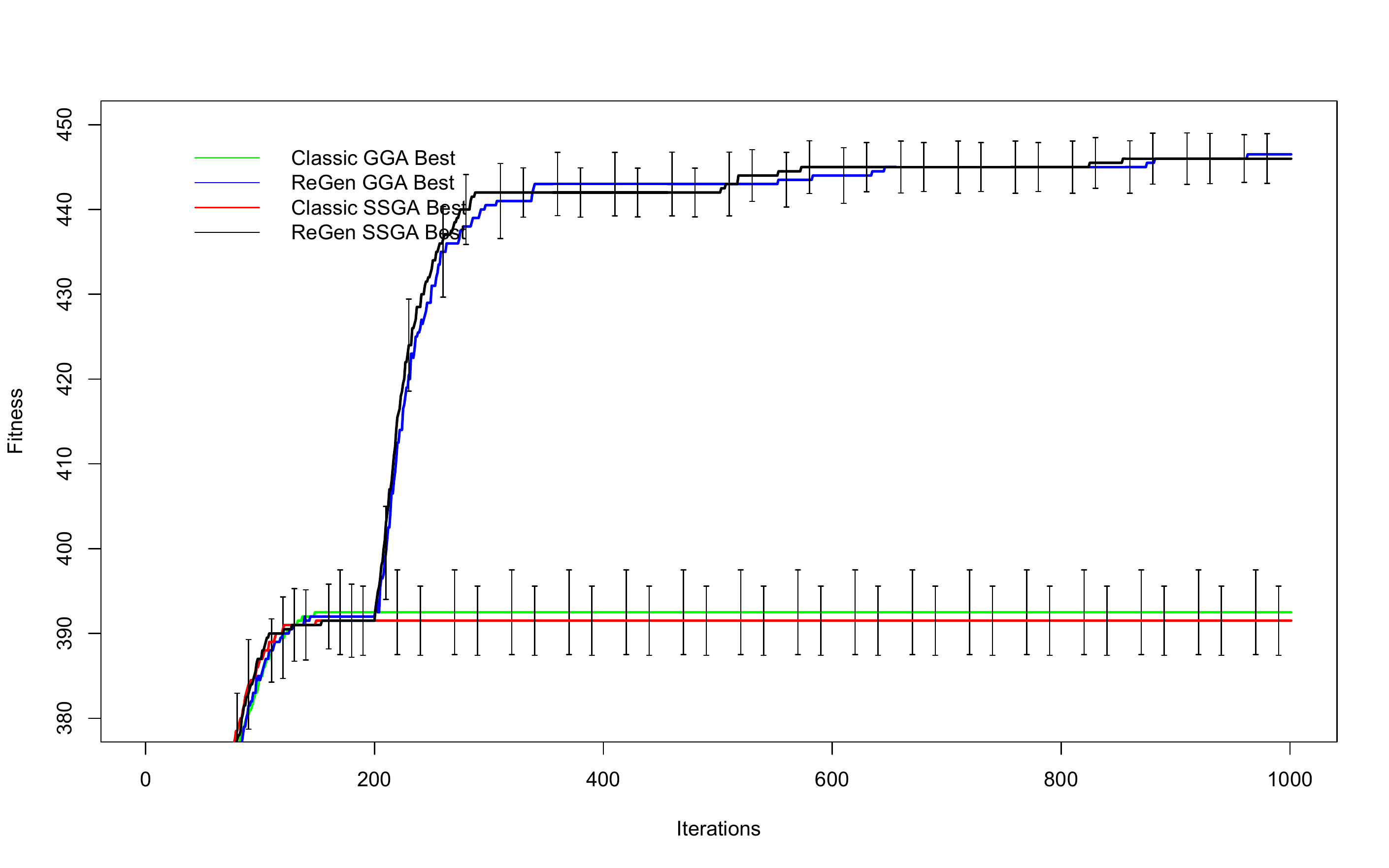}
\includegraphics[width=1.9in]{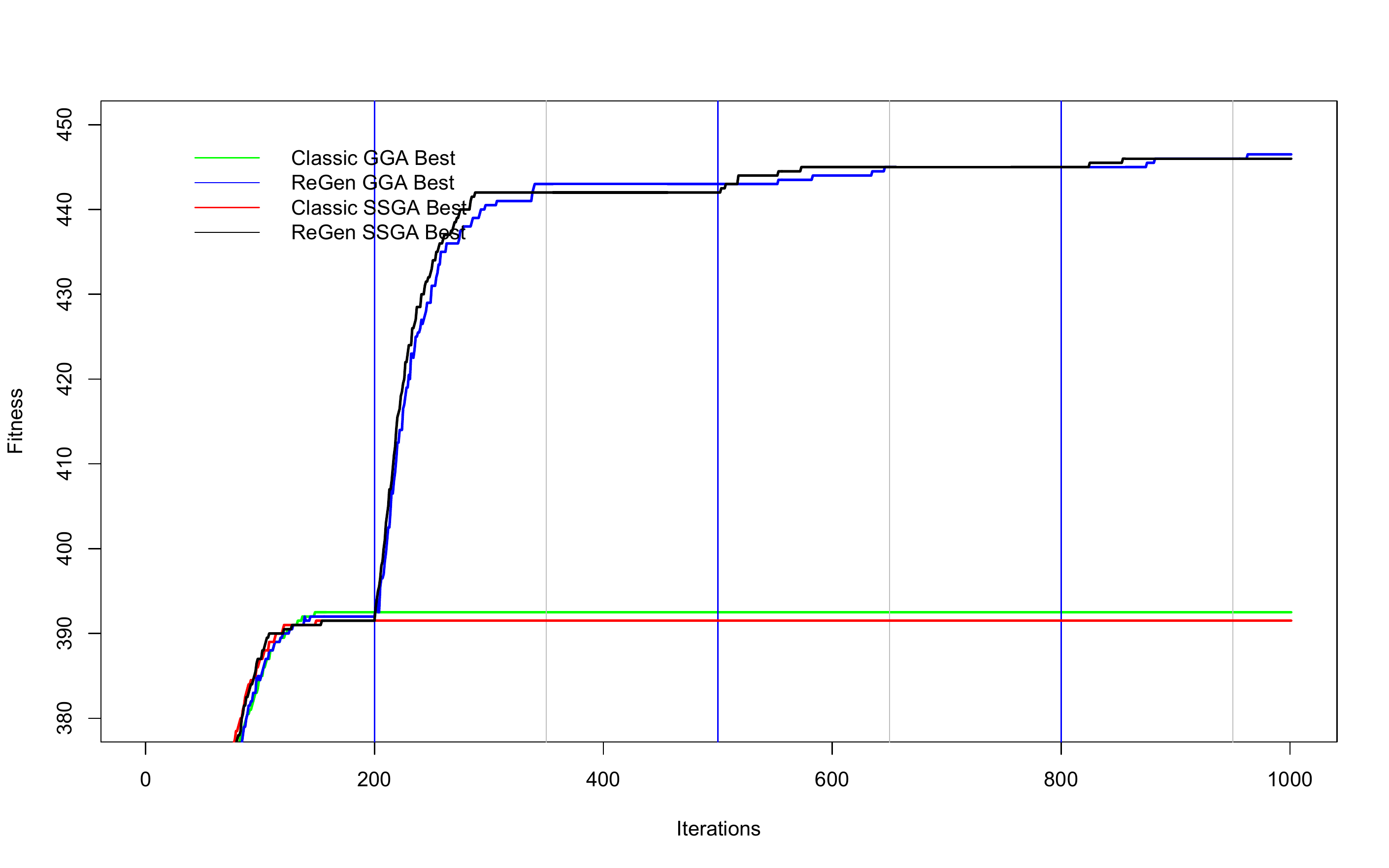}
\caption{Deceptive Order $4$. Generational replacement (GGA) and Steady State replacement (SSGA). From top to bottom, crossover rates from $0.6$ to $1.0$.}
\label{c4fig2}
\end{figure}

\begin{figure}[H]
\centering
\includegraphics[width=1.9in]{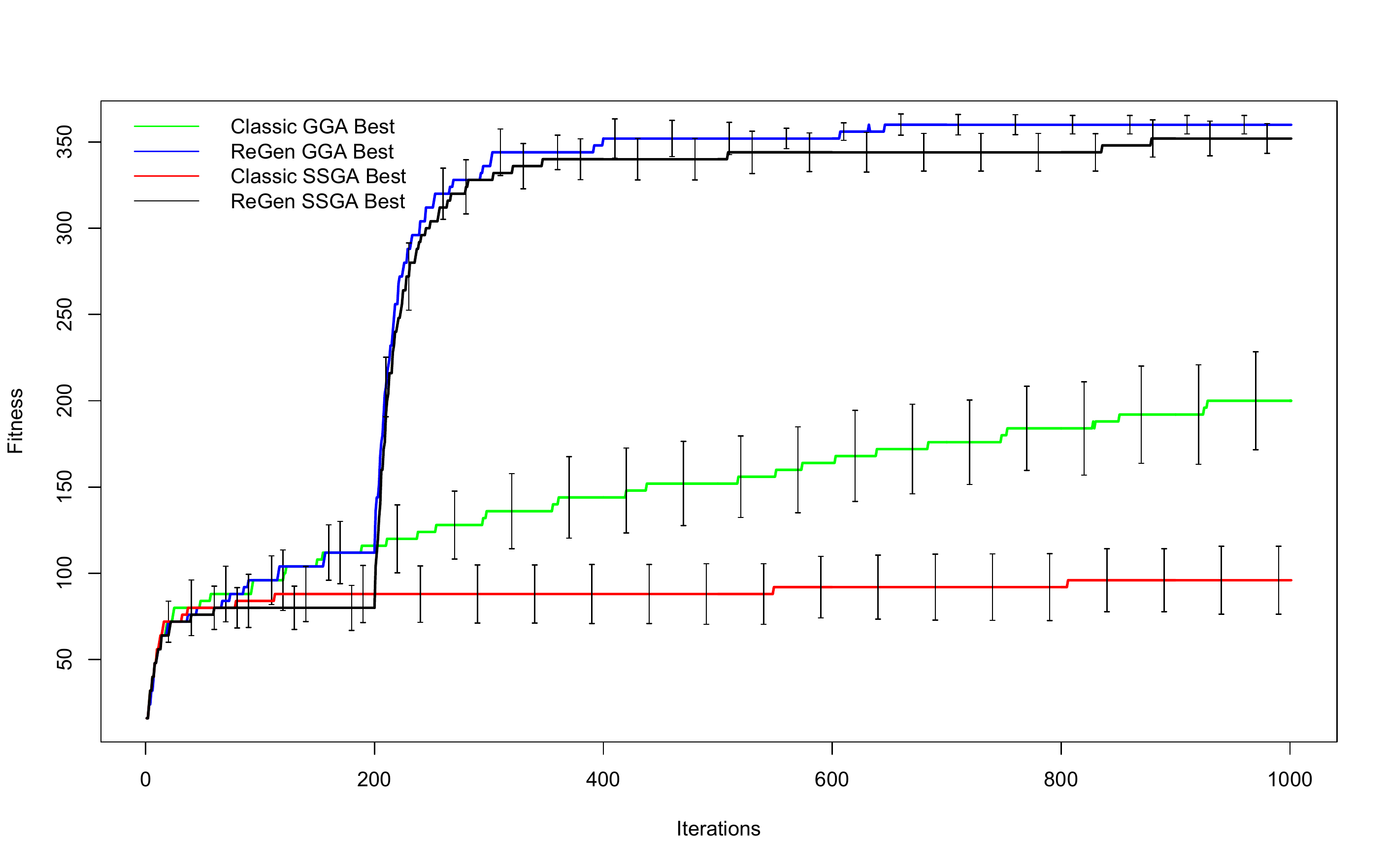}
\includegraphics[width=1.9in]{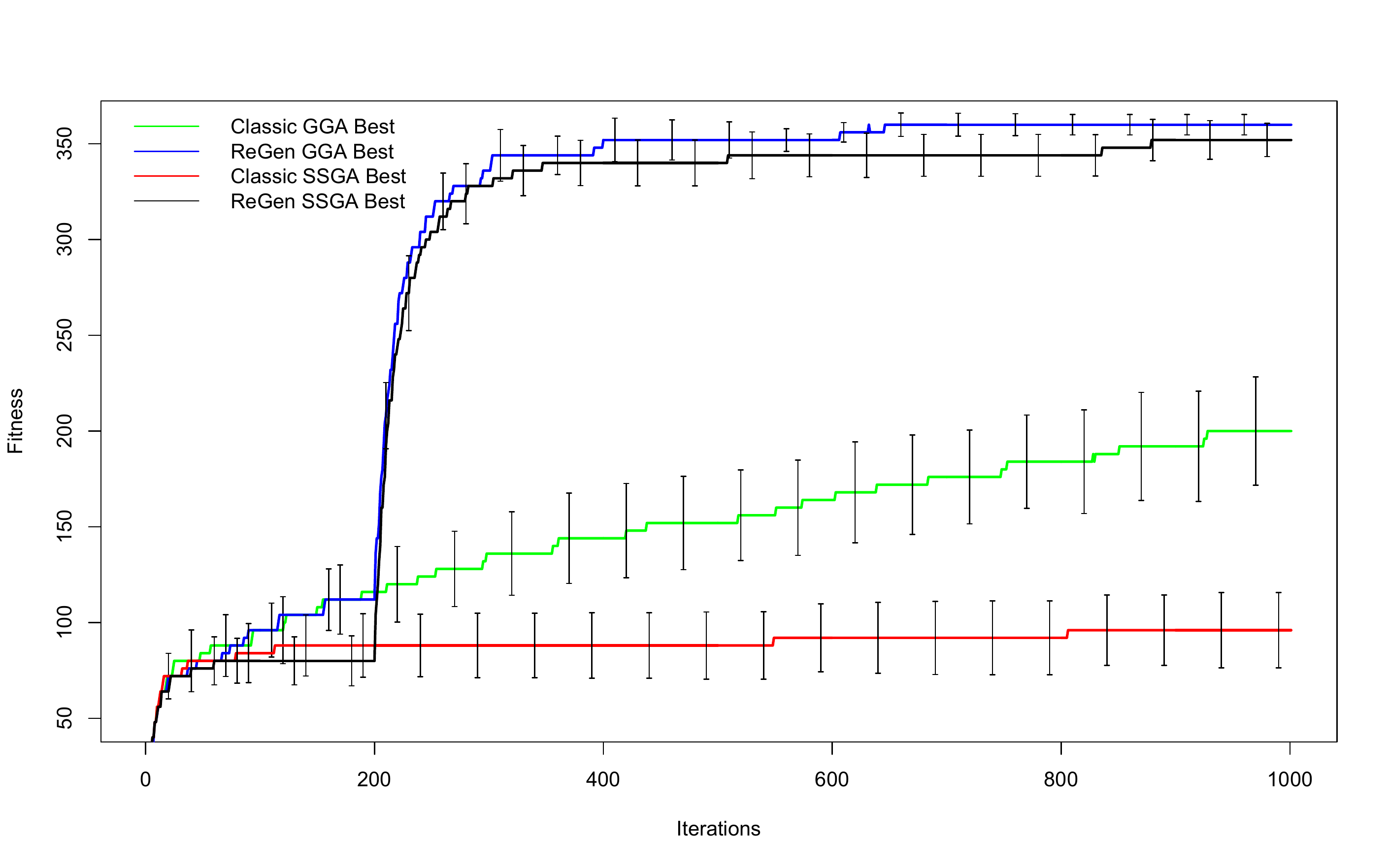}
\includegraphics[width=1.9in]{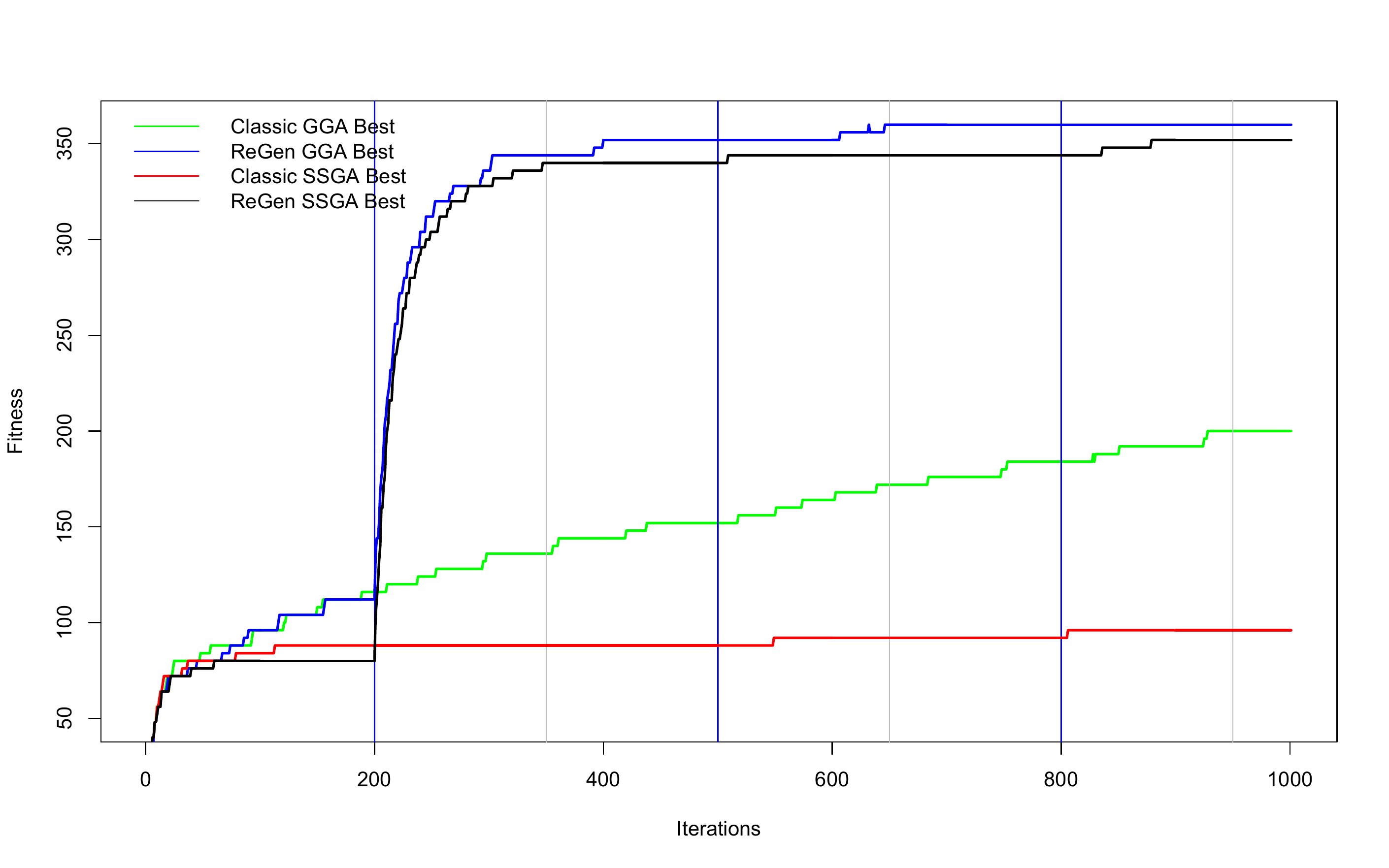}
\includegraphics[width=1.9in]{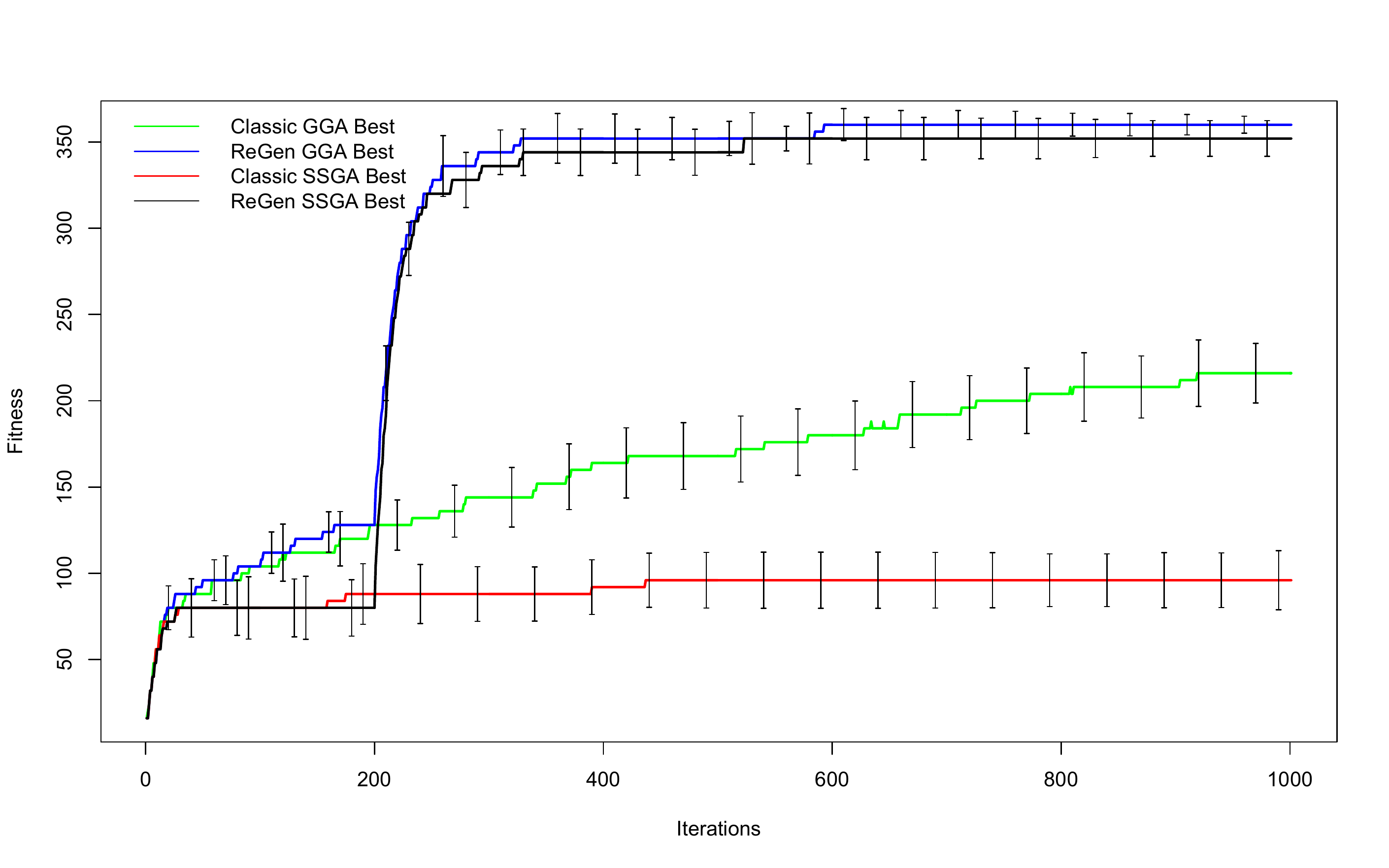}
\includegraphics[width=1.9in]{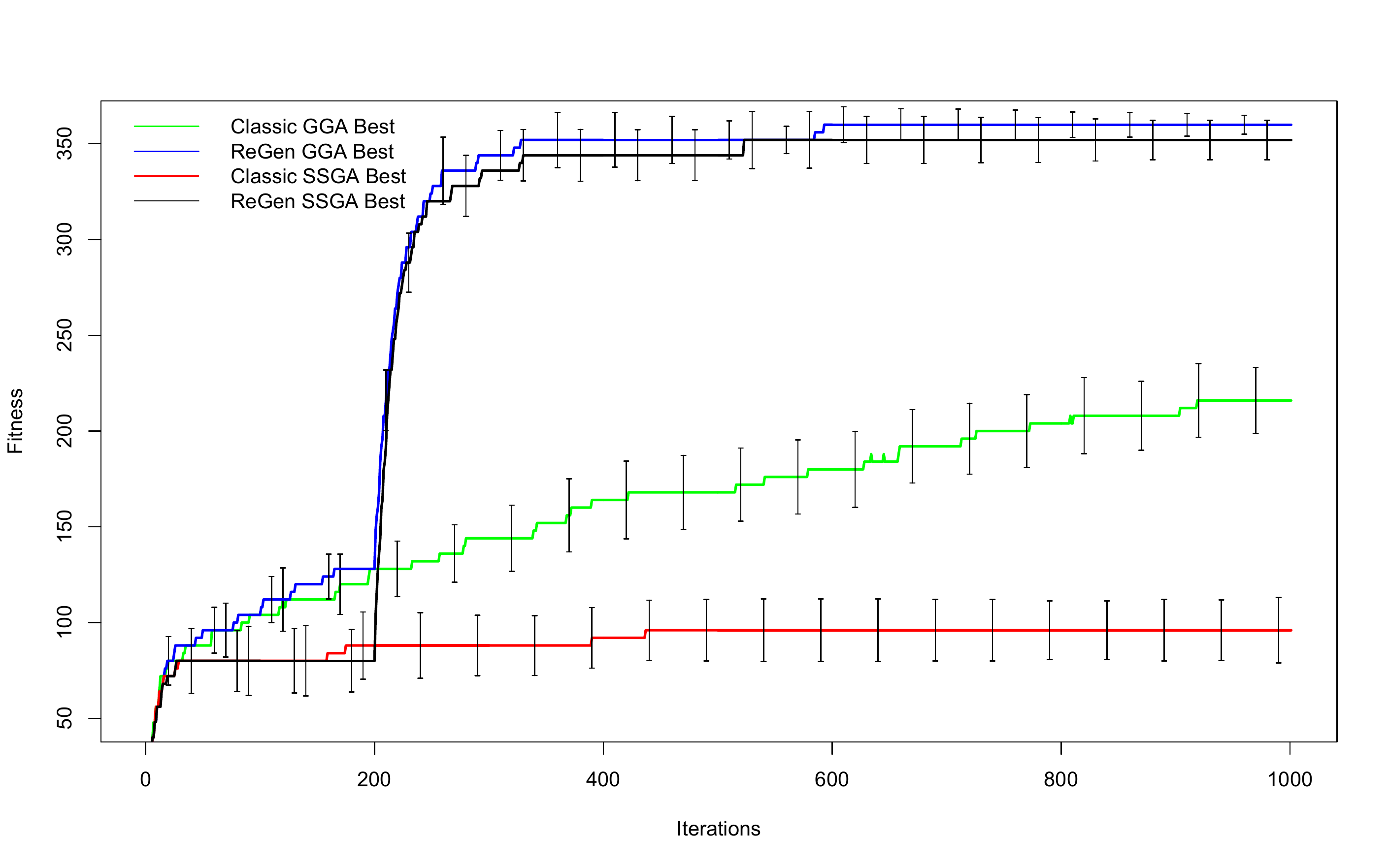}
\includegraphics[width=1.9in]{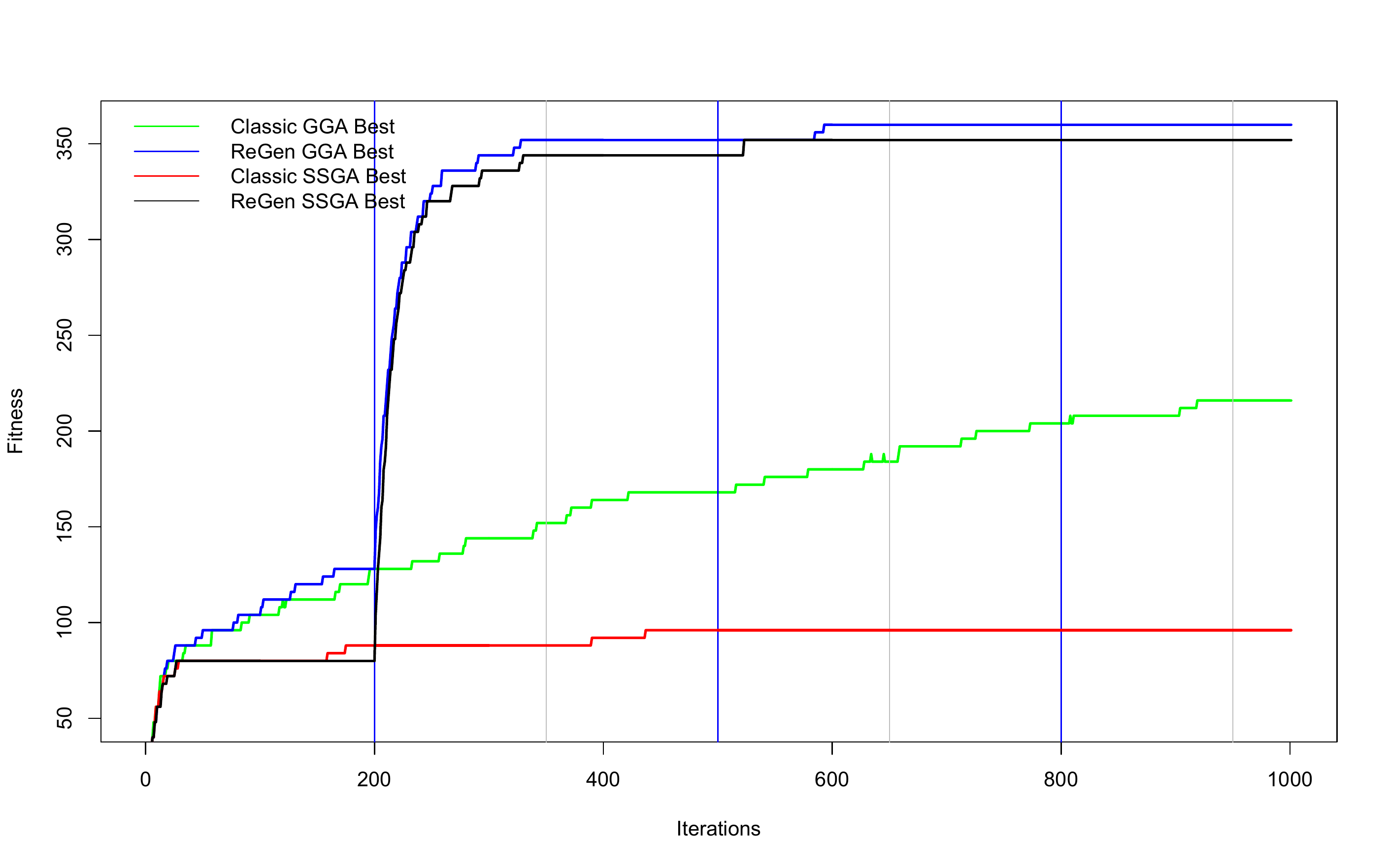}
\includegraphics[width=1.9in]{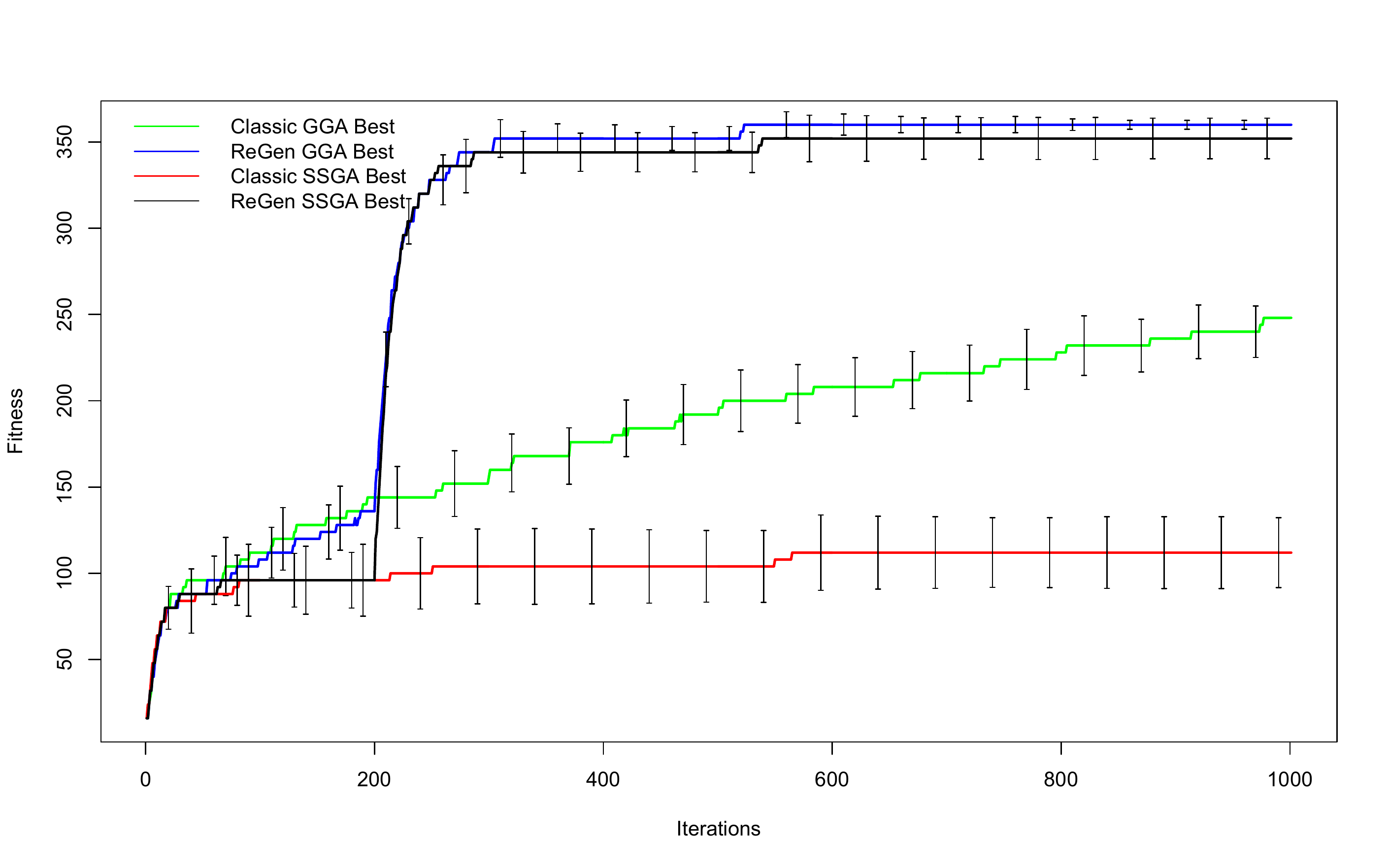}
\includegraphics[width=1.9in]{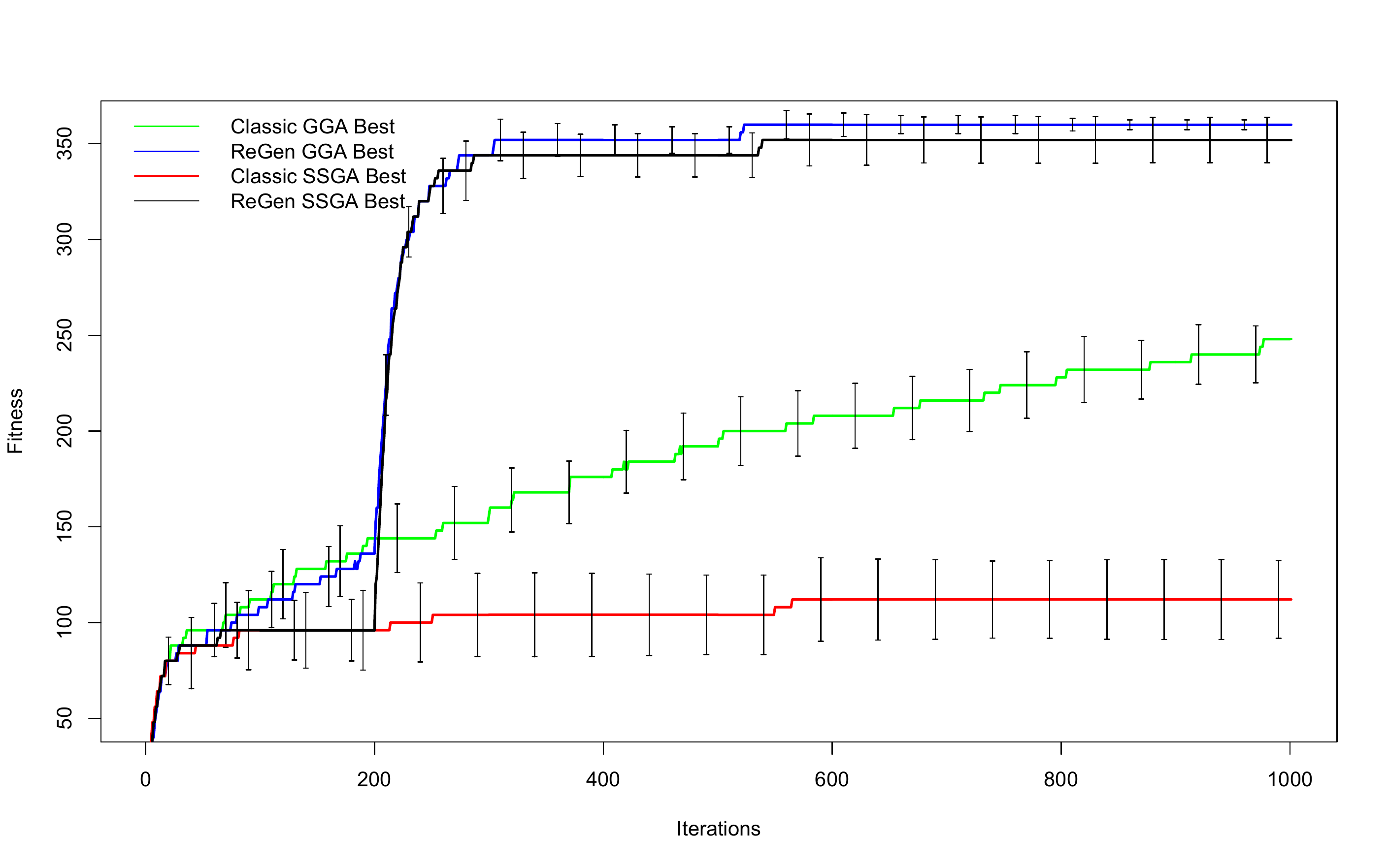}
\includegraphics[width=1.9in]{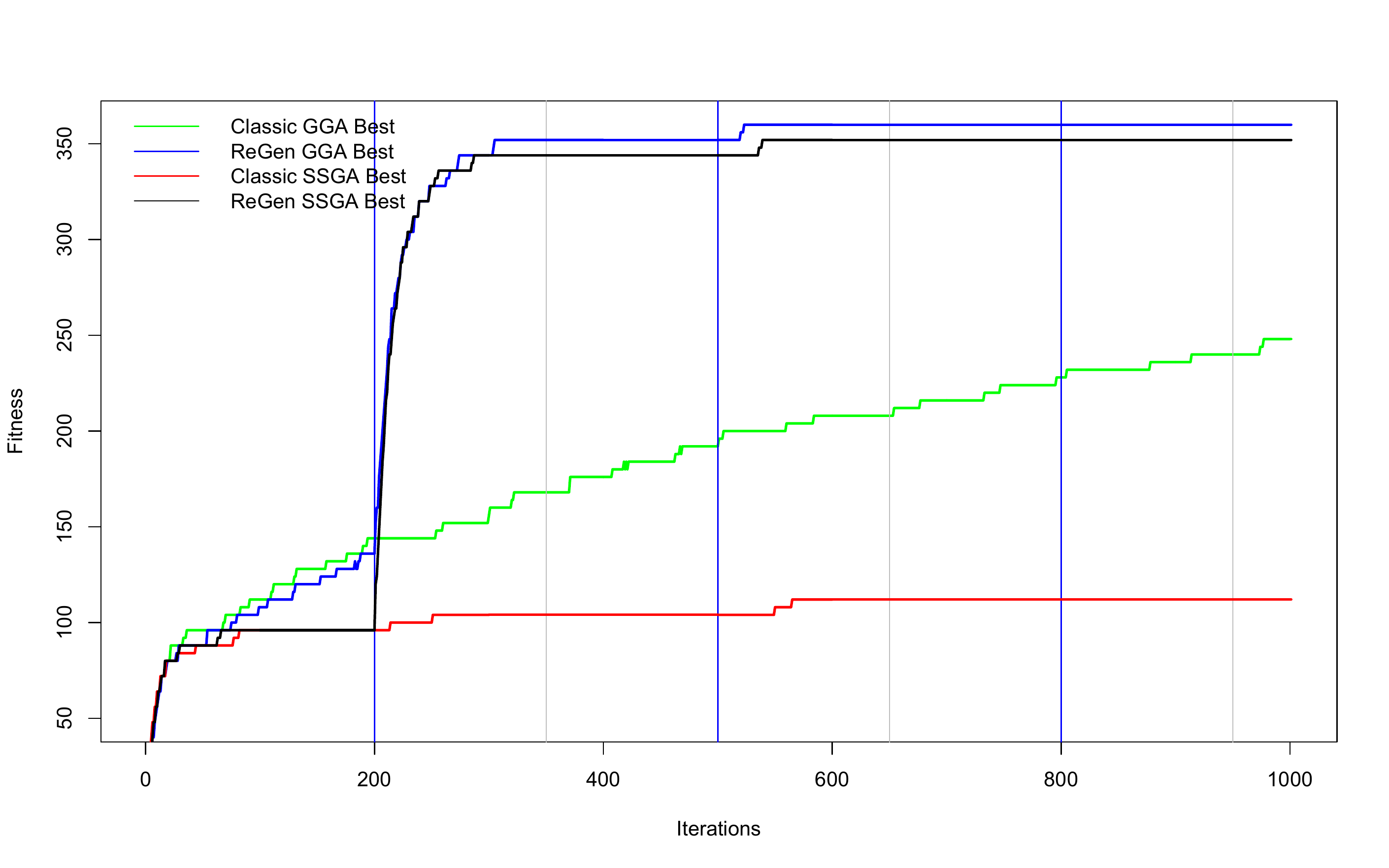}
\includegraphics[width=1.9in]{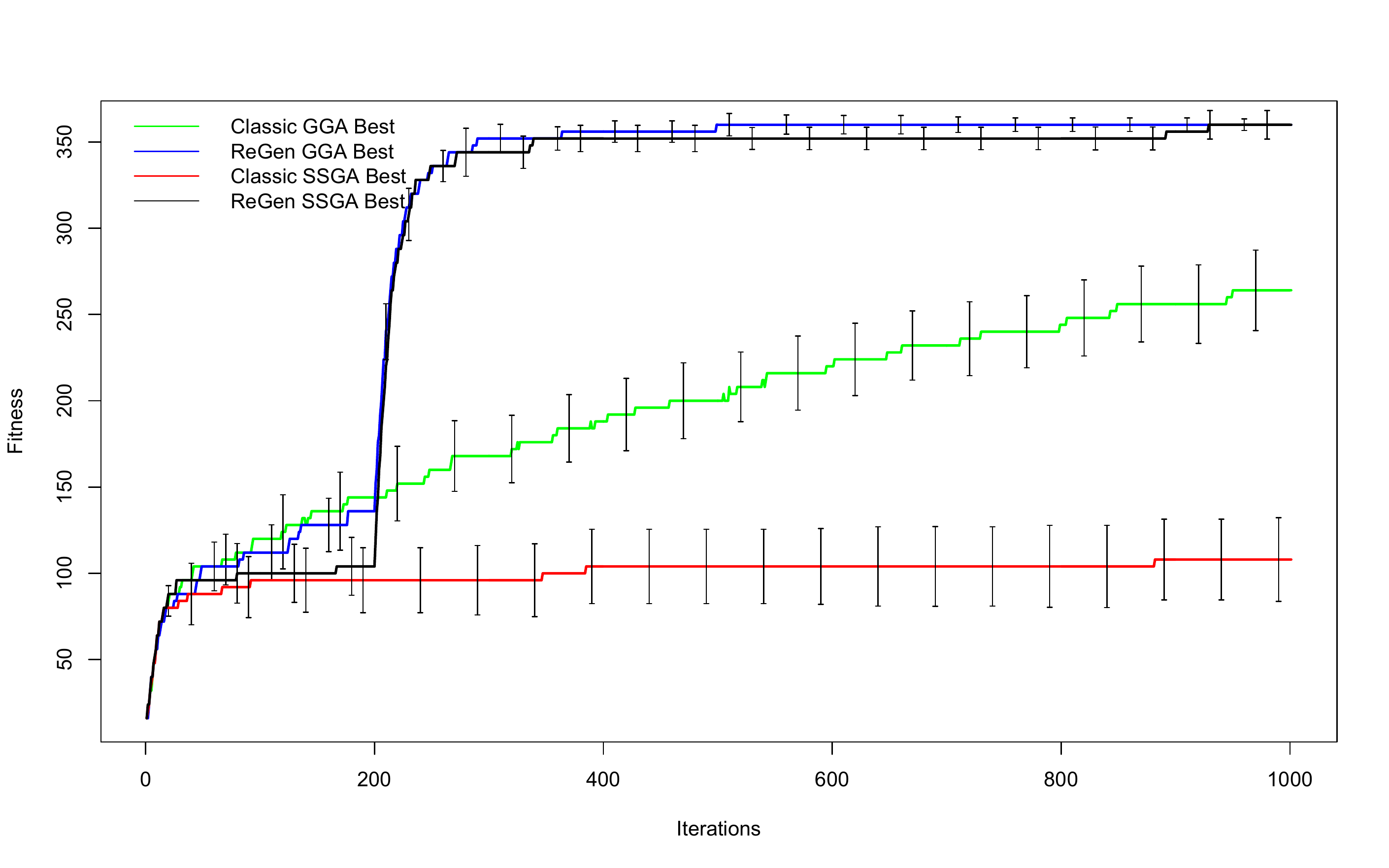}
\includegraphics[width=1.9in]{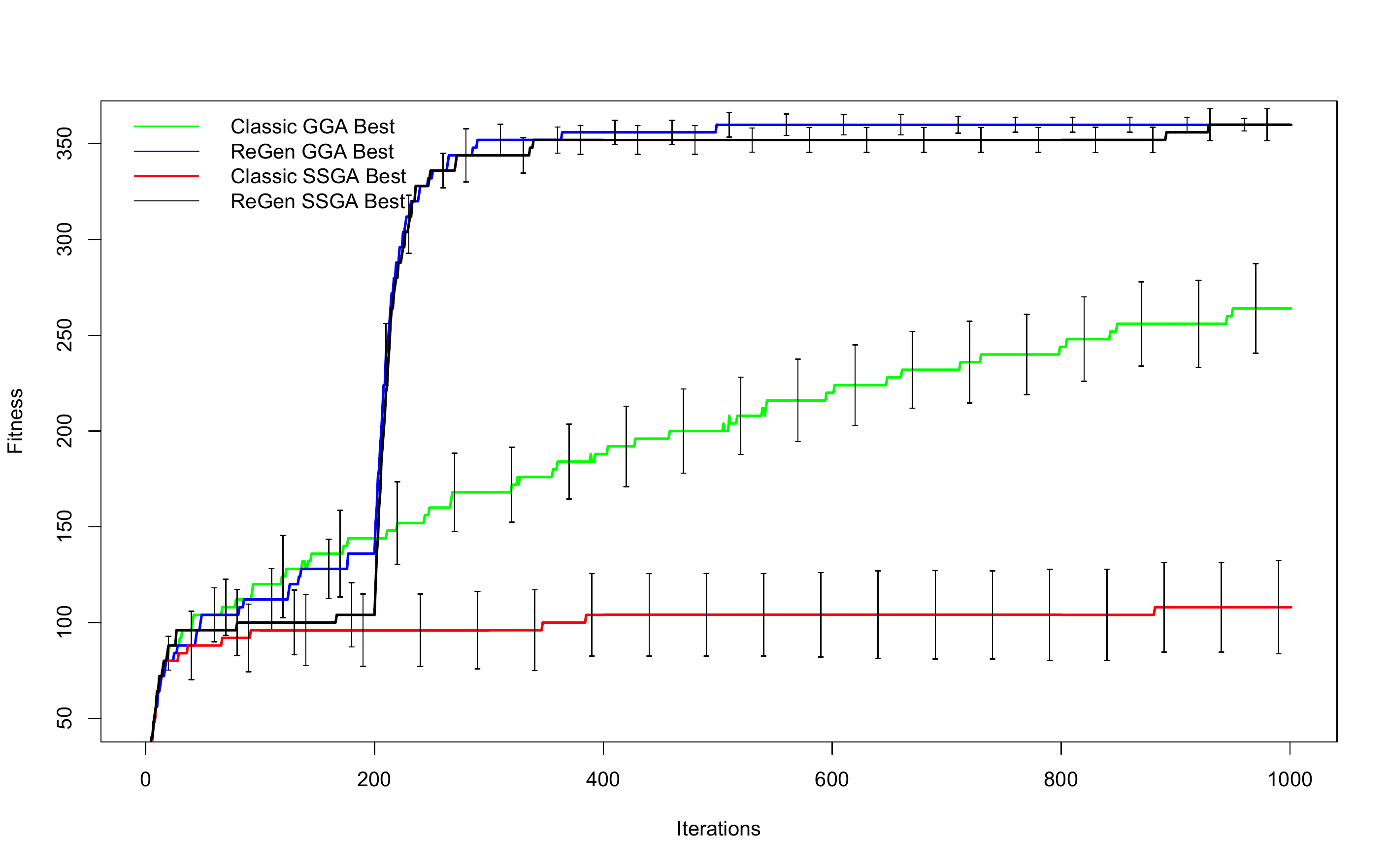}
\includegraphics[width=1.9in]{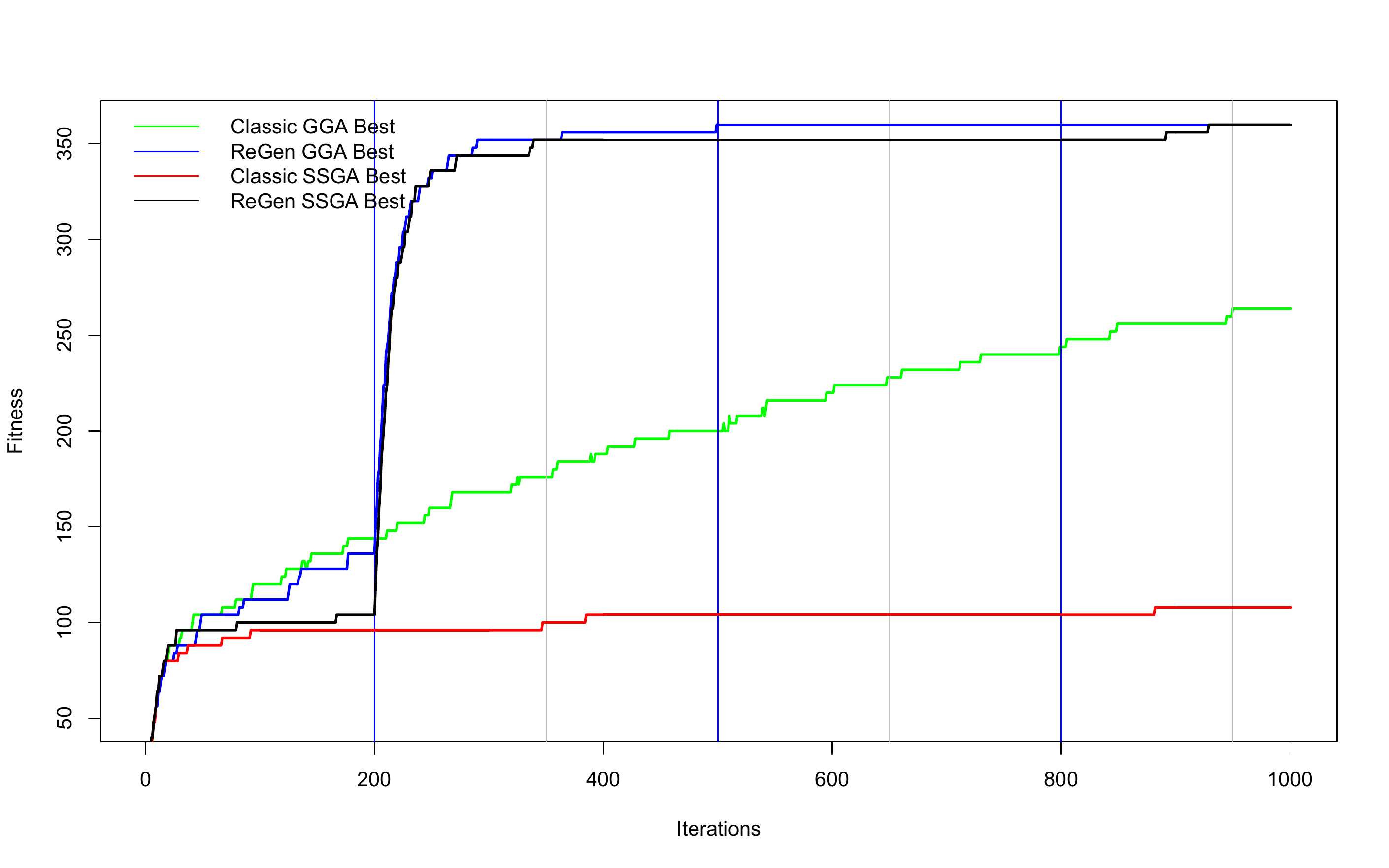}
\includegraphics[width=1.9in]{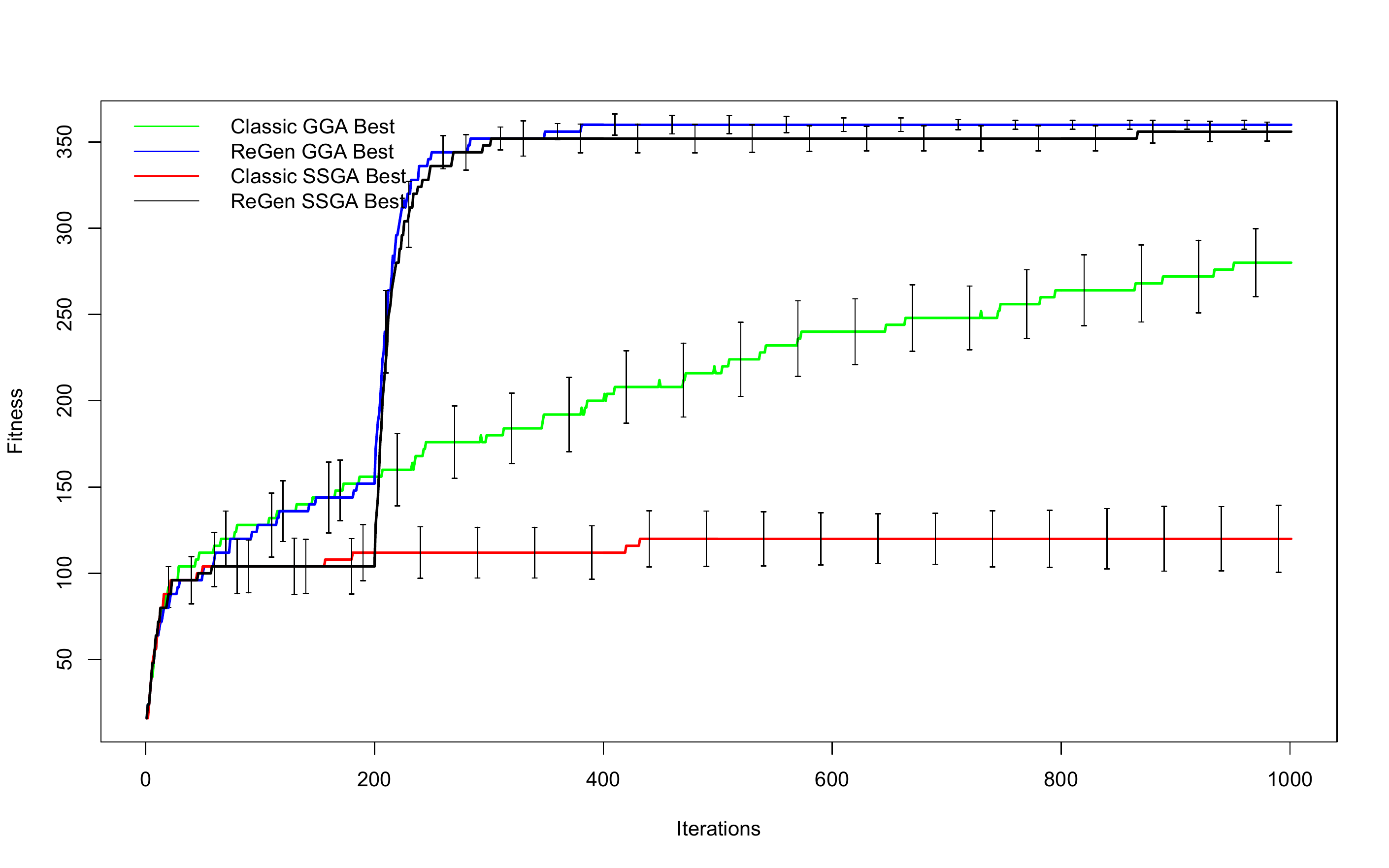}
\includegraphics[width=1.9in]{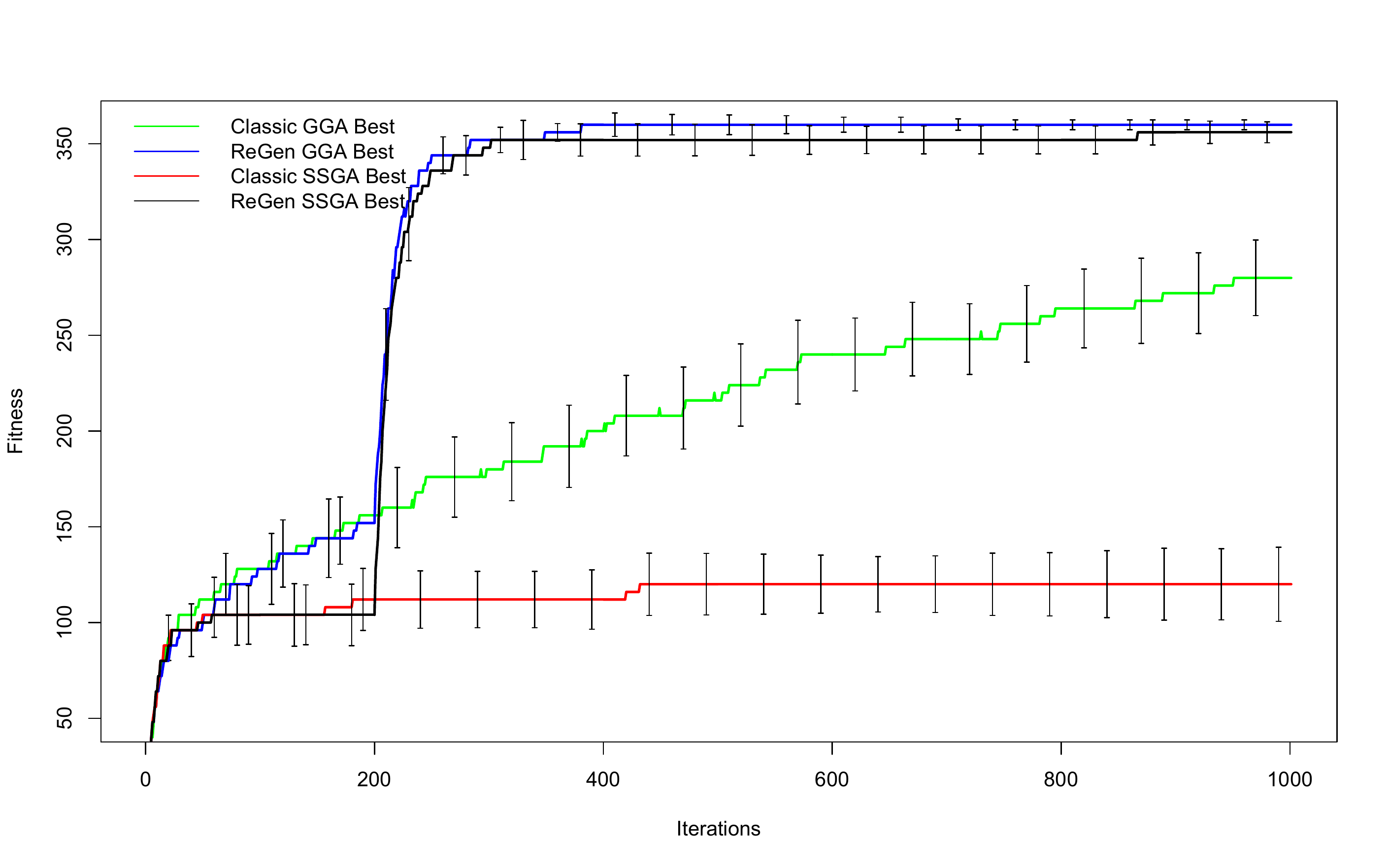}
\includegraphics[width=1.9in]{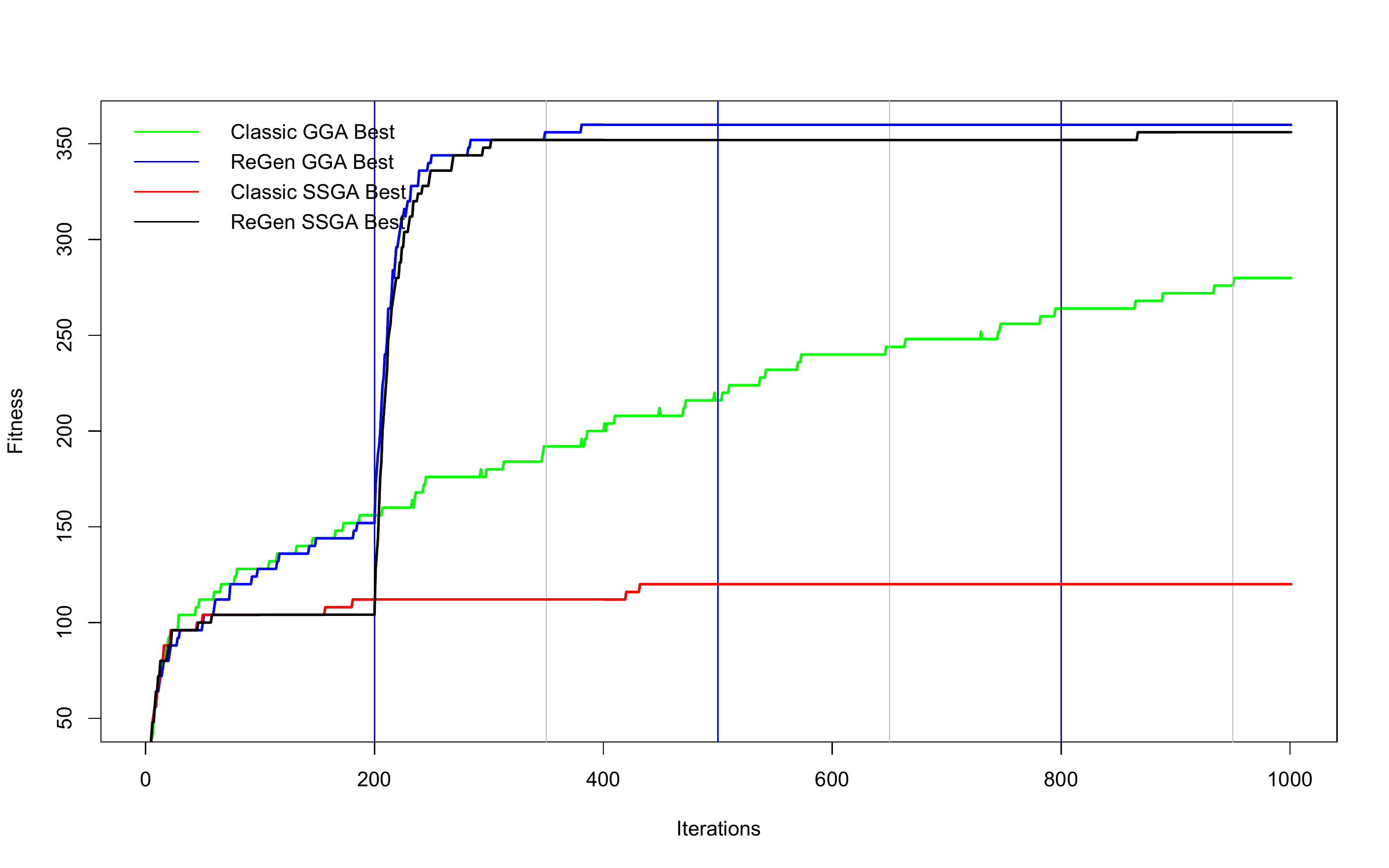}
\caption{Royal Road. Generational replacement (GGA) and Steady State replacement (SSGA). From top to bottom, crossover rates from $0.6$ to $1.0$.}
\label{c4fig3}
\end{figure}

\begin{figure}[H]
\centering
\includegraphics[width=1.9in]{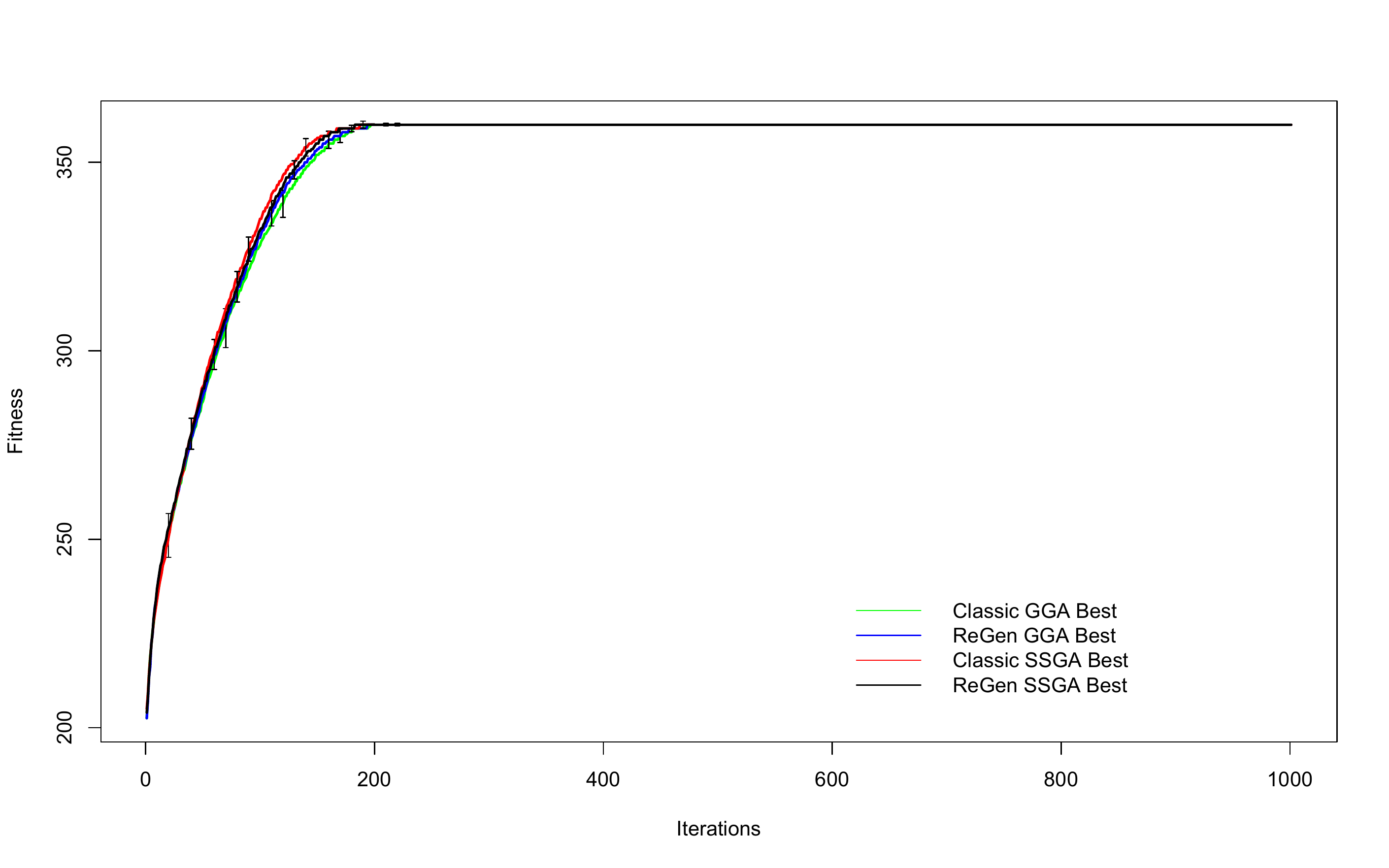}
\includegraphics[width=1.9in]{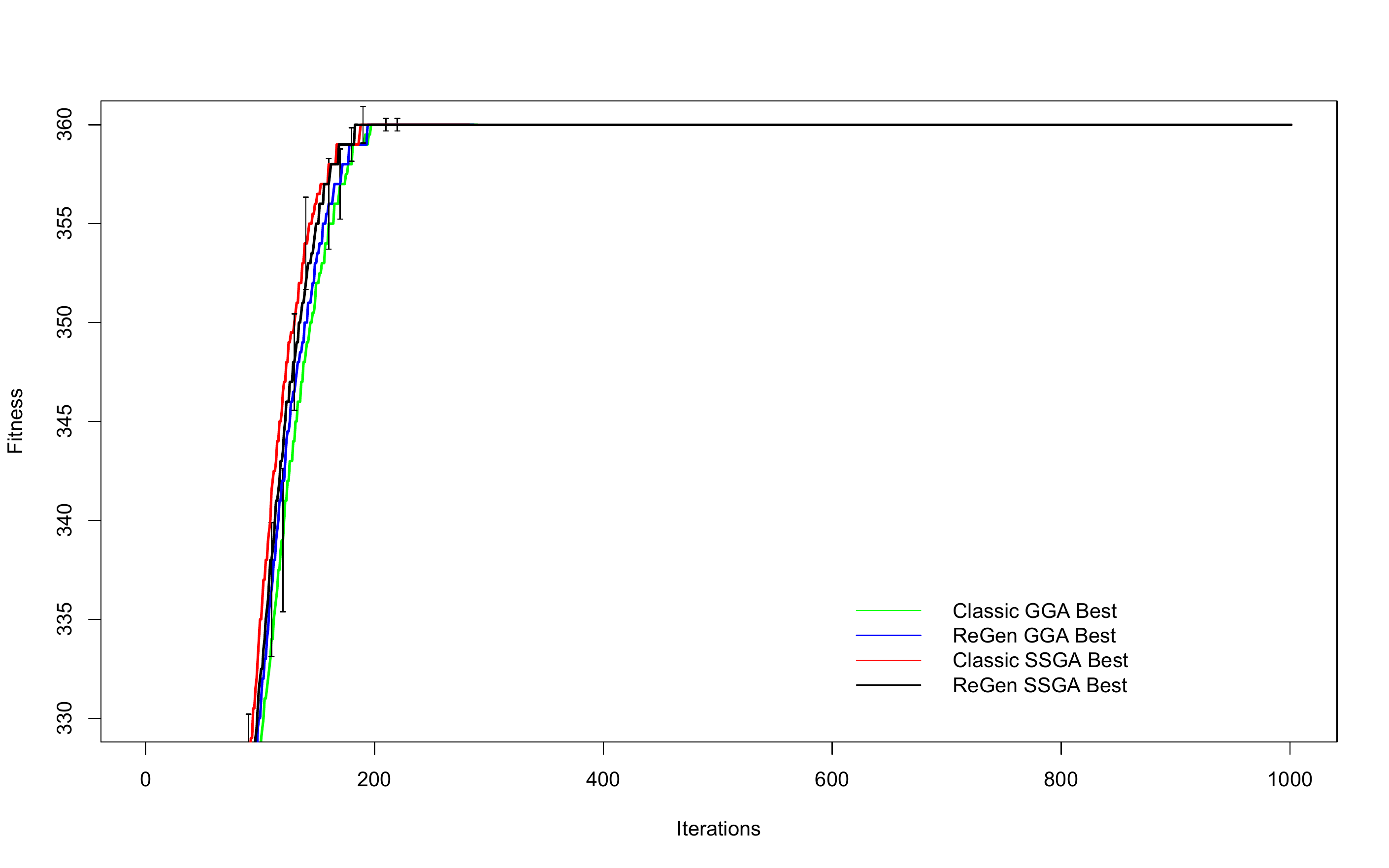}
\includegraphics[width=1.9in]{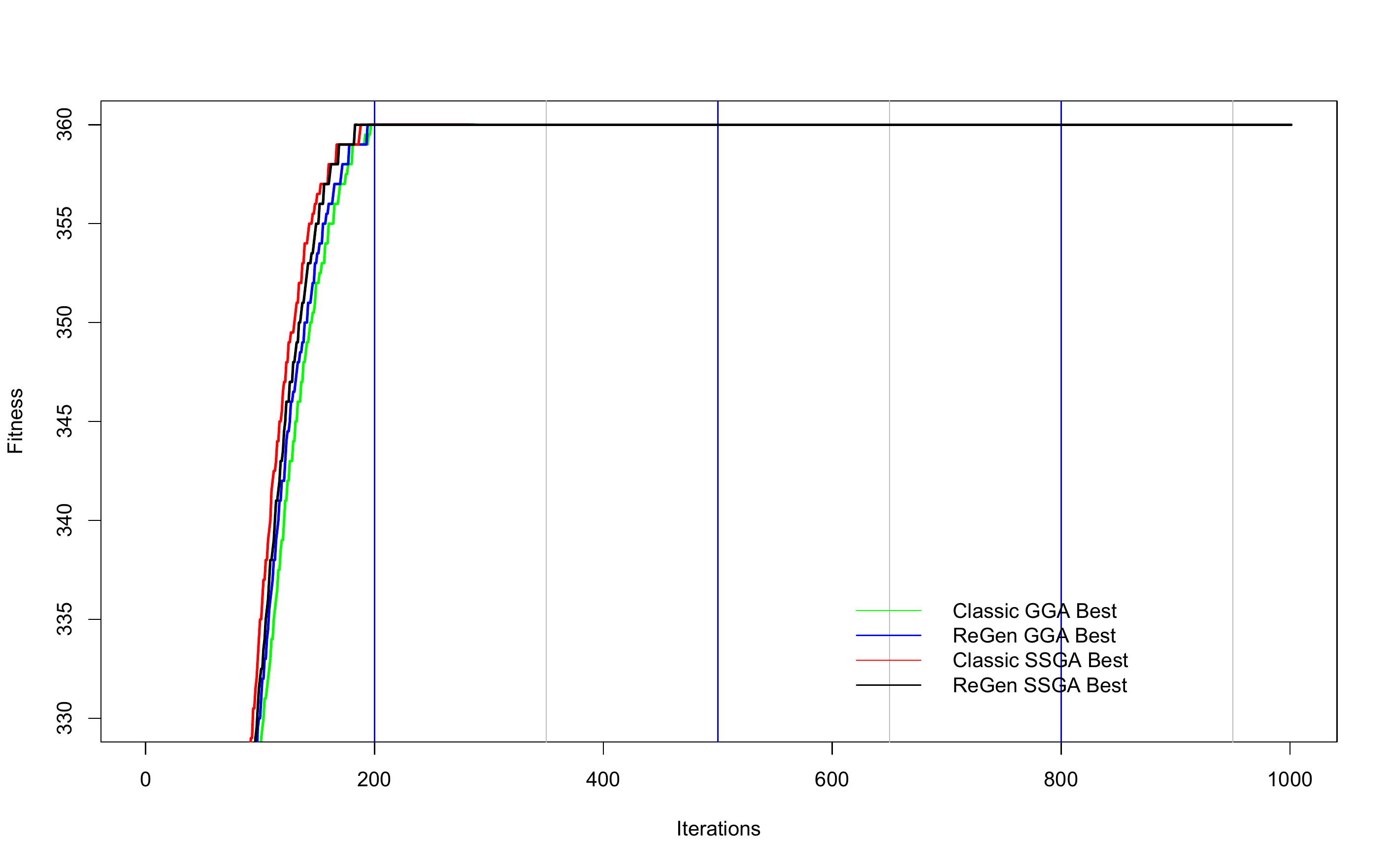}
\includegraphics[width=1.9in]{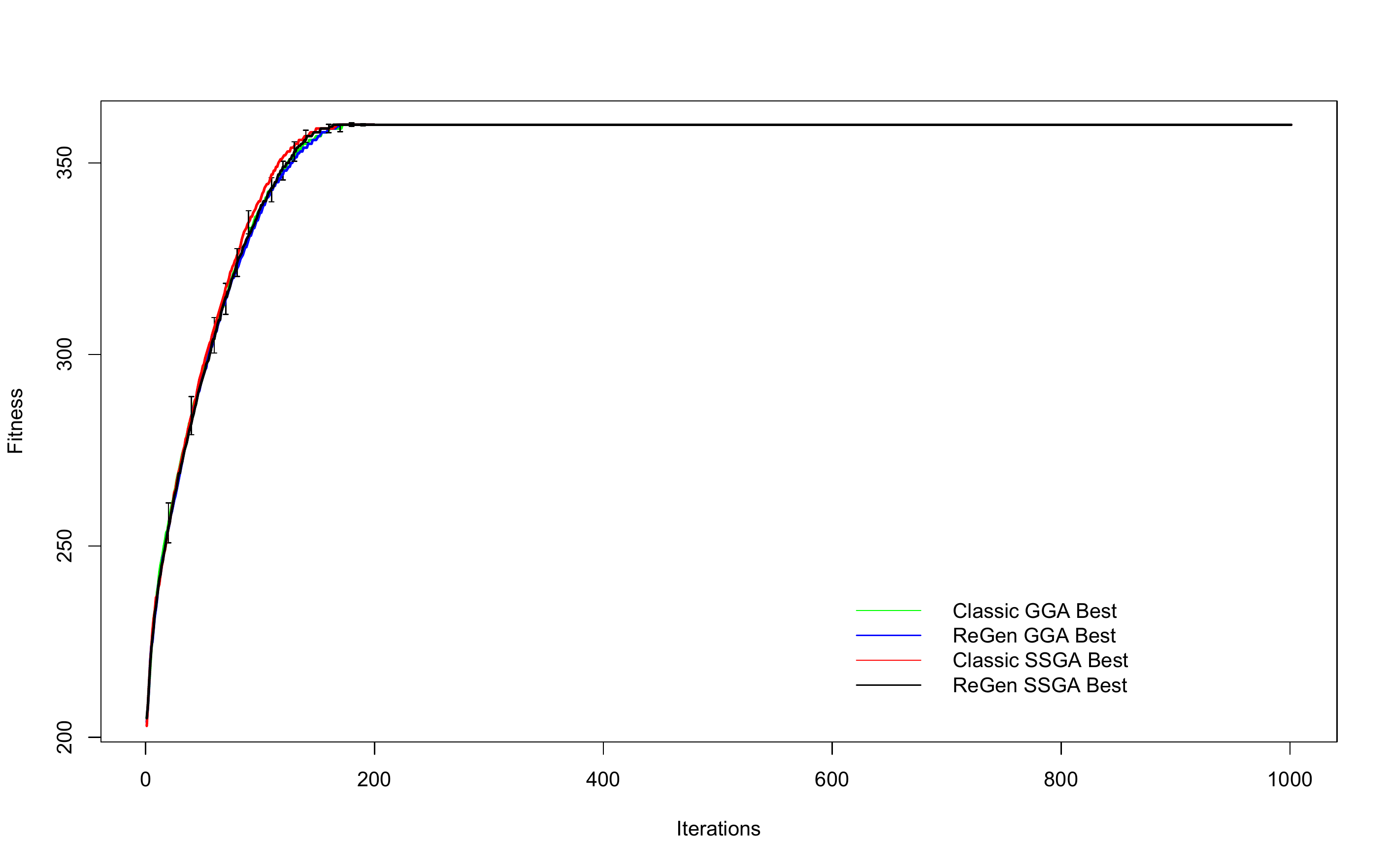}
\includegraphics[width=1.9in]{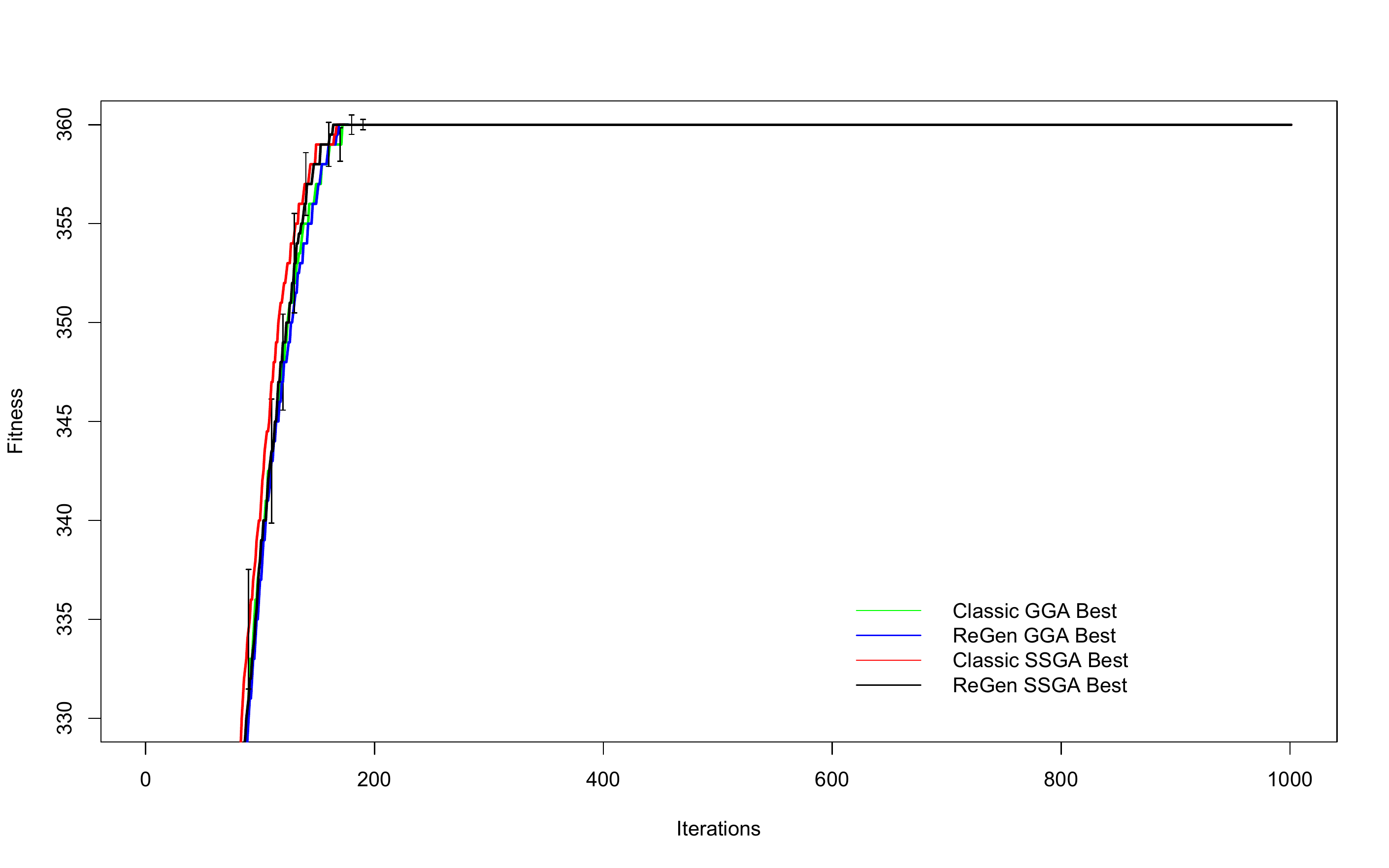}
\includegraphics[width=1.9in]{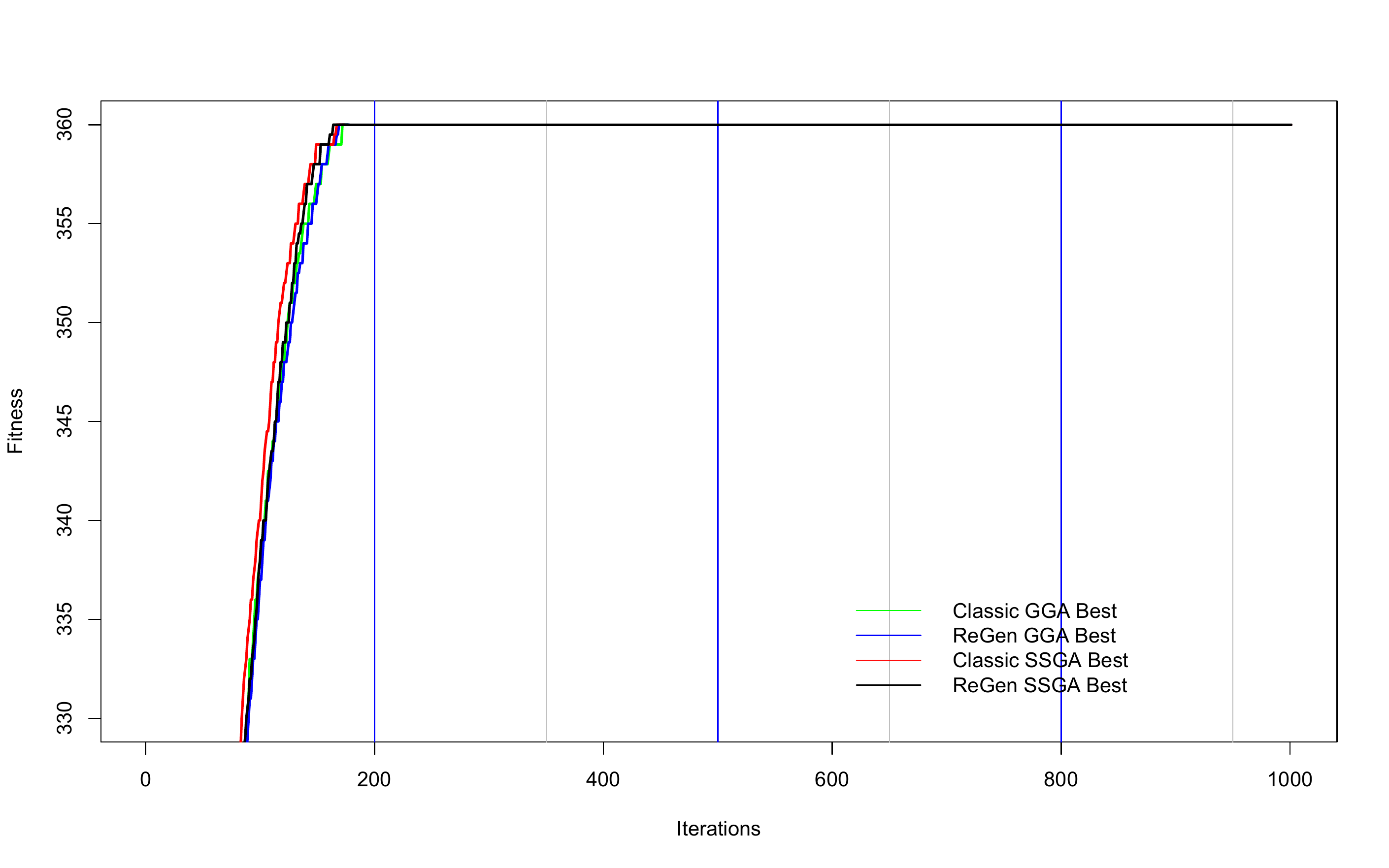}
\includegraphics[width=1.9in]{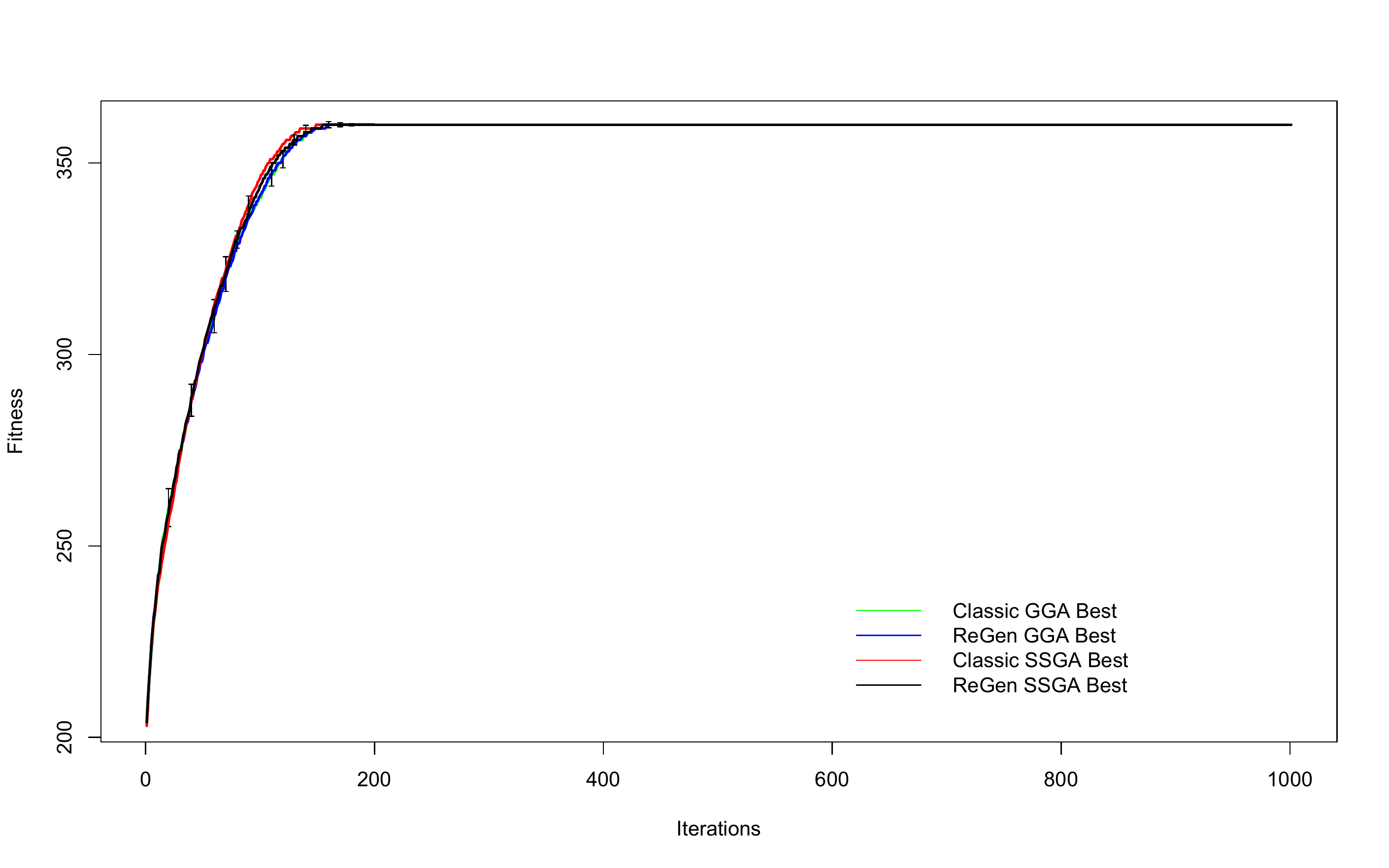}
\includegraphics[width=1.9in]{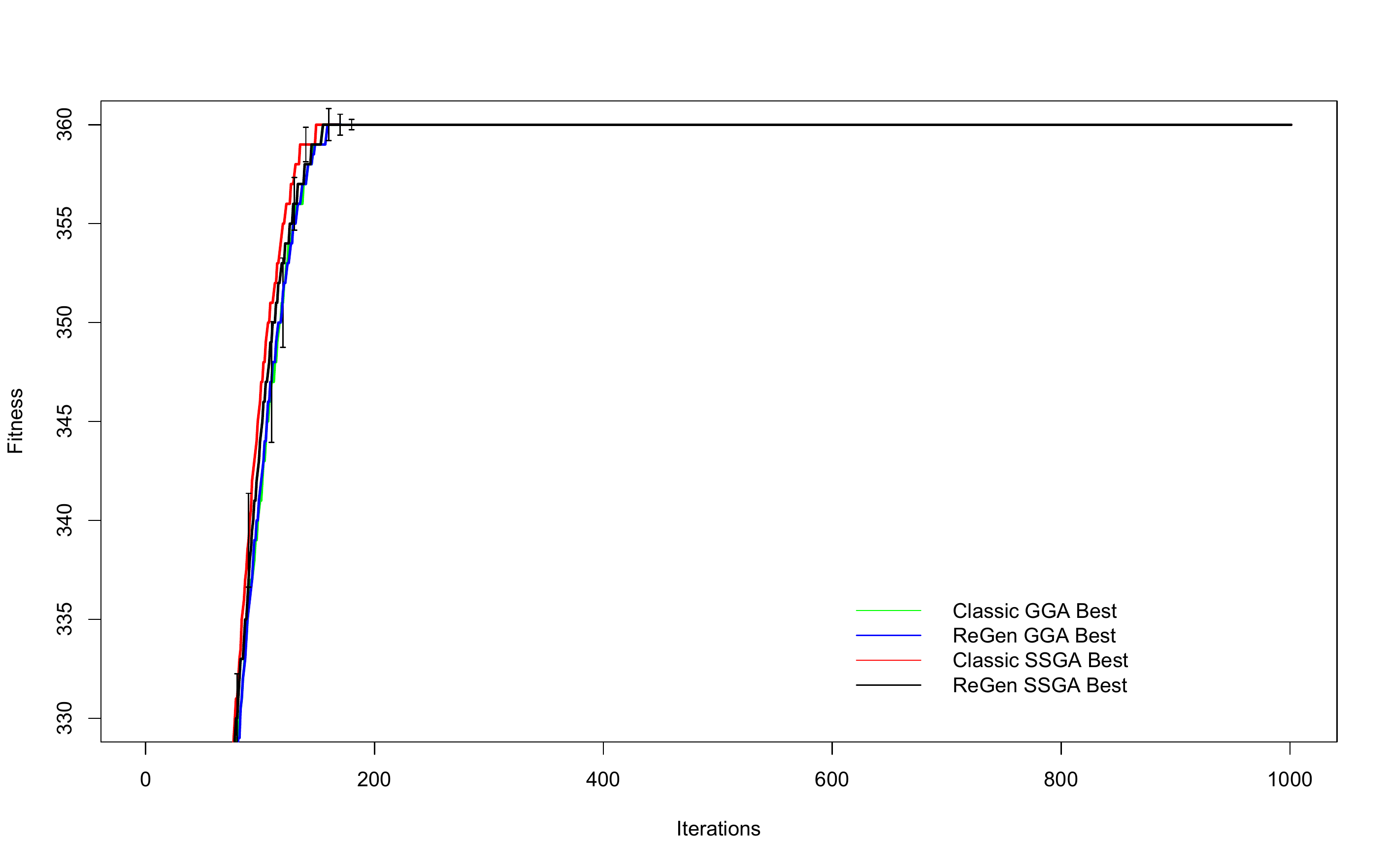}
\includegraphics[width=1.9in]{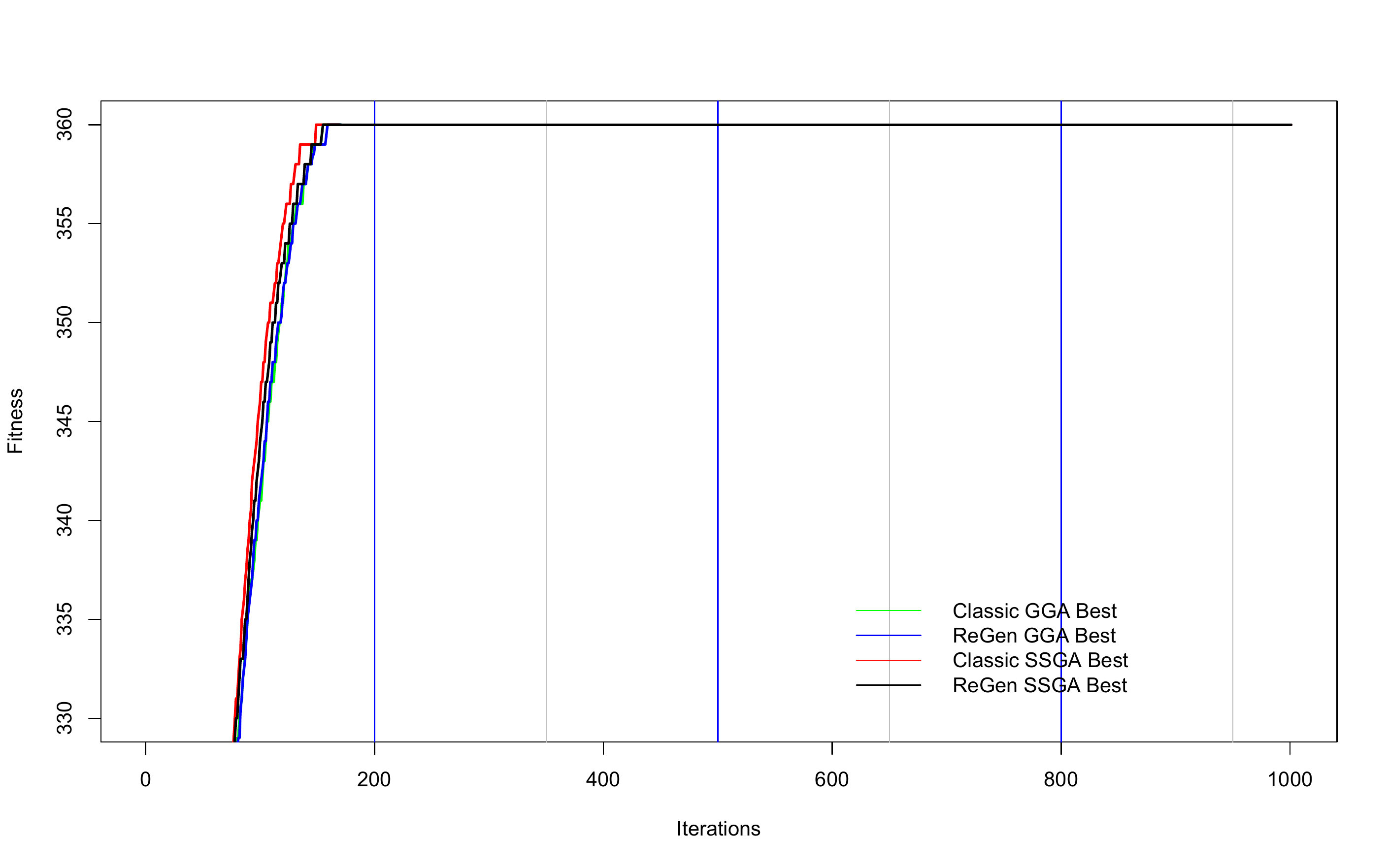}
\includegraphics[width=1.9in]{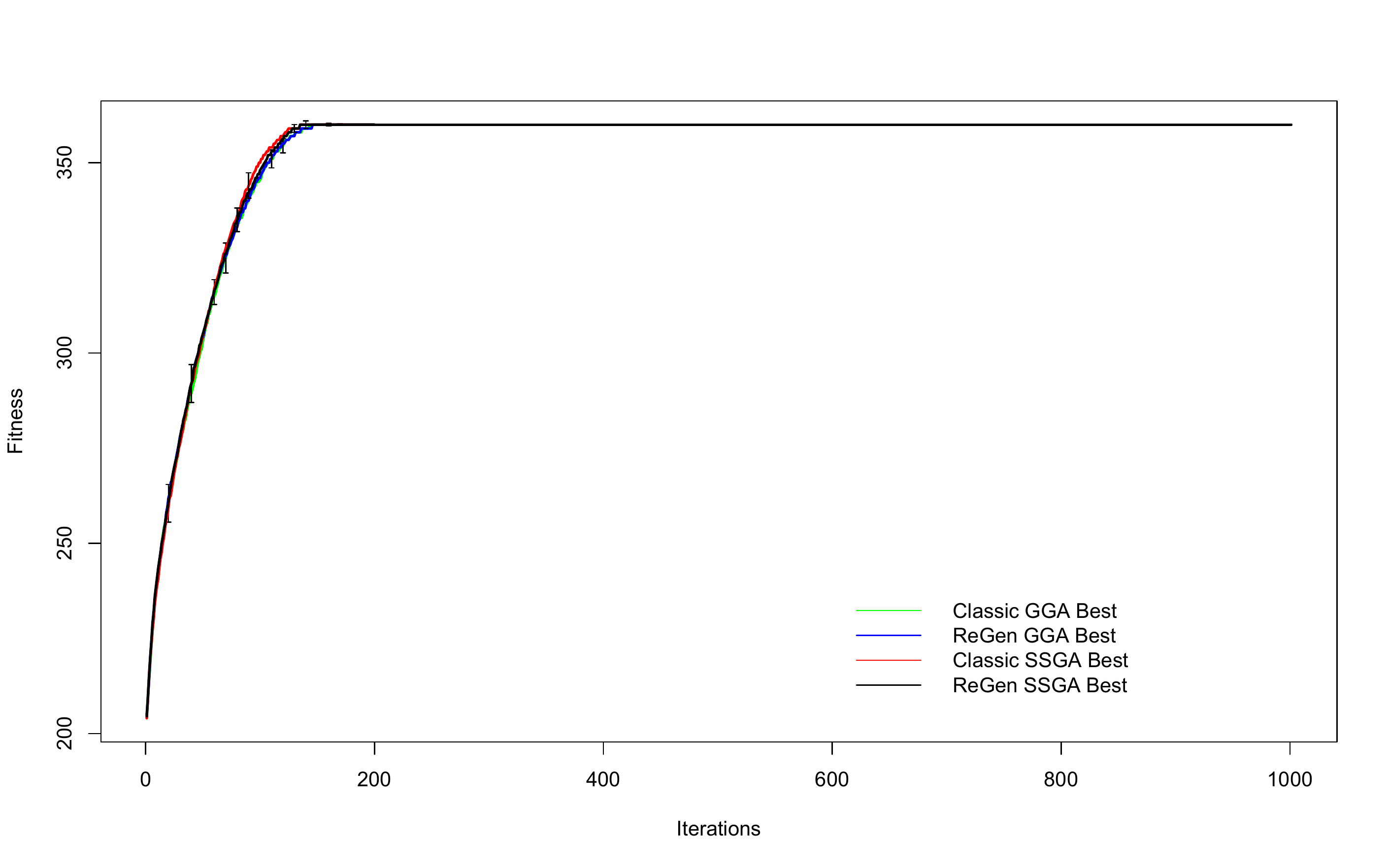}
\includegraphics[width=1.9in]{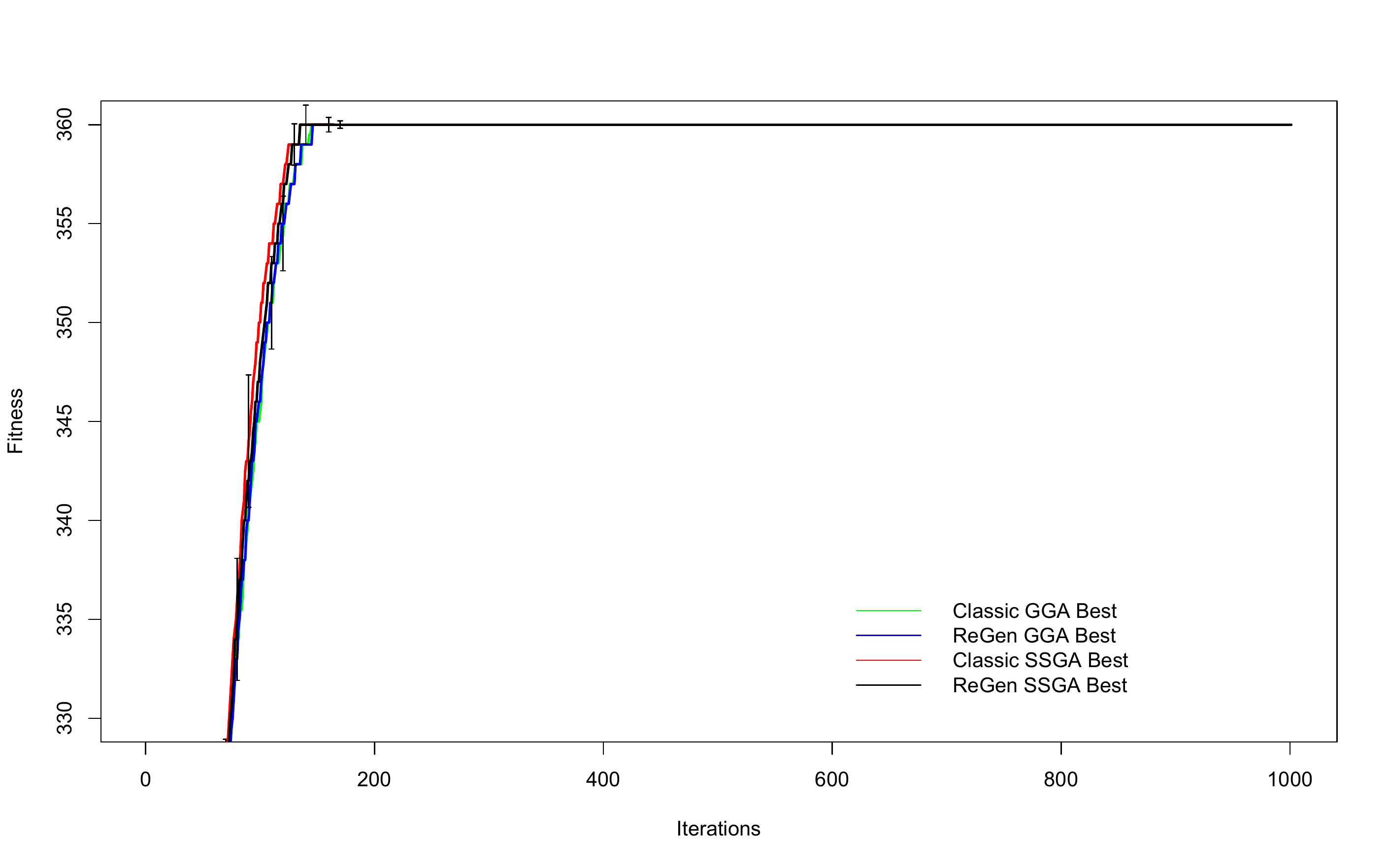}
\includegraphics[width=1.9in]{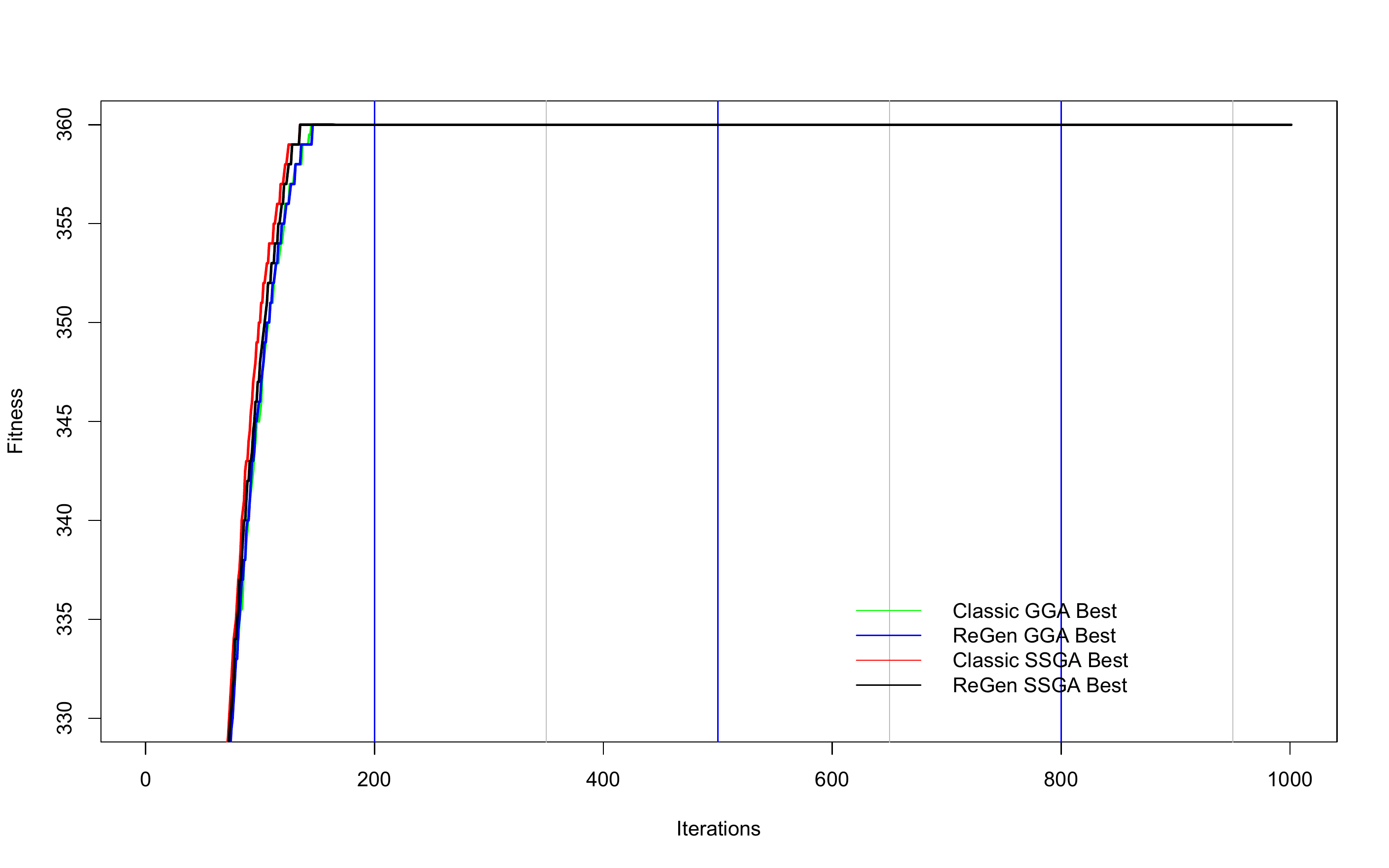}
\includegraphics[width=1.9in]{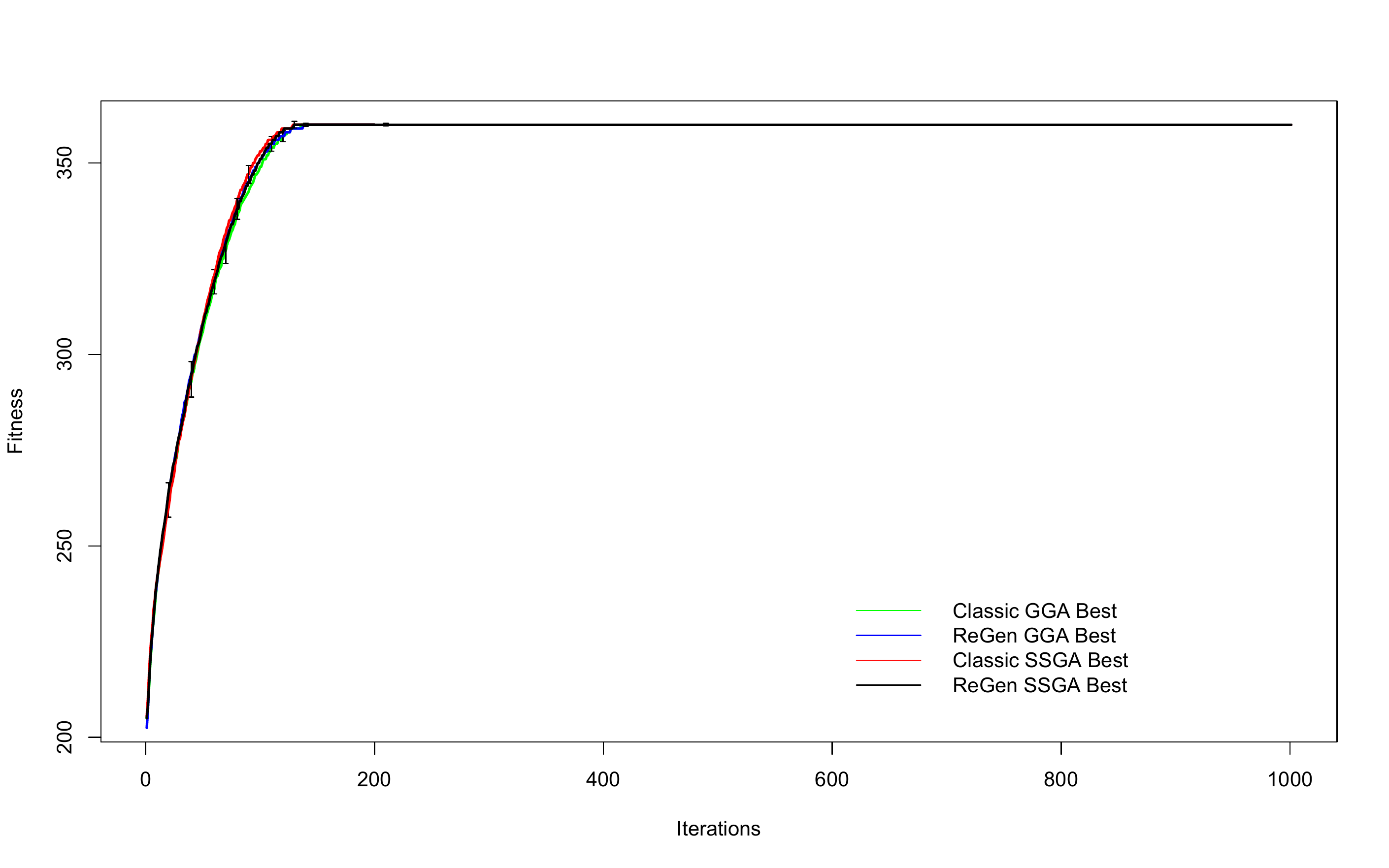}
\includegraphics[width=1.9in]{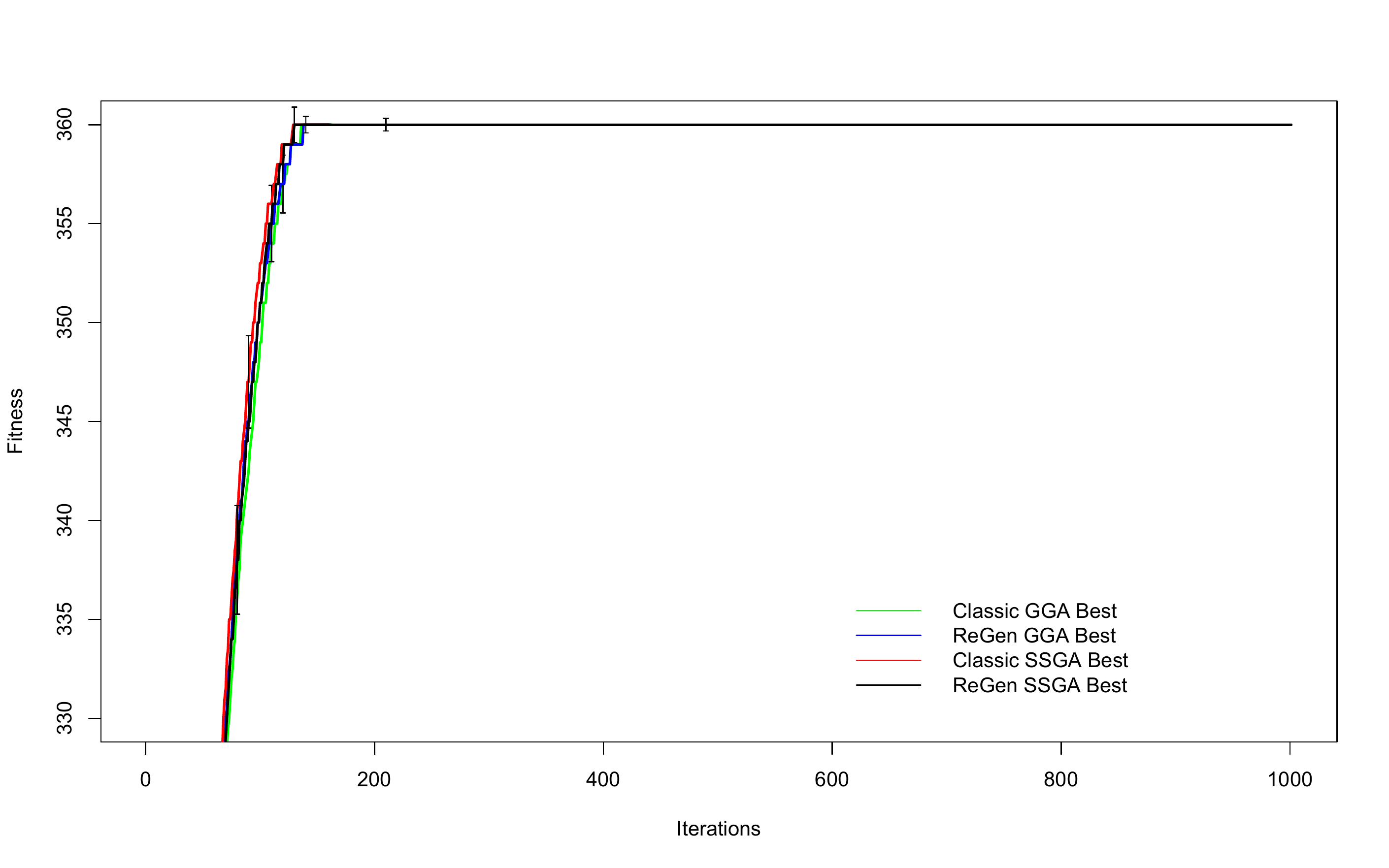}
\includegraphics[width=1.9in]{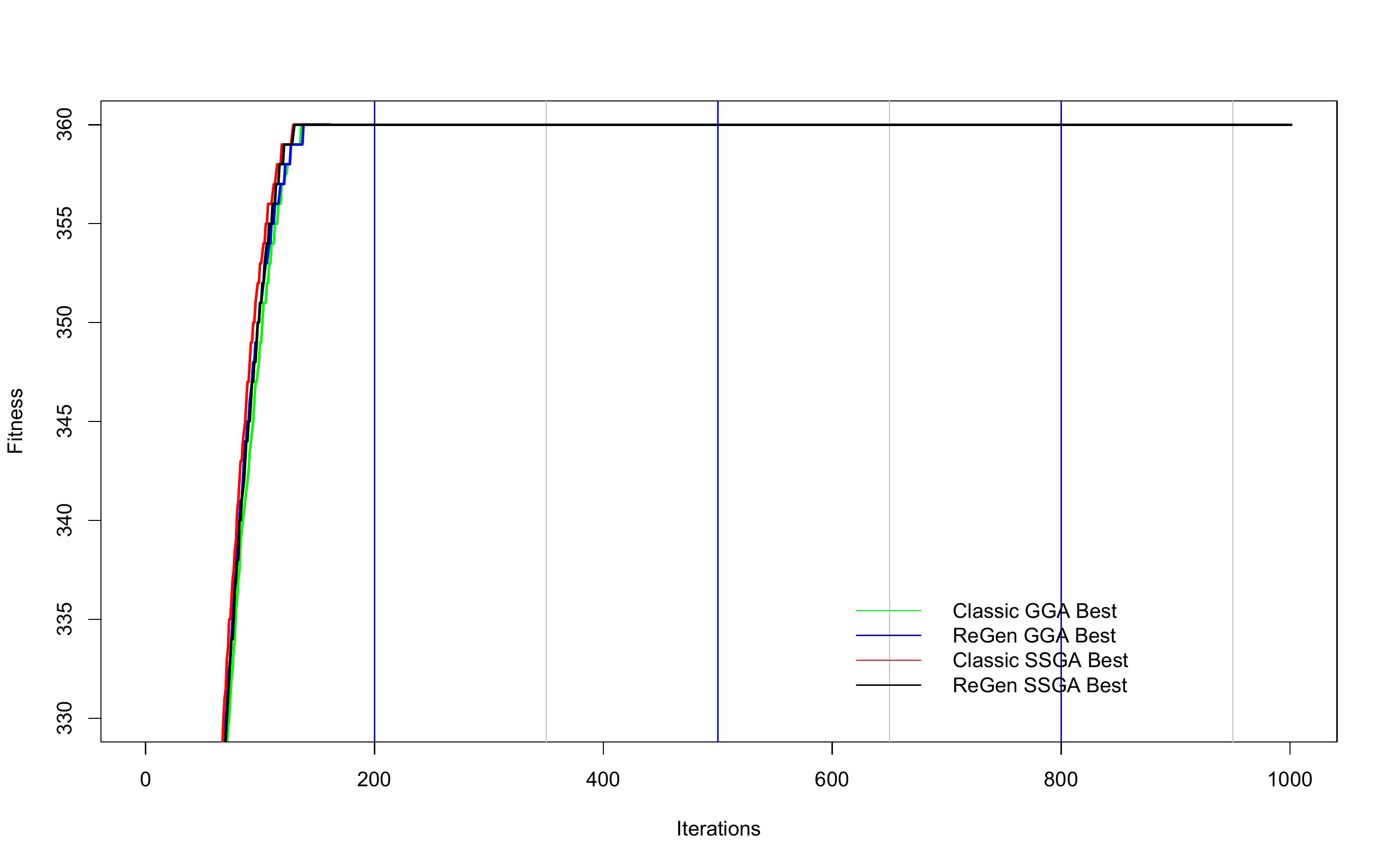}
\caption{Max Ones. Generational replacement (GGA) and Steady State replacement (SSGA). From top to bottom, crossover rates from $0.6$ to $1.0$.}
\label{c4fig4}
\end{figure}

Graphs from Fig.~\ref{c4fig1} to Fig.~\ref{c4fig4} illustrate the fitness of best individuals of populations in the experiments, reported fitnesses are based on the maximum median performance. Each graph shows the tendency of best individuals per technique. For ReGen GA and Classic GA, two methods are applied: steady state and generational population replacements. The fitness evolution of individuals can be appreciated by tracking green and red lines that depict the best individual's fitness for classic GAs. Blue and black lines trace the best individual's fitness for ReGen GAs. From top to bottom, each figure displays individuals' behavior with crossover rates from $0.6$ to $1.0$. Figures on the right corner show defined marking periods. Vertical lines in blue depict the starting of a marking period, lines in gray delimit the end of such periods. 

\newpage

\begin{table}[H]
  \centering
\caption{Results of the experiments for Generational and Steady replacements: Deceptive Order 3}
\label{c4table5}
\begin{tabular}{p{1cm}cccccl}
 \hline
\multirow{2}{5cm}{\textbf{Rate}} & \multicolumn{4}{c}{\textbf{ Deceptive Order 3}} \\
\cline{2-5} & \textbf{Classic GGA} & \textbf{Classic SSGA} & \textbf{ReGen GGA} & \textbf{ReGen SSGA} \\
\hline
0.6 & $3436 \pm11.86 [206]$ & $3430 \pm10.80 [192]$ & $3578 \pm11.70 [894]$ & $3573 \pm12.60 [926]$\\
0.7 & $3429 \pm09.50 [846]$  & $3433 \pm09.06 [263]$  & $3577 \pm13.31 [919]$ & $3576 \pm14.77 [929]$\\
0.8 & $3436 \pm10.31 [224]$ & $3438 \pm12.38 [171]$ & $3580 \pm16.20 [911]$ & $3582 \pm13.90 [884]$\\
0.9 & $3438 \pm11.12 [171]$ & $3435 \pm10.60 [163]$ & $3582 \pm15.54 [928]$ & $3586 \pm11.76 [918]$\\
1.0 & $3435 \pm09.33 [218]$  & $3437 \pm12.51 [145]$ & $3584 \pm13.49 [957]$ & $3587 \pm11.83 [854]$\\
\hline
\end{tabular}
\end{table}

\begin{table}[H]
  \centering
\caption{Results of the experiments for Generational and Steady replacements: Deceptive Order 4}
\label{c4table6}
\begin{tabular}{p{1cm}cccccl}
 \hline
\multirow{2}{5cm}{\textbf{Rate}} & \multicolumn{4}{c}{\textbf{ Deceptive Order 4}} \\
\cline{2-5} & \textbf{Classic GGA} & \textbf{Classic SSGA} & \textbf{ReGen GGA} & \textbf{ReGen SSGA} \\
\hline
0.6 &  $388.0 \pm4.62 [175]$ & $387.5\pm4.84 [273]$ & $445.0 \pm3.32 [916]$ & $443.5 \pm2.86 [909] $\\
0.7 &  $389.0 \pm4.82 [155]$ & $387.0 \pm3.37  [191]$ & $446.0 \pm1.83 [900]$ & $446.0 \pm2.74 [846] $\\
0.8 &  $390.0 \pm3.88 [156]$ & $390.0 \pm4.32  [158]$ & $444.5\pm3.94 [898]$ & 
$445.0 \pm3.57 [608] $\\
0.9 &  $390.0 \pm3.40 [139]$ & $ 388.5\pm5.20 [131]$ & $ 445.5\pm2.16 [943]$ & $446.0 \pm2.14 [897] $\\
1.0 &  $392.5\pm4.82 [148]$& $ 391.5\pm4.07 [155]$ & $ 446.5\pm2.89 [963]$ & $446.0 \pm3.12 [854]$ \\\hline
\end{tabular}
\end{table}

\begin{table}[H]
  \centering
\caption{Results of the experiments for Generational and Steady replacements: Royal Road}
\label{c4table7}
\begin{tabular}{p{1cm}cccccl}
 \hline
\multirow{2}{5cm}{\textbf{Rate}} & \multicolumn{4}{c}{\textbf{ Royal Road}} \\
\cline{2-5} & \textbf{Classic GGA} & \textbf{Classic SSGA} & \textbf{ReGen GGA} & \textbf{ReGen SSGA} \\
\hline
0.6 & $ 200 \pm28.34 [928]$ & $ 96.0 \pm18.25 [810]$ & $ 360 \pm6.64 [632]$ & $ 352 \pm10.81 [879]$\\
0.7 & $ 216 \pm19.19 [919]$ & $ 96.0 \pm15.72 [440]$ & $ 360 \pm9.27 [593]$ & $ 352 \pm14.92 [523]$\\
0.8 & $ 248 \pm15.93 [977]$ & $ 112 \pm21.83 [569]$ & $ 360 \pm7.86 [523]$ & $ 352 \pm13.45 [539]$\\
0.9 & $ 264 \pm23.76 [950]$ & $ 180 \pm23.37 [887]$ & $ 360 \pm6.64 [499]$ & $ 360 \pm08.27 [929]$\\
1.0 & $ 280 \pm20.58 [951]$ & $ 188 \pm16.74 [436]$ & $ 360 \pm6.12 [381]$ & $ 356 \pm07.20 [867] $\\\hline
\end{tabular}
\end{table}

\begin{table}[H]
  \centering
\caption{Results of the experiments for Generational and Steady replacements: Max Ones}
\label{c4table8}
\begin{tabular}{p{1cm}cccccl}
 \hline
\multirow{2}{5cm}{\textbf{Rate}} & \multicolumn{4}{c}{\textbf{ Max Ones}} \\
\cline{2-5} & \textbf{Classic GGA} & \textbf{Classic SSGA} & \textbf{ReGen GGA} & \textbf{ReGen SSGA} \\
\hline
0.6 & $ 360 \pm0.92 [197]$  &  $ 360 \pm0.74 [192]$  & $ 360 \pm0.89 [194]$ & $ 360 \pm0.87 [183]$\\
0.7 & $ 360 \pm1.03 [172]$  &  $ 360 \pm0.66 [174]$  & $ 360 \pm0.98 [169]$ & $ 360 \pm0.98 [164]$\\
0.8 & $ 360 \pm1.27 [155]$  &  $ 360 \pm0.83 [158]$  & $ 360 \pm0.87 [159]$ & $ 360 \pm1.06 [155]$\\
0.9 & $ 360 \pm0.89 [145]$  &  $ 360 \pm0.92 [143]$  & $ 360 \pm0.89 [146]$ & $ 360 \pm1.00 [135]$\\
1.0 & $ 360 \pm0.89 [136]$  &  $ 360 \pm0.87 [138]$  & $ 360 \pm0.83 [138]$ & $ 360 \pm0.89 [130]$\\\hline
\end{tabular}
\end{table}

\newpage

Based on tabulated results in Table~\ref{c4table5} for Deceptive Order Three, it can be noted that ReGen GA performs better than the classic GA. ReGen GA is able to discover varied optimal solutions until achieving the total of configured iterations. However, it does not report solutions with the global optimum ($3600$); the best solutions are close to the peak value; no better solutions than the reported are reached. In Fig.~\ref{c4fig1} is noticeable that the pressure applied on chromosomes, at iterations $200$, $500$, and $800$, does cause a change in the evolution of individuals. After starting the marking period, the fitness improves to be closer to the optimum, and populations improve their performance once tags are added. The ReGen GA found a variety of suited solutions during the evolution process, exposing the proposed approach's ability to discover novelties that are not identified by the classic GA. Fig.~\ref{c4fig1} also shows that classic GA performance is under ReGen GA performance in all crossover rates levels.

Tabulated results in Table~\ref{c4table6} for Deceptive Order Four Trap function show that ReGen GA performs better than the classic GA. ReGen GA solutions surpass the local maximum of $440$, but do not reach the global optimum ($450$). No better solutions than the reported are reached. In Fig.~\ref{c4fig2} is notable that the pressure applied on chromosomes at iteration $200$ produces a change in the evolution of individuals. After starting the marking period, the fitness raises near the optimum, and populations improve their performance once individuals' chromosomes are marked. Fig.~\ref{c4fig2} also shows that classic GA performance is under ReGen GA performance in all crossover rates levels. ReGen GA solutions reach a local optimum above $400$; in contrast, classic GA solutions are under $400$ for all crossover rates.

Next in order, tabulated results in Table~\ref{c4table7} for Royal Road function reflect that ReGen GA does reach solutions with the global optimum ($360$) in each crossover rate for generational replacement. Nevertheless, for steady state replacement, solutions with the absolute maximum are reported only for one crossover rate. In Fig.~\ref{c4fig3} is noticeable that the pressure applied on chromosomes at iterations $200$, $500$, and $800$ causes a significant change in the evolution of individuals. After starting the marking period, the fitness improves to reach the optimum, populations improve their performance once the tags are added. Fig.~\ref{c4fig3} also shows that the classic GA performance is under ReGen GA performance in all crossover rates levels. The classic GA does not reach suitable solutions for this experiment. Additionally, the maximum fitnesses are gotten in late iterations. ReGen GA obtained better solutions in earlier iterations.

Finally, tabulated results in Table~\ref{c4table8} for Max Ones' objective function display that both ReGen GA and classic GA have a similar performance. Experiments show that for the Max Ones' function, optimal solutions are found in both implementations. In Fig.~\ref{c4fig4}, the pressure applied to chromosomes during marking periods does not cause any change in the evolution of individuals. The reason is that before the marking period started, individual scores are near the optimum, or the global optimum is already found. After starting the marking period, the fitnesses keep stable for the best individuals. Fig.~\ref{c4fig4} also shows that both performances are similar in all crossover rates levels.

\subsection{Statistical Analysis}\label{c4s2ss3}

Three different tests are performed, One-Way ANOVA test, Pairwise Student’s t-test, and Paired Samples Wilcoxon Test (also known as Wilcoxon signed-rank test). The data set ReGen EAs Samples in Appendix \ref{appendB} is used, the samples contain twenty EA implementations for each of the following functions: Deceptive Order Three, Deceptive Order Four Trap, Royal Road, and Max Ones. The samples refer to the best fitness of a solution found in each run, the number of executions per algorithm is $30$. Different implementations involve classic GAs and ReGen GAs with Generational (G) and Steady State (SS) population replacements, and crossover rates from $0.6$ to $1.0$. 

The null hypothesis is a type of conjecture used in statistics that proposes that there is no difference between specific characteristics of a data-generating process. ANOVA test is being performed to evaluate the null hypothesis. The ANOVA test is an analysis of variance that is used to determine if a statistically significant difference exists in the performance of various EAs. If the given p-value for each combination of EA variations is smaller than $0.05$ (alpha value), then variances differ, such that there is a statistically significant difference between algorithms. When the null hypothesis is false, it brings up the alternative hypothesis, which proposes that there is a difference. When significant differences between groups (EAs) are found, Student’s T-test is used to interpret the result of one-way ANOVA tests. Multiple pairwise-comparison T-test helps to determine which pairs of EAs are different. The T-test concludes if the mean difference between specific pairs of EAs is statistically significant. In order to identify any significant difference in the median fitness, between two experimental conditions (classic GAs and ReGen GAs), Wilcoxon signed-rank test is performed. For the Wilcoxon test, crossover rates are ignored, and EAs are classified into four groups: GGAs vs. ReGen GGAs and SSGAs vs. ReGen SSGAs.

Based on the ReGen EAs Samples in Appendix \ref{appendB}, the analysis of variance is computed to know the difference between evolutionary algorithms with different implementations. Variations include classic GAs and ReGen GAs, replacement strategies (Generational and Steady State), and crossover rates from $0.6$ to $1.0$, algorithms are twenty in total. Table~\ref{c4table9} shows a summary for each algorithm and function. The summary presents the number of samples per algorithm ($30$), the sum of the fitness, the average fitness, and their variances. Results of the ANOVA single factor are tabulated in Table~\ref{c4table10}.

\begin{landscape}
\begin{table}[H]
\centering
\caption{Anova Single Factor: SUMMARY}
\label{c4table9}
\scriptsize
\begin{tabular}{llllllllllllll}
\hline
\multicolumn{2}{l}{} & \multicolumn{3}{c}{\textbf{Deceptive Order Three}} & \multicolumn{3}{c}{\textbf{Deceptive Order Four Trap}} & \multicolumn{3}{c}{\textbf{Royal Road}} & \multicolumn{3}{c}{\textbf{Max Ones}} \\
Groups & Count & Sum & Average & Variance & Sum & Average & Variance & Sum & Average & Variance & Sum & Average & Variance \\\hline
GGAX06 & 30 & 103036 & 3434.533333 & 119.4298851 & 11695 & 389.8333333 & 20.55747126 & 6040 & 201.3333333 & 852.2298851 & 10800 & 360 & 0 \\
GGAX07 & 30 & 102898 & 3429.933333 & 88.96091954 & 11662 & 388.7333333 & 20.54712644 & 6616 & 220.5333333 & 294.3264368 & 10800 & 360 & 0 \\
GGAX08 & 30 & 103016 & 3433.866667 & 102.3264368 & 11701 & 390.0333333 & 16.3091954 & 7384 & 246.1333333 & 210.4643678 & 10800 & 360 & 0 \\
GGAX09 & 30 & 103164 & 3438.8 & 113.2689655 & 11713 & 390.4333333 & 10.87471264 & 7880 & 262.6666667 & 490.2988506 & 10800 & 360 & 0 \\
GGAX10 & 30 & 103080 & 3436 & 83.31034483 & 11757 & 391.9 & 24.50689655 & 8504 & 283.4666667 & 325.2229885 & 10800 & 360 & 0 \\
SSGAX06 & 30 & 102898 & 3429.933333 & 107.1678161 & 11626 & 387.5333333 & 23.42988506 & 2976 & 99.2 & 364.5793103 & 10800 & 360 & 0 \\
SSGAX07 & 30 & 103046 & 3434.866667 & 78.53333333 & 11611 & 387.0333333 & 11.68850575 & 2928 & 97.6 & 288.662069 & 10800 & 360 & 0 \\
SSGAX08 & 30 & 103038 & 3434.6 & 129.9724138 & 11686 & 389.5333333 & 17.42988506 & 3304 & 110.1333333 & 395.8436782 & 10800 & 360 & 0 \\
SSGAX09 & 30 & 103020 & 3434 & 100.9655172 & 11649 & 388.3 & 25.38965517 & 3392 & 113.0666667 & 563.7885057 & 10800 & 360 & 0 \\
SSGAX10 & 30 & 103084 & 3436.133333 & 148.9471264 & 11739 & 391.3 & 16.56206897 & 3696 & 123.2 & 364.5793103 & 10800 & 360 & 0 \\
ReGenGGAX06 & 30 & 107326 & 3577.533333 & 124.6022989 & 13340 & 444.6666667 & 10.43678161 & 10720 & 357.3333333 & 19.12643678 & 10800 & 360 & 0 \\
ReGenGGAX07 & 30 & 107364 & 3578.8 & 146.3724138 & 13383 & 446.1 & 2.644827586 & 10744 & 358.1333333 & 20.67126437 & 10800 & 360 & 0 \\
ReGenGGAX08 & 30 & 107328 & 3577.6 & 200.3862069 & 13324 & 444.1333333 & 12.32643678 & 10784 & 359.4666667 & 4.11954023 & 10800 & 360 & 0 \\
ReGenGGAX09 & 30 & 107372 & 3579.066667 & 175.6505747 & 13378 & 445.9333333 & 4.616091954 & 10760 & 358.6666667 & 9.195402299 & 10800 & 360 & 0 \\
ReGenGGAX10 & 30 & 107412 & 3580.4 & 163.6965517 & 13373 & 445.7666667 & 7.840229885 & 10776 & 359.2 & 5.95862069 & 10800 & 360 & 0 \\
ReGenSSGAX06 & 30 & 107234 & 3574.466667 & 161.291954 & 13310 & 443.6666667 & 7.609195402 & 10480 & 349.3333333 & 67.67816092 & 10800 & 360 & 0 \\
ReGenSSGAX07 & 30 & 107202 & 3573.4 & 211.0758621 & 13367 & 445.5666667 & 7.21954023 & 10496 & 349.8666667 & 101.2229885 & 10800 & 360 & 0 \\
ReGenSSGAX08 & 30 & 107394 & 3579.8 & 176.3724138 & 13337 & 444.5666667 & 7.564367816 & 10504 & 350.1333333 & 135.4298851 & 10800 & 360 & 0 \\
ReGenSSGAX09 & 30 & 107486 & 3582.866667 & 119.3609195 & 13375 & 445.8333333 & 4.281609195 & 10648 & 354.9333333 & 41.85747126 & 10800 & 360 & 0 \\
ReGenSSGAX10 & 30 & 107544 & 3584.8 & 115.2 & 13351 & 445.0333333 & 7.688505747 & 10656 & 355.2 & 29.13103448 & 10800 & 360 & 0 \\\hline
\end{tabular}
\end{table}
\end{landscape}

\begin{table}[H]
\centering
\caption{Anova Single Factor: ANOVA}
\label{c4table10}
\begin{tabular}{lllllll}
\hline
\multicolumn{7}{c}{\textbf{Deceptive Order Three}} \\
Source of Variation & SS & df & MS & F & P-value & F crit \\\hline
Between Groups & 3141837.193 & 19 & 165359.8523 & 1240.0941 & 0 & 1.60449 \\
Within Groups & 77339.86667 & 580 & 133.3445977 &  &  &  \\
 &  &  &  &  &  &  \\
Total & 3219177.06 & 599 &  &  &  &  \\\hline
\multicolumn{7}{c}{\textbf{Deceptive Order Four Trap}} \\
Source of Variation & SS & df & MS & F & P-value & F crit \\\hline
Between Groups & 465618.6183 & 19 & 24506.24307 & 1888.5604 & 0 & 1.60449 \\
Within Groups & 7526.166667 & 580 & 12.97614943 &  &  &  \\
 &  &  &  &  &  &  \\
Total & 473144.785 & 599 &  &  &  &  \\\hline
\multicolumn{7}{c}{\textbf{Royal Road}} \\
Source of Variation & SS & df & MS & F & P-value & F crit \\\hline
Between Groups & 6329162.56 & 19 & 333113.8189 & 1453.2537 & 0 & 1.60449 \\
Within Groups & 132947.2 & 580 & 229.2193103 &  &  &  \\
 &  &  &  &  &  &  \\
Total & 6462109.76 & 599 &  &  &  &  \\\hline
\multicolumn{7}{c}{\textbf{Max Ones}} \\
Source of Variation & SS & df & MS & F & P-value & F crit \\\hline
Between Groups & 5.53E-25 & 19 & 2.91E-26 & 1 & 0.459 & 1.60449 \\
Within Groups & 1.69E-23 & 580 & 2.91E-26 &  &  &  \\
 &  &  &  &  &  &  \\
Total & 1.74E-23 & 599 &  &  &  & \\\hline
\end{tabular}
\end{table}


As P-values for Deceptive Order Three, Deceptive Order Four Trap, and Royal Road functions are less than the significance level $0.05$, the results allow concluding that there are significant differences between groups, as shown in Table~\ref{c4table10} ({\em P-value} columns). In one-way ANOVA tests, significant P-values indicate that some group means are different, but it is not evident which pairs of groups are different. In order to interpret one-way ANOVA test' results, multiple pairwise-comparison with Student's t-test is performed to determine if the mean difference between specific pairs of the group is statistically significant. Also, paired-sample Wilcoxon tests are computed.

ANOVA test for Max Ones' samples shows that the P-value is higher than the significance level 0.05, this result means that there are no significant differences between algorithms (EAs) listed above in the model summary Table~\ref{c4table9}. Therefore, no multiple pairwise-comparison Student's t-tests between means of groups are performed; neither, paired-sample Wilcoxon test is computed.

\begin{landscape}
\begin{figure}[H]
\centering
\includegraphics[width=4.2in]{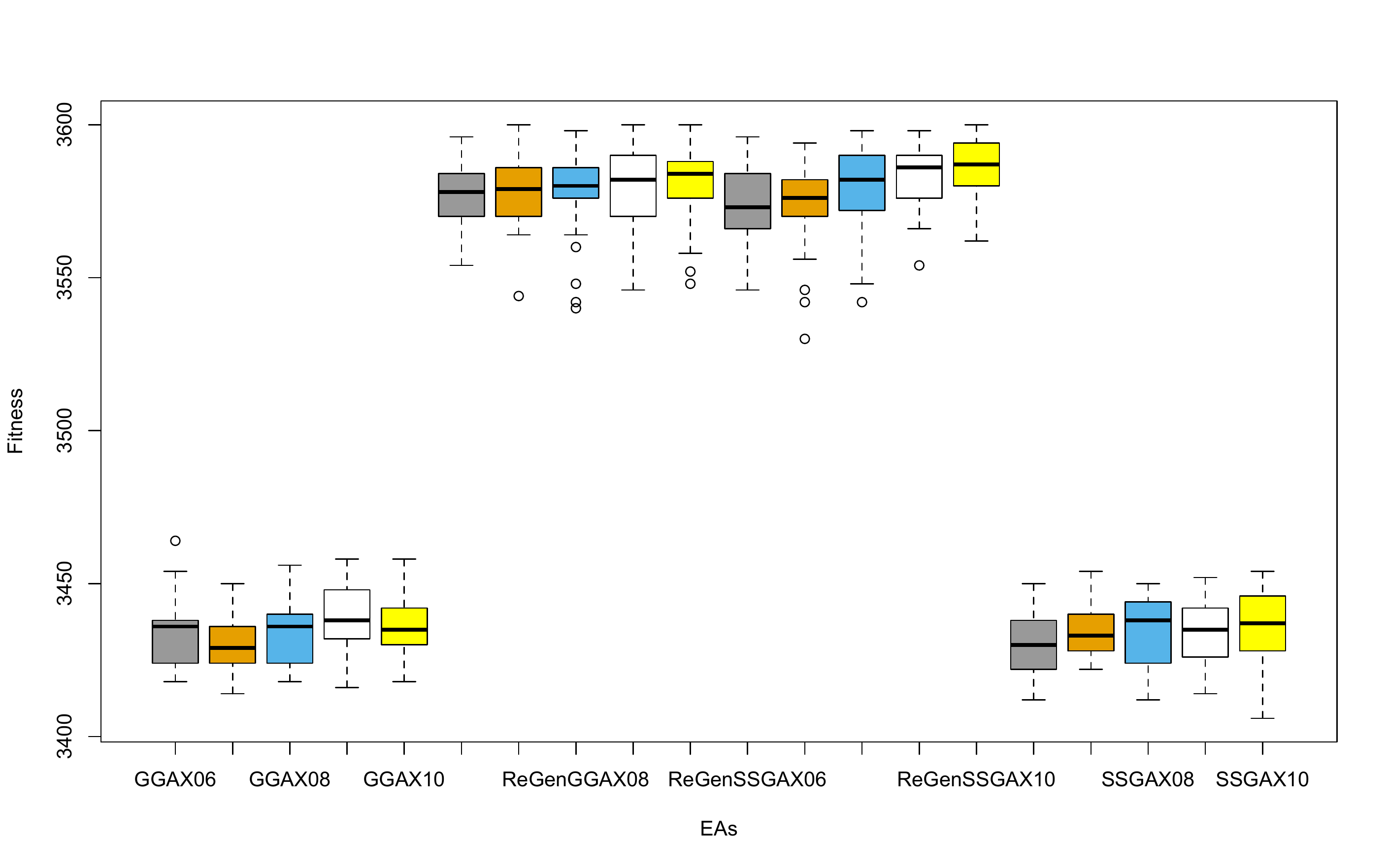}
\includegraphics[width=4.2in]{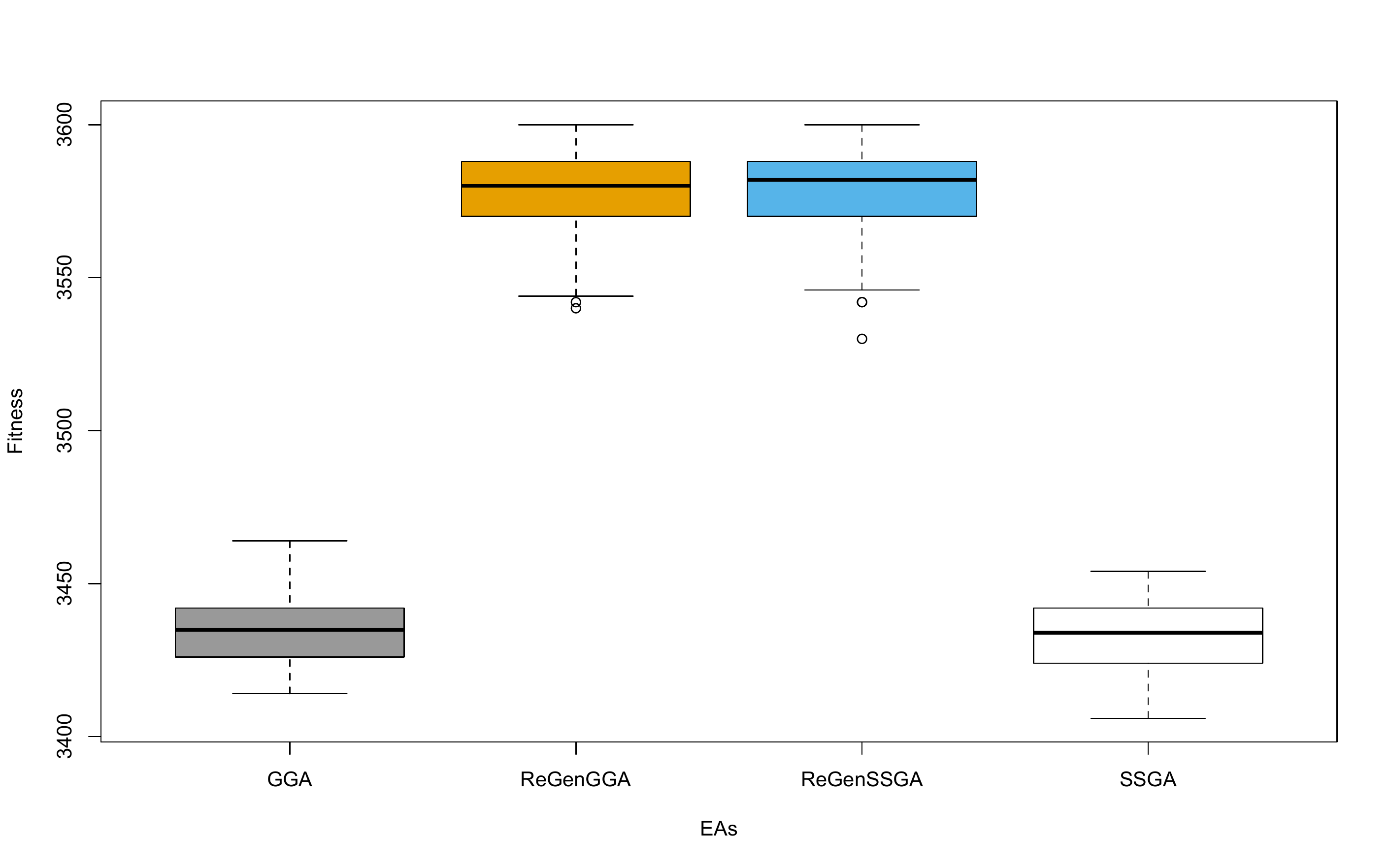}
\includegraphics[width=4.2in]{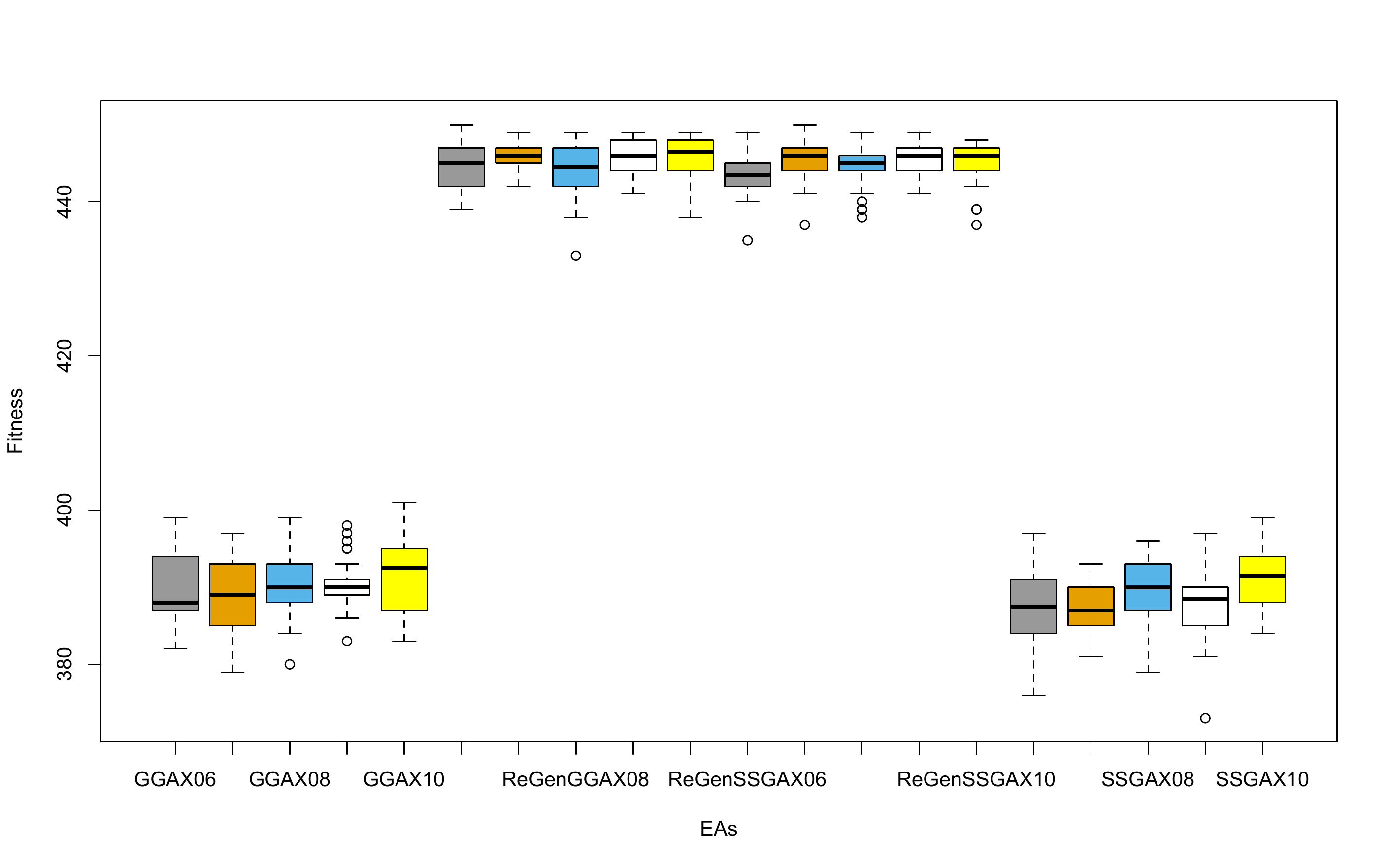}
\includegraphics[width=4.2in]{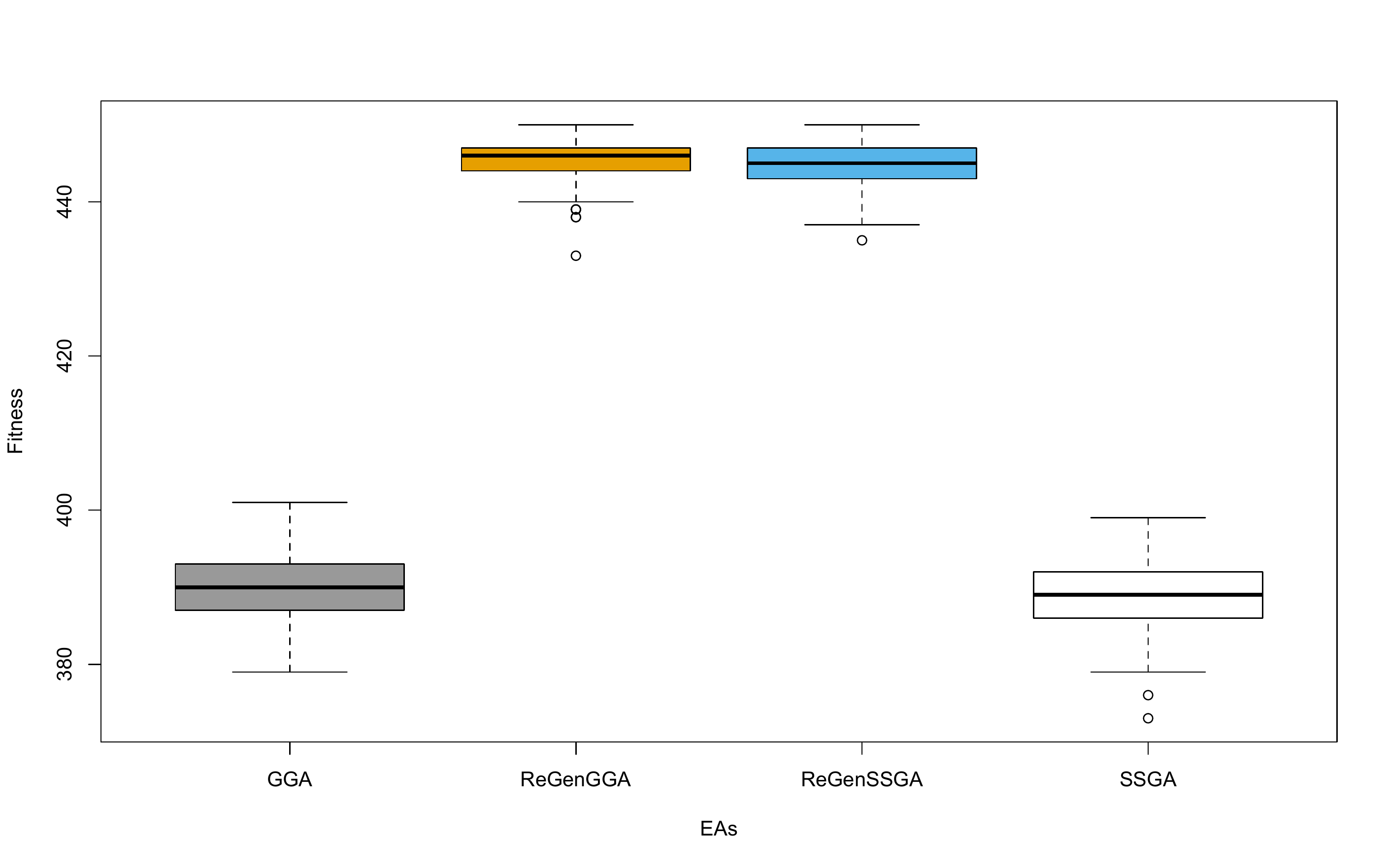}
\caption{From top to bottom: Deceptive Order Three and Deceptive Order Four Trap Functions. On the left, EAs with Generational replacement (GGA) and Steady State replacement (SSGA) with Crossover rates from $0.6$ to $1.0$. On the right, EAs grouped by Generational replacement (GGA) and Steady State replacement (SSGA).}
\label{c4fig5}
\end{figure}
\end{landscape}

\begin{figure}[H]
\centering
\includegraphics[width=4.2in]{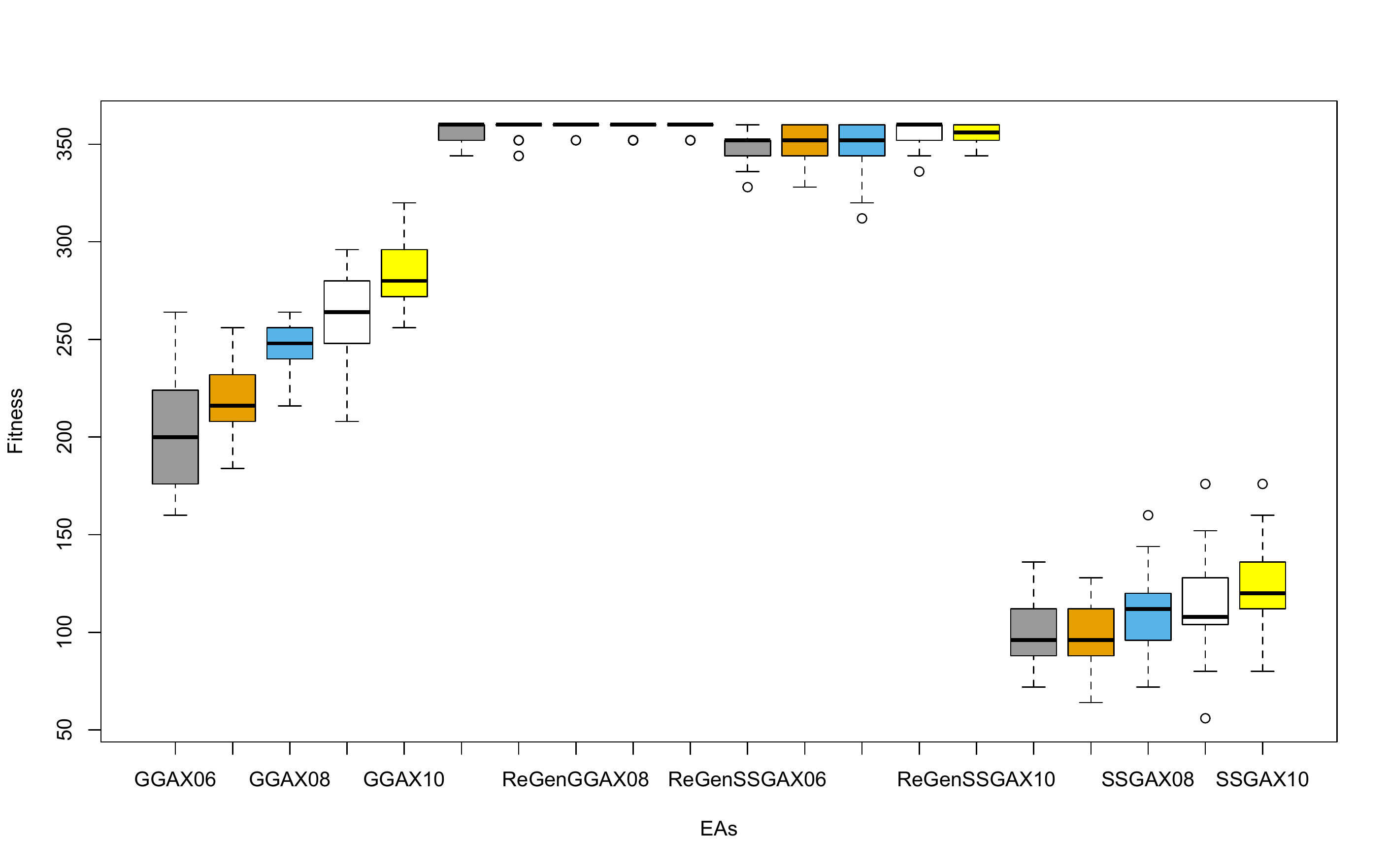}
\includegraphics[width=4.2in]{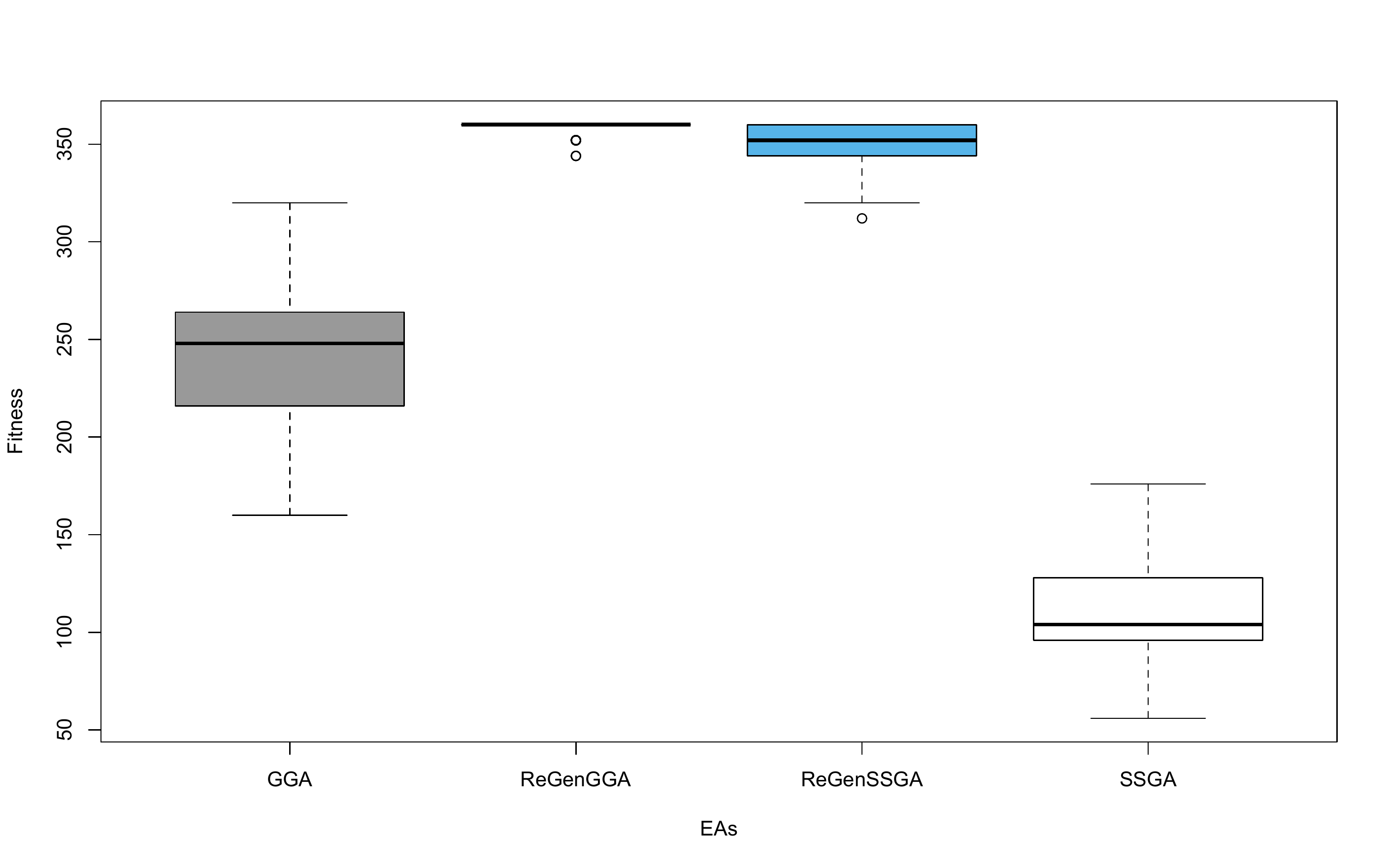}
\caption{Royal Road Function. On top, EAs with Generational (GGA) and Steady State (SSGA) replacements with Crossover rates from $0.6$ to $1.0$. On the bottom, EAs grouped by Generational replacement (GGA) and Steady State replacement (SSGA).}
\label{c4fig6}
\end{figure}

Box plots in Fig.~\ref{c4fig5} and Fig.~\ref{c4fig6} depict the median fitness of EAs' best solutions (ReGen EAs Samples in Appendix \ref{appendB}). On the left, twenty EAs' variations with different crossover rates: Gray ($0.6$), Orange ($0.7$), Blue ($0.8$), White ($0.9$), and Yellow ($1.0$). On the right, figures illustrate the median fitness of classic and epigenetic EAs, which are grouped by population replacement type: Gray (GGA), Orange (ReGen GGA), Blue (ReGen SSGA), and White (SSGA). For Deceptive Order Three function, the median fitness for each Epigenetic EA is close to the global optimum ($3600$), while the median fitnesses for classic GAs are under the local optimum ($3450$). On the other hand, Deceptive Order Four Trap median fitness is above $440$ for all Epigenetic implementations; in contrast, for classic GAs, the median fitness does not exceed $400$. The same occurs for Royal Road function; the median fitness reported for epigenetic evolutionary algorithms outpoints local optimum ($320$), while traditional GAs median fitness maximum value is $320$. So, based on these data, it seems that Epigenetic GAs find better solutions than classic GAs. However, it is needed to determine whether this finding is statistically significant.

\begin{landscape}
\begin{table}[H]
\centering
\caption{D3 Student T-tests pairwise comparisons with pooled standard deviation. Benjamini Hochberg (BH) as p-value adjustment method.}
\label{c4table11}
\tiny
\begin{tabular}{llllllllllllllllllll}
\hline
EAs & GGAX06 & GGAX07 & GGAX08 & GGAX09 & GGAX10 & ReGenGGAX06 & ReGenGGAX07 & ReGenGGAX08 & ReGenGGAX09 & ReGenGGAX10 \\\hline
GGAX07 & 0.17764919 & - & - & - & - & - & - & - & - & - \\
GGAX08 & 0.87373395 & 0.24754713 & - & - & - & - & - & - & - & - \\
GGAX09 & 0.21528009 & \cellcolor[HTML]{EFEFEF}0.00554289 & 0.14859587 & - & - & - & - & - & - & - \\
GGAX10 & 0.71303353 & 0.0705529 & 0.57069862 & 0.43796354 & - & - & - & - & - & - \\
ReGenGGAX06 & \cellcolor[HTML]{EFEFEF}1.48E-203 & \cellcolor[HTML]{EFEFEF}1.40E-209 & \cellcolor[HTML]{EFEFEF}1.94E-204 & \cellcolor[HTML]{EFEFEF}1.24E-197 & \cellcolor[HTML]{EFEFEF}1.54E-201 & - & - & - & - & - \\
ReGenGGAX07 & \cellcolor[HTML]{EFEFEF}3.23E-205 & \cellcolor[HTML]{EFEFEF}3.71E-211 & \cellcolor[HTML]{EFEFEF}4.47E-206 & \cellcolor[HTML]{EFEFEF}2.13E-199 & \cellcolor[HTML]{EFEFEF}2.72E-203 & 0.75006919 & - & - & - & - \\
ReGenGGAX08 & \cellcolor[HTML]{EFEFEF}1.23E-203 & \cellcolor[HTML]{EFEFEF}1.22E-209 & \cellcolor[HTML]{EFEFEF}1.62E-204 & \cellcolor[HTML]{EFEFEF}1.01E-197 & \cellcolor[HTML]{EFEFEF}1.26E-201 & 0.98736535 & 0.76386956 & - & - & - \\
\hline
\end{tabular}
\end{table}

\begin{table}[H]
\centering
\caption{D3 Student T-tests pairwise comparisons with pooled standard deviation. Benjamini Hochberg (BH) as p-value adjustment method.}
\label{c4table12}
\tiny
\begin{tabular}{lllllllllllllllllll}
\hline
 EAs & ReGenSSGAX06 & ReGenSSGAX07 & ReGenSSGAX08 & ReGenSSGAX09 & ReGenSSGAX10 & SSGAX06 & SSGAX07 & SSGAX08 & SSGAX09 \\\hline
GGAX07 & - & - & - & - & - & - & - & - & - \\
GGAX08 & - & - & - & - & - & - & - & - & - \\
GGAX09 & - & - & - & - & - & - & - & - & - \\
GGAX10 & - & - & - & - & - & - & - & - & - \\
ReGenGGAX06 & - & - & - & - & - & - & - & - & - \\
ReGenGGAX07 & - & - & - & - & - & - & - & - & - \\
ReGenGGAX08 & - & - & - & - & - & - & - & - & - \\
\hline
\end{tabular}
\end{table}

\begin{table}[H]
\centering
\caption{D4 Student T-tests pairwise comparisons with pooled standard deviation. Benjamini Hochberg (BH) as p-value adjustment method.}
\label{c4table13}
\tiny
\begin{tabular}{llllllllllllllllllll}
\hline
EAs & GGAX06 & GGAX07 & GGAX08 & GGAX09 & GGAX10 & ReGenGGAX06 & ReGenGGAX07 & ReGenGGAX08 & ReGenGGAX09 & ReGenGGAX10 \\\hline
GGAX07 & 0.29677823 & - & - & - & - & - & - & - & - & - \\
GGAX08 & 0.85224562 & 0.21774467 & - & - & - & - & - & - & - & - \\
GGAX09 & 0.59061606 & 0.09801927 & 0.71632051 & - & - & - & - & - & - & - \\
GGAX10 & \cellcolor[HTML]{EFEFEF}0.04153318 & \cellcolor[HTML]{EFEFEF}0.00125756 & 0.06818451 & 0.15999358 & - & - & - & - & - & - \\
ReGenGGAX06 & \cellcolor[HTML]{EFEFEF}1.17E-246 & \cellcolor[HTML]{EFEFEF}8.61E-251 & \cellcolor[HTML]{EFEFEF}6.80E-246 & \cellcolor[HTML]{EFEFEF}2.34E-244 & \cellcolor[HTML]{EFEFEF}1.43E-238 & - & - & - & - & - \\
ReGenGGAX07 & \cellcolor[HTML]{EFEFEF}5.36E-252 & \cellcolor[HTML]{EFEFEF}6.26E-256 & \cellcolor[HTML]{EFEFEF}2.87E-251 & \cellcolor[HTML]{EFEFEF}8.11E-250 & \cellcolor[HTML]{EFEFEF}3.13E-244 & 0.16928761 & - & - & - & - \\
ReGenGGAX08 & \cellcolor[HTML]{EFEFEF}1.31E-244 & \cellcolor[HTML]{EFEFEF}8.13E-249 & \cellcolor[HTML]{EFEFEF}7.53E-244 & \cellcolor[HTML]{EFEFEF}2.76E-242 & \cellcolor[HTML]{EFEFEF}2.03E-236 & 0.63324079 & 0.05347637 & - & - & - \\
\hline
\end{tabular}
\end{table}

\begin{table}[H]
\centering
\caption{D4 Student T-tests pairwise comparisons with pooled standard deviation. Benjamini Hochberg (BH) as p-value adjustment method.}
\label{c4table14}
\tiny
\begin{tabular}{lllllllllllllllllll}
\hline
EAs & ReGenSSGAX06 & ReGenSSGAX07 & ReGenSSGAX08 & ReGenSSGAX09 & ReGenSSGAX10 & SSGAX06 & SSGAX07 & SSGAX08 & SSGAX09 \\\hline
GGAX07 & - & - & - & - & - & - & - & - & - \\
GGAX08 & - & - & - & - & - & - & - & - & - \\
GGAX09 & - & - & - & - & - & - & - & - & - \\
GGAX10 & - & - & - & - & - & - & - & - & - \\
ReGenGGAX06 & - & - & - & - & - & - & - & - & - \\
ReGenGGAX07 & - & - & - & - & - & - & - & - & - \\
ReGenGGAX08 & - & - & - & - & - & - & - & - & - \\
\hline
\end{tabular}
\end{table}

\begin{table}[H]
\centering
\caption{RR Student T-tests pairwise comparisons with pooled standard deviation. Benjamini Hochberg (BH) as p-value adjustment method.}
\label{c4table15}
\tiny
\begin{tabular}{llllllllllllllllllll}
\hline
EAs & GGAX06 & GGAX07 & GGAX08 & GGAX09 & GGAX10 & ReGenGGAX06 & ReGenGGAX07 & ReGenGGAX08 & ReGenGGAX09 & ReGenGGAX10 \\\hline
GGAX07 & \cellcolor[HTML]{EFEFEF}1.64E-06 & - & - & - & - & - & - & - & - & - \\
GGAX08 & \cellcolor[HTML]{EFEFEF}2.21E-27 & \cellcolor[HTML]{EFEFEF}1.83E-10 & - & - & - & - & - & - & - & - \\
GGAX09 & \cellcolor[HTML]{EFEFEF}2.49E-46 & \cellcolor[HTML]{EFEFEF}1.19E-24 & \cellcolor[HTML]{EFEFEF}3.78E-05 & - & - & - & - & - & - & - \\
GGAX10 & \cellcolor[HTML]{EFEFEF}4.37E-73 & \cellcolor[HTML]{EFEFEF}2.66E-48 & \cellcolor[HTML]{EFEFEF}5.20E-20 & \cellcolor[HTML]{EFEFEF}2.08E-07 & - & - & - & - & - & - \\
ReGenGGAX06 & \cellcolor[HTML]{EFEFEF}5.59E-168 & \cellcolor[HTML]{EFEFEF}1.09E-144 & \cellcolor[HTML]{EFEFEF}8.81E-112 & \cellcolor[HTML]{EFEFEF}8.33E-90 & \cellcolor[HTML]{EFEFEF}3.21E-62 & - & - & - & - & - \\
ReGenGGAX07 & \cellcolor[HTML]{EFEFEF}6.43E-169 & \cellcolor[HTML]{EFEFEF}1.10E-145 & \cellcolor[HTML]{EFEFEF}7.83E-113 & \cellcolor[HTML]{EFEFEF}7.12E-91 & \cellcolor[HTML]{EFEFEF}2.93E-63 & 0.86524143 & - & - & - & - \\
ReGenGGAX08 & \cellcolor[HTML]{EFEFEF}1.81E-170 & \cellcolor[HTML]{EFEFEF}2.51E-147 & \cellcolor[HTML]{EFEFEF}1.42E-114 & \cellcolor[HTML]{EFEFEF}1.20E-92 & \cellcolor[HTML]{EFEFEF}5.40E-65 & 0.6320313 & 0.7738968 & - & - & - \\
\hline
\end{tabular}
\end{table}

\begin{table}[H]
\centering
\caption{RR Student T-tests pairwise comparisons with pooled standard deviation. Benjamini Hochberg (BH) as p-value adjustment method.}
\label{c4table16}
\tiny
\begin{tabular}{lllllllllllllllllll}
\hline
EAs & ReGenSSGAX06 & ReGenSSGAX07 & ReGenSSGAX08 & ReGenSSGAX09 & ReGenSSGAX10 & SSGAX06 & SSGAX07 & SSGAX08 & SSGAX09 \\\hline
GGAX07 & - & - & - & - & - & - & - & - & - \\
GGAX08 & - & - & - & - & - & - & - & - & - \\
GGAX09 & - & - & - & - & - & - & - & - & - \\
GGAX10 & - & - & - & - & - & - & - & - & - \\
ReGenGGAX06 & - & - & - & - & - & - & - & - & - \\
ReGenGGAX07 & - & - & - & - & - & - & - & - & - \\
ReGenGGAX08 & - & - & - & - & - & - & - & - & - \\
\hline
\end{tabular}
\end{table}

\end{landscape}





\paragraph{\em{Multiple pairwise t-test:}}
Multiple pairwise-comparison between means of EA groups is performed. In the one-way ANOVA test described above, significant p-values indicate that some group means are different. In order to know which pairs of groups are different, multiple pairwise-comparison is performed for Deceptive Order Three (D3), Deceptive Order Four Trap (D4), and Royal Road (RR) best solutions samples. Tables (\ref{c4table11}, \ref{c4table12}, \ref{c4table13}, \ref{c4table14}, \ref{c4table15}, and \ref{c4table16}) present Pairwise comparisons using t-tests with pooled standard deviation (SD) with their respective p-values. The test adjusts p-values with the Benjamini-Hochberg method. Pairwise comparisons show that only highlighted values in gray between two algorithms are significantly different ($p < 0.05$). Therefore, the alternative hypothesis is true.

Now, to find out any significant difference between the median fitness of individuals in the two experimental groups (classic GAs and GAs with regulated genes), the Wilcoxon test is conducted.

\paragraph{\em{Paired Samples Wilcoxon Test:}}  
For this test, algorithms are grouped per population replacement strategy, without taking into account the crossover rates. Wilcoxon signed rank test for generational EAs (GGA and ReGen GGA) and Wilcoxon signed rank test for steady state EAs (SSGA and ReGen SSGA). The test assesses classic EAs versus Epigenetic EAs. 

\begin{itemize}
    \item Deceptive Order Three (D3)
    \begin{enumerate}
      \item \par Wilcoxon signed rank test with continuity correction for generational EAs uses all data-set samples from GGAs and ReGen GGAs. The {\em P-value} is equal to $2.256122e-26$, which is less than the significance level alpha ($0.05$).
      
      \item \par Wilcoxon signed rank test with continuity correction for steady state EAs uses all data-set samples from SSGAs and ReGen SSGAs. {\em P-value} is equal to $2.250642e-26$, which is less than the significance level $alpha = 0.05$.
      \end{enumerate}
      
    \item Deceptive Order Four Trap (D4)
    \begin{enumerate}
      \item \par Wilcoxon signed rank test with continuity correction for generational EAs uses all data-set samples from GGAs and ReGen GGAs. The P-value is equal to $2.163978e-26$, which is less than the significance level alpha ($0.05$).
      
      \item \par Wilcoxon signed rank test with continuity correction for steady state EAs uses all data-set samples from SSGAs and ReGen SSGAs. {\em P-value} is equal to $2.217806e-26$, which is less than the significance level $alpha = 0.05$.
      \end{enumerate}
    \newpage
    \item Royal Road (RR)
    \begin{enumerate}
      \item \par Wilcoxon signed rank test with continuity correction for generational EAs uses all data-set samples from GGAs and ReGen GGAs. The P-value is equal to $2.135633e-26$, which is less than the significance level $alpha = 0.05$.
      
      \item \par Wilcoxon signed rank test with continuity correction for steady state EAs uses all data-set samples from SSGAs and ReGen SSGAs. {\em P-value} is equal to $1.948245e-26$, which is less than the significance level alpha ($0.05$). 
      \end{enumerate}
      
\end{itemize}

The above leads to conclude that median fitnesses of solutions found by classic generational genetic algorithms (GGAs) are significantly different from median fitnesses of solutions found by generational genetic algorithms with regulated genes (ReGen GGAs) with p-values equal to $2.256122e-26$ (D3 samples), $2.163978e-26$ (D4 samples), and $2.135633e-26$ (RR samples). So, the alternative hypothesis is true.

The median fitness of solutions found by classic steady state genetic algorithms (SSGAs) is significantly different from the median fitness of solutions found by steady state genetic algorithms with regulated genes (ReGen SSGAs) with p-values equal to $2.250642e-26$ (D3 sampling fitness), $217806e-26$ (D4 sampling fitness), and $1.948245e-26$ (RR sampling fitness). As p-values are less than the significance level $0.05$, it may be concluded that there are significant differences between the two EAs groups in each Wilcoxon Test. 



\newpage
\section{Real Problems}\label{c4s3}

The real problems have been encoded as binary strings. The individuals are initialized with randomized binary strings of $(d \cdot n)$, where $d$ is the number of dimensions of the problem and $n$ the length in bits of the binary representation for a real value. The process to obtain real values from binary strings of $32$ bits is done by taking its representation as an integer number and then applying a decoding function. Equation \ref{c4eq3} and Equation \ref{c4eq4} define the encoding/decoding schema \cite{BODENHOFER}.

In the general form for an arbitrary interval [a, b] the coding function is defined as:

\begin{align} 
 C_n,\left[a, b\right] : \left[a, b\right] 
 &\longrightarrow\lbrace0,1\rbrace^{n} {\nonumber}\\
 x &\longmapsto bin_n\left(round\left((2^{n}-1)\cdot \frac{x-a}{b-a}\right)\right)
\label{c4eq3}
\end{align}


where $bin_n$ is the function which converts a number from $\lbrace0,..., 2^{n}-1\rbrace$ to its binary representation of length $n$ \cite{BODENHOFER}. The corresponding decoding function is defined as follows:

\begin{align} 
\widetilde C_n,[a,b]: \{0,1\}^{n} & \longrightarrow [a, b] {\nonumber} \\
s & \longmapsto a + bin_{n}^{-1}(s)\cdot \frac{b-a}{2^{n}-1}
\label{c4eq4}
\end{align}

Now, applying the above decoding function to the interval [$-5.12, 5.11$] with $n = 32$, where the total size of the search space is $2^{32} = 4.294.967.296$, that is $\lbrace0,..., 4294967295\rbrace$, the $32$ bits string is equal to {\em11111111111111111111111111111111}, and the bit string representation as integer number is $4294967295$. The decoding function yields:

\begin{eqnarray*}
s \longmapsto -5.12 + 4294967295 \cdot \frac{5.11 - (-5.12)}{4294967295} = 5.11
\label{c4eq5}
\end{eqnarray*}

\subsection{Experiments}\label{c4s3ss1}

Experiments using real definition are performed to determine the proposed technique applicability. For the selected problems with real definition, a vector with binary values encodes the problem's solution. The real functions explained in this section are used as testbeds. For all functions, the problem dimension is fixed to $n = 10$; each real value is represented with a binary of $32$-bits.

\subsubsection{Rastrigin}
The Rastrigin function has several local minima, it is highly multimodal, but locations of the minima are regularly distributed. Among its features: the function is continuous, convex, defined on n-dimensional space, multimodal, differentiable, and separable. The function is usually evaluated on the hypercube $x_i \in [-5.12, 5.12]$ for $i = 1, ..., n$. The global minimum $f(\textbf{x}^{\ast}) = 0$ at $\textbf{x}^{\ast} = (0, ..., 0)$ \cite{BENCHMARKS, BENCHMARKS2}. On an n-dimensional domain, it is defined by Equation \ref{c4eq6} as:

\begin{eqnarray}
f(x, y)=10n + \sum_{i=1}^{n}(x_i^2 - 10cos(2\pi x_i))
\label{c4eq6}
\end{eqnarray}

\subsubsection{Rosenbrock}
The Rosenbrock function, also referred to as the Valley or Banana function, is a popular test problem for gradient-based optimization algorithms. Among its features: the function is continuous, convex, defined on n-dimensional space, multimodal, differentiable, and non-separable. The function is usually evaluated on the hypercube $x_i \in [-5, 10]$ for $i = 1, ..., n$, although it may be restricted to the hypercube $x_i \in [-2.048, 2.048]$ for $i = 1, ..., n$. The global minimum $f(\textbf{x}^{\ast}) = 0$ at $\textbf{x}^{\ast} = (1, ..., 1)$. In Equation \ref{c4eq7}, the parameters \em{a} and \em{b} are constants and are generally set to $a = 1$ and $b = 100$ \cite{BENCHMARKS, BENCHMARKS2}. On an n-dimensional domain, it is defined by:

\begin{eqnarray}
f(x, y)=\sum_{i=1}^{n}[b (x_{i+1} - x_i^2)^ 2 + (a - x_i)^2]
\label{c4eq7}
\end{eqnarray}

\subsubsection{Schwefel}
The Schwefel function is complex, with many local minima. Among its features: the function is continuous, not convex, multimodal, and can be defined on n-dimensional space. The function can be defined on any input domain but it is usually evaluated on the hypercube $x_i \in [-500, 500]$ for $i = 1, ..., n$. The global minimum $f(\textbf{x}^{\ast}) = 0$ at $\textbf{x}^{\ast} = (420.9687, ..., 420.9687)$ \cite{BENCHMARKS, BENCHMARKS2}. On an n-dimensional domain, it is defined by Equation \ref{c4eq8} as:

\begin{eqnarray}
f(\textbf{x}) = f(x_1, x_2, ..., x_n) = 418.9829n -{\sum_{i=1}^{n} x_i sin\left(\sqrt{|x_i|}\right)}
\label{c4eq8}
\end{eqnarray}

\subsubsection{Griewank}
The Griewank function has many widespread local minima, which are regularly distributed. Among its features: this function is continuous, not convex, can be defined on n-dimensional space, and is unimodal. This function can be defined on any input domain but it is usually evaluated on $x_i \in [-600, 600]$ for $i = 1, ..., n$. The global minimum $f(\textbf{x}^{\ast}) = 0$ at $\textbf{x}^{\ast} = (0, ..., 0)$ \cite{BENCHMARKS, BENCHMARKS2}. On an n-dimensional domain, it is defined by Equation \ref{c4eq9} as:

\begin{eqnarray}
f(\textbf{x}) = f(x_1, ..., x_n) = 1 + \sum_{i=1}^{n} \frac{x_i^{2}}{4000} - \prod_{i=1}^{n}cos\left(\frac{x_i}{\sqrt{i}}\right)
\label{c4eq9}
\end{eqnarray}

\subsection{Results}\label{c4s3ss2}

Based on the defined configuration, both classic and ReGen GA are compared to identify the tags' behavior during individuals' evolution. Results are tabulated from Table~\ref{c4table17} to Table~\ref{c4table20}, these tables present real defined functions: Rastrigin (RAS), Rosenbrock (ROSE), Schwefel (SCHW), and Griewank (GRIE). Both EA implementations with generational (GGA) and steady state (SSGA) replacements, and five crossover rates per technique. For each rate, the best fitness based on the minimum median performance is reported, following the standard deviation of the observed value, and the iteration where the reported fitness is found. The latter is enclosed in square brackets.

\begin{figure}[H]
\centering
\includegraphics[width=2in]{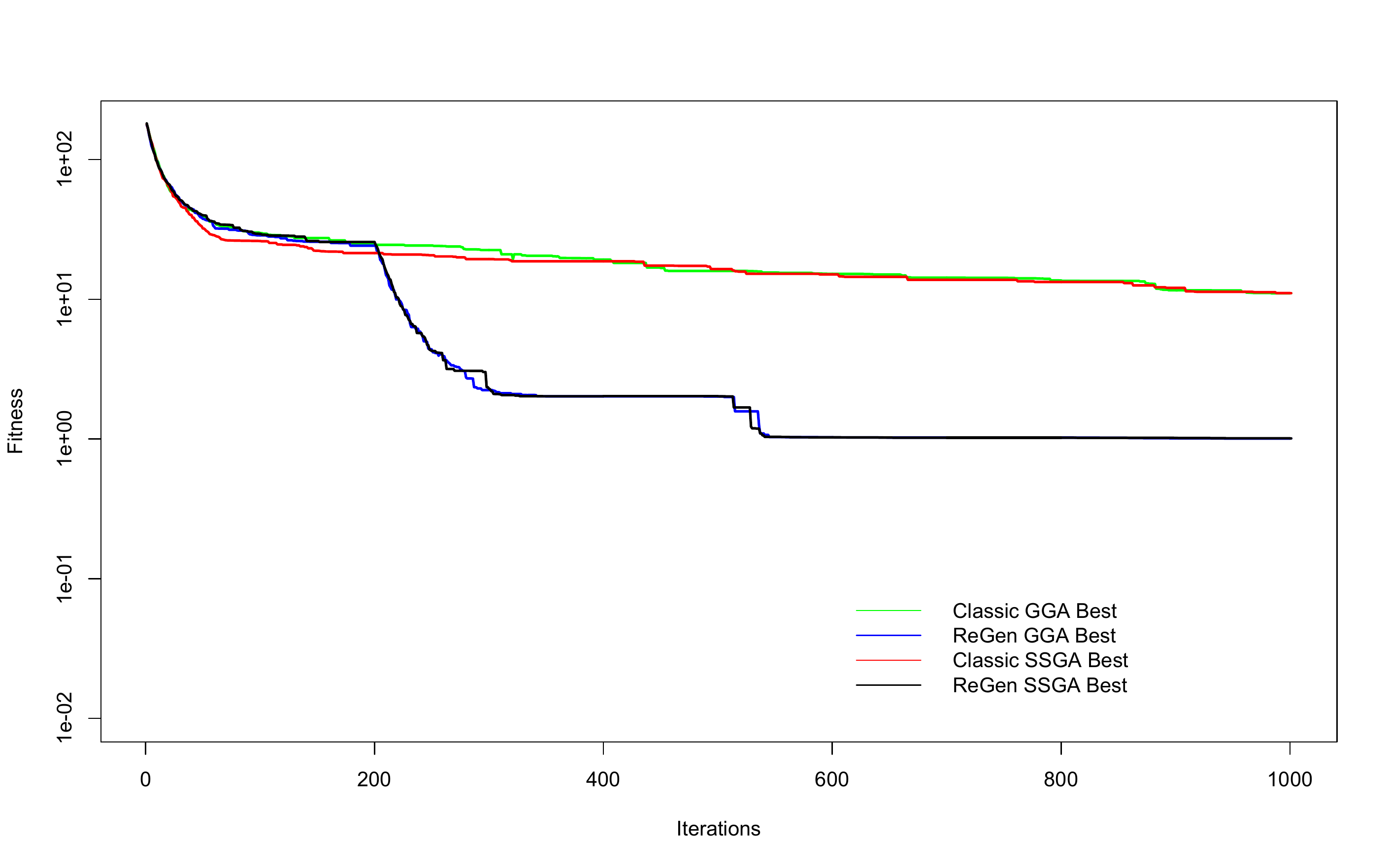}
\includegraphics[width=2in]{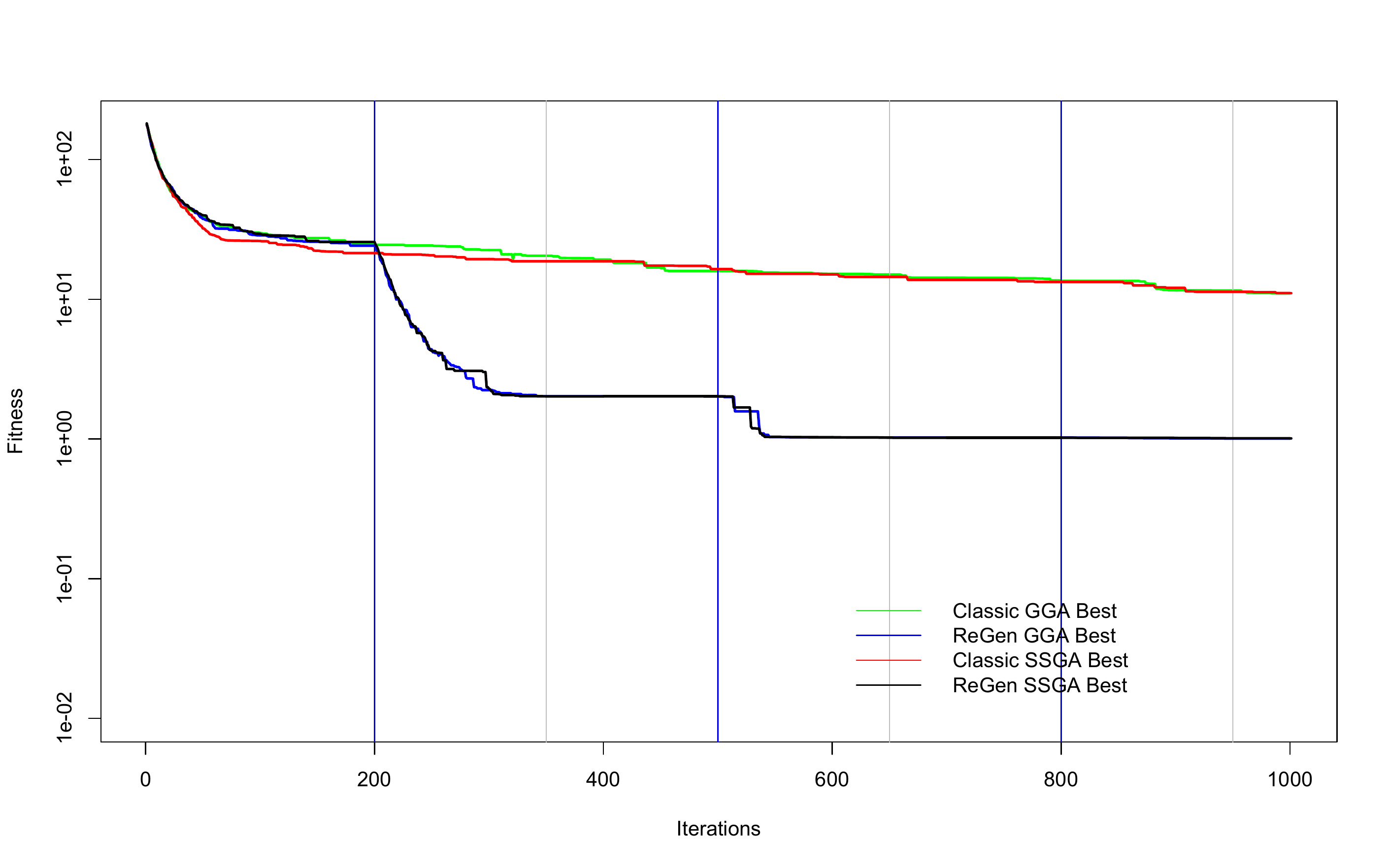}
\includegraphics[width=2in]{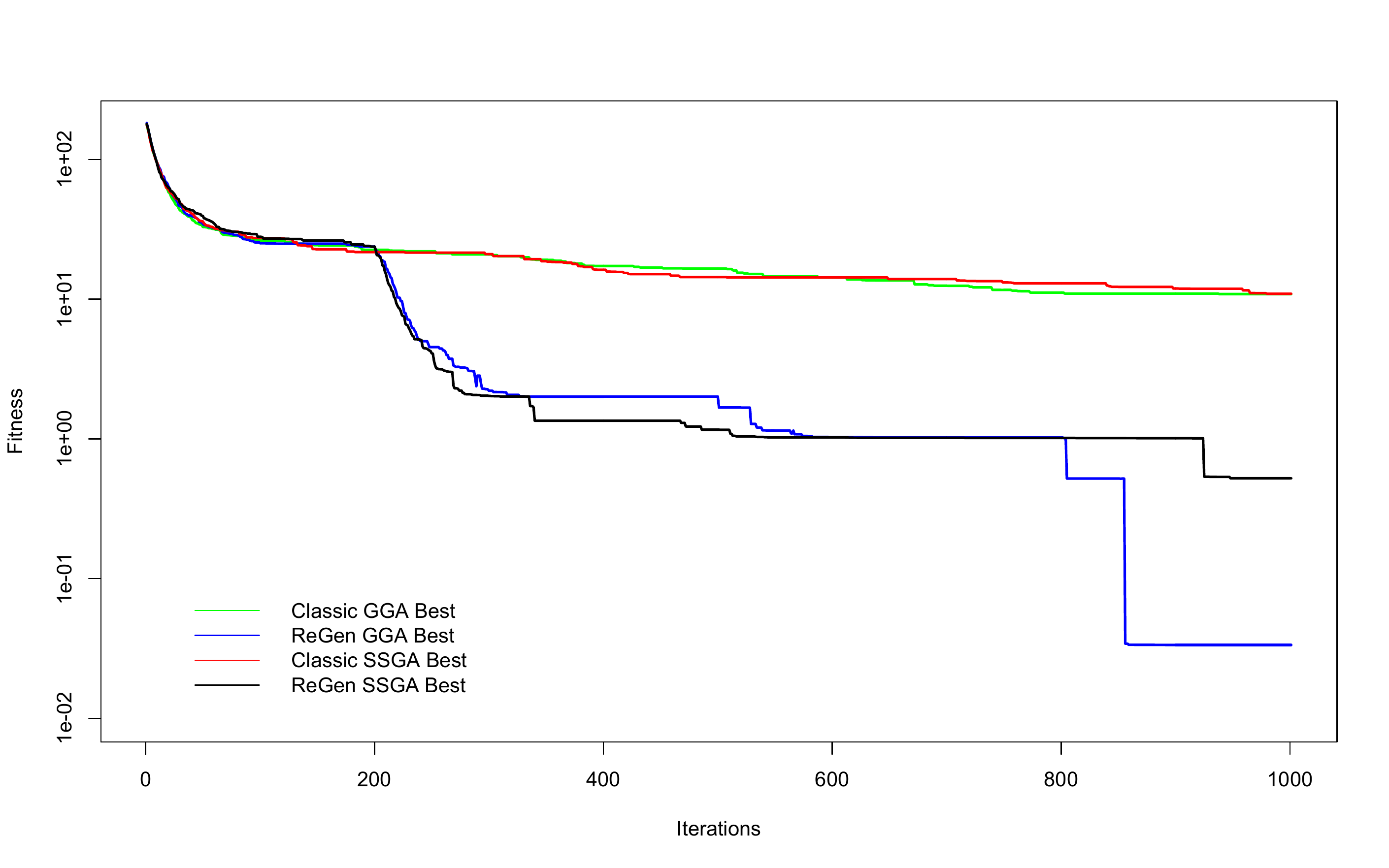}
\includegraphics[width=2in]{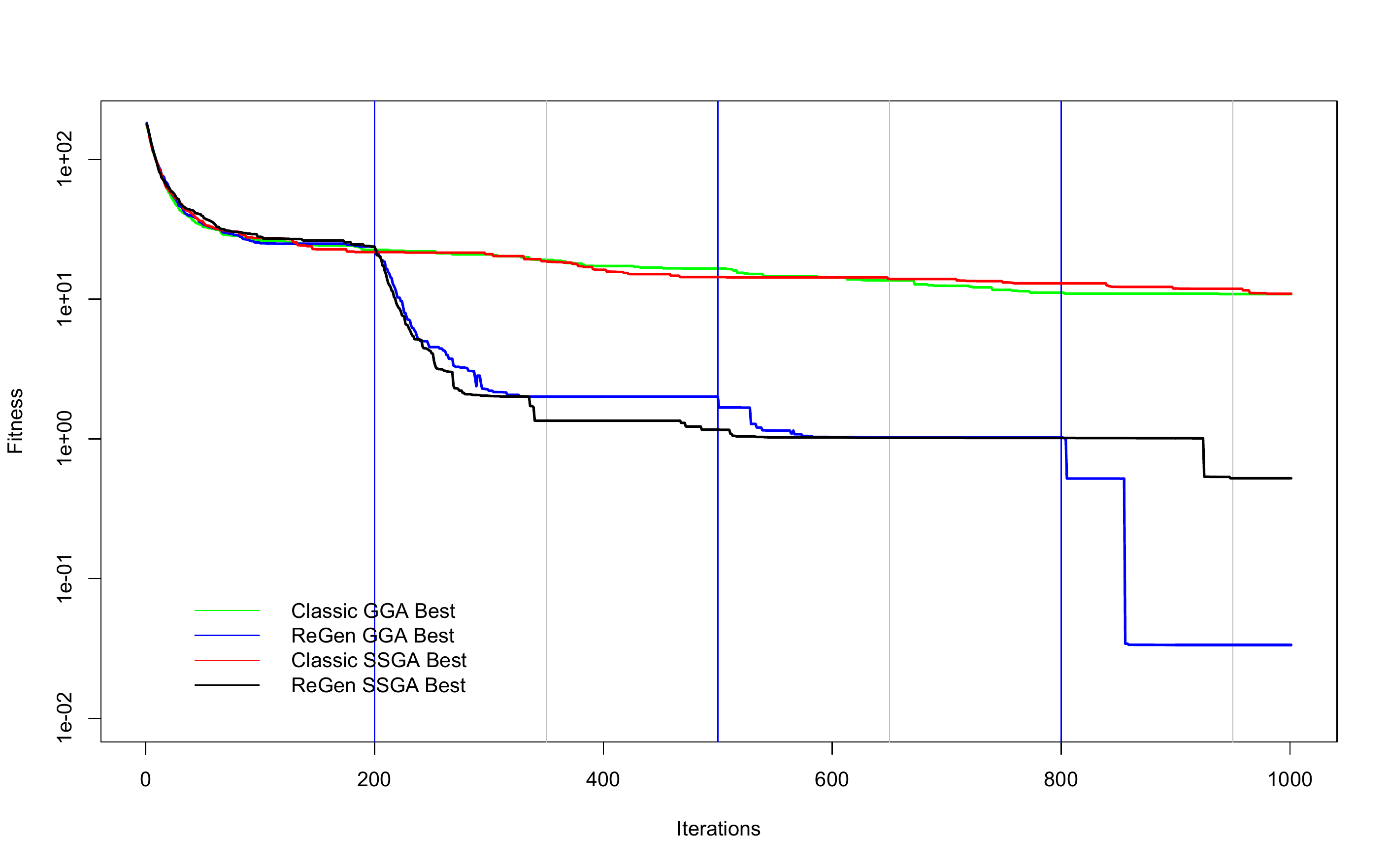}
\includegraphics[width=2in]{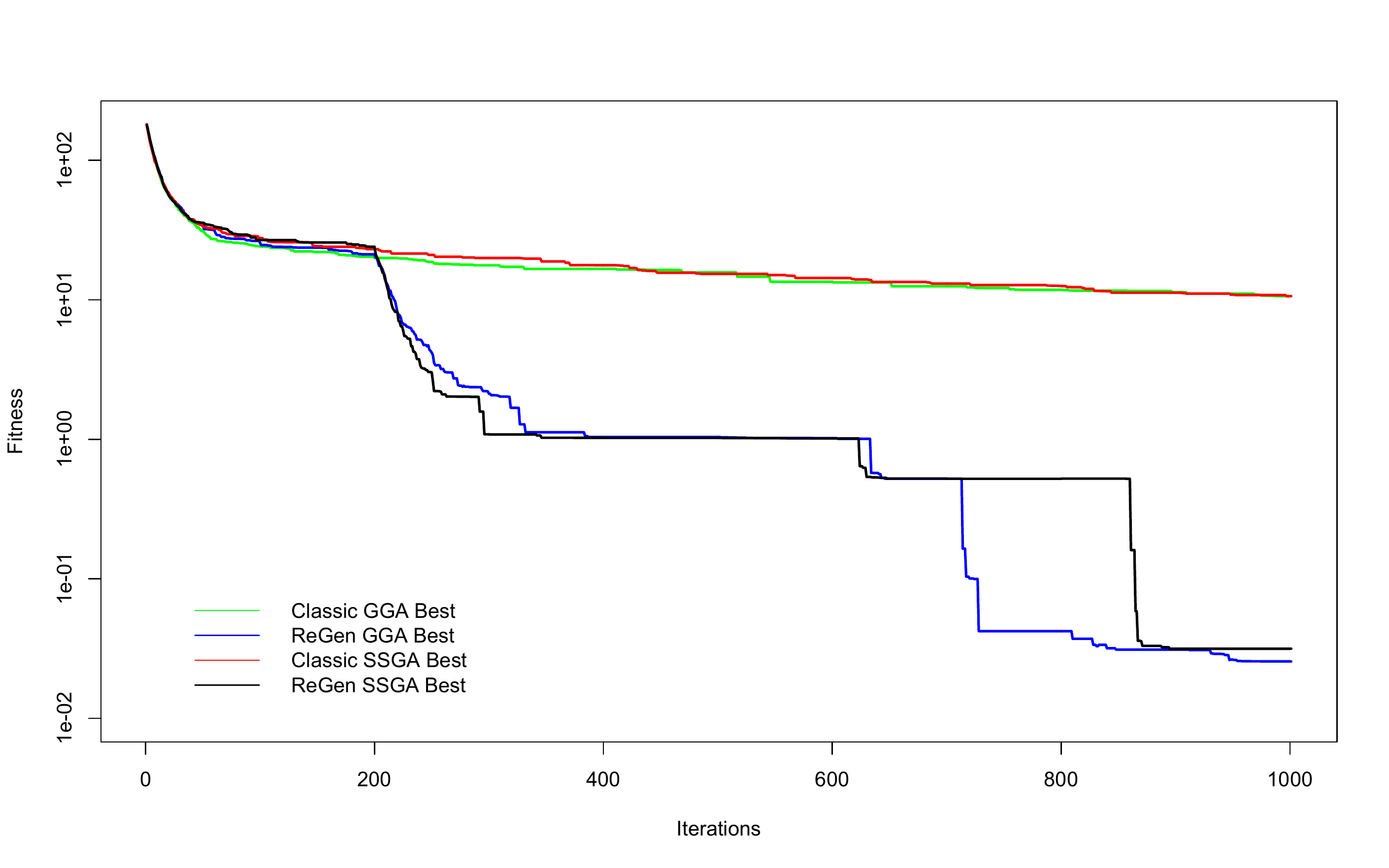}
\includegraphics[width=2in]{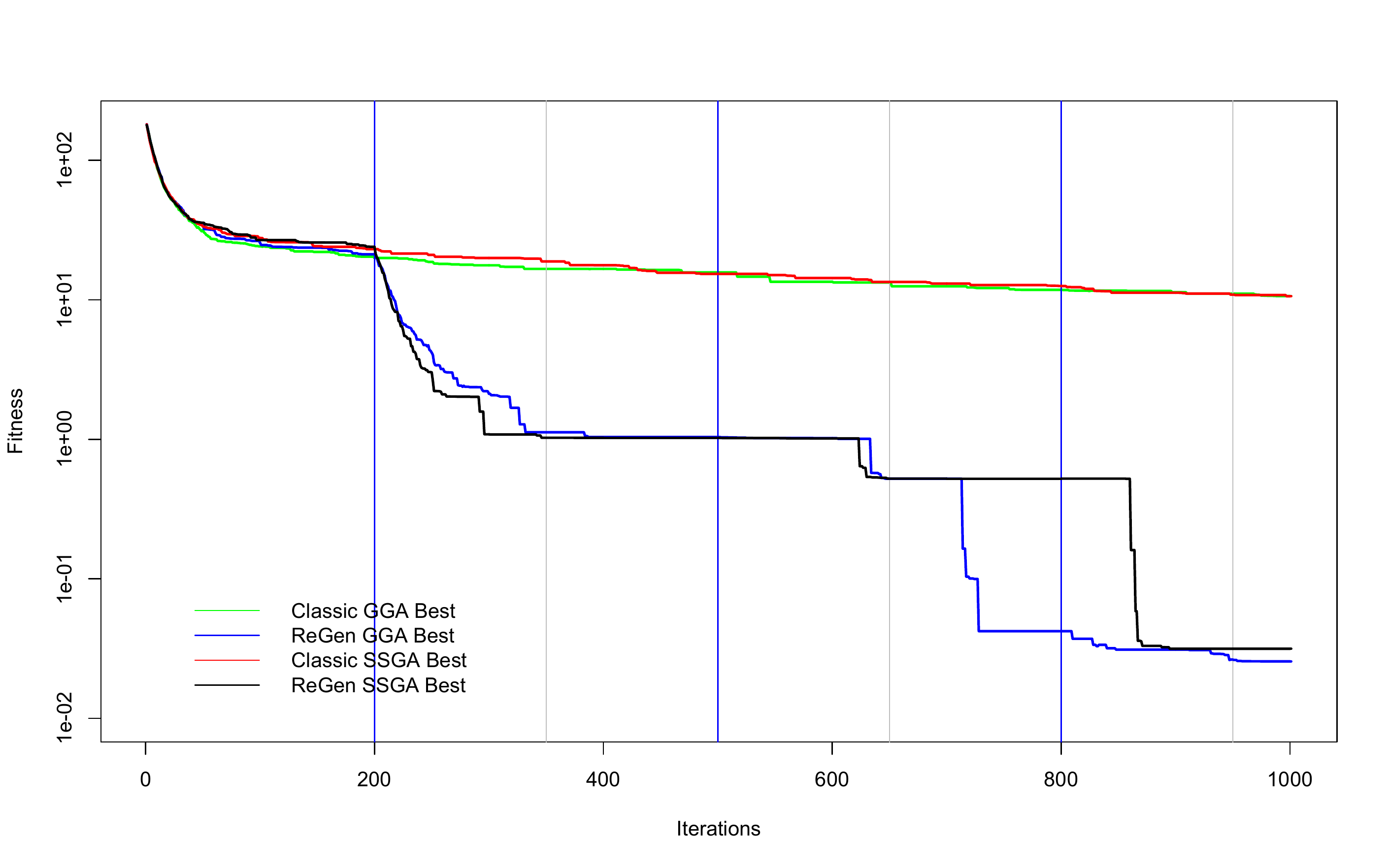}
\includegraphics[width=2in]{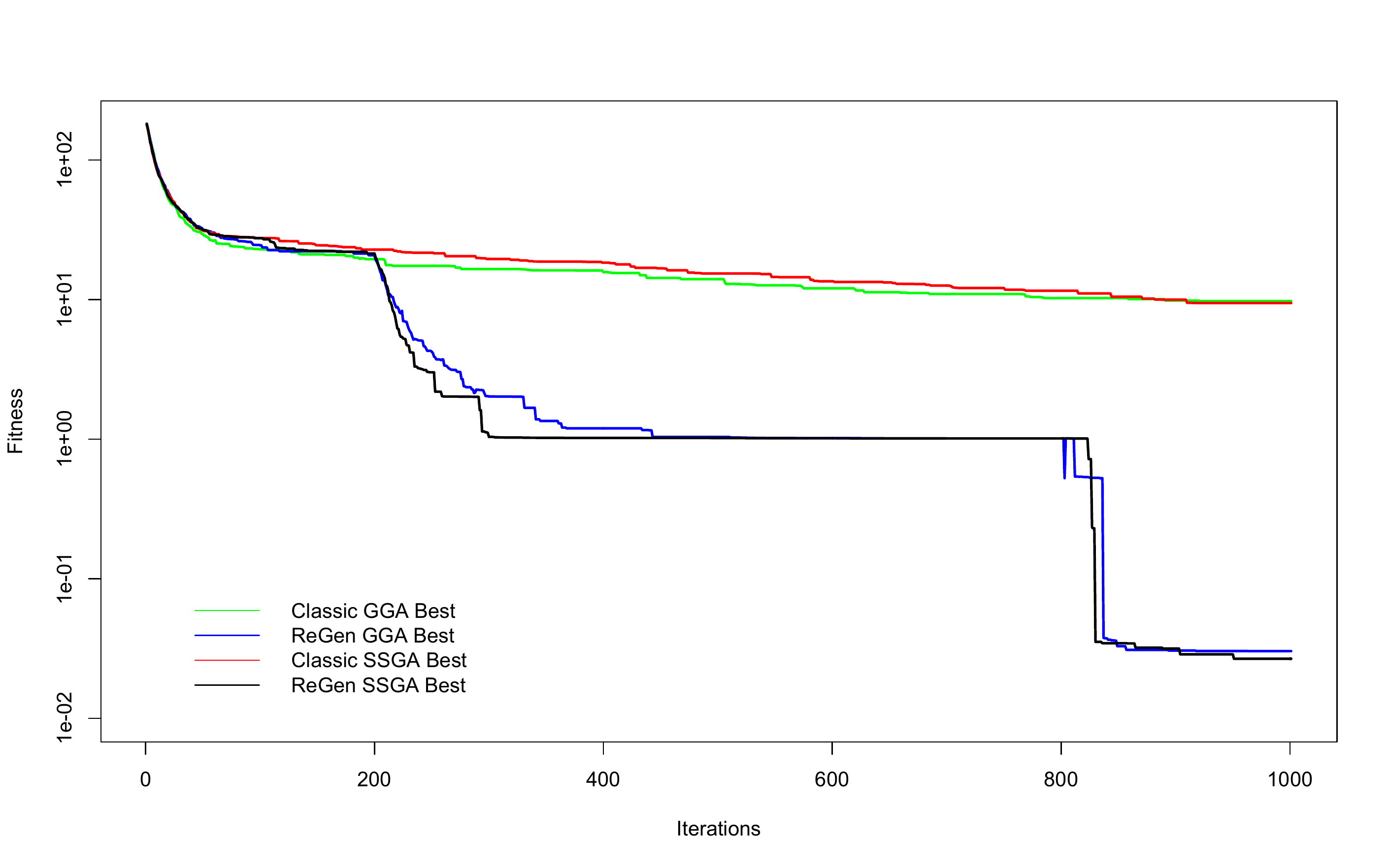}
\includegraphics[width=2in]{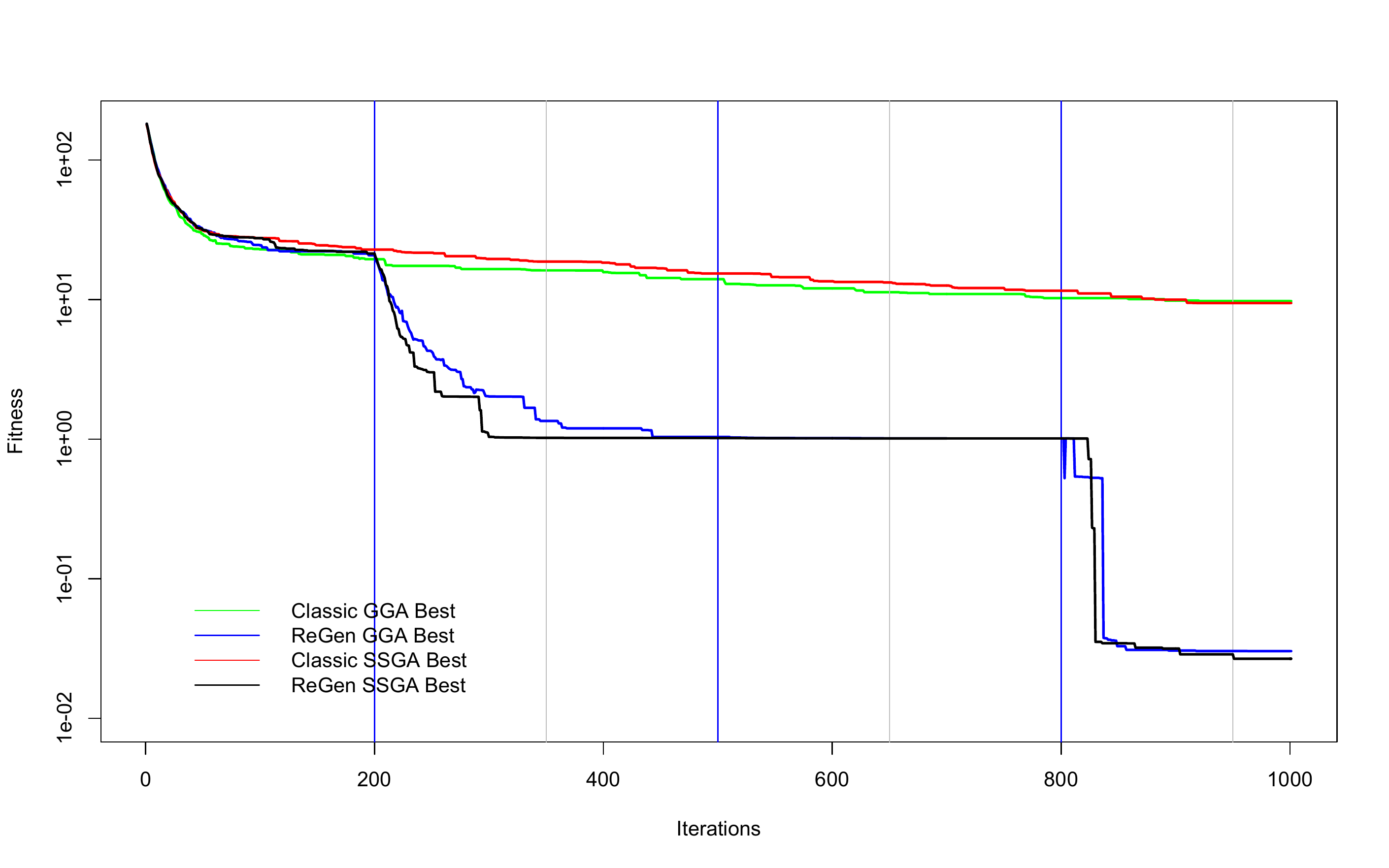}
\includegraphics[width=2in]{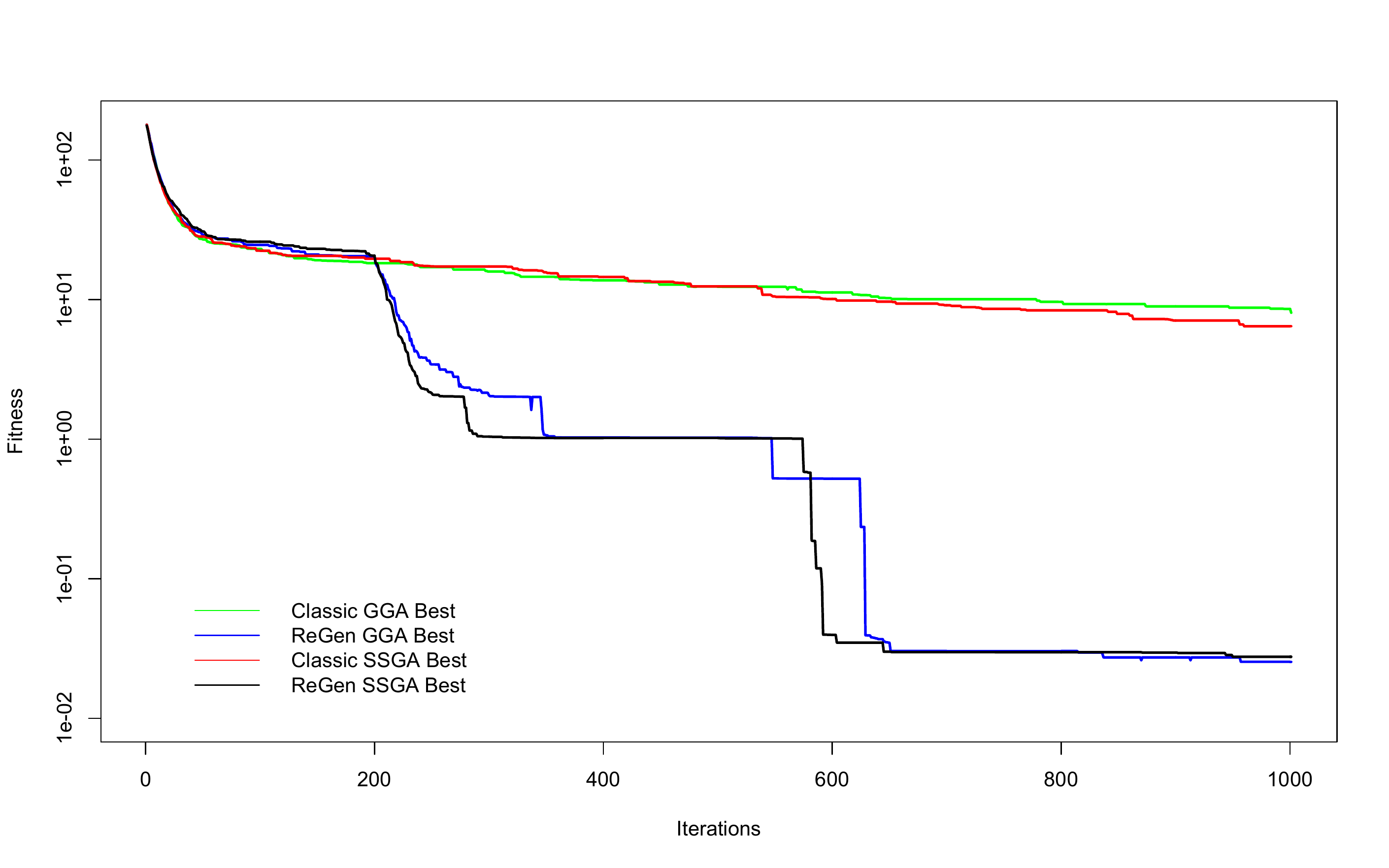}
\includegraphics[width=2in]{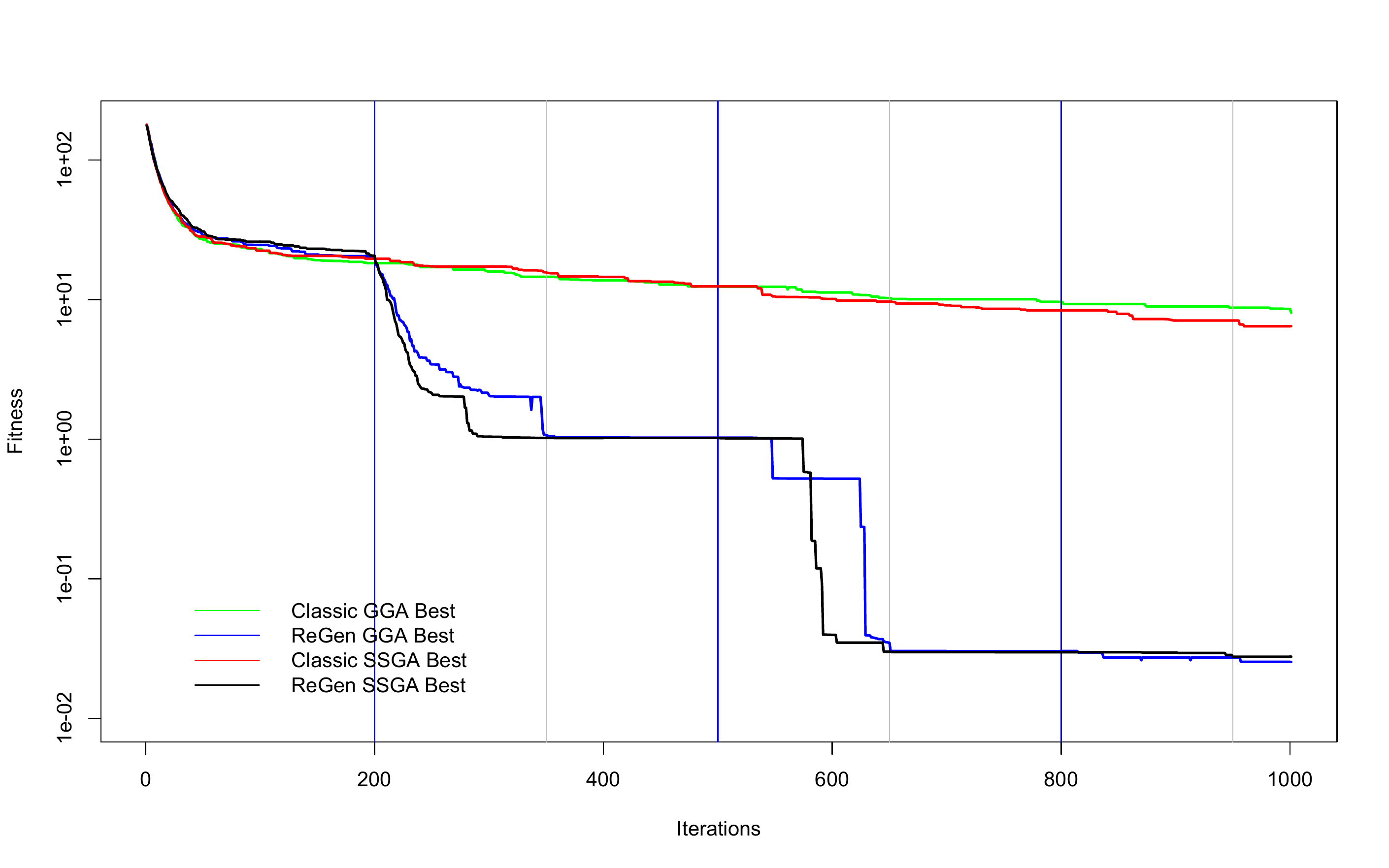}
\caption{Rastrigin. Generational replacement (GGA) and Steady State replacement (SSGA). From top to bottom, crossover rates from $0.6$ to $1.0$.}
\label{c4fig7}
\end{figure}

\begin{figure}[H]
\centering
\includegraphics[width=2in]{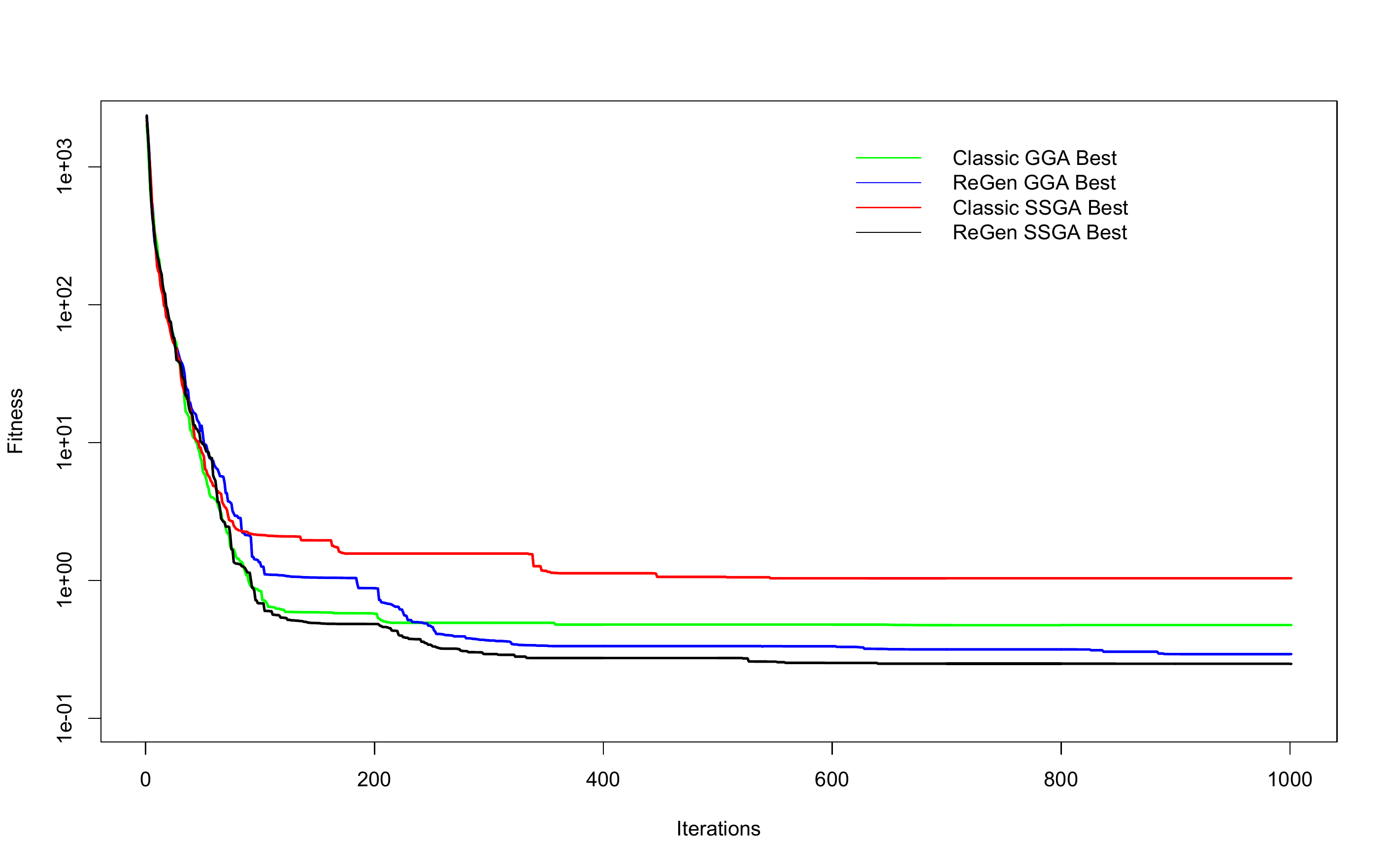}
\includegraphics[width=2in]{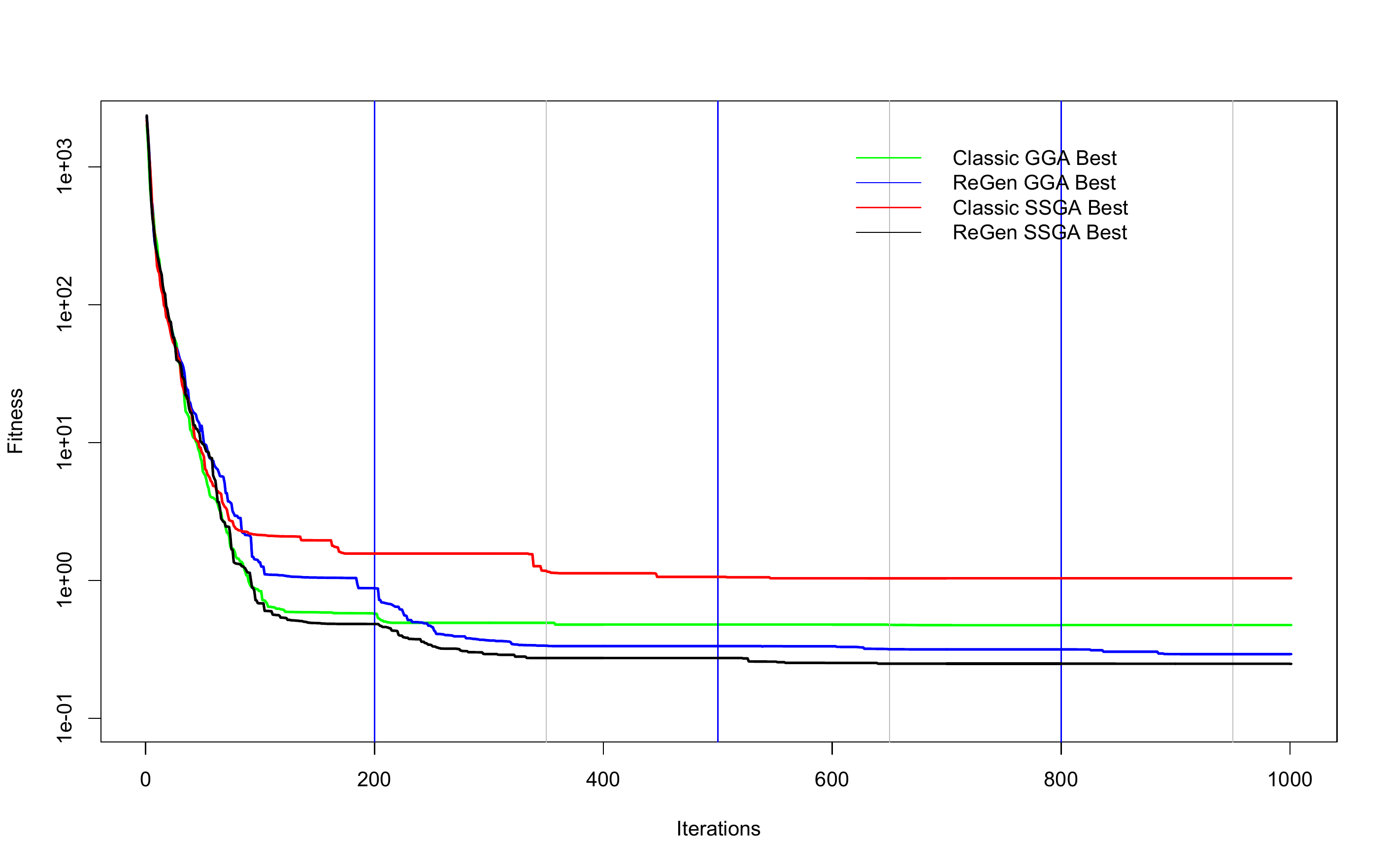}
\includegraphics[width=2in]{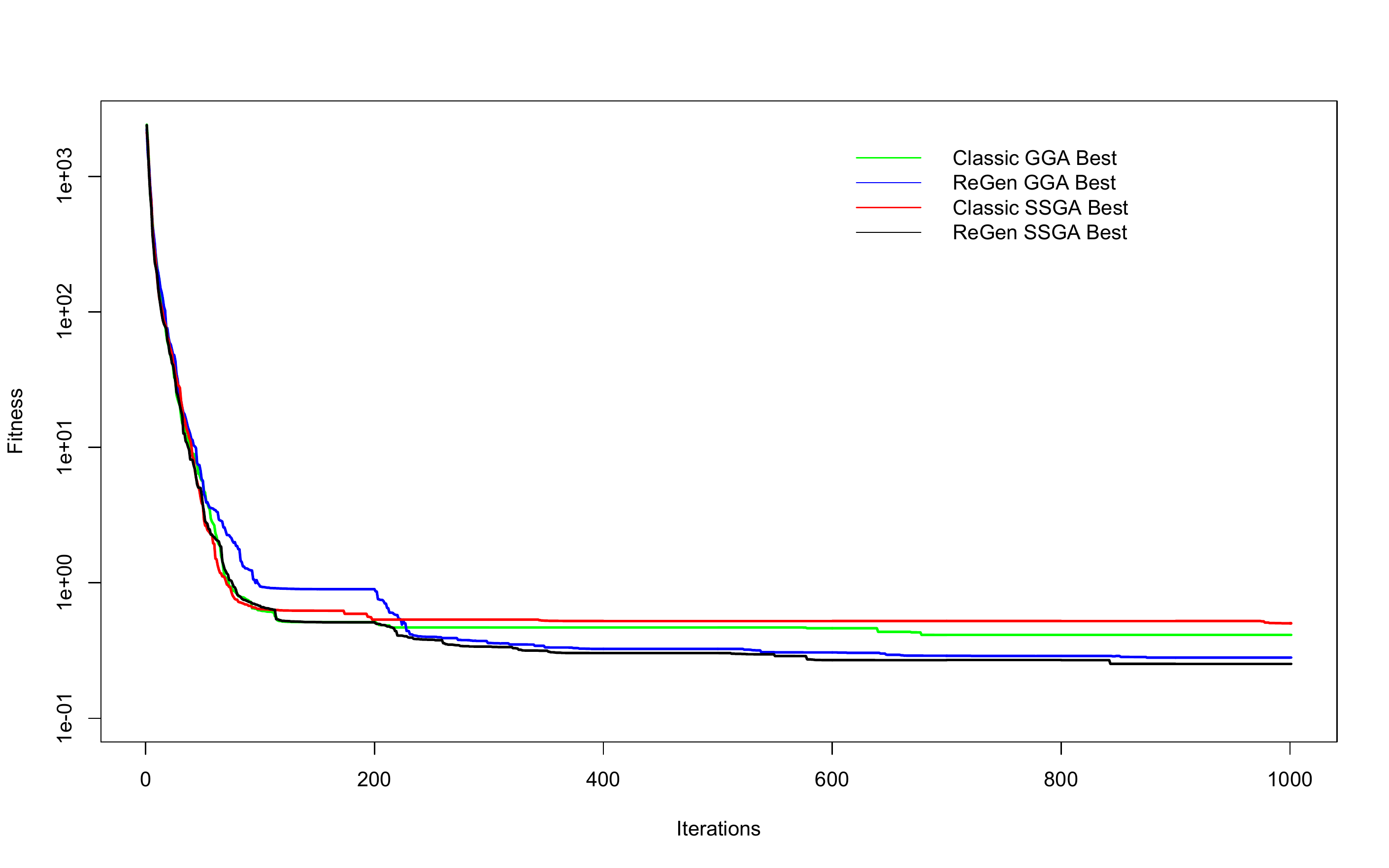}
\includegraphics[width=2in]{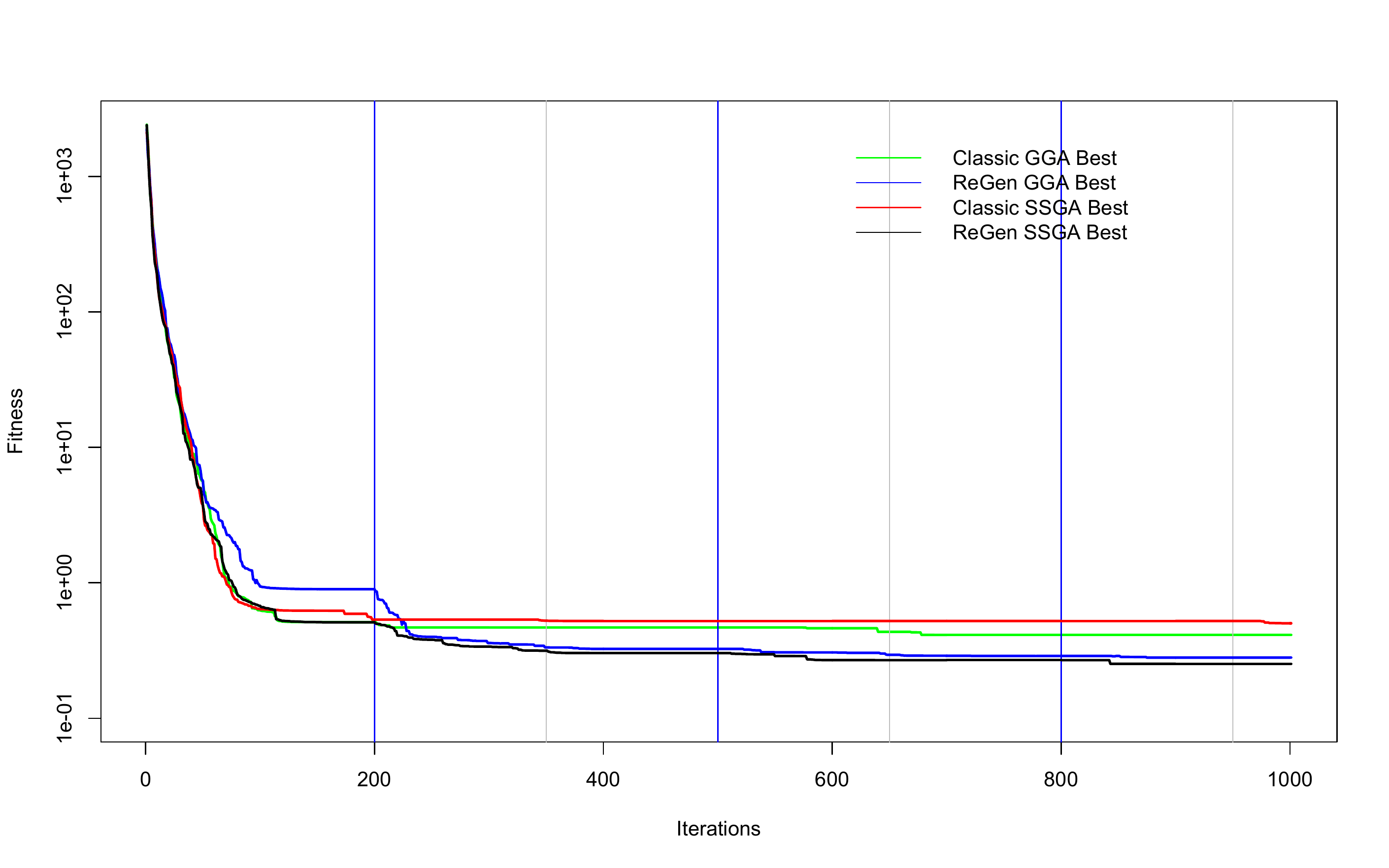}
\includegraphics[width=2in]{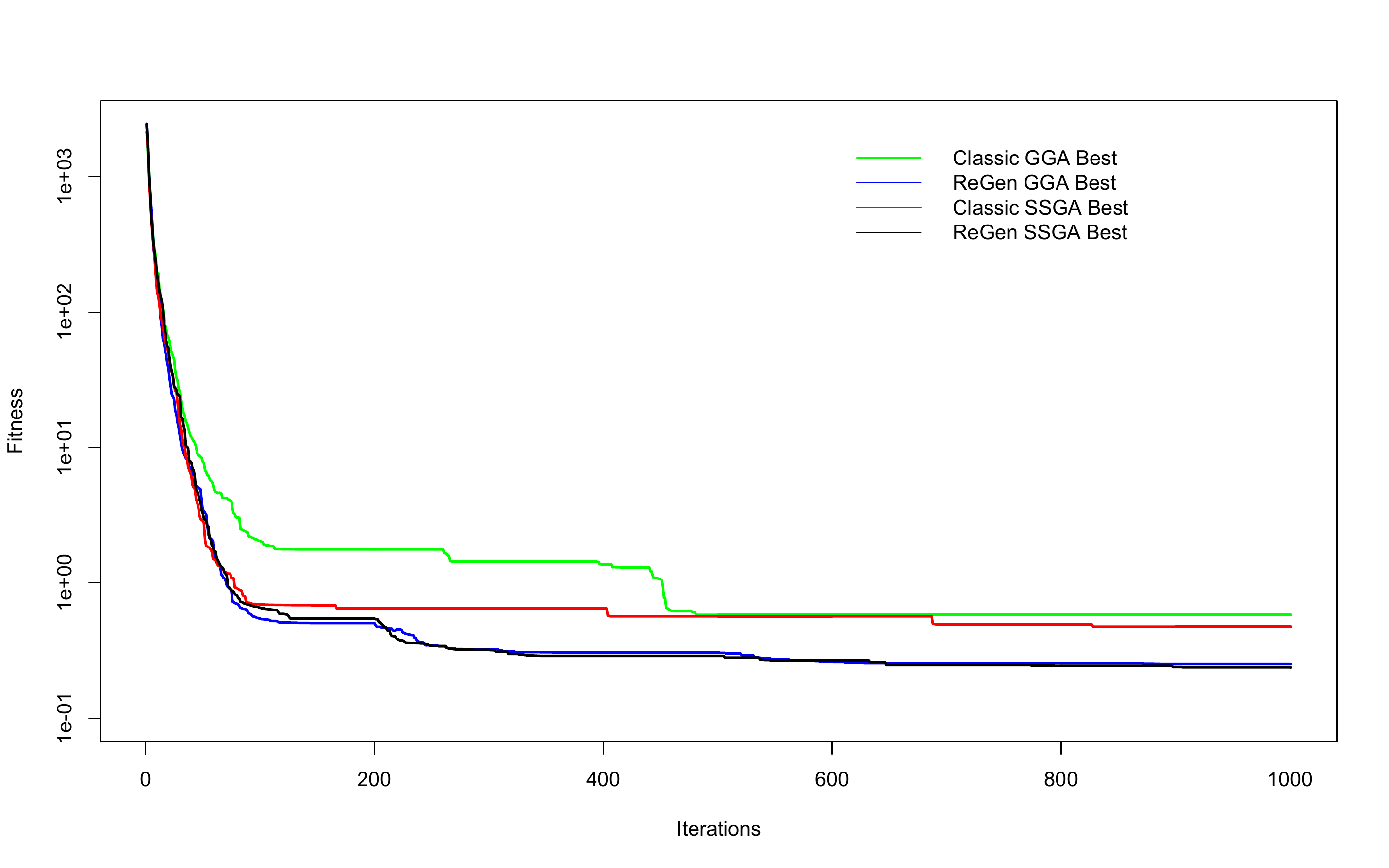}
\includegraphics[width=2in]{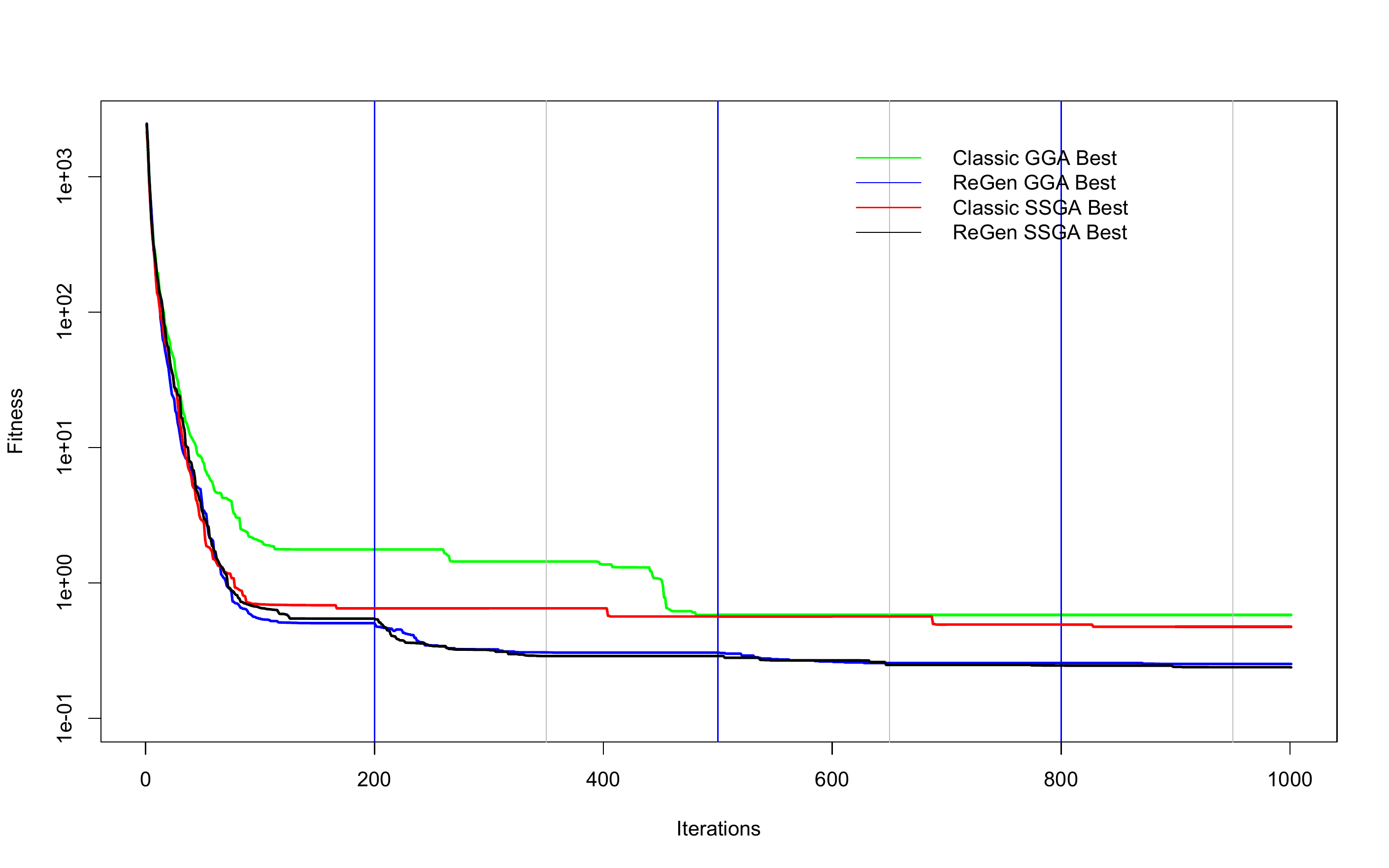}
\includegraphics[width=2in]{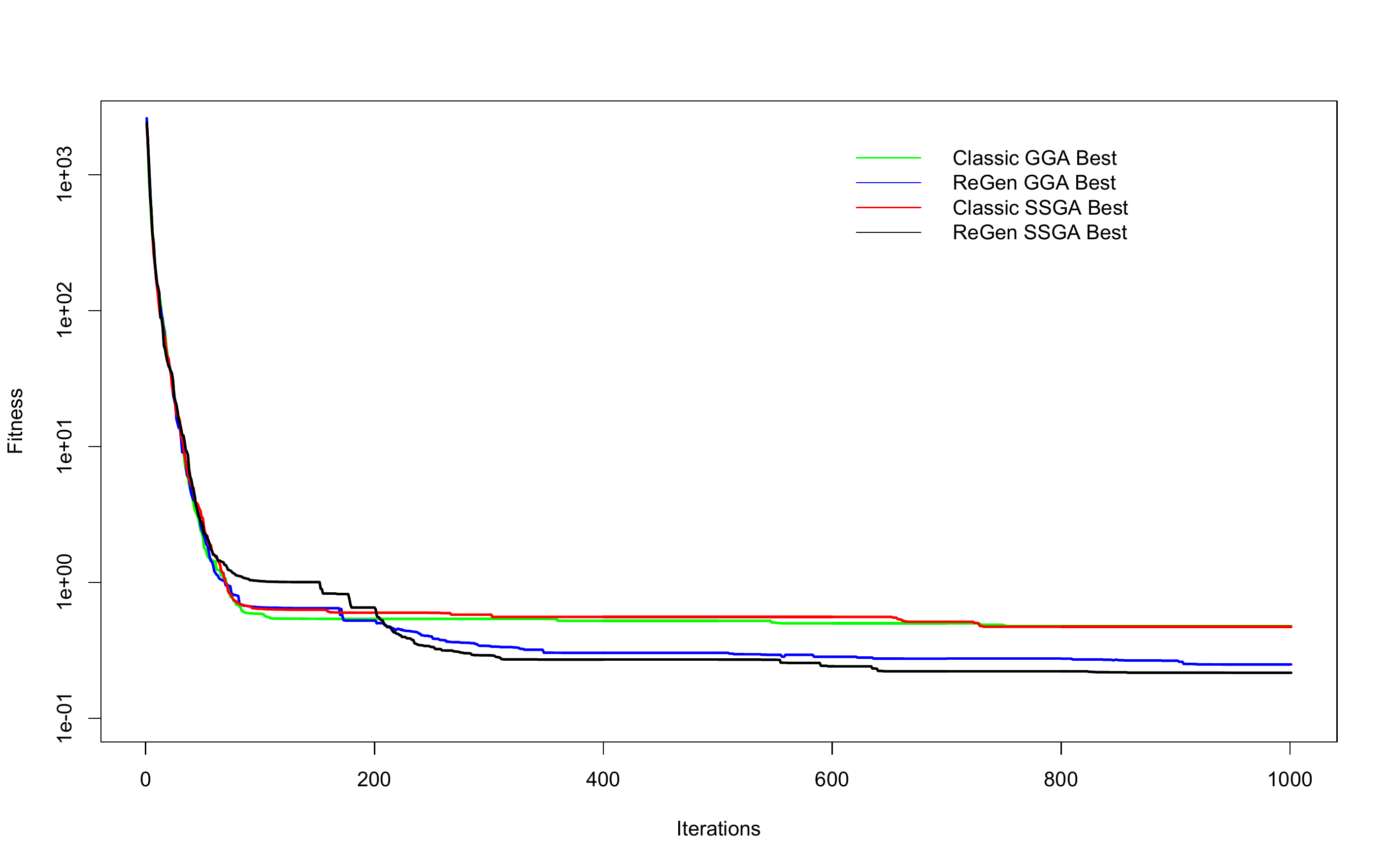}
\includegraphics[width=2in]{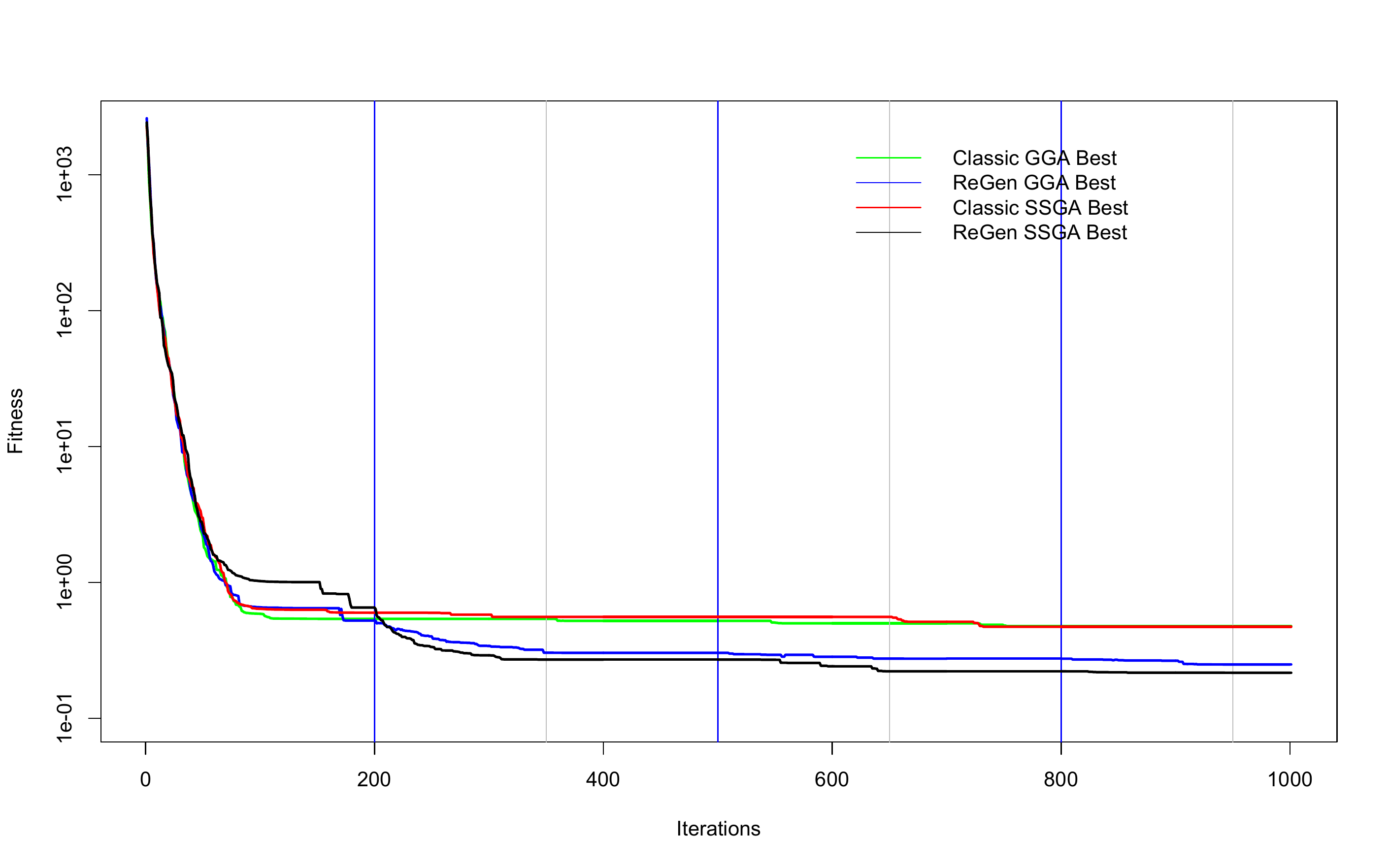}
\includegraphics[width=2in]{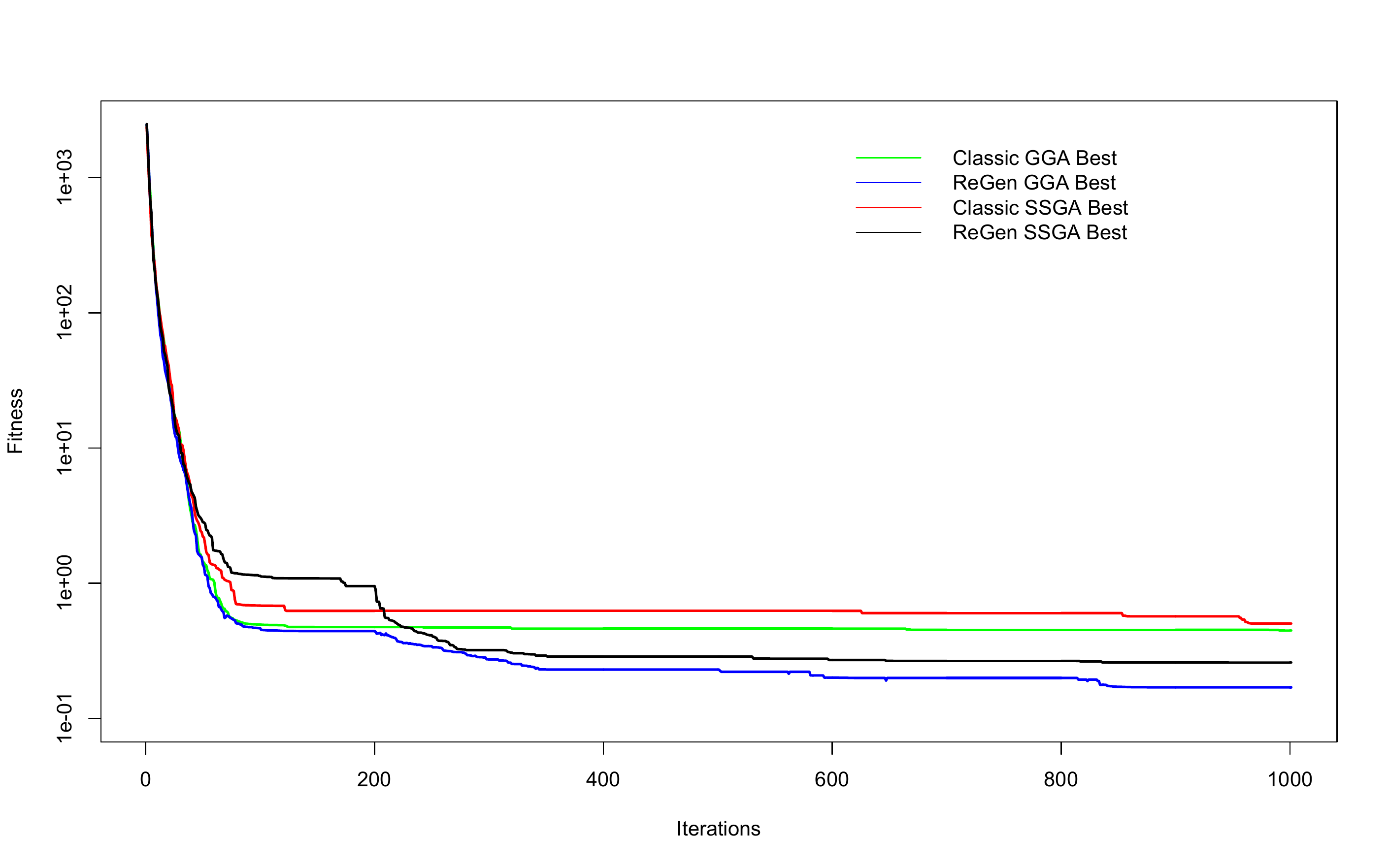}
\includegraphics[width=2in]{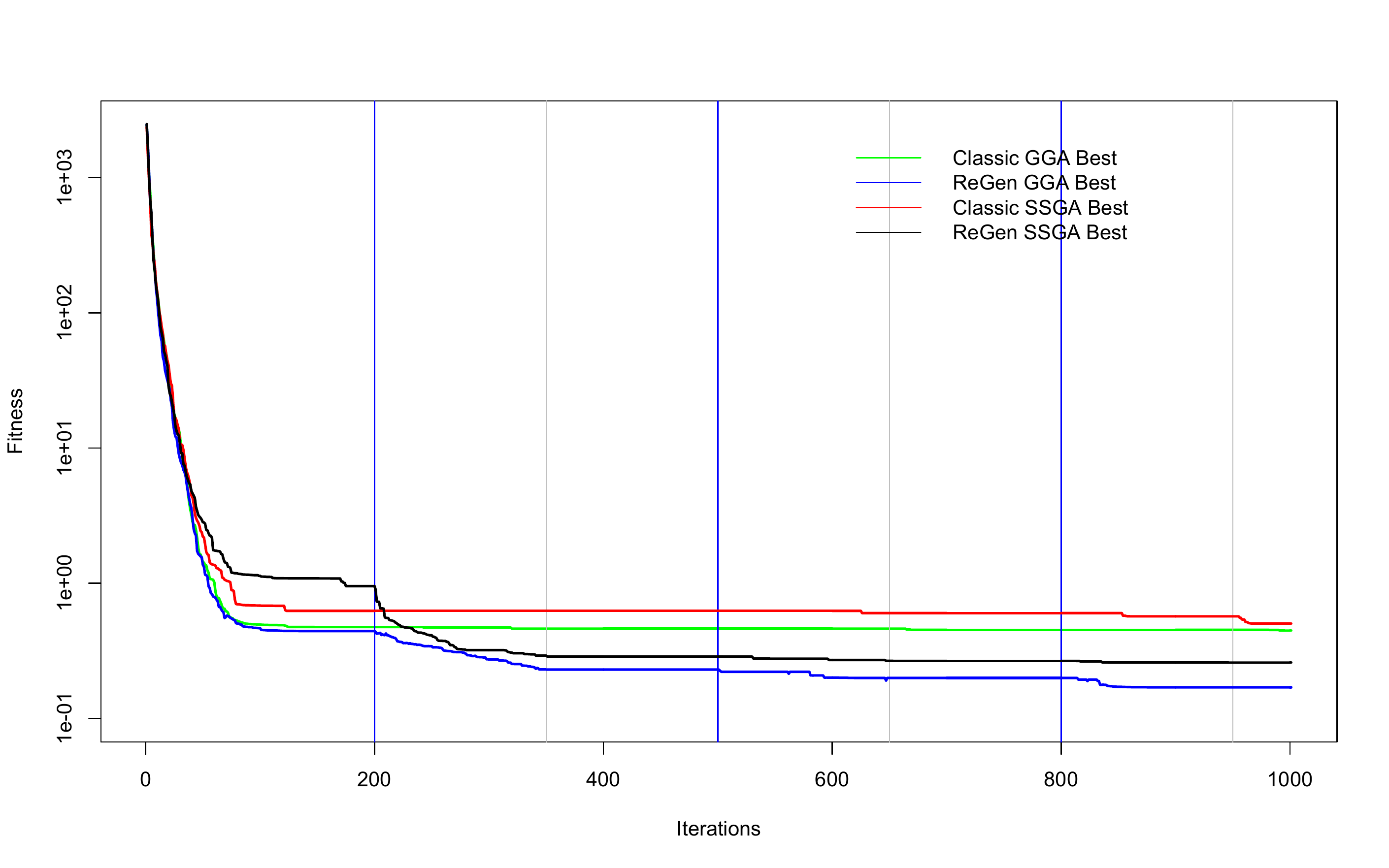}
\caption{Rosenbrock. Generational replacement (GGA) and Steady State replacement (SSGA). From top to bottom, crossover rates from $0.6$ to $1.0$.}
\label{c4fig8}
\end{figure}

\begin{figure}[H]
\centering
\includegraphics[width=2in]{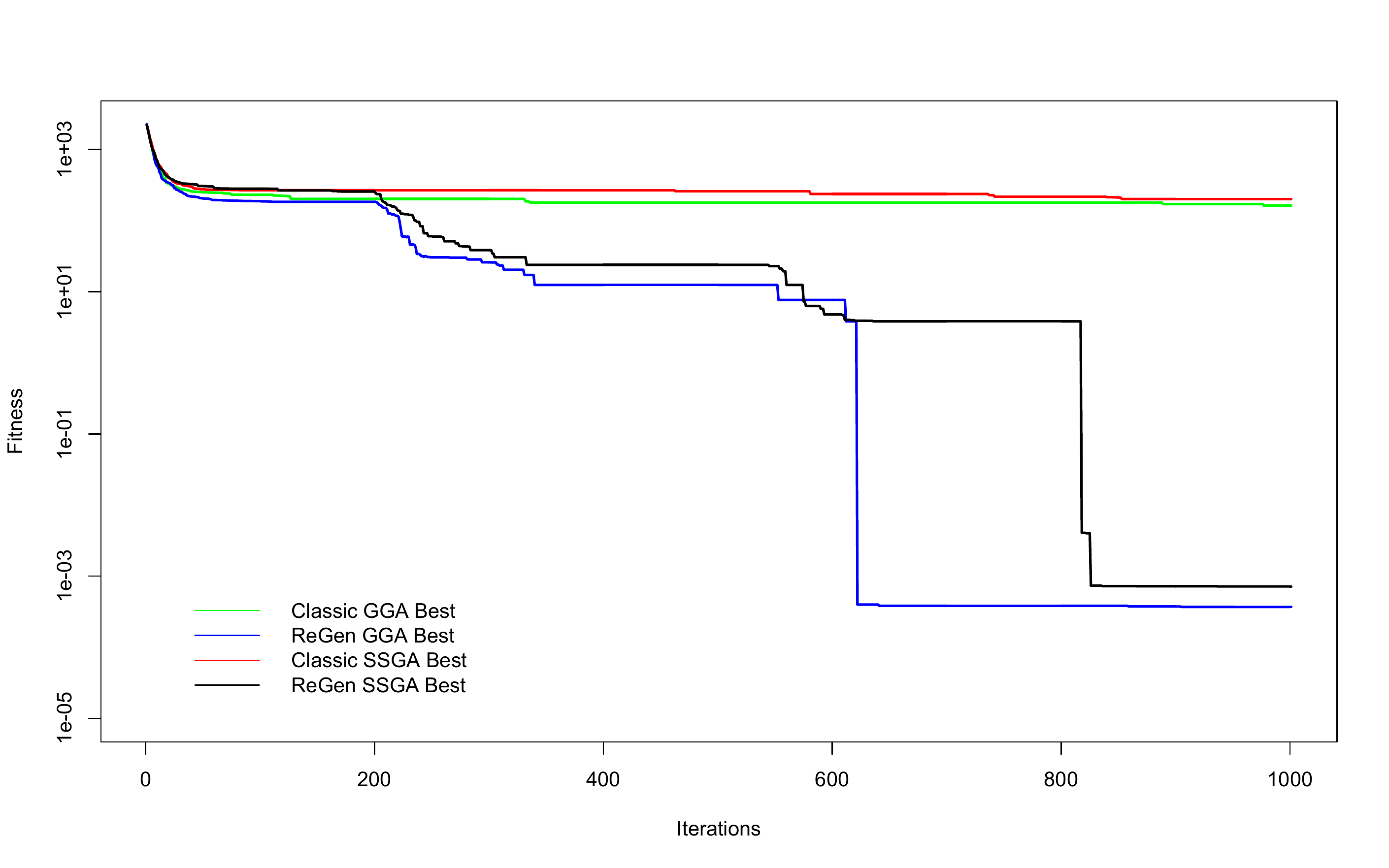}
\includegraphics[width=2in]{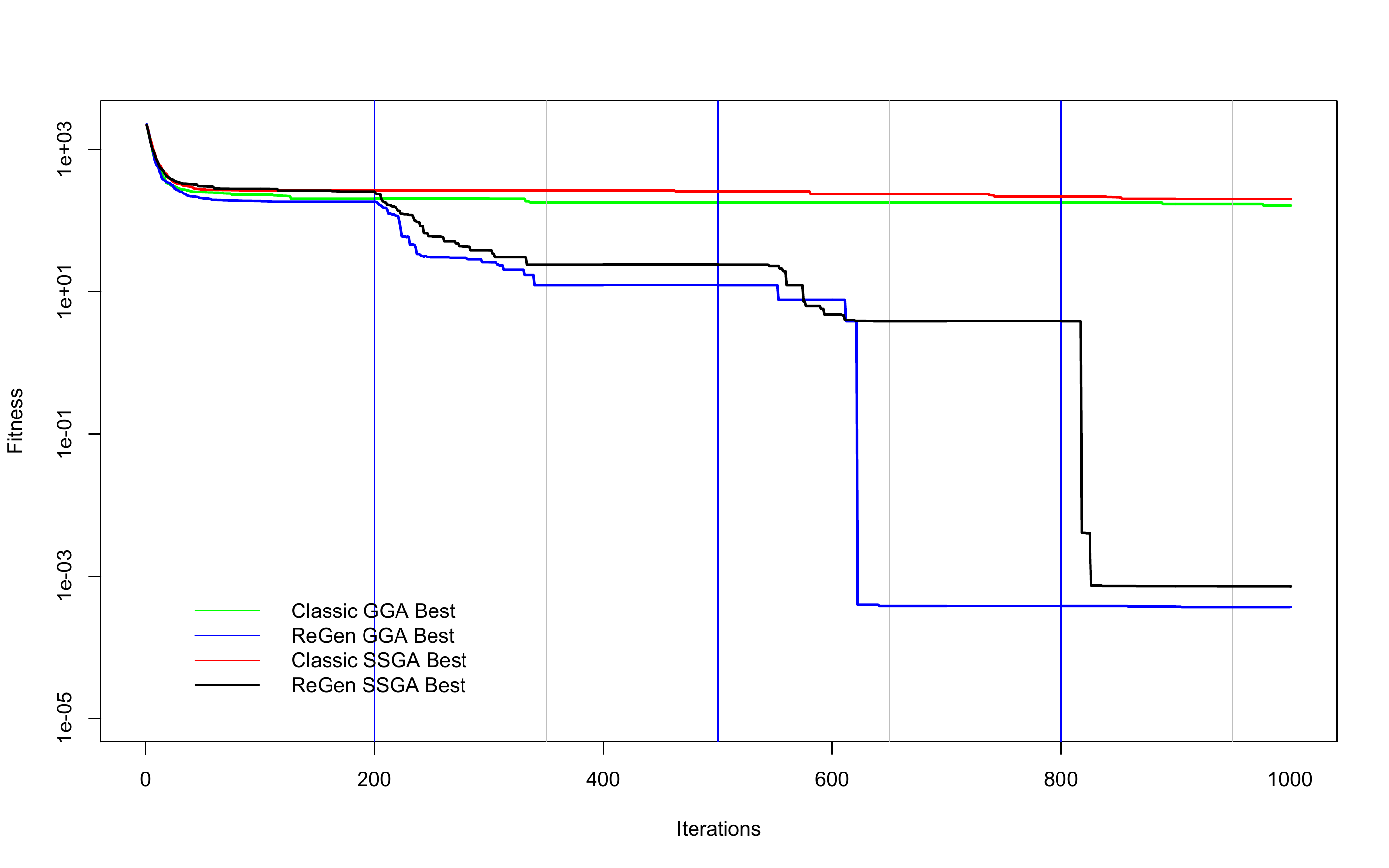}
\includegraphics[width=2in]{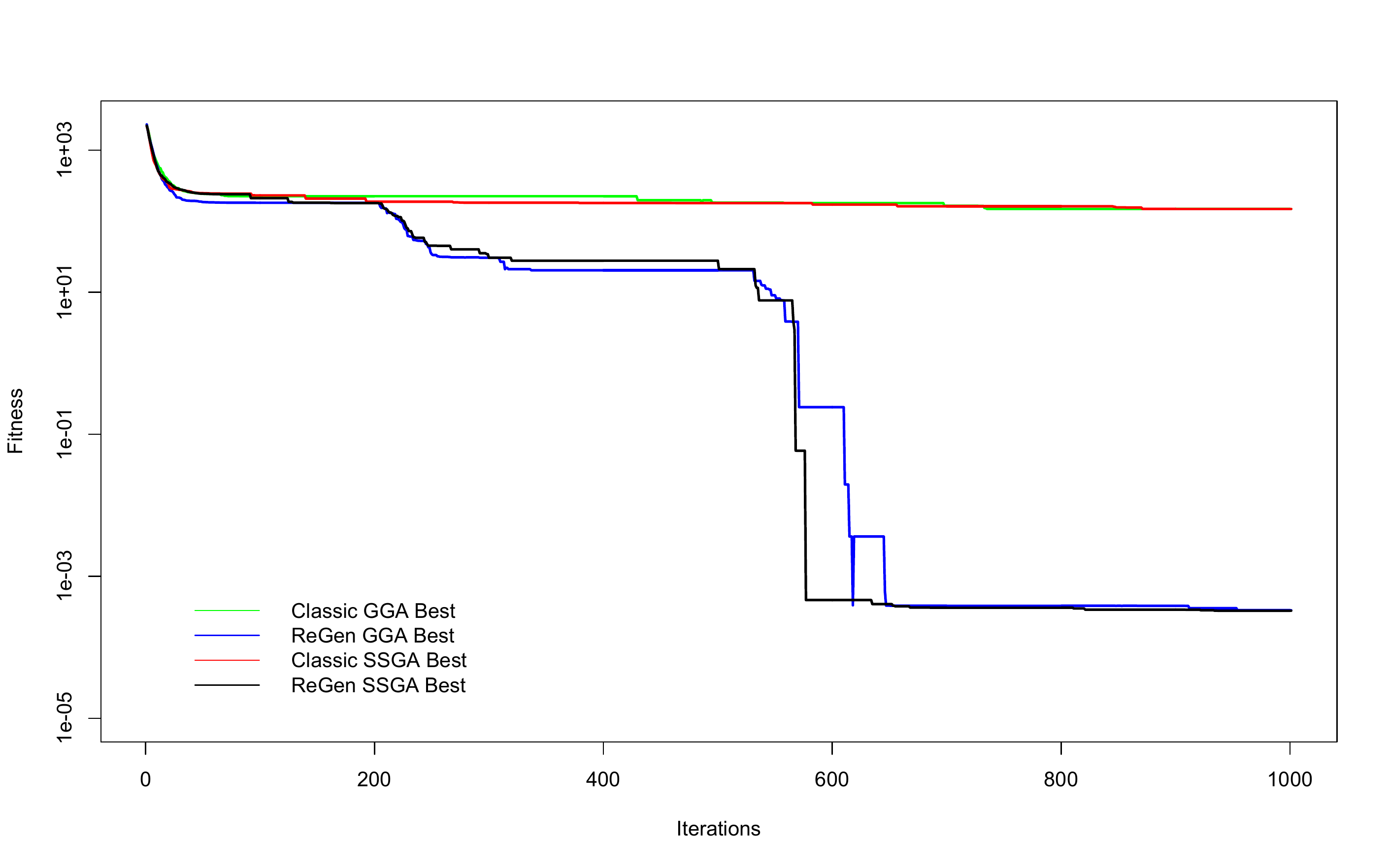}
\includegraphics[width=2in]{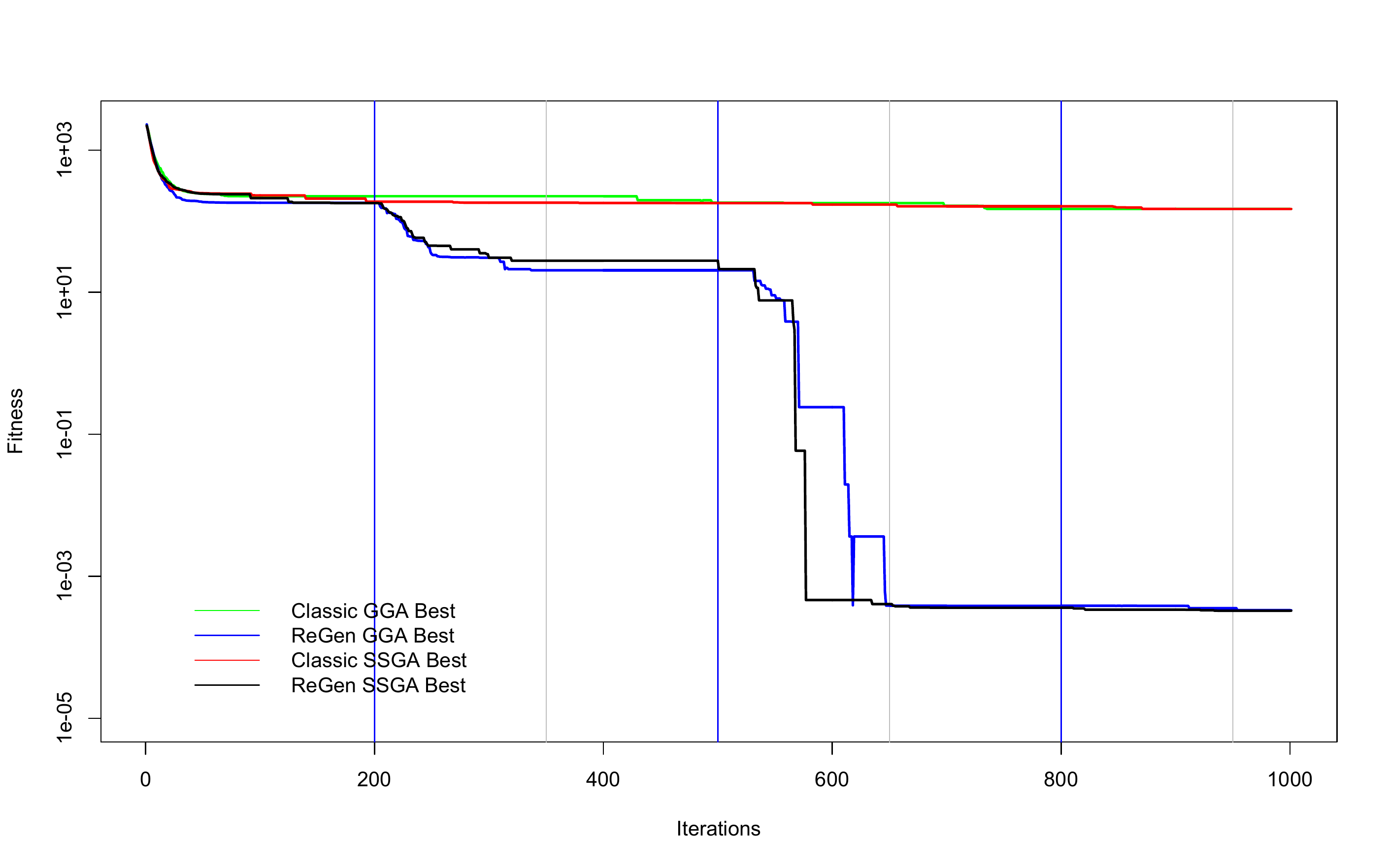}
\includegraphics[width=2in]{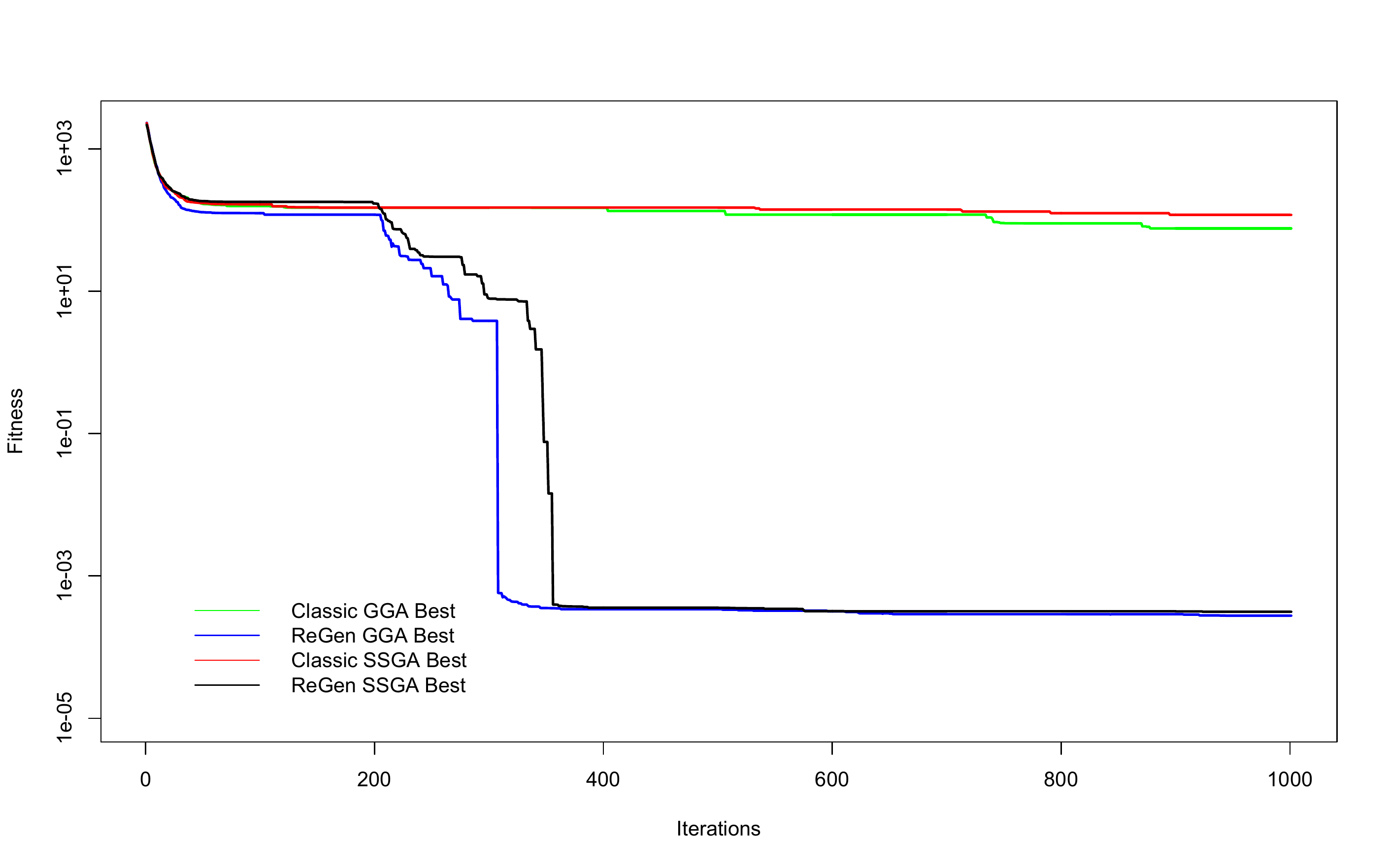}
\includegraphics[width=2in]{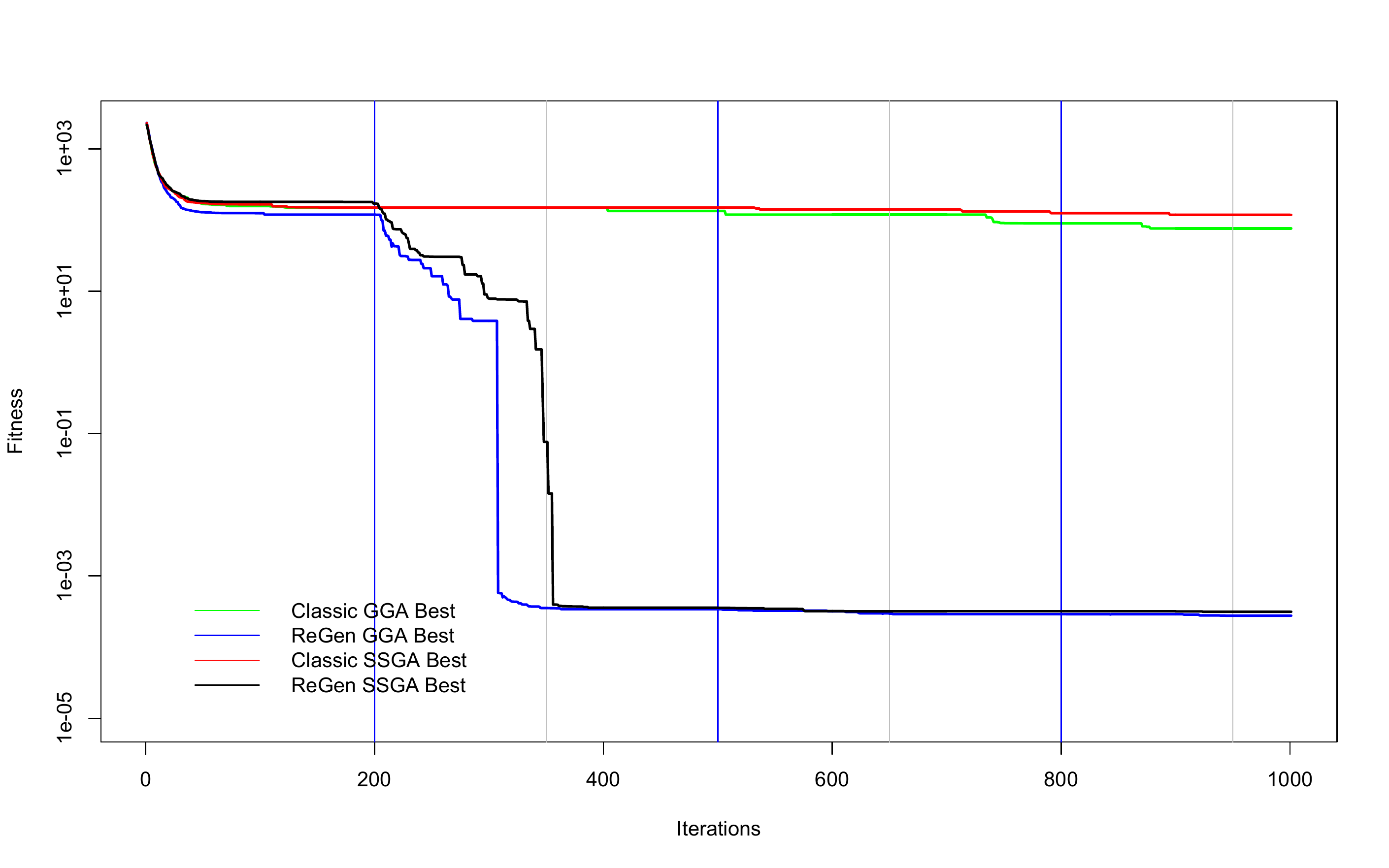}
\includegraphics[width=2in]{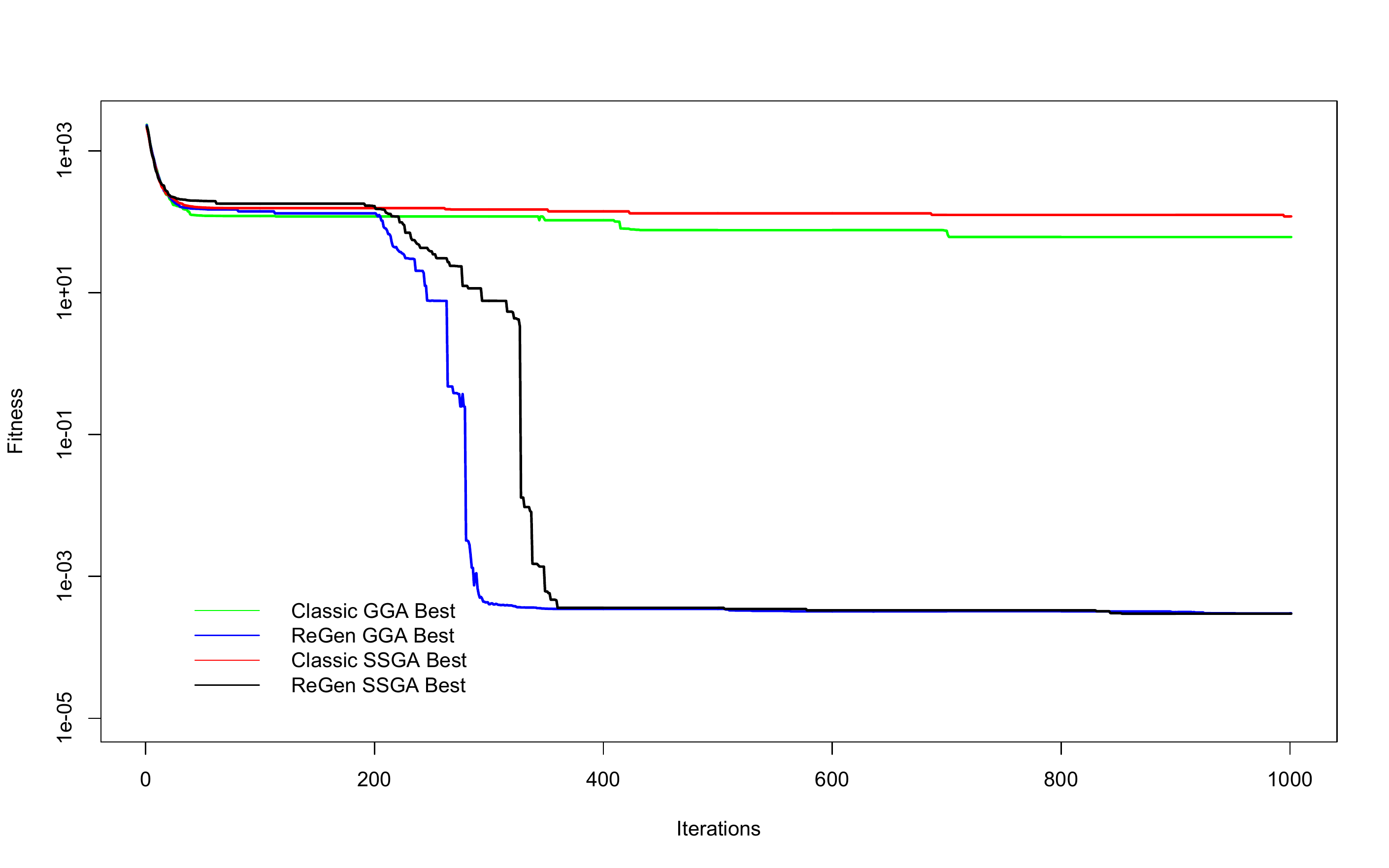}
\includegraphics[width=2in]{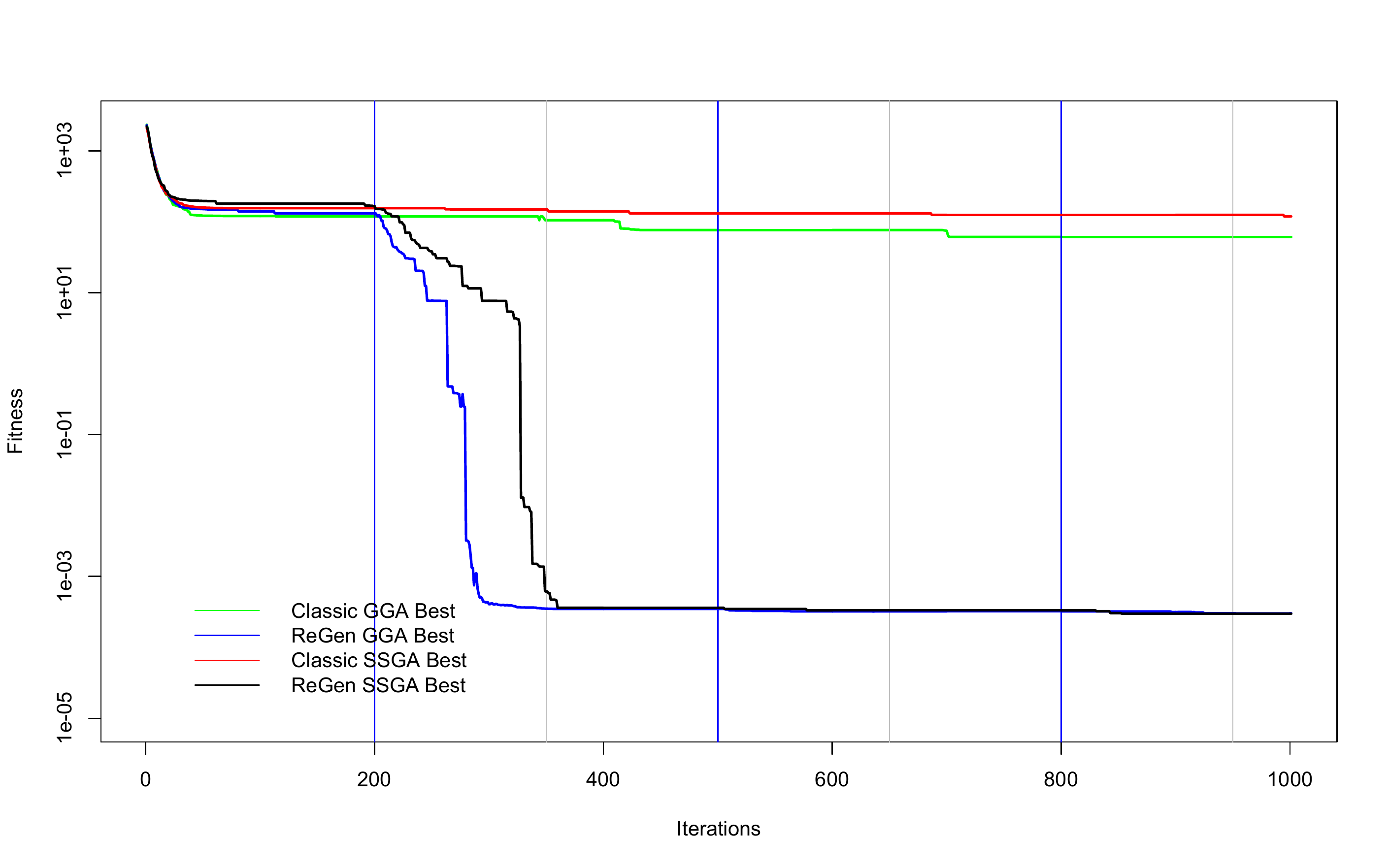}
\includegraphics[width=2in]{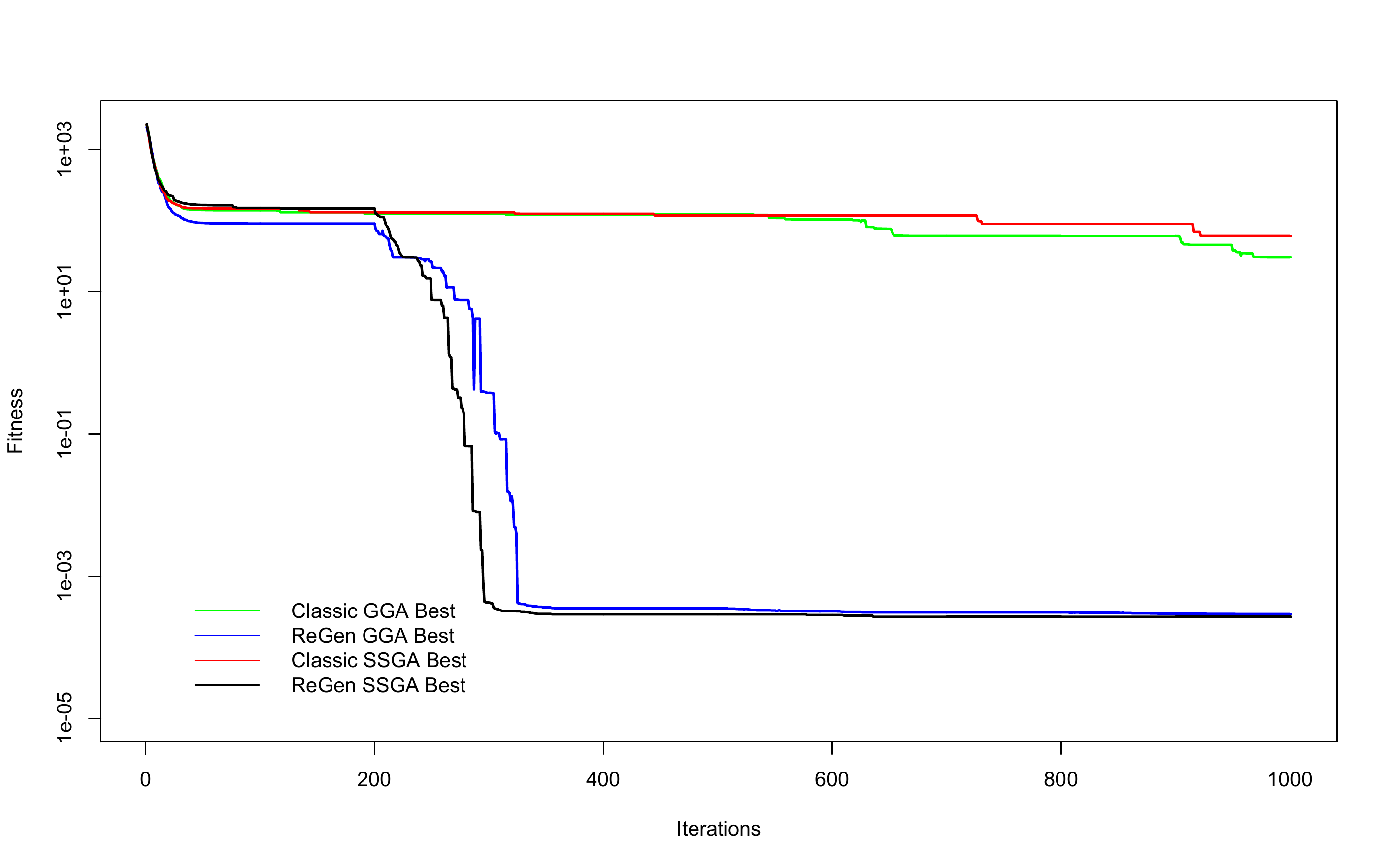}
\includegraphics[width=2in]{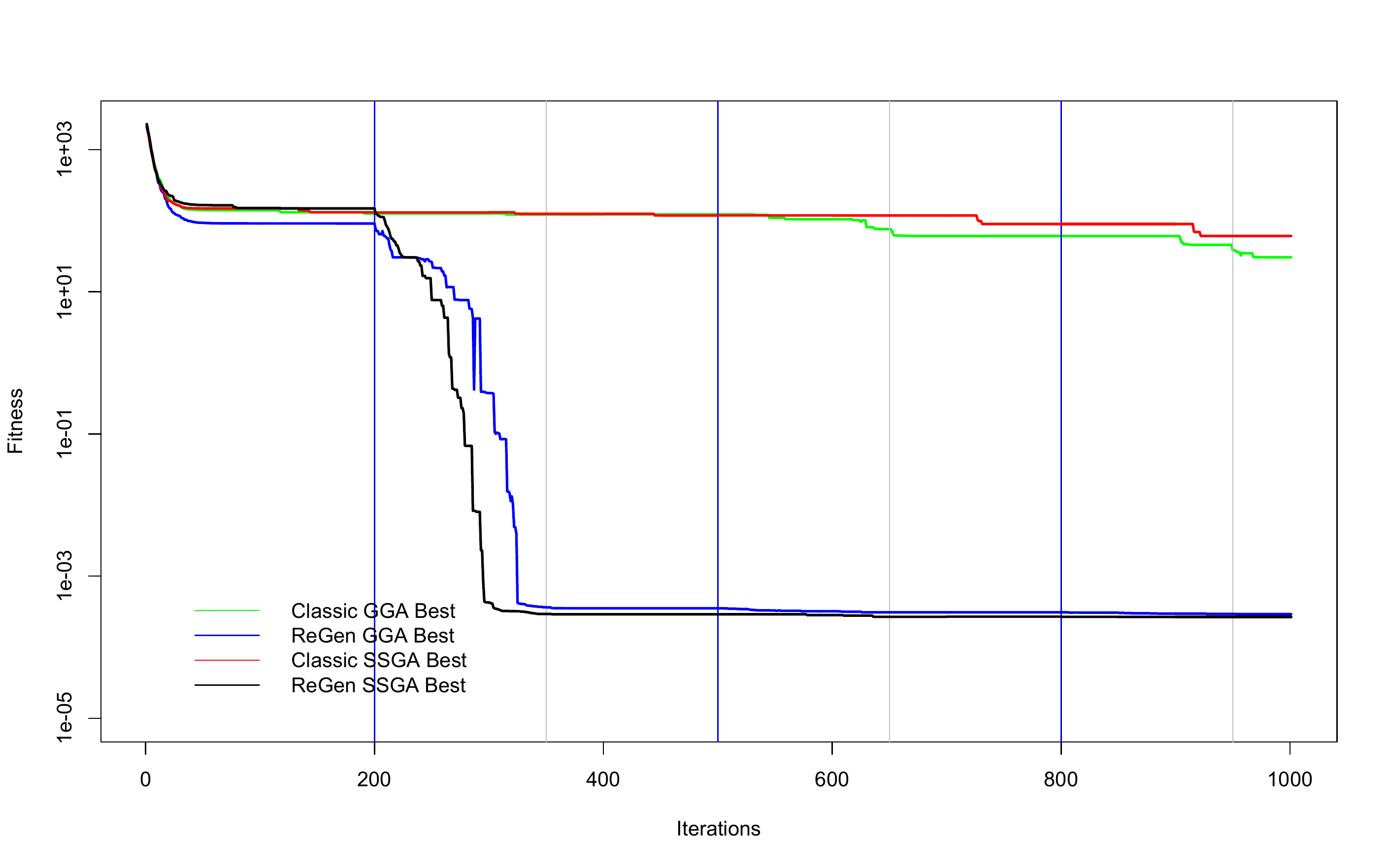}
\caption{Schwefel. Generational replacement (GGA) and Steady State replacement (SSGA). From top to bottom, crossover rates from $0.6$ to $1.0$.}
\label{c4fig9}
\end{figure}

\begin{figure}[H]
\centering
\includegraphics[width=2in]{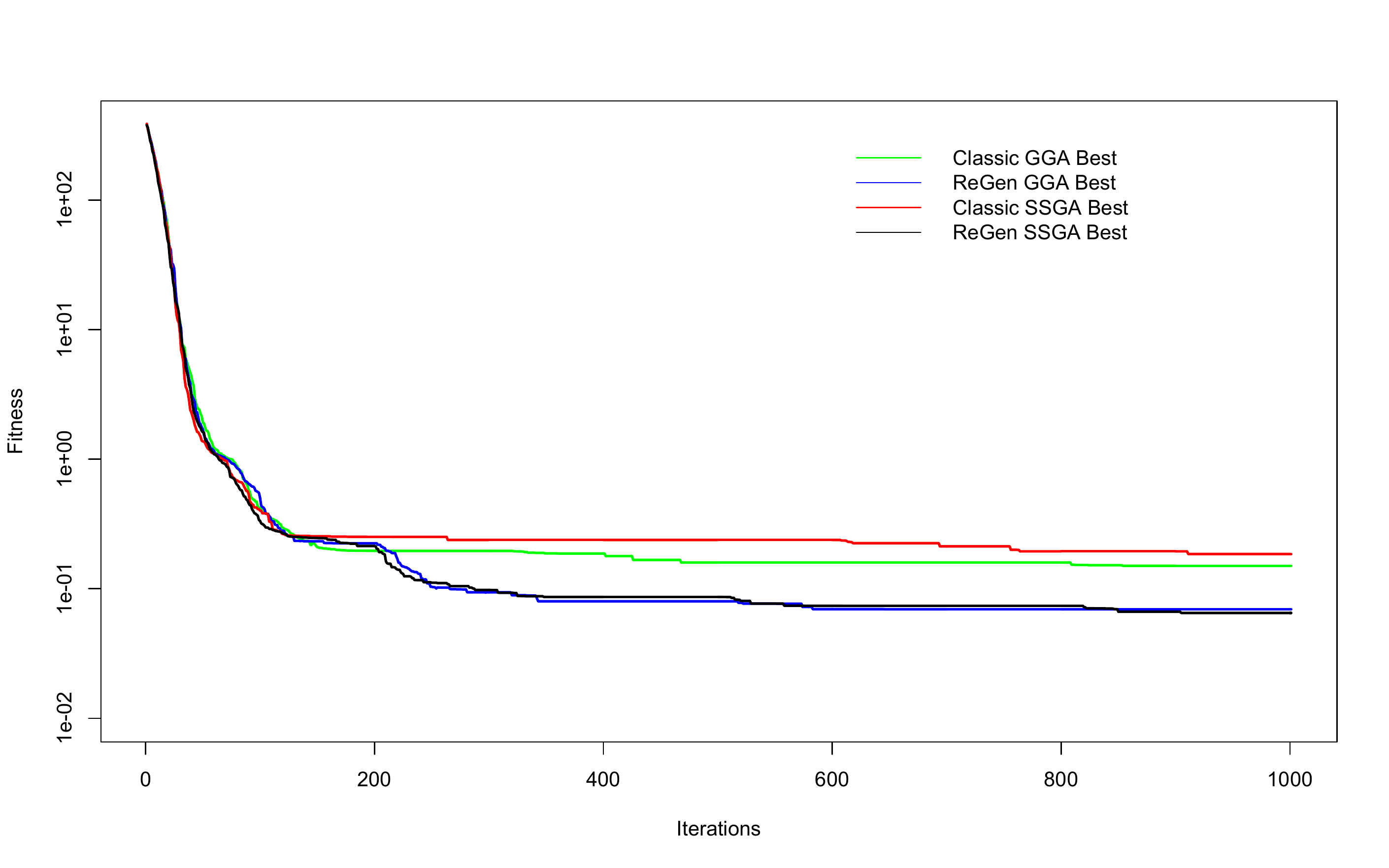}
\includegraphics[width=2in]{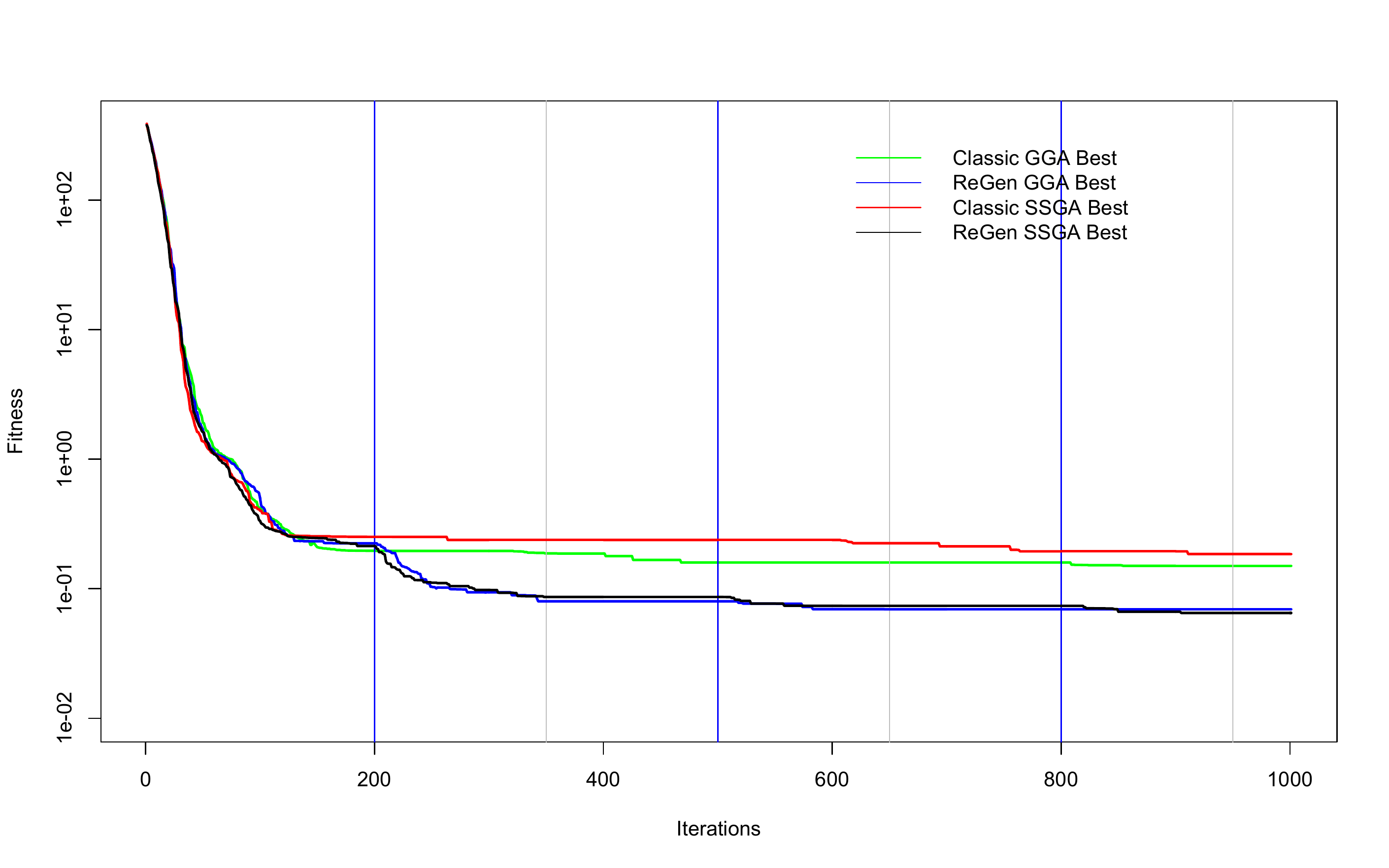}
\includegraphics[width=2in]{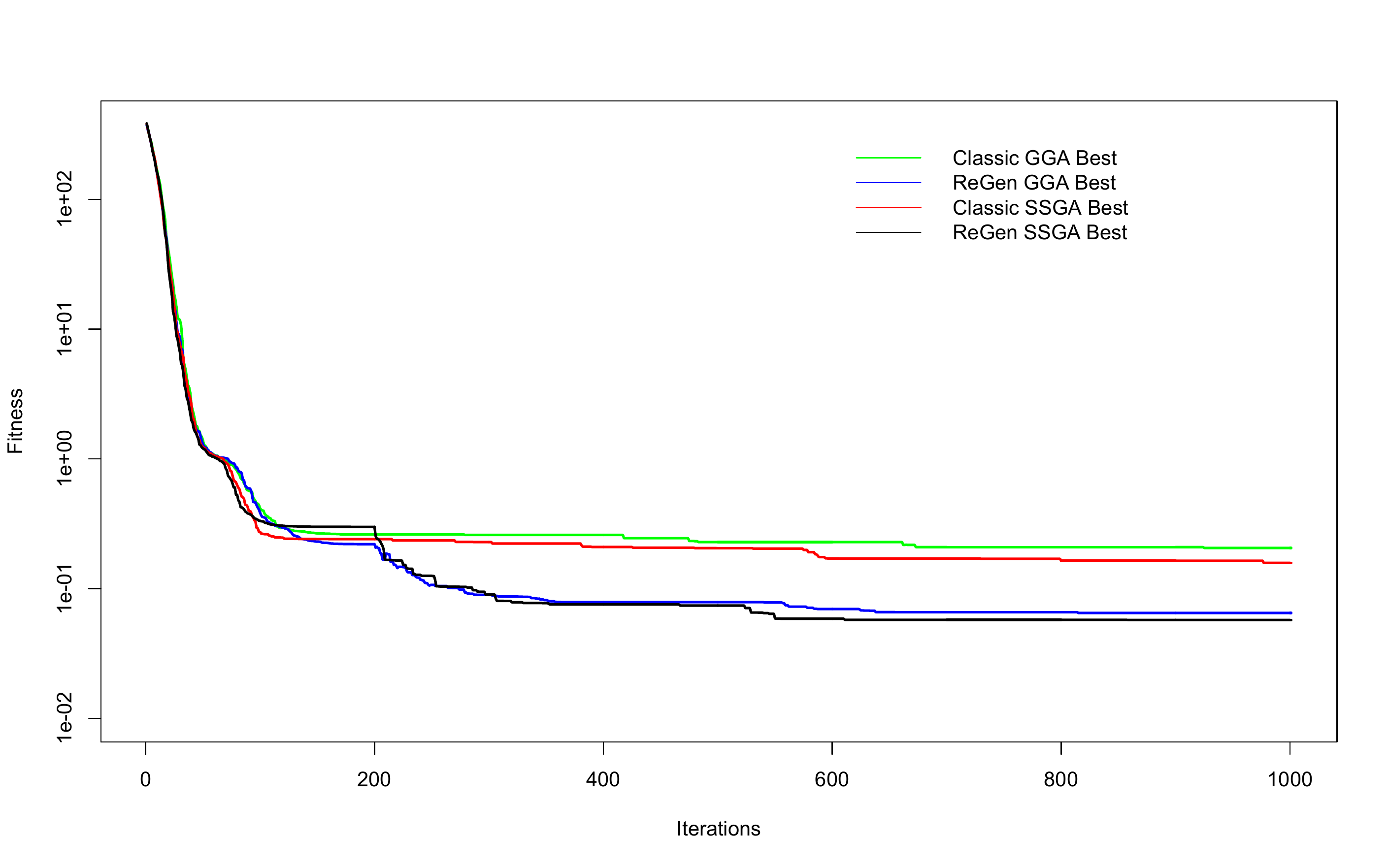}
\includegraphics[width=2in]{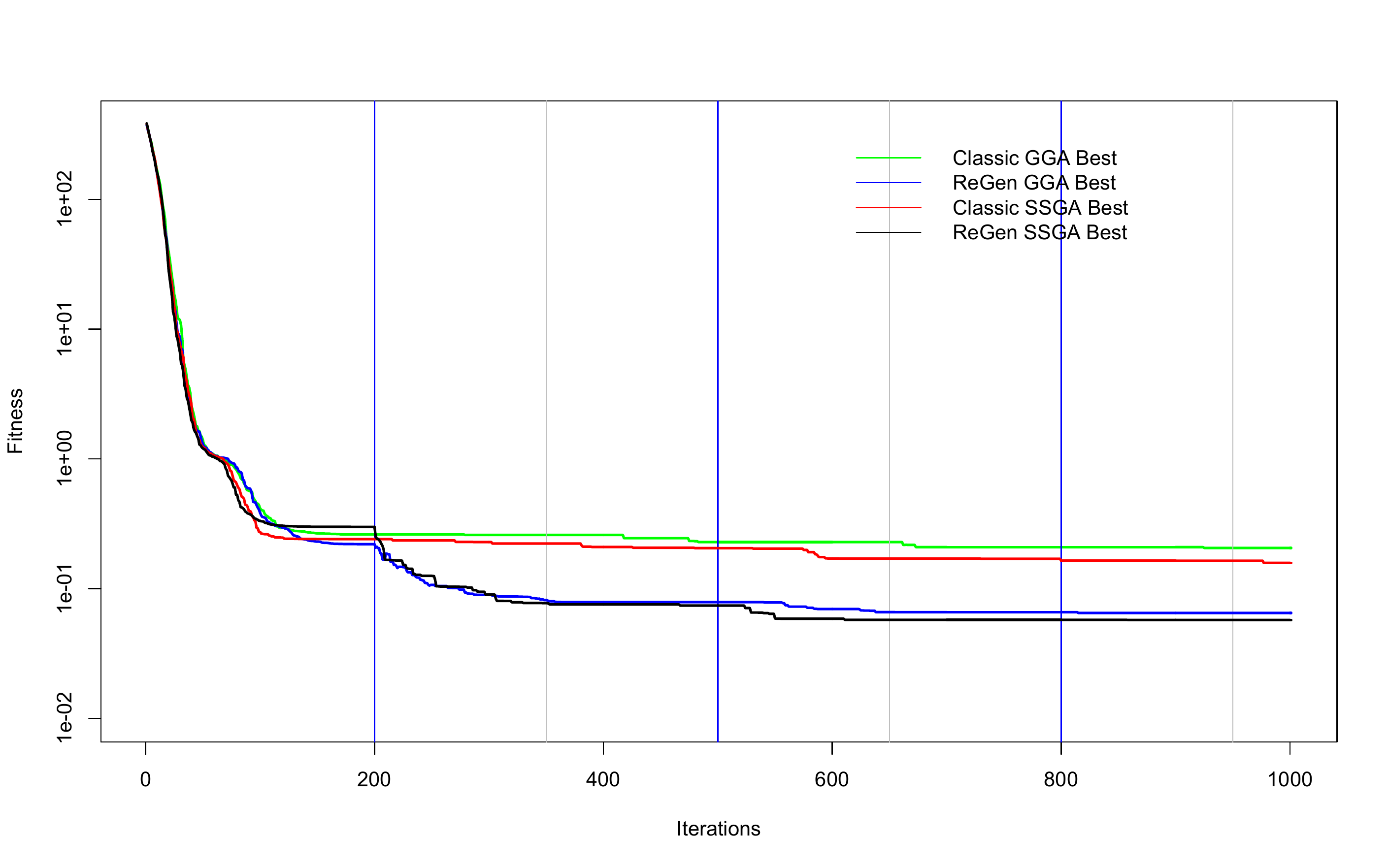}
\includegraphics[width=2in]{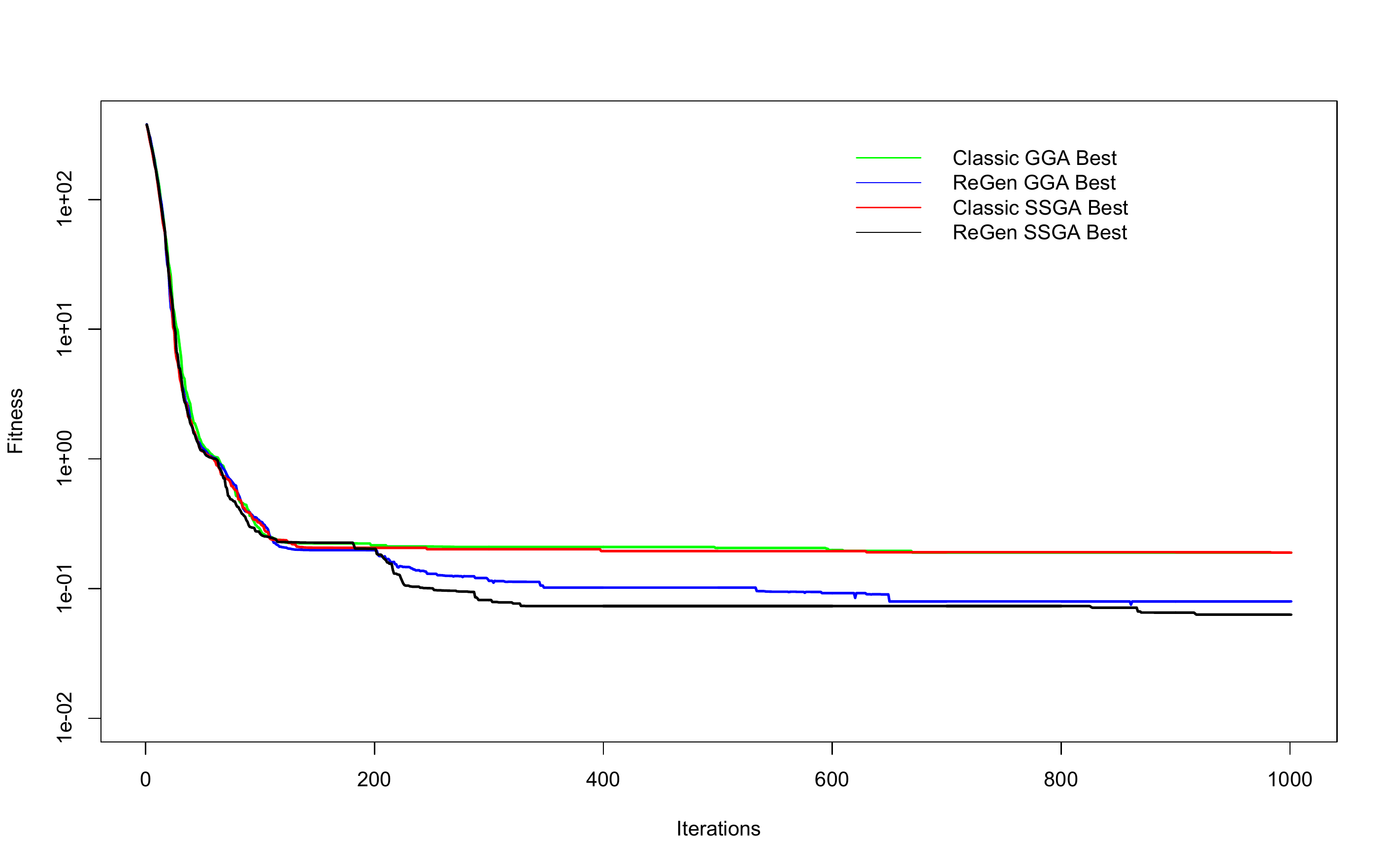}
\includegraphics[width=2in]{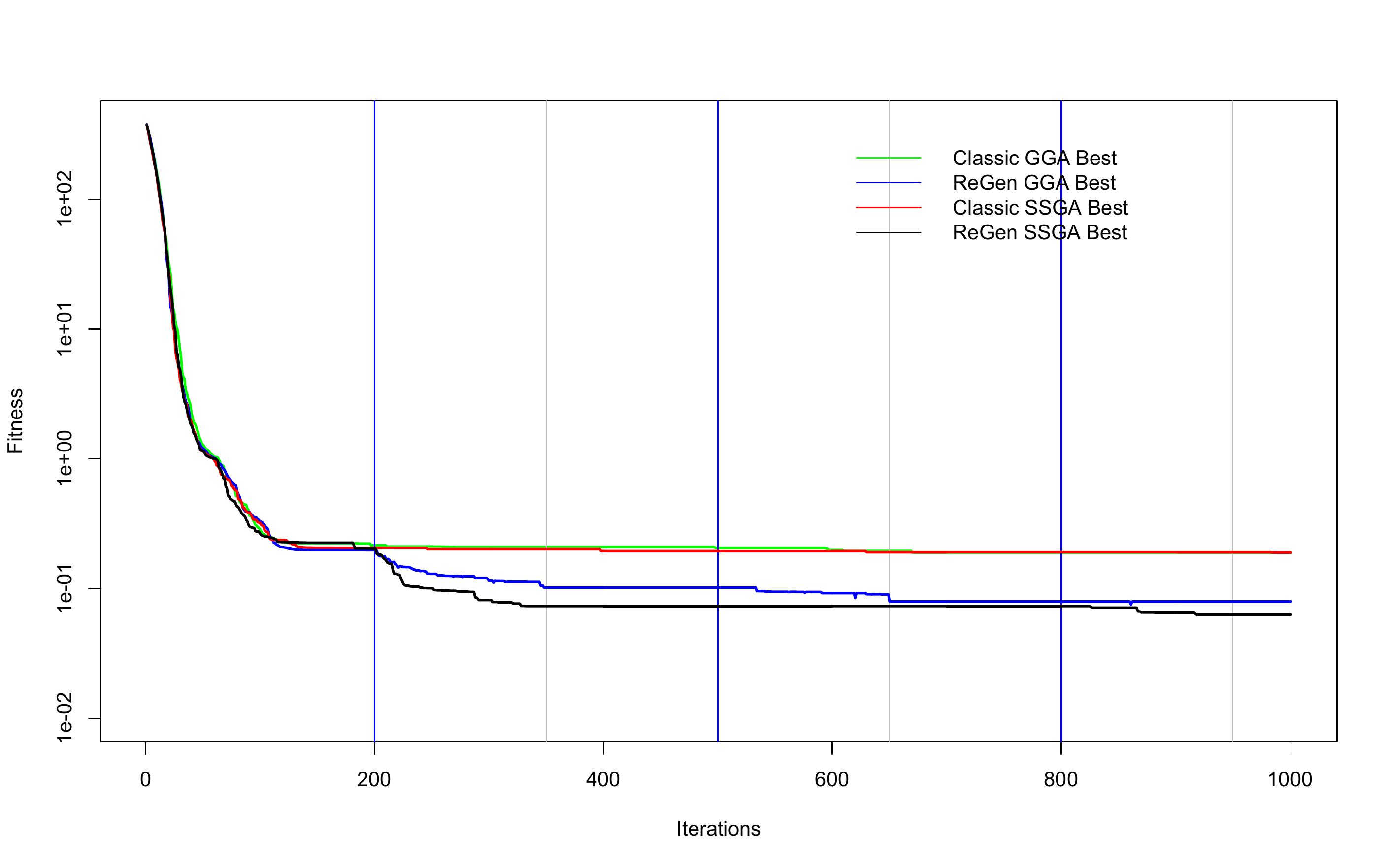}
\includegraphics[width=2in]{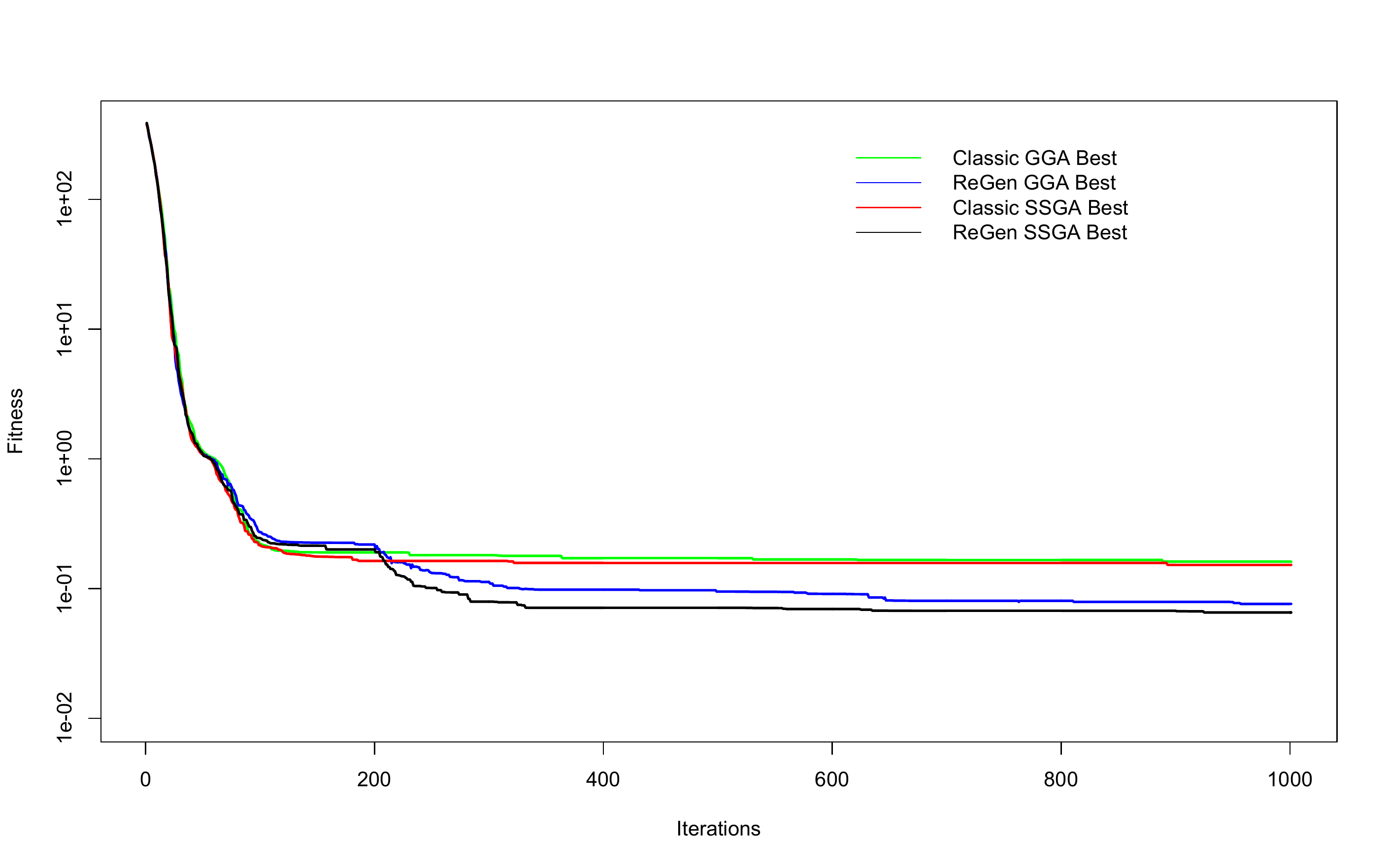}
\includegraphics[width=2in]{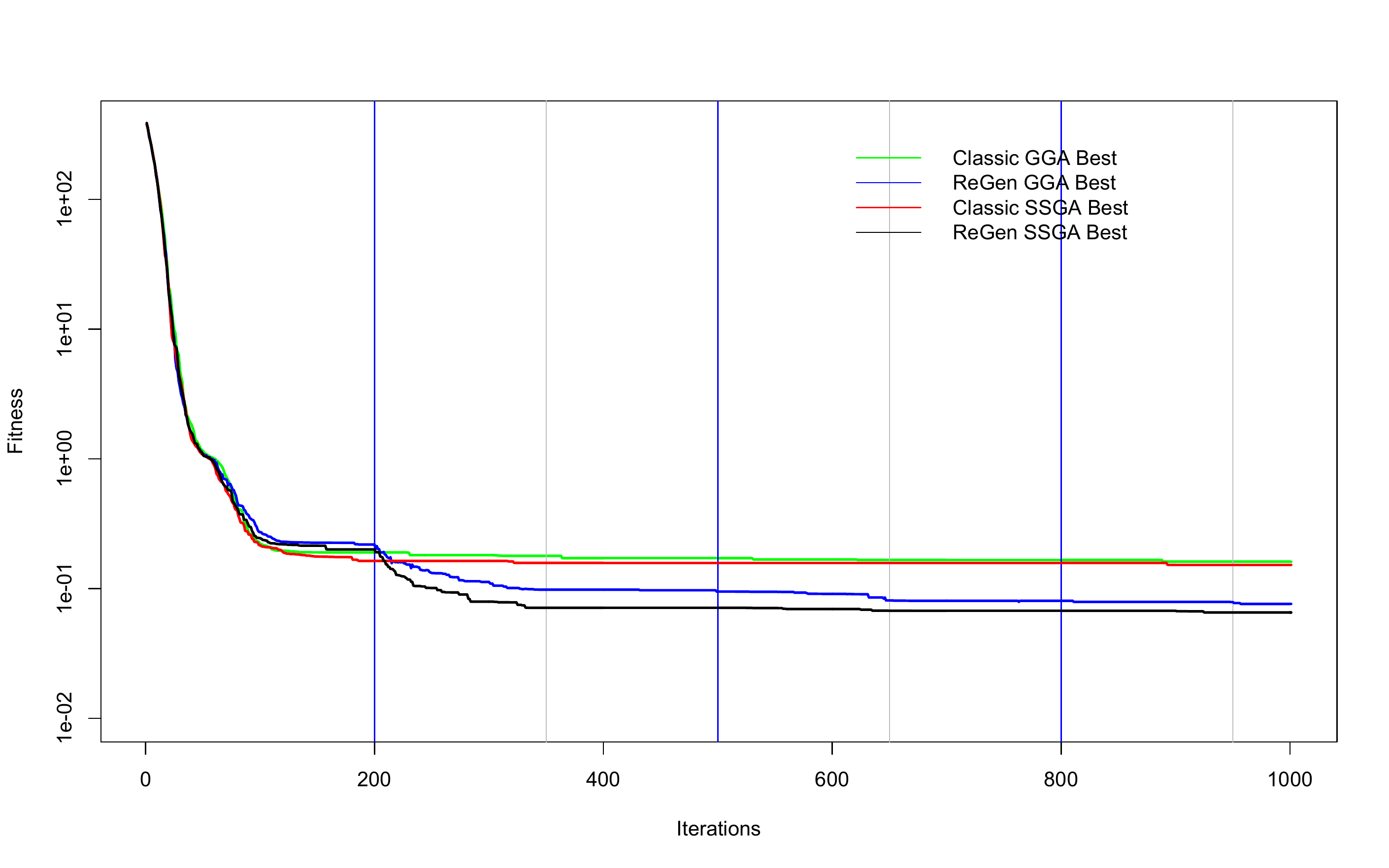}
\includegraphics[width=2in]{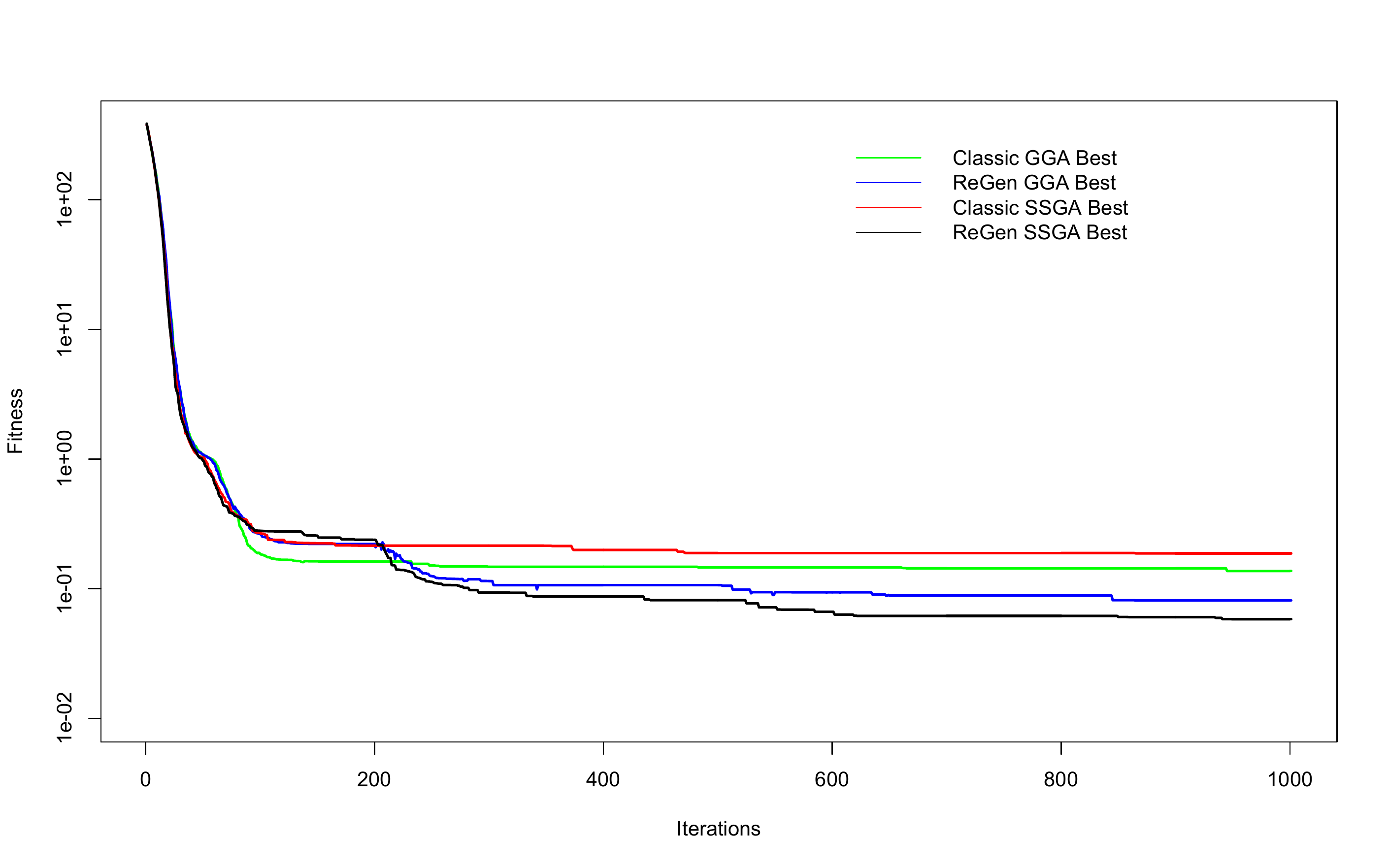}
\includegraphics[width=2in]{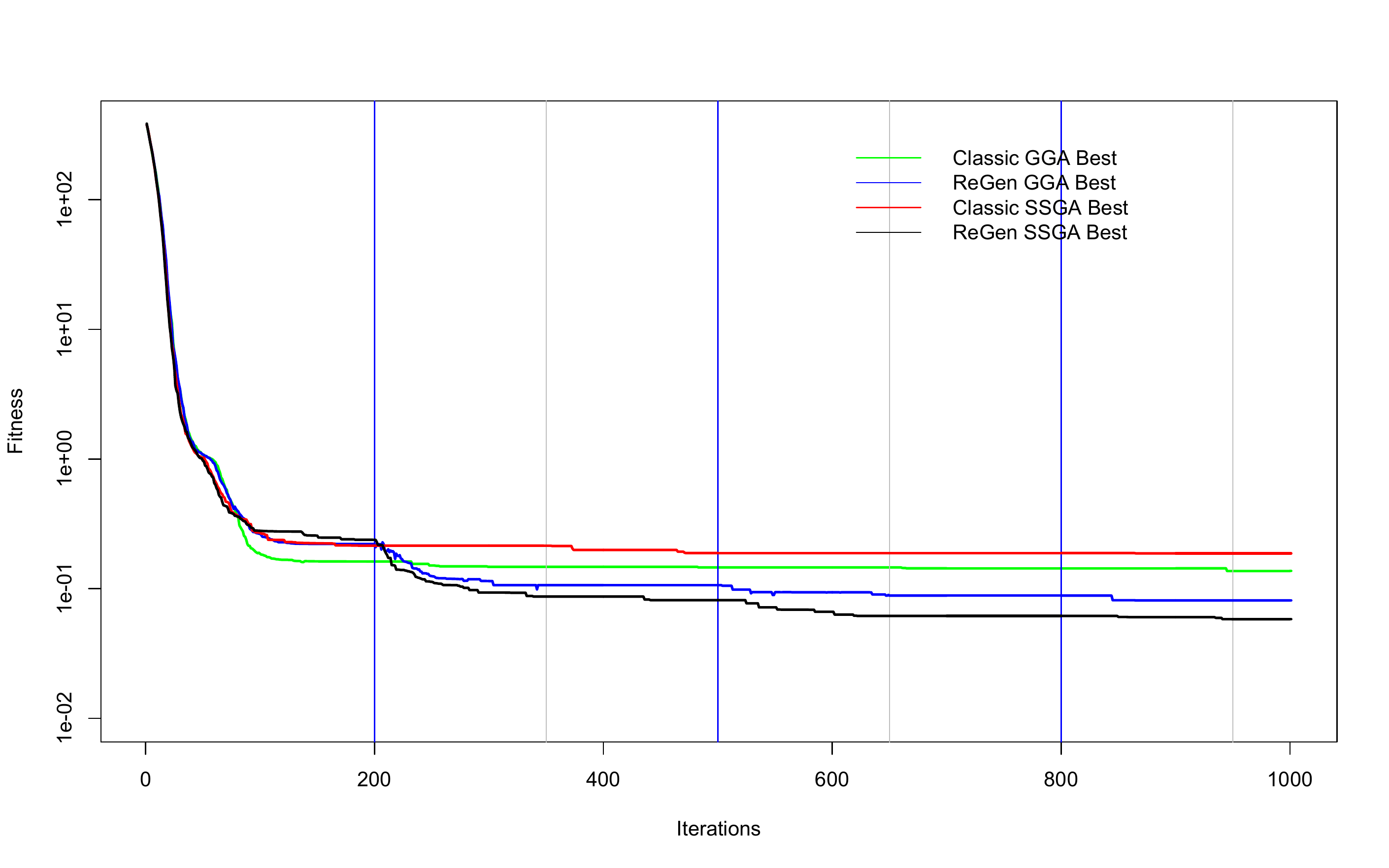}
\caption{Griewank. Generational replacement (GGA) and Steady State replacement (SSGA). From top to bottom, crossover rates from $0.6$ to $1.0$.}
\label{c4fig10}
\end{figure}

Graphs from Fig.~\ref{c4fig7} to Fig.~\ref{c4fig10} illustrate the best individuals' fitness in performed experiments, reported fitnesses are based on the minimum median performance. Each figure shows the tendency of the best individuals per technique. For ReGen GA and Classic GA, two methods are applied: steady state and generational population replacements. The fitness evolution of individuals can be appreciated by tracking green and red lines that depict the best individual's fitness for classic GA. Blue and black lines trace the best individual's fitness for ReGen GA. From top to bottom, each figure displays individuals' behavior with crossover rates from $0.6$ to $1.0$. Figures on the right side show defined marking periods. Vertical lines in blue depict the starting of a marking period, lines in gray delimit the end of such periods.

\newpage

\begin{table}[H]
  \centering
\caption{Results of the experiments for Generational and Steady replacements: Rastrigin}
\label{c4table17}
\begin{tabular}{p{1cm}lllll}
 \hline
\multirow{2}{5cm}{\textbf{Rate}} & \multicolumn{4}{c}{\textbf{Rastrigin}} \\
\cline{2-5} & \textbf{Classic GGA} & \textbf{Classic SSGA} & \textbf{ReGen GGA} & \textbf{ReGen SSGA} \\
\hline
0.6 &  $ 11.018\pm4.43 [998]$  &  $ 11.074 \pm4.42 [990]$ & $ 1.005 \pm0.77 [943]$  & $ 1.011 \pm1.18 [1000]$\\
0.7 &  $ 10.878\pm5.07 [997]$  &  $ 10.909 \pm4.34 [980]$ & $ 0.033 \pm1.00 [947]$  & $ 0.521 \pm0.70 [948]$\\
0.8 &  $ 10.638\pm4.91 [1000]$ &  $ 10.592 \pm3.68 [1000]$& $ 0.025 \pm0.88 [1000]$ & $ 0.031 \pm1.00 [946]$\\
0.9 &  $ 09.748 \pm4.47 [1000]$ &  $ 09.435 \pm3.71  [919]$ & $ 0.030 \pm0.85 [995] $ & $ 0.026 \pm0.72 [951]$\\
1.0 &  $ 08.038 \pm3.88 [1000]$ &  $ 06.422 \pm3.79  [960]$ & $ 0.025 \pm0.67 [964]$  & $ 0.027 \pm0.60 [954]$\\
\hline
\end{tabular}
\end{table}

\begin{table}[H]
  \centering
\caption{Results of the experiments for Generational and Steady replacements: Rosenbrock}
\label{c4table18}
\begin{tabular}{p{1cm}lllll}
 \hline
\multirow{2}{5cm}{\textbf{Rate}} & \multicolumn{4}{c}{\textbf{Rosenbrock}} \\
\cline{2-5} & \textbf{Classic GGA} & \textbf{Classic SSGA} & \textbf{ReGen GGA} & \textbf{ReGen SSGA} \\
\hline

0.6 &  $ 0.473 \pm3.95 [850]$ &  $ 1.035 \pm3.71 [736]$  & $ 0.291 \pm0.42 [1000]$ & $ 0.248 \pm0.86 [987]$\\
0.7 &  $ 0.412 \pm2.79 [804]$ &  $ 0.502 \pm3.45 [1000]$ & $ 0.280 \pm0.19 [938]$  & $ 0.252 \pm0.67 [939]$\\
0.8 &  $ 0.580 \pm5.24 [481]$ &  $ 0.474 \pm2.88 [829]$  & $ 0.251 \pm0.25 [998]$  & $ 0.238 \pm0.46 [999]$\\
0.9 &  $ 0.476 \pm3.92 [753]$ &  $ 0.471 \pm3.77 [732]$  & $ 0.248 \pm0.29 [1000]$ & $ 0.216 \pm0.20 [950]$\\
1.0 &  $ 0.445 \pm2.84 [1000]$ & $ 0.503 \pm3.67 [994]$  & $ 0.169 \pm0.18 [999]$  & $ 0.258 \pm0.31 [1000]$\\
\hline
\end{tabular}
\end{table}

\begin{table}[H]
  \centering
\caption{Results of the experiments for Generational and Steady replacements: Schwefel}
\label{c4table19}
\begin{tabular}{p{1cm}lllll}
 \hline
\multirow{2}{5cm}{\textbf{Rate}} & \multicolumn{4}{c}{\textbf{Schwefel}} \\
\cline{2-5} & \textbf{Classic GGA} & \textbf{Classic SSGA} & \textbf{ReGen GGA} & \textbf{ReGen SSGA} \\
\hline

0.6 &  $161.9\pm179.5 [977]$ &  $ 201.2 \pm117.4 [853]$  & $ 3.6e-4 \pm26.4 [952]$ & $ 7.1e-4 \pm55.0 [952]$\\
0.7 &  $148.9\pm126.5 [979]$ &  $ 148.9 \pm147.5 [915]$  & $ 3.3e-4 \pm44.2 [957]$ & $ 3.2e-4 \pm55.4 [941]$\\
0.8 &  $ 76.20\pm93.80 [879]$ &  $ 118.4 \pm124.4 [982]$  & $ 2.7e-4 \pm35.1 [965]$ & $ 3.1e-4 \pm23.5 [976]$\\
0.9 &  $ 60.90\pm121.6 [889]$ &  $ 118.4 \pm103.1 [995]$  & $ 3.0e-4 \pm10.7 [998]$ & $ 2.9e-4 \pm66.3 [858]$\\
1.0 &  $ 30.40\pm84.6 [1000]$ &  $ 60.90 \pm78.80 [922]$  & $ 2.9e-4 \pm3.4 [1000]$ & $ 2.6e-4 \pm1.40 [880]$\\
\hline
\end{tabular}
\end{table}

\begin{table}[H]
  \centering
\caption{Results of the experiments for Generational and Steady replacements: Griewank}
\label{c4table20}
\begin{tabular}{p{1cm}lllll}
 \hline
\multirow{2}{5cm}{\textbf{Rate}} & \multicolumn{4}{c}{\textbf{Griewank}} \\
\cline{2-5} & \textbf{Classic GGA} & \textbf{Classic SSGA} & \textbf{ReGen GGA} & \textbf{ReGen SSGA} \\
\hline

0.6 &  $ 0.150 \pm0.12 [929]$ &  $ 0.185 \pm0.16 [911]$ & $ 0.069 \pm0.05 [990]$ & $ 0.064 \pm0.04 [910]$\\
0.7 &  $ 0.205 \pm0.09 [973]$ &  $ 0.157 \pm0.09 [977]$ & $ 0.064 \pm0.04 [942]$ & $ 0.057 \pm0.04 [991]$\\
0.8 &  $ 0.189 \pm0.11 [847]$ &  $ 0.189 \pm0.11 [984]$ & $ 0.075 \pm0.06 [861]$ & $ 0.063 \pm0.04 [961]$\\
0.9 &  $ 0.161 \pm0.07 [1000]$&  $ 0.152 \pm0.07 [893]$ & $ 0.076 \pm0.06 [957]$ & $ 0.065 \pm0.05 [1000]$\\
1.0 &  $ 0.136 \pm0.08 [945]$ &  $ 0.187 \pm0.08 [865]$ & $ 0.081 \pm0.04 [944]$ & $ 0.058 \pm0.04 [989]$\\
\hline
\end{tabular}
\end{table}

Based on tabulated results in Table~\ref{c4table17} for Rastrigin function, it can be noted that ReGen GA performs better than the classic GA. ReGen GA is able to discover varied optimal solutions until achieving the total of configured iterations. Even though the global minimum ($0.0$) is not reached, it achieves suitable solutions in general. ReGen GA reports solutions with local minimum under $1.0$; in contrast, classic GA solutions are above $5.0$. In Fig.~\ref{c4fig7} is observable that marking periods applied on chromosomes at iterations $200$, $500$, and $800$ produce significant changes in the evolution of individuals. After starting the first marking period, the fitness improves to be closer to the optimum, and populations improve their performance once tags are added to individuals. Fig.~\ref{c4fig7} also shows that classic GA performance is under ReGen GA performance in all crossover rates levels. Generational and steady replacements performed similarly to this problem.

Tabulated results in Table~\ref{c4table18} for Rosenbrock show that ReGen GA accomplishes better solutions than the classic GA. However, not much difference is evident in results. ReGen GA solutions are a bit closer to the global minimum ($0.0$) than solutions reported by the classic GA. In Fig.~\ref{c4fig8} is noticeable that the pressure applied on chromosomes at iteration $200$ does cause a change in the evolution of individuals. After starting the marking period, the fitness slightly improves, and populations improve their performance once tags are bound. ReGen GA reports better local minima than GA, and generational and steady replacements have almost similar results for all crossover rates.

On the other hand, tabulated results in Table~\ref{c4table19} for Schwefel, evidence that ReGen GA performs much better than the classic GA for current experiments. ReGen GA reports suitable solutions nearer the global minimum ($0.0$) than GA solutions. It can be appreciated that the best solutions are close to the optima for all crossover rates. On the contrary, individuals' fitness for classic GA does not reach the same local optima. Fig.~\ref{c4fig9} remarks that the pressure applied on chromosomes during defined marking periods introduces a great change in the evolution of individuals. After starting the first marking period, the fitness improves to be closer to the optimum, and populations improve their performance once tags are attached. The ReGen GA reaches a variety of good solutions during the evolution process, exposing the ability of the proposed approach to discover novelties that are not identified by the classic GA. Fig.~\ref{c4fig9} also shows that classic GA performance is below ReGen GA performance in all crossover rates levels; the classic GA does not find suitable solutions for this experiment. Generational replacement performed better than the steady replacement for this problem.

As well, tabulated results in Table~\ref{c4table20} for Griewank objective function, show that both ReGen GA and classic GA have a small margin of difference on their performances; still, ReGen GA produces better solutions than the classic GA. Both reached local optima under $1.0$. In Fig.~\ref{c4fig10} is evident that the marking process at iteration $200$ generates a change in the evolution of individuals. After starting the marking period, fitness improves and keeps stable for the best individuals. Both generational and steady replacements performed slightly similar for all crossover rates.

Continuing with the analysis, Table~\ref{c4table21} gives an outline of the best solutions found by ReGen GA as reported previously and best solutions reported by Gomez \cite{GOMEZa, GOMEZb} on GA implementations with Real encoding \footnote{Gomez implements four GAs with Single Point Real Crossover (X), Gaussian (G), and Uniform (U) Mutation as genetic operators in order to compare their performance with \haea. Two generational GAs (GGA(XG) and GGA(XU)), and two steady state GAs (SSGA(XG) and SSGA(XU)). The GAs uses a tournament size of four as parent selection method. For steady state implementations, the worst individual of the population is replaced with the best child generated after crossover and mutation occurs. The reported results are performed with a mutation rate of $0.5$ and  crossover rate of $0.7$.}.

\begin{table}[H]
\centering
  \caption{Solutions found by different EAs on real functions}
  \label{c4table21}
  \begin{tabular}{lllll} \hline
\textbf{EA} & \textbf{Rosenbrock} & \textbf{Schwefel} & \textbf{Rastrigin} & \textbf{Griewank} \\\hline
ReGen GGA & $0.16954\pm0.18$ & $0.00027\pm35.15$ & $0.02539\pm0.67$ & $0.06481\pm0.04$\\
ReGen SSGA & $0.21634\pm0.20$ & $0.00026\pm1.420$ & $0.02669\pm0.72$ & $0.05725\pm0.04$\\
GGA(XU) & $0.17278\pm0.11$ & $2.00096\pm1.210$ & $0.26500\pm0.15$ & $0.63355\pm0.24$\\
GGA(XG) & $0.03852\pm0.03$ & $378.479\pm222.4$ & $12.1089\pm5.01$ & $0.05074\pm0.02$\\
SSGA(XU) & $0.06676\pm0.08$ & $0.88843\pm0.570$ & $0.12973\pm0.07$ & $0.32097\pm0.13$\\
SSGA(XG) & $0.04842\pm0.04$ & $659.564\pm277.3$ & $19.7102\pm7.80$ & $0.04772\pm0.02$\\
Digalakis \cite{DIGALAKIS} & $0.40000000$ & - & $10.000$ & $0.7000$\\
Patton \cite{PATTON} & - & - & $4.8970$ & $0.0043$\\\hline
\end{tabular}
\end{table}

ReGen GA, in general, has better performance for Schwefel (ReGen GGA with crossover $0.8$; ReGen SSGA with crossover $1.0$) and Rastrigin (ReGen GGA with crossover $1.0$; ReGen SSGA with crossover $0.9$) functions. Nevertheless, for Rosenbrock (ReGen GGA with crossover $1.0$; ReGen SSGA with crossover $0.9$) and Griewank (with crossover $0.7$ for both implementations) functions, it obtained suitable solutions but not always better than the ones reported by listed EAs.

\subsection{Statistical Analysis} \label{c4s3ss3}

The statistical analysis presented in this subsection follows the same scheme from binary problems section; therefore, some descriptions are omitted, refer to subsection~\ref{c4s2ss3} for more details. Three different tests are performed, One-Way ANOVA test, Pairwise Student’s t-test, and Paired Samples Wilcoxon Test (also known as Wilcoxon signed-rank test). The data set ReGen EAs Samples in Appendix \ref{appendB} is used, the samples contain twenty EAs implementations for each of the following functions: Ratrigin, Rosenbrock, Schwefel, and Griewank. The samples refer to the best fitness of a solution found in each run, the number of executions per algorithm is $30$. Different implementations involve classic GAs and ReGen GAs with Generational (G) and Steady State (SS) population replacements, and crossover rates from $0.6$ to $1.0$.

\begin{landscape}
\begin{table}[H]
\centering
\caption{Anova Single Factor: SUMMARY}
\label{c4table22}
\scriptsize
\begin{tabular}{llllllllllllll}
\hline
\multicolumn{2}{l}{} & \multicolumn{3}{c}{\textbf{Rastrigin}} & \multicolumn{3}{c}{\textbf{Rosenbrock}} & \multicolumn{3}{c}{\textbf{Schwefel}} & \multicolumn{3}{c}{\textbf{Griewank}} \\
Groups & Count & Sum & Average & Variance & Sum & Average & Variance & Sum & Average & Variance & Sum & Average & Variance \\\hline
GGAX06 & 30 & 340.2427 & 11.3414 & 19.3266 & 52.58683 & 1.75289 & 13.96380 & 5859.204 & 195.3068 & 30075.83 & 5.33366 & 0.17779 & 0.00578 \\
GGAX07 & 30 & 342.8903 & 11.4297 & 24.1491 & 46.71745 & 1.55725 & 5.79244 & 5045.616 & 168.1872 & 14482.14 & 6.11428 & 0.20381 & 0.00946 \\
GGAX08 & 30 & 336.9313 & 11.2310 & 23.8248 & 97.34299 & 3.24477 & 17.80430 & 2816.054 & 93.8685 & 6880.24 & 5.85960 & 0.19532 & 0.01076 \\
GGAX09 & 30 & 310.2216 & 10.3407 & 19.6901 & 69.00354 & 2.30012 & 11.97435 & 2676.372 & 89.2124 & 13007.02 & 5.13625 & 0.17121 & 0.00449 \\
GGAX10 & 30 & 260.1163 & 8.6705 & 14.5260 & 42.00588 & 1.40020 & 7.17257 & 2057.618 & 68.5873 & 5671.65 & 4.28734 & 0.14291 & 0.00598 \\
SSGAX06 & 30 & 352.4735 & 11.7491 & 19.0427 & 77.03410 & 2.56780 & 11.33058 & 6353.236 & 211.7745 & 13329.70 & 7.33354 & 0.24445 & 0.02425 \\
SSGAX07 & 30 & 335.8611 & 11.1954 & 18.6028 & 52.04329 & 1.73478 & 10.39958 & 4691.105 & 156.3702 & 19022.31 & 5.45293 & 0.18176 & 0.00758 \\
SSGAX08 & 30 & 303.3327 & 10.1111 & 13.3066 & 48.09139 & 1.60305 & 7.03069 & 4213.878 & 140.4626 & 15033.45 & 6.57597 & 0.21920 & 0.01325 \\
SSGAX09 & 30 & 292.4265 & 9.7475 & 11.4686 & 62.27272 & 2.07576 & 11.61922 & 3689.997 & 122.9999 & 10995.33 & 5.11276 & 0.17043 & 0.00483 \\
SSGAX10 & 30 & 228.0086 & 7.6003 & 12.6033 & 58.61603 & 1.95387 & 11.33277 & 2557.336 & 85.2445 & 5461.39 & 5.83554 & 0.19452 & 0.00733 \\
ReGenGGAX06 & 30 & 21.1354 & 0.7045 & 0.5303 & 11.54441 & 0.38481 & 0.16875 & 344.900 & 11.4967 & 559.56 & 2.29069 & 0.07636 & 0.00237 \\
ReGenGGAX07 & 30 & 19.2442 & 0.6415 & 0.6273 & 9.33206 & 0.31107 & 0.03705 & 601.931 & 20.0644 & 1538.00 & 1.90437 & 0.06348 & 0.00186 \\
ReGenGGAX08 & 30 & 13.3463 & 0.4449 & 0.6015 & 8.84472 & 0.29482 & 0.06261 & 302.420 & 10.0807 & 673.19 & 2.44422 & 0.08147 & 0.00326 \\
ReGenGGAX09 & 30 & 13.9836 & 0.4661 & 0.5390 & 8.97137 & 0.29905 & 0.08245 & 93.521 & 3.1174 & 104.54 & 2.55778 & 0.08526 & 0.00372 \\
ReGenGGAX10 & 30 & 8.2175 & 0.2739 & 0.2014 & 6.59899 & 0.21997 & 0.03338 & 24.872 & 0.8291 & 11.52 & 2.65118 & 0.08837 & 0.00197 \\
ReGenSSGAX06 & 30 & 24.7609 & 0.8254 & 1.3584 & 16.02581 & 0.53419 & 0.66559 & 891.330 & 29.7110 & 2116.61 & 2.14538 & 0.07151 & 0.00158 \\
ReGenSSGAX07 & 30 & 19.3034 & 0.6434 & 0.4815 & 14.03180 & 0.46773 & 0.40673 & 854.721 & 28.4907 & 2233.54 & 1.96305 & 0.06544 & 0.00216 \\
ReGenSSGAX08 & 30 & 17.8044 & 0.5935 & 0.6824 & 11.06604 & 0.36887 & 0.20173 & 210.062 & 7.0021 & 505.22 & 2.08242 & 0.06941 & 0.00241 \\
ReGenSSGAX09 & 30 & 10.9318 & 0.3644 & 0.3144 & 8.10569 & 0.27019 & 0.03800 & 820.391 & 27.3464 & 2287.48 & 2.08257 & 0.06942 & 0.00260 \\
ReGenSSGAX10 & 30 & 9.6430 & 0.3214 & 0.2819 & 9.05231 & 0.30174 & 0.09498 & 0.00824998 & 0.00027 & 1.41E-09 & 2.05718 & 0.06857 & 0.00186
\\\hline
\end{tabular}
\end{table}

\end{landscape}

Based on the ReGen EAs Samples in Appendix \ref{appendB}, the analysis of variance is computed to know the difference between evolutionary algorithms with different implementations. Variations include classic GAs and ReGen GAs, replacement strategies (Generational and Steady State), and crossover rates from $0.6$ to $1.0$, algorithms are twenty in total. Table~\ref{c4table22} shows a summary for each algorithm and function; the summary presents the number of samples per algorithm ($30$), the sum of the fitness, the average fitness, and their variances. Results of the ANOVA single factor are tabulated in Table~\ref{c4table23}.

\begin{table}[H]
\centering
\caption{Anova Single Factor: ANOVA}
\label{c4table23}
\begin{tabular}{lllllll}
\hline
\multicolumn{7}{c}{\textbf{Rastrigin}} \\
Source of Variation & SS & df & MS & F & P-value & F crit \\\hline
Between Groups & 14947.3266 & 19 & 786.70140 & 86.3753 & 4.8815E-155 & 1.60449 \\
Within Groups & 5282.60586 & 580 & 9.10794 &  &  &  \\
 &  &  &  &  &  &  \\
Total & 20229.9325 & 599 &  &  &  & 
\\\hline
\multicolumn{7}{c}{\textbf{Rosenbrock}} \\
Source of Variation & SS & df & MS & F & P-value & F crit \\\hline
Between Groups & 507.02716 & 19 & 26.68564 & 4.8426 & 1.30194E-10 & 1.60449 \\
Within Groups & 3196.1356 & 580 & 5.51057 &  &  &  \\
 &  &  &  &  &  &  \\
Total & 3703.1628 & 599 &  &  &  &
 \\\hline
\multicolumn{7}{c}{\textbf{Schwefel}} \\
Source of Variation & SS & df & MS & F & P-value & F crit \\\hline
Between Groups & 2832064.8 & 19 & 149056.042 & 20.7038 & 2.4055E-53 & 1.60449 \\
Within Groups & 4175672.6 & 580 & 7199.43552 &  &  &  \\
 &  &  &  &  &  &  \\
Total & 7007737.3 & 599 &  &  &  &
 \\\hline
\multicolumn{7}{c}{\textbf{Griewank}} \\
Source of Variation & SS & df & MS & F & P-value & F crit \\\hline
Between Groups & 2.26223 & 19 & 0.11906 & 20.2641 & 2.6291E-52 & 1.60449 \\
Within Groups & 3.40786 & 580 & 0.00587 &  &  &  \\
 &  &  &  &  &  &  \\
Total & 5.67009 & 599 &  &  &  & 
 \\\hline
\end{tabular}
\end{table}


As P-values for Rastrigin, Rosenbrock, Schwefel, and Griewank functions are less than the significance level $0.05$, results allow concluding that there are significant differences between groups, as shown in Table~\ref{c4table23}. In one-way ANOVA tests, significant P-values indicate that some group means are different, but it is not evident which pairs of groups are different. In order to interpret one-way ANOVA test' results, multiple pairwise-comparison with Student's t-test is performed to determine if the mean difference between specific pairs of the group is statistically significant. Also, paired-sample Wilcoxon tests are computed.

\begin{landscape}
\begin{figure}[H]
\centering
\includegraphics[width=4.2in]{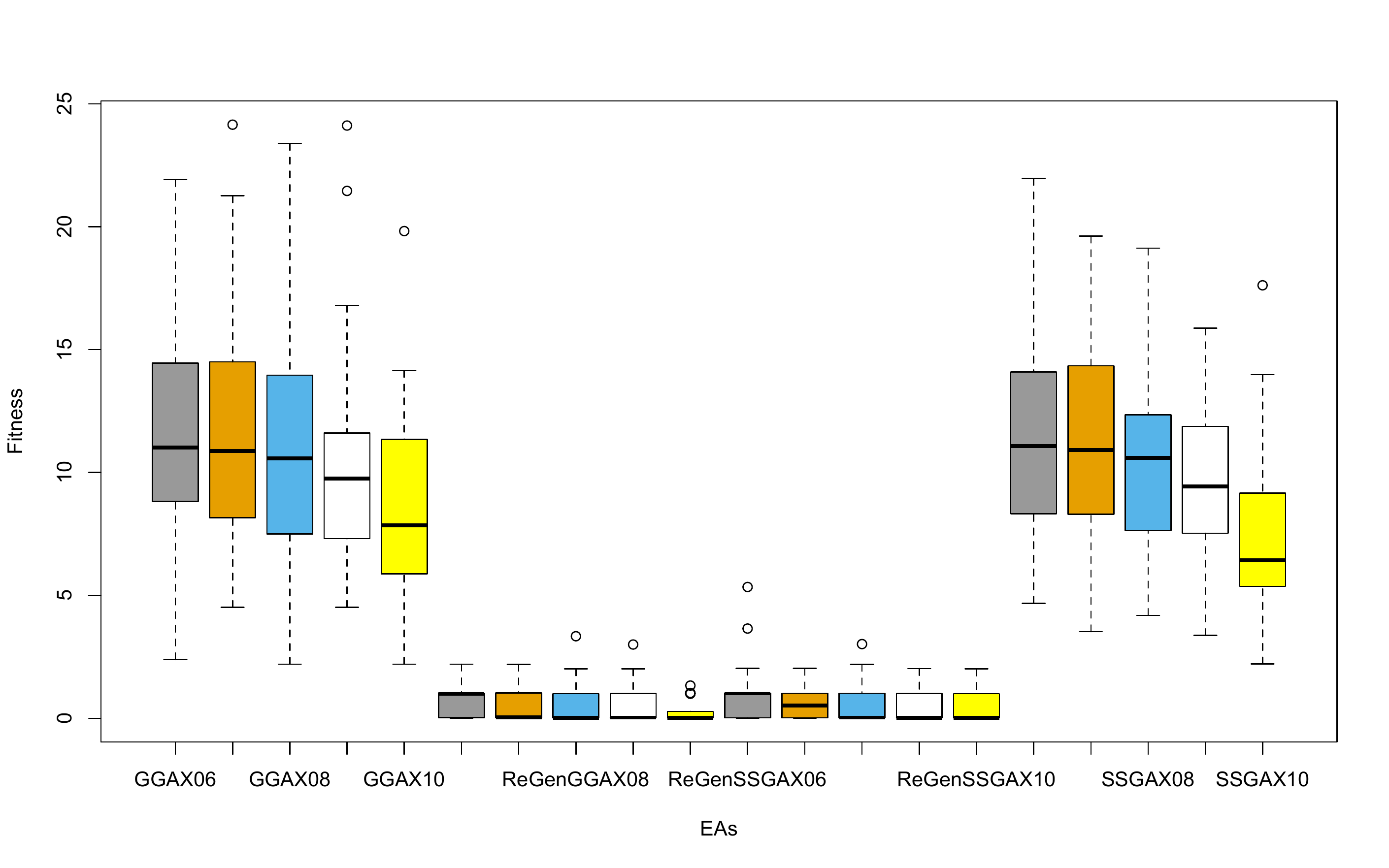}
\includegraphics[width=4.2in]{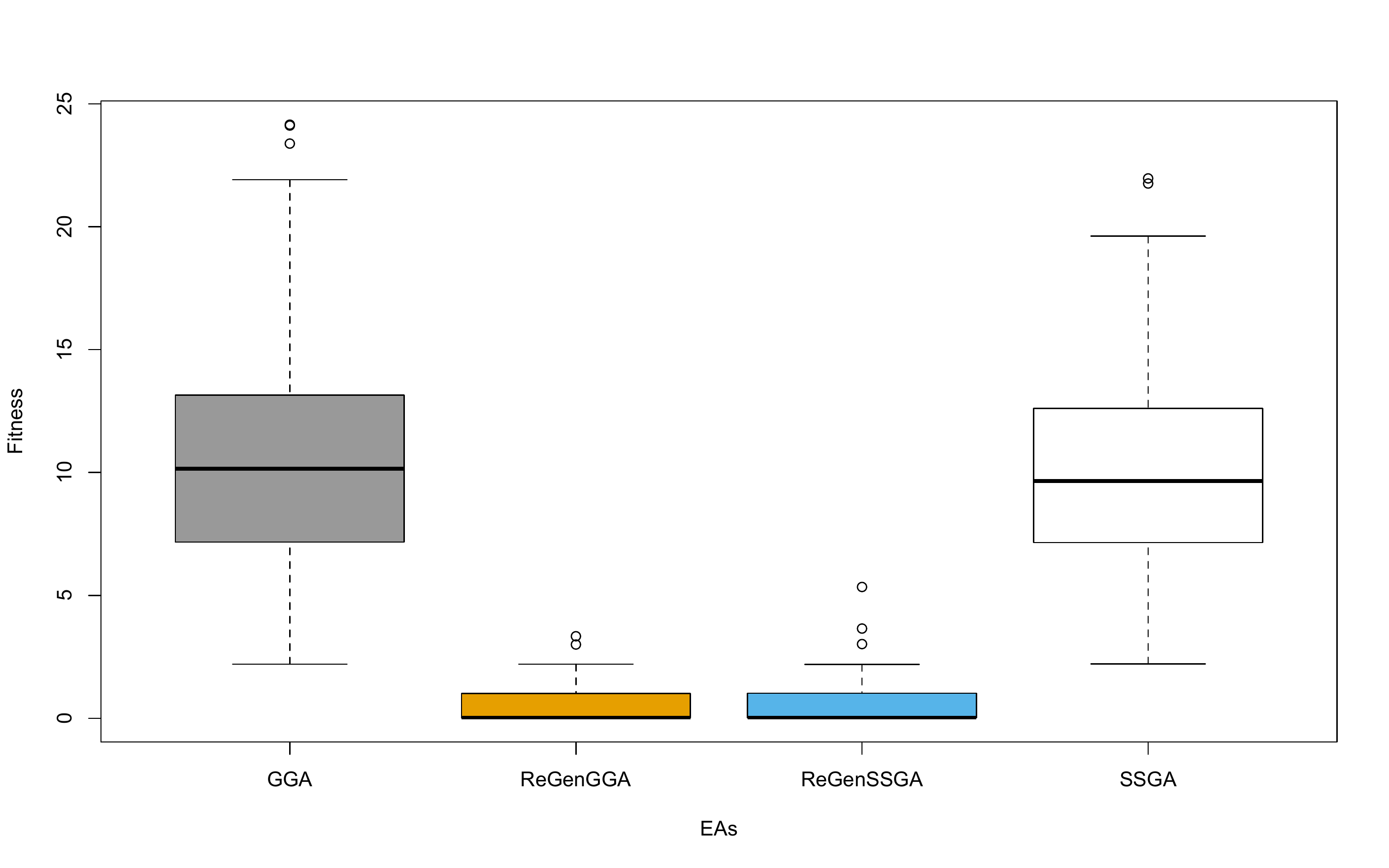}
\includegraphics[width=4.2in]{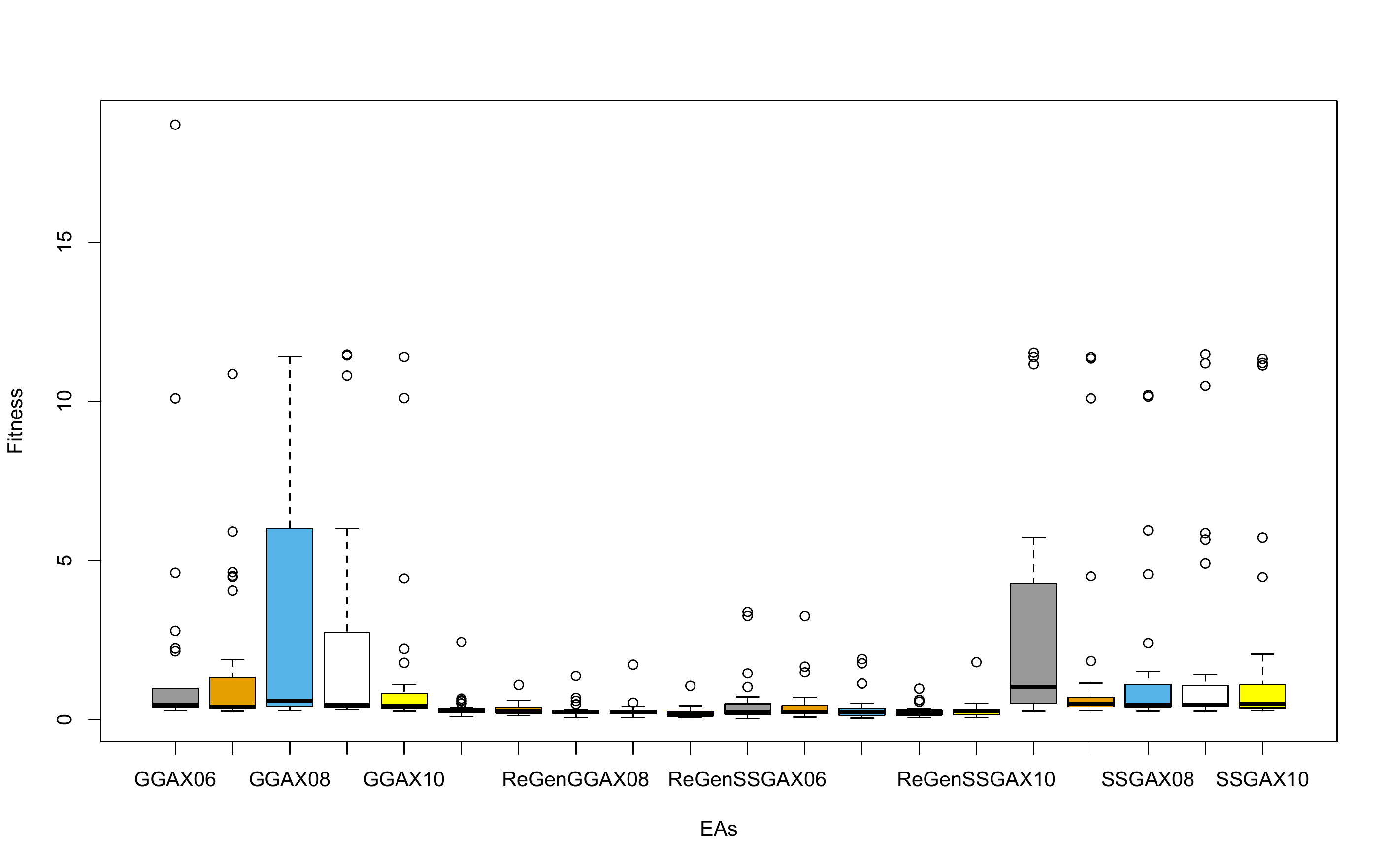}
\includegraphics[width=4.2in]{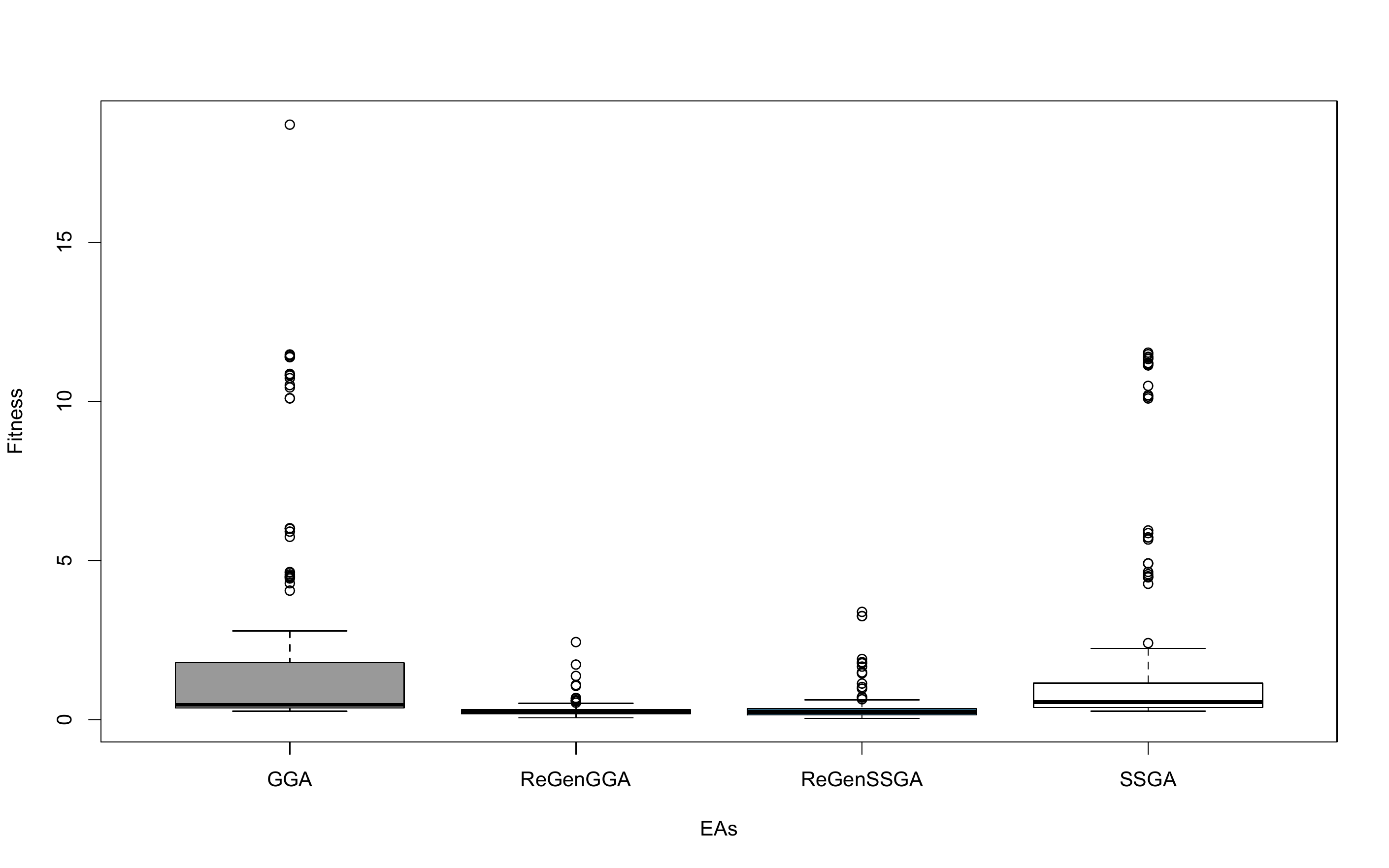}
\caption{From top to bottom: Rastrigin and Rosenbrock Functions. On the left, EAs with Generational replacement (GGA) and Steady State replacement (SSGA) with Crossover rates from $0.6$ to $1.0$. On the right, EAs grouped by Generational replacement (GGA) and Steady State replacement (SSGA).}
\label{c4fig11}
\end{figure}

\begin{figure}[H]
\centering
\includegraphics[width=4.6in]{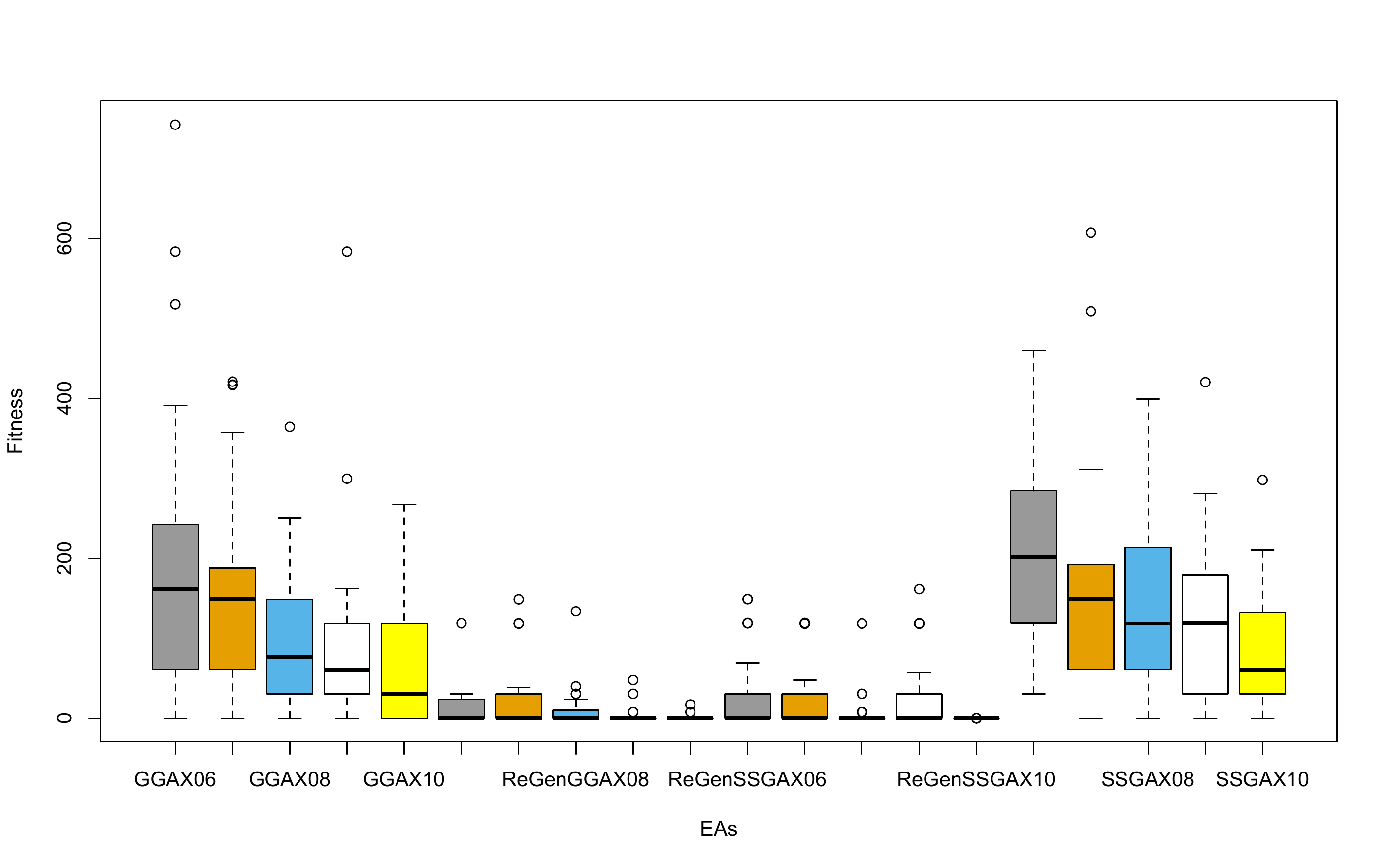}
\includegraphics[width=4.6in]{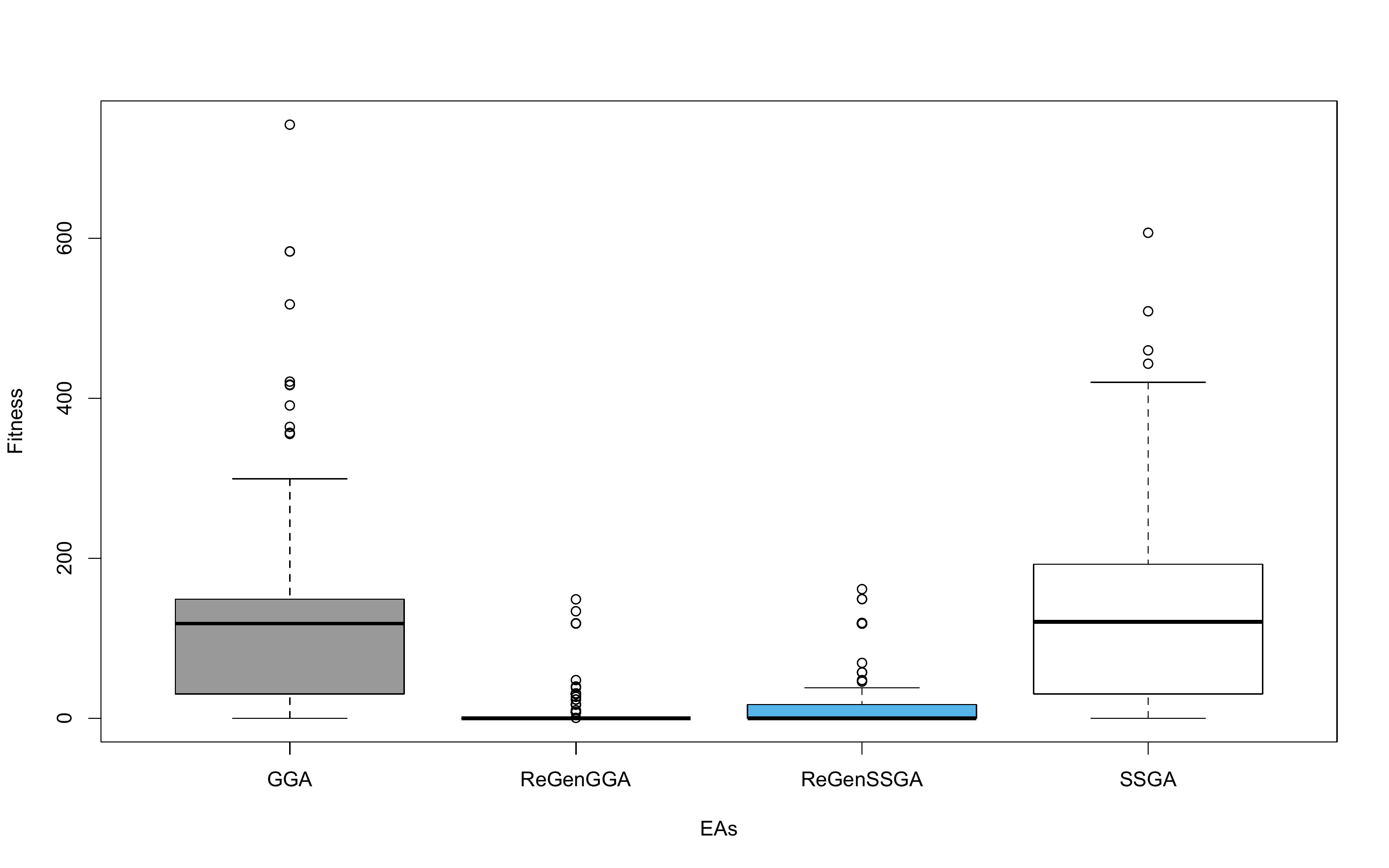}
\includegraphics[width=4.6in]{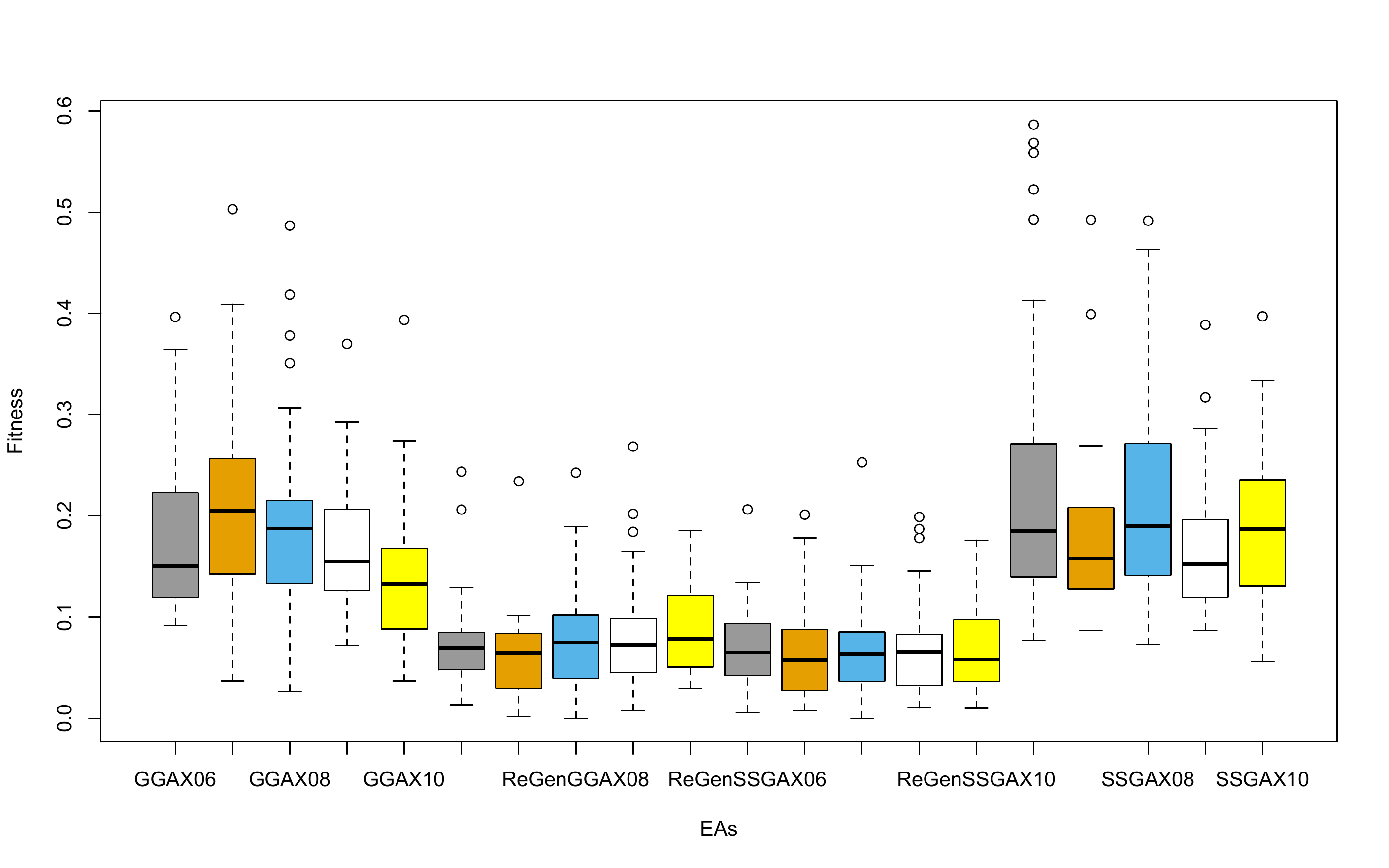}
\includegraphics[width=4.6in]{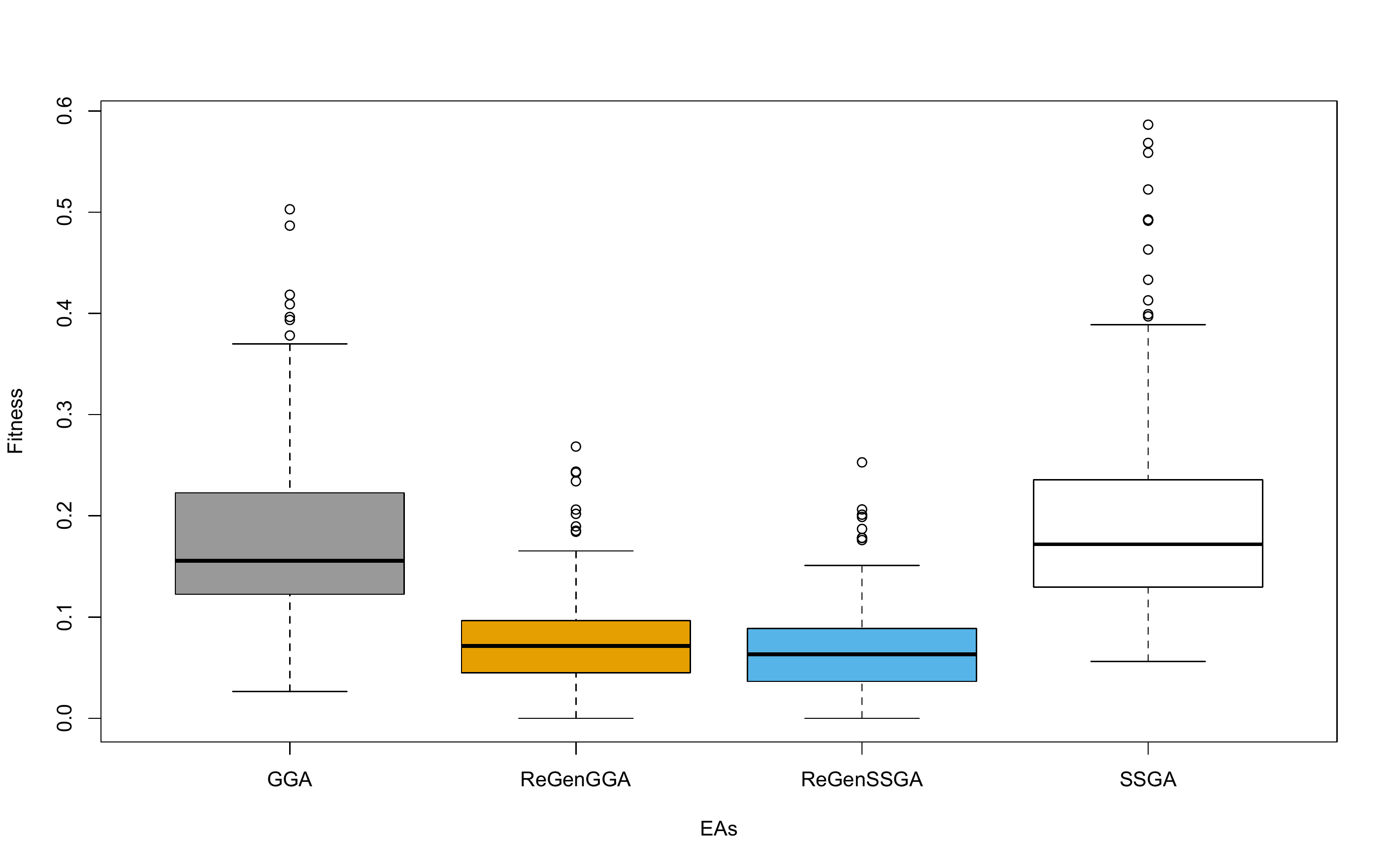}
\caption{From top to bottom: Schwefel and Griewank Functions. On the left, EAs with Generational replacement (GGA) and Steady State replacement (SSGA) with Crossover rates from $0.6$ to $1.0$. On the right, EAs grouped by Generational replacement (GGA) and Steady State replacement (SSGA).}
\label{c4fig12}
\end{figure}
\end{landscape}

Box plots in Fig.~\ref{c4fig11} and Fig.~\ref{c4fig12} depict the median fitness of EAs' best solutions (ReGen EAs Samples in Appendix \ref{appendB}). On the left, twenty EAs' variations with different crossover rates: Gray ($0.6$), Orange ($0.7$), Blue ($0.8$), White ($0.9$), and Yellow ($1.0$). On the right, figures illustrate the median fitness of classic and epigenetic EAs, which are grouped by population replacement type: Gray (GGA), Orange (ReGen GGA), Blue (ReGen SSGA), and White (SSGA). For Rastrigin function, the median fitness for each Epigenetic EA is under the local minima ($1.0$), while median fitnesses for classic GAs are over the local optimum ($5.0$). On the other hand, Rosenbrock's median fitness is less than $0.5$ for all Epigenetic implementations; in contrast, for standard GAs, the median fitness does exceed $1.0$. Epigenetic EAs for Schwefel achieved median fitness inferior to $0.1$; conversely, GAs median fitnesses are greater than $30$. For Griewank function's box plots, depicted median fitnesses are below the local optimum $0.1$ for epigenetic evolutionary algorithms, while traditional GAs median fitness values are above $0.1$. So, based on these data, it seems that Epigenetic GAs find better solutions than classic GAs. However, it is needed to determine whether these findings are statistically significant.

\begin{landscape}
\begin{table}[H]
\centering
\caption{RAS Student T-tests pairwise comparisons with pooled standard deviation. Benjamini Hochberg (BH) as p-value adjustment method.}
\label{c4table24}
\tiny
\begin{tabular}{llllllllllllllllllll}
\hline
EAs & GGAX06 & GGAX07 & GGAX08 & GGAX09 & GGAX10 & ReGenGGAX06 & ReGenGGAX07 & ReGenGGAX08 & ReGenGGAX09 & ReGenGGAX10 \\\hline
GGAX07 & 0.96041246 & - & - & - & - & - & - & - & - & - \\
GGAX08 & 0.95257897 & 0.92910961 & - & - & - & - & - & - & - & - \\
GGAX09 & 0.29168935 & 0.24549691 & 0.36794632 & - & - & - & - & - & - & - \\
GGAX10 & \cellcolor[HTML]{EFEFEF}0.0011367 & \cellcolor[HTML]{EFEFEF}0.00076506 & \cellcolor[HTML]{EFEFEF}0.0018616 & 0.05322976 & - & - & - & - & - & - \\
ReGenGGAX06 & \cellcolor[HTML]{EFEFEF}2.66E-36 & \cellcolor[HTML]{EFEFEF}9.11E-37 & \cellcolor[HTML]{EFEFEF}9.78E-36 & \cellcolor[HTML]{EFEFEF}7.19E-31 & \cellcolor[HTML]{EFEFEF}2.44E-22 & - & - & - & - & - \\
ReGenGGAX07 & \cellcolor[HTML]{EFEFEF}1.24E-36 & \cellcolor[HTML]{EFEFEF}4.69E-37 & \cellcolor[HTML]{EFEFEF}4.64E-36 & \cellcolor[HTML]{EFEFEF}3.45E-31 & \cellcolor[HTML]{EFEFEF}1.23E-22 & 0.97139031 & - & - & - & - \\
ReGenGGAX08 & \cellcolor[HTML]{EFEFEF}1.45E-37 & \cellcolor[HTML]{EFEFEF}6.45E-38 & \cellcolor[HTML]{EFEFEF}4.69E-37 & \cellcolor[HTML]{EFEFEF}3.12E-32 & \cellcolor[HTML]{EFEFEF}1.36E-23 & 0.90599934 & 0.92910961 & - & - & - \\
\hline
\end{tabular}
\end{table}

\begin{table}[H]
\centering
\caption{RAS Student T-tests pairwise comparisons with pooled standard deviation. Benjamini Hochberg (BH) as p-value adjustment method.}
\label{c4table25}
\tiny
\begin{tabular}{lllllllllllllllllll}
\hline
EAs & ReGenSSGAX06 & ReGenSSGAX07 & ReGenSSGAX08 & ReGenSSGAX09 & ReGenSSGAX10 & SSGAX06 & SSGAX07 & SSGAX08 & SSGAX09 \\\hline
GGAX07 & - & - & - & - & - & - & - & - & - \\
GGAX08 & - & - & - & - & - & - & - & - & - \\
GGAX09 & - & - & - & - & - & - & - & - & - \\
GGAX10 & - & - & - & - & - & - & - & - & - \\
ReGenGGAX06 & - & - & - & - & - & - & - & - & - \\
ReGenGGAX07 & - & - & - & - & - & - & - & - & - \\
ReGenGGAX08 & - & - & - & - & - & - & - & - & - \\
\hline
\end{tabular}
\end{table}

\begin{table}[H]
\centering
\caption{ROSE Student T-tests pairwise comparisons with pooled standard deviation. Benjamini Hochberg (BH) as p-value adjustment method.}
\label{c4table26}
\tiny
\begin{tabular}{llllllllllllllllllll}
\hline
EAs & GGAX06 & GGAX07 & GGAX08 & GGAX09 & GGAX10 & ReGenGGAX06 & ReGenGGAX07 & ReGenGGAX08 & ReGenGGAX09 & ReGenGGAX10 \\\hline
GGAX07 & 0.97208658 & - & - & - & - & - & - & - & - & - \\
GGAX08 & \cellcolor[HTML]{EFEFEF}0.04881414 & \cellcolor[HTML]{EFEFEF}0.02555667 & - & - & - & - & - & - & - & - \\
GGAX09 & 0.57153304 & 0.36485797 & 0.20856668 & - & - & - & - & - & - & - \\
GGAX10 & 0.83562136 & 0.97544713 & \cellcolor[HTML]{EFEFEF}0.01549184 & 0.23648078 & - & - & - & - & - & - \\
ReGenGGAX06 & 0.06582199 & 0.10908846 & \cellcolor[HTML]{EFEFEF}7.08E-05 & \cellcolor[HTML]{EFEFEF}0.01126426 & 0.16925516 & - & - & - & - & - \\
ReGenGGAX07 & 0.05376269 & 0.08887445 & \cellcolor[HTML]{EFEFEF}5.27E-05 & \cellcolor[HTML]{EFEFEF}0.00799463 & 0.13573914 & 0.99094875 & - & - & - & - \\
ReGenGGAX08 & 0.05286811 & 0.08631238 & \cellcolor[HTML]{EFEFEF}5.27E-05 & \cellcolor[HTML]{EFEFEF}0.00788007 & 0.13321453 & 0.99094875 & 0.9964496 & - & - & - \\
\hline
\end{tabular}
\end{table}

\begin{table}[H]
\centering
\caption{ROSE Student T-tests pairwise comparisons with pooled standard deviation. Benjamini Hochberg (BH) as p-value adjustment method.}
\label{c4table27}
\tiny
\begin{tabular}{lllllllllllllllllll}
\hline
EAs & ReGenSSGAX06 & ReGenSSGAX07 & ReGenSSGAX08 & ReGenSSGAX09 & ReGenSSGAX10 & SSGAX06 & SSGAX07 & SSGAX08 & SSGAX09 \\\hline
GGAX07 & - & - & - & - & - & - & - & - & - \\
GGAX08 & - & - & - & - & - & - & - & - & - \\
GGAX09 & - & - & - & - & - & - & - & - & - \\
GGAX10 & - & - & - & - & - & - & - & - & - \\
ReGenGGAX06 & - & - & - & - & - & - & - & - & - \\
ReGenGGAX07 & - & - & - & - & - & - & - & - & - \\
ReGenGGAX08 & - & - & - & - & - & - & - & - & - \\
ReGenGGAX09 & - & - & - & - & - & - & - & - & - \\
ReGenGGAX10 & - & - & - & - & - & - & - & - & - \\
ReGenSSGAX06 & - & - & - & - & - & - & - & - & - \\
ReGenSSGAX07 & 0.99094875 & - & - & - & - & - & - & - & - \\
ReGenSSGAX08 & 0.97544713 & 0.99094875 & - & - & - & - & - & - & - \\
ReGenSSGAX09 & 0.93354867 & 0.97208658 & 0.99094875 & - & - & - & - & - & - \\
ReGenSSGAX10 & 0.95886064 & 0.97544713 & 0.99094875 & 0.9964496 & - & - & - & - & - \\
SSGAX06 & \cellcolor[HTML]{EFEFEF}0.00745631 & \cellcolor[HTML]{EFEFEF}0.00569756 & \cellcolor[HTML]{EFEFEF}0.00347469 & \cellcolor[HTML]{EFEFEF}0.00256288 & \cellcolor[HTML]{EFEFEF}0.00256288 & - & - & - & - \\
SSGAX07 & 0.10151582 & 0.085759 & 0.06582199 & 0.05236195 & 0.05376269 & 0.2855976 & - & - & - \\
SSGAX08 & 0.14452907 & 0.12182725 & 0.09212144 & 0.07258523 & 0.07947888 & 0.19702636 & 0.99094875 & - & - \\
SSGAX09 & \cellcolor[HTML]{EFEFEF}0.04186549 & \cellcolor[HTML]{EFEFEF}0.03244222 & \cellcolor[HTML]{EFEFEF}0.02448551 & \cellcolor[HTML]{EFEFEF}0.01789277 & \cellcolor[HTML]{EFEFEF}0.01918161 & 0.63931284 & 0.83884205 & 0.66236082 & - \\
SSGAX10 & 0.05531127 & \cellcolor[HTML]{EFEFEF}0.04910282 & \cellcolor[HTML]{EFEFEF}0.03549543 & \cellcolor[HTML]{EFEFEF}0.02555667 & \cellcolor[HTML]{EFEFEF}0.02790432 & 0.4973954 & 0.96054387 & 0.83562136 & 0.99094875
\\\hline
\end{tabular}
\end{table}

\begin{table}[H]
\centering
\caption{SCHW Student T-tests pairwise comparisons with pooled standard deviation. Benjamini Hochberg (BH) as p-value adjustment method.}
\label{c4table28}
\tiny
\begin{tabular}{llllllllllllllllllll}
\hline
EAs & GGAX06 & GGAX07 & GGAX08 & GGAX09 & GGAX10 & ReGenGGAX06 & ReGenGGAX07 & ReGenGGAX08 & ReGenGGAX09 & ReGenGGAX10 \\\hline
GGAX07 & 0.29349563 & - & - & - & - & - & - & - & - & - \\
GGAX08 & \cellcolor[HTML]{EFEFEF}1.32E-05 & \cellcolor[HTML]{EFEFEF}0.00154568 & - & - & - & - & - & - & - & - \\
GGAX09 & \cellcolor[HTML]{EFEFEF}4.96E-06 & \cellcolor[HTML]{EFEFEF}0.00074085 & 0.88288414 & - & - & - & - & - & - & - \\
GGAX10 & \cellcolor[HTML]{EFEFEF}4.79E-08 & \cellcolor[HTML]{EFEFEF}1.91E-05 & 0.32852579 & 0.44530689 & - & - & - & - & - & - \\
ReGenGGAX06 & \cellcolor[HTML]{EFEFEF}5.04E-15 & \cellcolor[HTML]{EFEFEF}1.89E-11 & \cellcolor[HTML]{EFEFEF}0.00042857 & \cellcolor[HTML]{EFEFEF}0.00089786 & \cellcolor[HTML]{EFEFEF}0.01566357 & - & - & - & - & - \\
ReGenGGAX07 & \cellcolor[HTML]{EFEFEF}7.65E-14 & \cellcolor[HTML]{EFEFEF}2.05E-10 & \cellcolor[HTML]{EFEFEF}0.00165974 & \cellcolor[HTML]{EFEFEF}0.00325874 & \cellcolor[HTML]{EFEFEF}0.04229624 & 0.76682348 & - & - & - & - \\
ReGenGGAX08 & \cellcolor[HTML]{EFEFEF}3.60E-15 & \cellcolor[HTML]{EFEFEF}1.27E-11 & \cellcolor[HTML]{EFEFEF}0.00034065 & \cellcolor[HTML]{EFEFEF}0.0007295 & \cellcolor[HTML]{EFEFEF}0.01320684 & 0.9634304 & 0.73372792 & - & - & - \\
\hline
\end{tabular}
\end{table}

\begin{table}[H]
\centering
\caption{SCHW Student T-tests pairwise comparisons with pooled standard deviation. Benjamini Hochberg (BH) as p-value adjustment method.}
\label{c4table29}
\tiny
\begin{tabular}{lllllllllllllllllll}
\hline
EAs & ReGenSSGAX06 & ReGenSSGAX07 & ReGenSSGAX08 & ReGenSSGAX09 & ReGenSSGAX10 & SSGAX06 & SSGAX07 & SSGAX08 & SSGAX09 \\\hline
GGAX07 & - & - & - & - & - & - & - & - & - \\
GGAX08 & - & - & - & - & - & - & - & - & - \\
GGAX09 & - & - & - & - & - & - & - & - & - \\
GGAX10 & - & - & - & - & - & - & - & - & - \\
ReGenGGAX06 & - & - & - & - & - & - & - & - & - \\
ReGenGGAX07 & - & - & - & - & - & - & - & - & - \\
ReGenGGAX08 & - & - & - & - & - & - & - & - & - \\
\hline
\end{tabular}
\end{table}

\begin{table}[H]
\centering
\caption{GRIE Student T-tests pairwise comparisons with pooled standard deviation. Benjamini Hochberg (BH) as p-value adjustment method.}
\label{c4table30}
\tiny
\begin{tabular}{llllllllllllllllllll}
\hline
EAs & GGAX06 & GGAX07 & GGAX08 & GGAX09 & GGAX10 & ReGenGGAX06 & ReGenGGAX07 & ReGenGGAX08 & ReGenGGAX09 & ReGenGGAX10 \\\hline
GGAX07 & 0.29214033 & - & - & - & - & - & - & - & - & - \\
GGAX08 & 0.510415 & 0.78361424 & - & - & - & - & - & - & - & - \\
GGAX09 & 0.83155928 & 0.15842627 & 0.32989657 & - & - & - & - & - & - & - \\
GGAX10 & 0.12649056 & \cellcolor[HTML]{EFEFEF}0.00396273 & \cellcolor[HTML]{EFEFEF}0.01463119 & 0.24076026 & - & - & - & - & - & - \\
ReGenGGAX06 & \cellcolor[HTML]{EFEFEF}1.06E-06 & \cellcolor[HTML]{EFEFEF}1.54E-09 & \cellcolor[HTML]{EFEFEF}1.51E-08 & \cellcolor[HTML]{EFEFEF}4.91E-06 & \cellcolor[HTML]{EFEFEF}0.00153207 & - & - & - & - & - \\
ReGenGGAX07 & \cellcolor[HTML]{EFEFEF}5.06E-08 & \cellcolor[HTML]{EFEFEF}4.12E-11 & \cellcolor[HTML]{EFEFEF}4.61E-10 & \cellcolor[HTML]{EFEFEF}2.45E-07 & \cellcolor[HTML]{EFEFEF}0.0001398 & 0.6486839 & - & - & - & - \\
ReGenGGAX08 & \cellcolor[HTML]{EFEFEF}3.48E-06 & \cellcolor[HTML]{EFEFEF}5.70E-09 & \cellcolor[HTML]{EFEFEF}5.65E-08 & \cellcolor[HTML]{EFEFEF}1.57E-05 & \cellcolor[HTML]{EFEFEF}0.00365564 & 0.87028325 & 0.49701859 & - & - & - \\
\hline
\end{tabular}
\end{table}

\begin{table}[H]
\centering
\caption{GRIE Student T-tests pairwise comparisons with pooled standard deviation. Benjamini Hochberg (BH) as p-value adjustment method.}
\label{c4table31}
\tiny
\begin{tabular}{lllllllllllllllllll}
\hline
EAs & ReGenSSGAX06 & ReGenSSGAX07 & ReGenSSGAX08 & ReGenSSGAX09 & ReGenSSGAX10 & SSGAX06 & SSGAX07 & SSGAX08 & SSGAX09 \\\hline
GGAX07 & - & - & - & - & - & - & - & - & - \\
GGAX08 & - & - & - & - & - & - & - & - & - \\
GGAX09 & - & - & - & - & - & - & - & - & - \\
GGAX10 & - & - & - & - & - & - & - & - & - \\
ReGenGGAX06 & - & - & - & - & - & - & - & - & - \\
ReGenGGAX07 & - & - & - & - & - & - & - & - & - \\
ReGenGGAX08 & - & - & - & - & - & - & - & - & - \\
\hline
\end{tabular}
\end{table}
\end{landscape}

\paragraph{\em{Multiple pairwise t-test:}}
Multiple pairwise-comparison between means of groups is performed. In the one-way ANOVA test described above, significant p-values indicate that some group means are different. In order to know which pairs of groups are different, multiple pairwise-comparison is performed for Rastrigin (RAS), Rosenbrock (ROSE), Schwefel (SCHW), and Griewank (GRIE) best solutions samples. Tables (\ref{c4table24}, \ref{c4table25}, \ref{c4table26}, \ref{c4table27}, \ref{c4table28}, \ref{c4table29}, \ref{c4table30}, and \ref{c4table31}) present Pairwise comparisons using t-tests with pooled standard deviation (SD) with their respective p-values. The test adjusts p-values with the Benjamini-Hochberg method. Pairwise comparisons show that only highlighted values in gray between two algorithms are significantly different ($p < 0.05$). Therefore, the alternative hypothesis is true.

Now, to find out any significant difference between the median fitness of individuals in the two experimental groups (classic GAs and GAs with regulated genes), the Wilcoxon test is conducted.

\paragraph{\em{Paired Samples Wilcoxon Test:}}  
For this test, algorithms are grouped per population replacement strategy, ignoring crossover rates. Wilcoxon signed rank test for generational EAs (GGA and ReGen GGA) and Wilcoxon signed rank test for steady state EAs (SSGA and ReGen SSGA). The test assesses classic EAs versus Epigenetic EAs. In the results, $V$ represents the total of the ranks assigned to differences with a positive sign, and {\em P-value} refers to the probability value. In statistical hypothesis testing, the p-value corresponds to the probability of obtaining test results as evidence to reject or confirm the null hypothesis.

\begin{itemize}
    \item  Rastrigin (RAS)
    \begin{enumerate}
      \item \par Wilcoxon signed rank test with continuity correction for generational EAs uses all data-set samples from GGAs and ReGen GGAs. $V = 11325$, {\em P-value} is equal to $2.322841e-26$, which is less than the significance level alpha ($0.05$).
      
      \item \par Wilcoxon signed rank test with continuity correction for steady state EAs uses all data-set samples from SSGAs and ReGen SSGAs. $V = 11325$, {\em P-value} is equal to $2.322841e-26$, which is less than the significance level $alpha = 0.05$.
      \end{enumerate}
      
    \item Rosenbrock (ROSE)
    \begin{enumerate}
      \item \par Wilcoxon signed rank test with continuity correction for generational EAs uses all data-set samples from GGAs and ReGen GGAs. $V = 10368$, {\em P-value} is equal to $1.068438e-18$, which is less than the significance level alpha ($0.05$).
      
      \item \par Wilcoxon signed rank test with continuity correction for steady state EAs uses all data-set samples from SSGAs and ReGen SSGAs. $V = 10114$, {\em P-value} is equal to $6.760613e-17$, which is less than the significance level $alpha = 0.05$.
      \end{enumerate}
      
      \newpage
      
    \item Schwefel (SCHW)
    \begin{enumerate}
      \item \par Wilcoxon signed rank test with continuity correction for generational EAs uses all data-set samples from GGAs and ReGen GGAs. $V = 11121$, {\em P-value} is equal to $1.305395e-24$, which is less than the significance level $alpha = 0.05$.
      
      \item \par Wilcoxon signed rank test with continuity correction for steady state EAs uses all data-set samples from SSGAs and ReGen SSGAs. $V = 10913$, {\em P-value} is equal to $6.836875e-23$, which is less than the significance level alpha ($0.05$). 
      \end{enumerate}
      
    \item Griewank (GRIE)
    \begin{enumerate}
      \item \par Wilcoxon signed rank test with continuity correction for generational EAs uses all data-set samples from GGAs and ReGen GGAs. $V = 10438$, {\em P-value} is equal to $3.275069e-19$, which is less than the significance level $alpha = 0.05$.
      
      \item \par Wilcoxon signed rank test with continuity correction for steady state EAs uses all data-set samples from SSGAs and ReGen SSGAs. $V = 10975$, {\em P-value} is equal to $2.134437e-23$, which is less than the significance level alpha ($0.05$). 
      \end{enumerate}
      
\end{itemize}

The above leads to conclude that median fitnesses of solutions found by classic generational genetic algorithms (GGAs) are significantly different from median fitnesses of solutions found by generational genetic algorithms with regulated genes (ReGen GGAs) with p-values equal to $2.322841e-26$ (RAS samples), $1.068438e-18$ (ROSE samples), $1.305395e-24$ (SCHW samples), and $3.275069e-19$ (GRIE samples). So, the alternative hypothesis is true.

The median fitness of solutions found by classic steady state genetic algorithms (SSGAs) is significantly different from the median fitness of solutions found by steady state genetic algorithms with regulated genes (ReGen SSGAs) with p-values equal to $2.322841e-26$ (RAS sampling fitness), $6.760613e-17$ (ROSE sampling fitness), $6.836875e-23$ (SCHW sampling fitness), and $2.134437e-23$ (GRIE sampling fitness). As p-values are less than the significance level $0.05$, it may be concluded that there are significant differences between the two EAs groups in each Wilcoxon Test.

\section{Summary}\label{c4s4}

The epigenetic technique is implemented on GAs to solve both binary and real encoding problems. For real encoding, the search space must be discretized by using a binary representation of real values. A decoding schema from binary to real value is performed in order to evaluate individuals' fitness. Results have shown that the marking process does impact the way the population evolves, and the fitness of individuals considerably improves to the optimum. The use of epigenetic tags revealed that they help the ReGen GA to find better solutions (although the optimum is not always reached). A better exploration and exploitation of the search space is evident; in addition, {\em Tags} are transmitted through generations, which leads to maintaining a notion of memory between generations. The statistical analysis helps to conclude that epigenetic implementations performed better than standard versions.
    \chapter{ReGen HAEA: Binary and Real Codification}\label{chapter5}

The Hybrid Adaptive Evolutionary Algorithm with Regulated Genes (ReGen \haea) is the implementation of the proposed epigenetic model on the standard \haea. This implementation is meant to address real and binary encoding problems. Experimental functions with binary and real encoding have been selected for determining the model applicability. In section~\ref{c5s1}, general settings for all experiments are described. In section~\ref{c5s2}, two binary experiments are presented, performing Deceptive order three and Deceptive order four trap functions to evidence tags effect on populations' behavior. Also, some experimental results and their analysis are exhibited in subsection~\ref{c5s2ss1} and subsection~\ref{c5s2ss2}. In section~\ref{c5s3}, three Real encoding problems are presented, implementing Rastrigin, Schwefel, and Griewank functions. Additionally, some experimental results and their analysis are exhibited in subsections~\ref{c5s3ss1} and~\ref{c5s3ss2}. In section~\ref{c5s4}, the statistical analysis of the results is described. At the end of this chapter, a summary is given in section~\ref{c5s5}.

Gomez in \cite{GOMEZa, GOMEZb} proposed an evolutionary algorithm that adapts operator rates while it is solving the optimization problem. \haea is a mixture of ideas borrowed from evolutionary strategies, decentralized control adaptation, and central control adaptation. Algorithm~\ref{c5algo1} presents the pseudo-code of \haea with the embedded epigenetic components.

\begin{algorithm}[htb!]%
\caption{Hybrid Adaptive Evolutionary Algorithm (\haea)}\label{c5algo1}%
$\haea(\text{fitness},\mu,\text{terminationCondition})$%
\begin{algorithmic}[1]%
   \State $t = 0$
   \State $P_0 = \textsc{initPopulation}(\mu)$
   \State $\text{evaluate}(P_0,\text{fitness})$
   \While {\big($\text{terminationCondition}(t,P_t,\text{fitness})$ is false\big)}
   		\State $P_{t+1} = \varnothing$
   		\For{\textbf{each} ind $\in P_t$}
			\State rates = extracRatesOper(ind)
			\State oper = \textsc{OpSelect}(operators, rates)
			\State parents = \textsc{ParentsSelection}\big($P_t$, ind, arity(oper)\big)
			\State offspring = apply(oper, parents)
			\If {\Call{markingPeriodON}{t}}
                \State \Call{applyMarking}{offspring}
            \EndIf
            \State offspring $\gets$ \textsc{decode}(\Call{epiGrowingFunction}{offspring})
            
            \If {steady}
            \State child = \textsc{Best}(offspring, ind)
            \Else
            \State child = \textsc{Best}(offspring)
            \EndIf
			\State $\delta = \text{random}(0,1)$ \Comment{learning rate}
			\If{$\big(\text{fitness}(\text{child}) > \text{fitness}(\text{ind})\big)$} 
			    \State rates[oper] = $(1.0 + \delta)\ *\ $rates[oper] \Comment{reward}
			\Else 
			    \State rates[oper] = $(1.0 - \delta)\ *\ $rates[oper] \Comment{punish}
			\EndIf
			\State normalizeRates(rates)
			\State setRates(child, rates)
			\State $P_{t+1} = P_{t+1}\ \cup$ \{child\}
   		\EndFor
   	    \State $t = t+1$
   	\EndWhile
\end{algorithmic}
\end{algorithm}


As can be noted, \haea does not generate a parent population to produce the next generation. Among the offspring produced by the genetic operator, only one individual is chosen as a child (lines 16 and 18) and will take the place of its parent in the next population (line 28). In order to be able to preserve competent individuals through evolution, \haea compares the parent individual against the offspring generated by the operator, for steady state replacement. For generational replacement, it chooses the best individual among the offspring (lines 15 and 17). 




At line 11, the marking period function has been embedded to initiate the marking process on individuals when defined periods are activated. Then, the epigenetic growing function (line 14) interprets markers on the chromosome structure of individuals with the purpose of generating phenotypes that will be evaluated by the objective function.

For all experiments in this chapter, three marking periods have been defined. There is not a particular reason why this number of periods has been chosen. Marking periods could be between different ranges of iterations, defined periods are just for testing purposes. Marking periods can be appreciated in figures of reported results delineated with vertical lines. Vertical lines in blue depict the starting point of marking periods and gray lines, the end of them.

\section{General Experimental Settings}\label{c5s1}

Following experimental settings apply to binary and real experiments reported in sections \ref{c5s2} and \ref{c5s3}. For the standard \haea, a population size of 100 is used and 1000 iterations. A tournament size of $4$ is implemented to select the parent of crossover. Reported results are the median over 30 different runs. For current tests, \haea only uses one genetic operator combination: the single-point mutation and the single-point crossover. The mutation operator always modifies the genome by randomly changing only one single bit with uniform distribution. The single-point crossover splits and combines parents' chromosome sections (left and right) using a randomly selected cutting point. The set up for the standard \haea also includes: generational (G\haea) and steady state (SS\haea) replacements to choose the fittest individuals for the new population. 

The ReGen \haea setup involves the same defined configuration for the standard \haea with an additional configuration for the epigenetic process as follows: a marking probability of $0.02$ (the probability to add a tag is $0.35$, to remove a tag is $0.35$, and to modify a tag is $0.3$) and three marking periods.

\section{Binary Problems}\label{c5s2}

Binary encoding experiments have been performed in order to determine the proposed approach applicability. In binary encoding, a vector with binary values encodes the problem's solution.

\subsection{Test Functions}\label{c5s2ss1}

Two well known binary problems deceptive order three and deceptive order four trap functions developed by Goldberg in 1989 \cite{GOLDBERG} have been selected. The genome length for each function is $360$, the global optimum for Deceptive order three is $3600$, and $450$ for Deceptive order four trap. A complete definition of these functions can be found in previous chapter \ref{chapter4}, section \ref{c4s2}. Also, in chapter \ref{chapter4}, a more in-depth explanation can be found regarding the implemented binary to real decoding mechanism.

\subsection{Results}\label{c5s2ss2}

Based on the defined configuration, both \haea and ReGen \haea are compared to identify tags' behavior during individuals' evolution. Results are tabulated in Table~\ref{c5table1}, the table presents binary functions: Deceptive order three and four with generational (G\haea) and steady state (SS\haea) replacements for standard and epigenetic implementations. Also, the table shows the best fitness based on the maximum median performance, following the standard deviation of the observed value, and the iteration where the reported fitness is found, which is enclosed in square brackets.

\begin{figure}[H]
\centering
\includegraphics[width=2in]{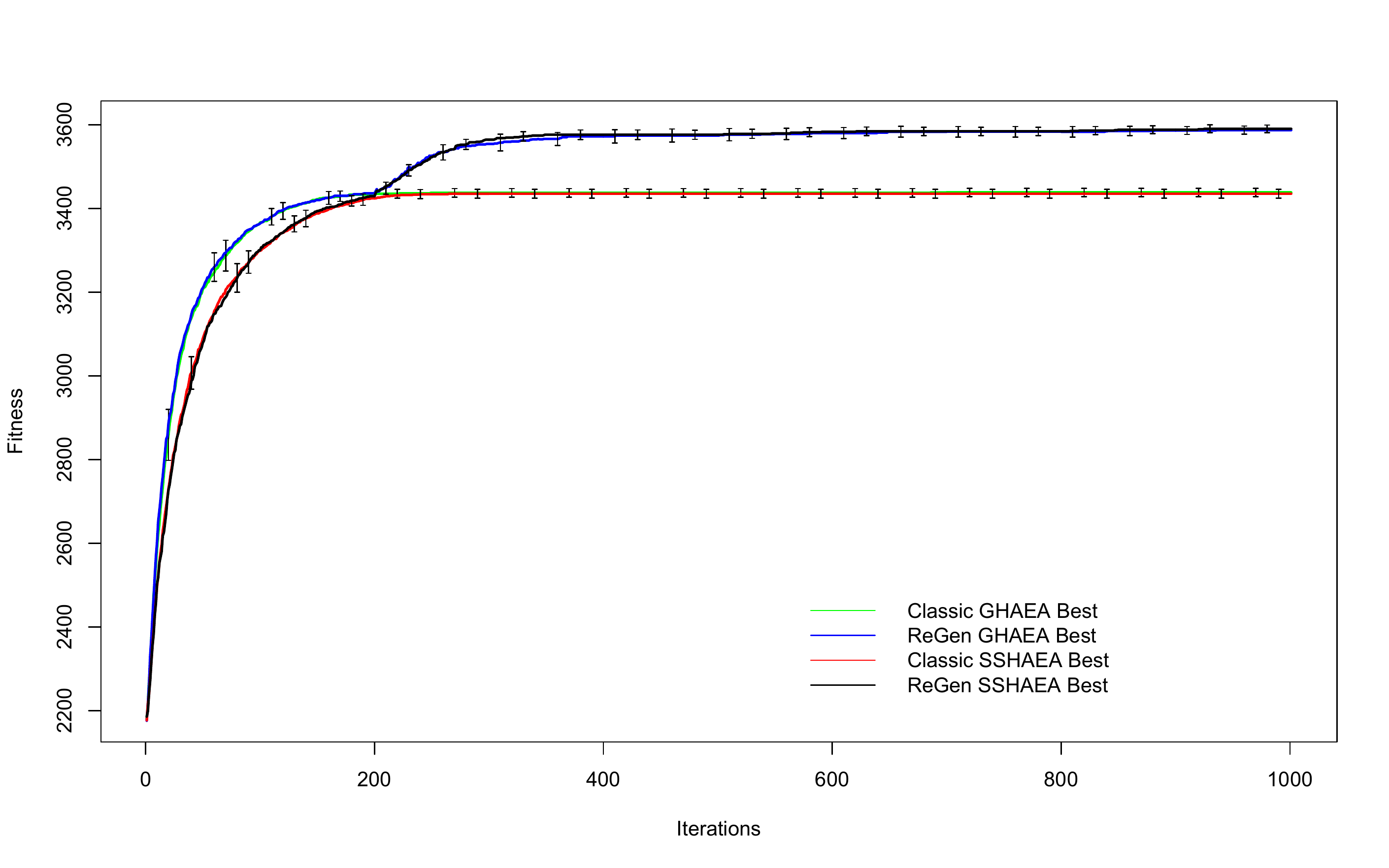}
\includegraphics[width=2in]{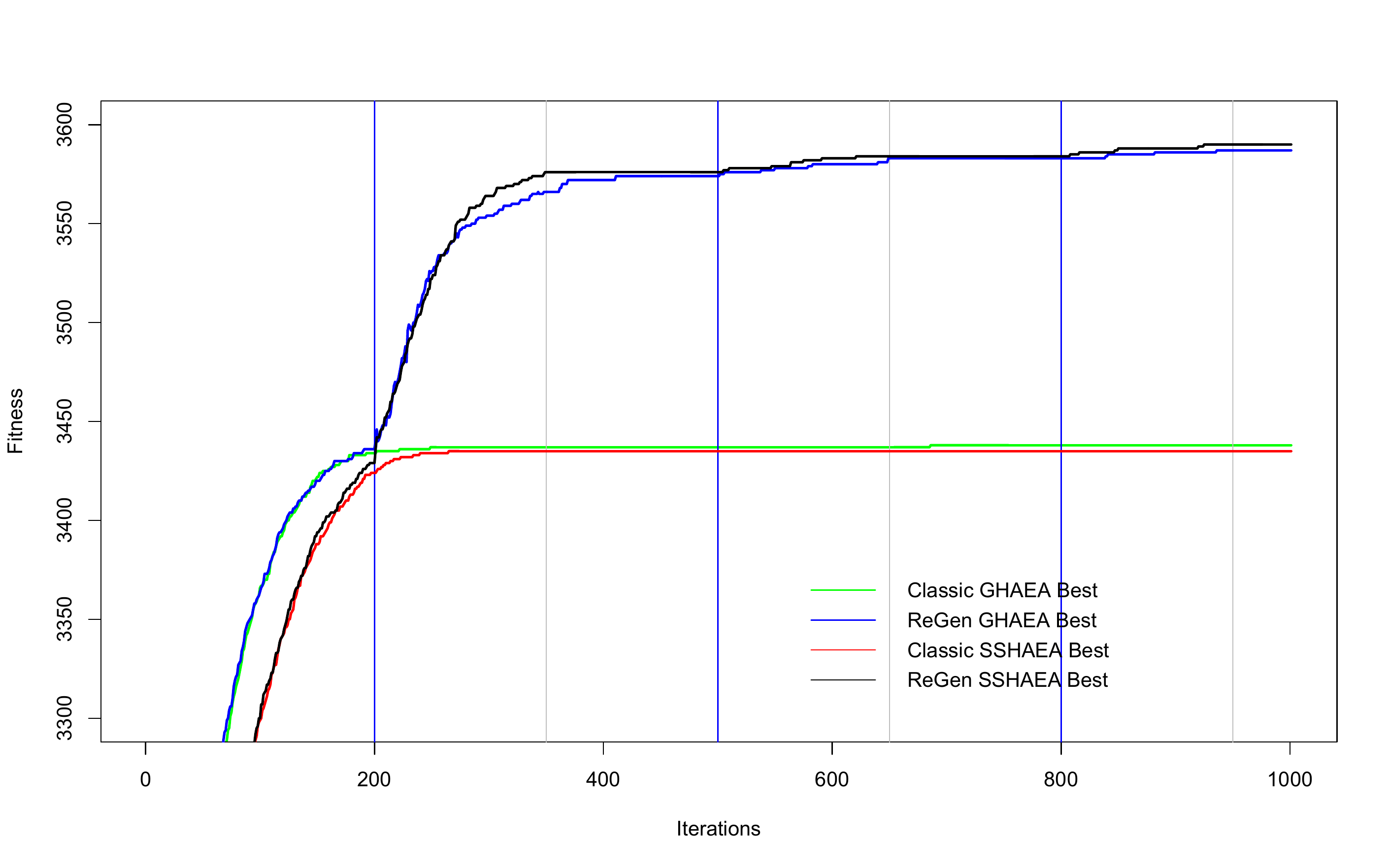}
\caption{Deceptive Order 3. Generational replacement (G\haea) and Steady state replacement (SS\haea).}
\label{c5fig1}
\end{figure}

\begin{figure}[H]
\centering
\includegraphics[width=2in]{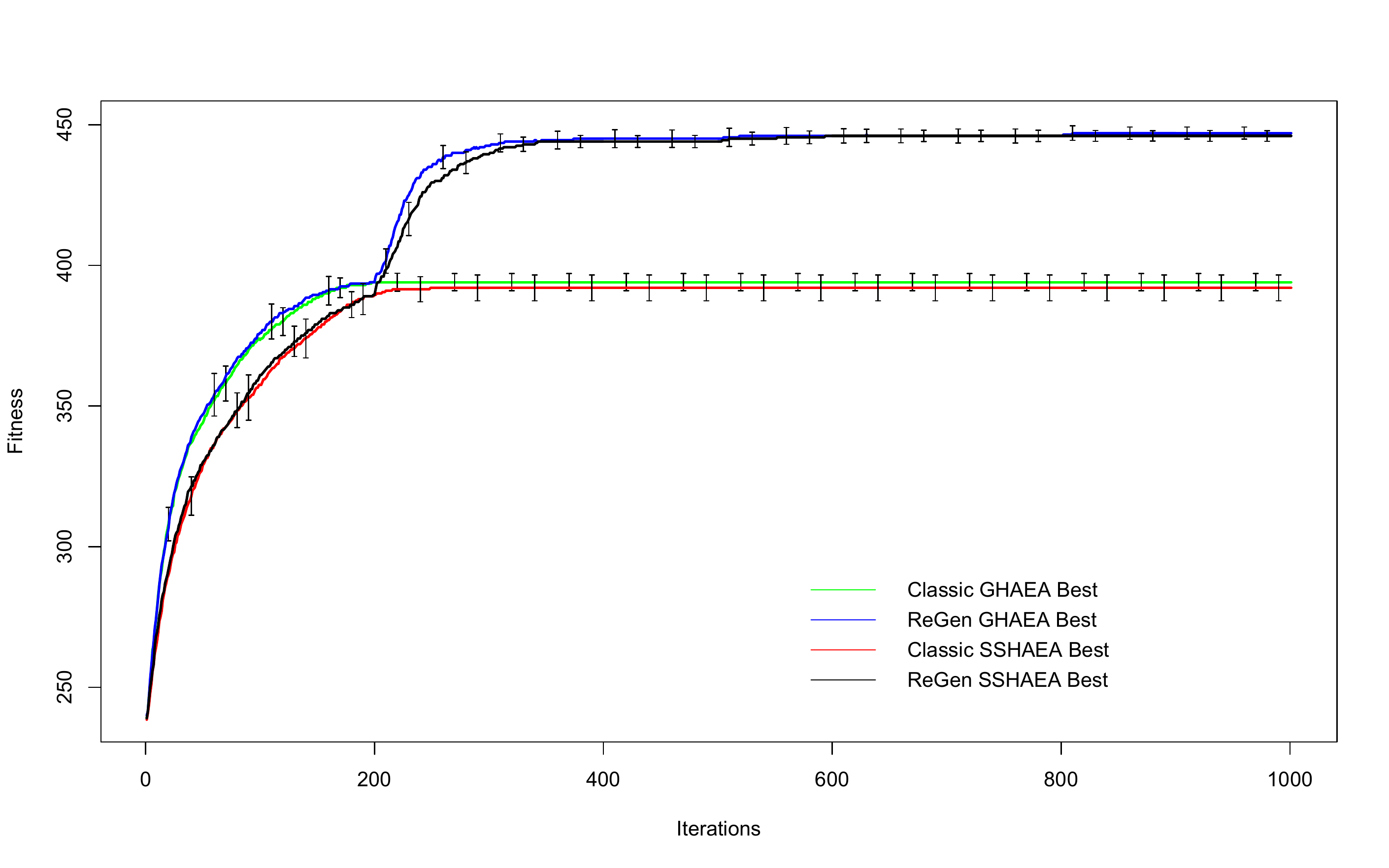}
\includegraphics[width=2in]{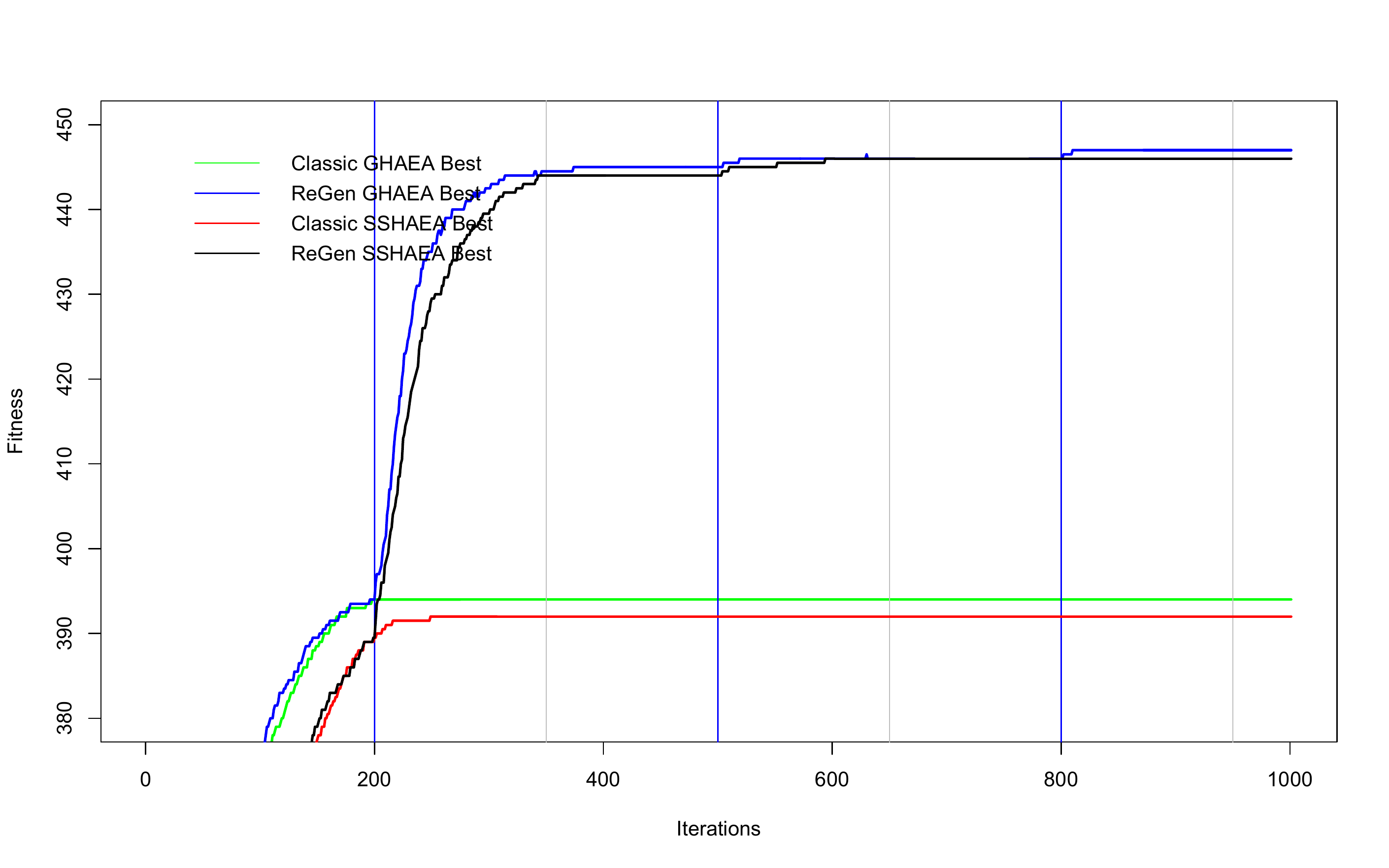}
\caption{Deceptive Order 4. Generational replacement (G\haea) and Steady state replacement (SS\haea).}

\label{c5fig2}
\end{figure}

Fig.~\ref{c5fig1} and Fig.~\ref{c5fig2} illustrate the fitness of best individuals in performed experiments, reported fitnesses are based on the maximum median performance. Each figure shows the tendency of the best individuals per technique. For \haea and ReGen \haea, two methods are applied: steady state and generational population replacements. The fitness evolution of individuals can be appreciated by tracking green and red lines that depict best individuals' fitness for the standard \haea. Blue and black lines trace best individuals' fitness for ReGen \haea. Figures on the right side show defined marking periods. Vertical lines in blue depict the starting of a marking period, lines in gray delimit the end of such periods. 

\begin{table}[H]
\centering
  \caption{Results of the experiments for Generational and Steady state replacements}
  \label{c5table1}
  \begin{tabular}{llll} \hline
      \textbf{EA} & \textbf{Deceptive Order 3} & \textbf{Deceptive Order 4} \\\hline
      G\haea  & $3438 \pm 10.16 [686]$ &  $394 \pm 3.16 [198]$\\
      SS\haea & $3435 \pm 10.96 [265]$ &  $392 \pm 4.55 [249]$\\ 
      ReGen G\haea  & $3587 \pm 09.89 [936]$ & $447 \pm 2.56 [810]$\\
      ReGen SS\haea & $3590 \pm 09.26 [925]$ & $446 \pm 2.39 [594]$\\ \hline
\end{tabular}
\end{table}

Tabulated results in Table~\ref{c5table1} for Deceptive Order Three and Deceptive Order Four Trap, show that ReGen \haea performs better than standard \haea implementations. ReGen \haea is able to discover varied optimal solutions until achieving the total of configured iterations, even though, it did not find the global optimum in performed experiments. In Fig.~\ref{c5fig1} and Fig.~\ref{c5fig2} is notable that the pressure applied on chromosomes at iteration $200$ does cause a change in the evolution of individuals. After starting the marking period, populations improve their performance once tags are added. Following the marking period at iteration $500$, a slight change is identified again, the individuals' fitness improves to some degree closer to the optimal solution. The same behavior can be appreciated in the next period, at iteration $800$.

\section{Real Problems}\label{c5s3}

Experiments using real definition are performed for determining the proposed technique applicability. For the selected problems with real coded definition, a vector with binary values encodes the problem's solution.

\subsection{Test Functions}\label{c5s3ss1}

Real functions shown in Table~\ref{c5table2} are used as testbeds. Each real value is represented with a binary string of 32-bits, for each function, the dimension of the problem is fixed to $n = 10$. A complete definition of these functions can be found in previous chapter \ref{chapter4}, section \ref{c4s3}. Also, a detailed description of the encoding/decoding scheme to obtain real values from binary strings of $32$ bits and its representation as integer numbers are presented in the same section.

\begin{table}[H]
\centering
\caption{Real functions tested}
\label{c5table2}
\begin{tabular}{ccccl}
\hline
 Name & Function & Feasible Region\\
\hline
   Rastrigin & 
   \large $f(\textbf{x}) = 10n + \sum_{i=1}^{n}(x_i^2 - 10cos(2\pi x_i))$ & -5.12 $ \geq  x_i \leq $  5.12\\
   Schwefel & \large $f(\textbf{x}) = 418.9829d -{\sum_{i=1}^{n} x_i sin(\sqrt{|x_i|})}$ & 
   -500 $ \geq  x_i \leq $  500\\
   Griewank & \large $f(\textbf{x}) = 1 + \sum_{i=1}^{n} \frac{x_i^{2}}{4000} - \prod_{i=1}^{n}cos(\frac{x_i}{\sqrt{i}})$  &  
   -600 $ \geq  x_i \leq $  600\\
\hline
\end{tabular}
\end{table}

\subsection{Results}\label{c5s3ss2}

Results are tabulated in Table~\ref{c5table3}, the table presents real encoded functions: Rastrigin, Schwefel, and Griewank with generational (G\haea) and steady state (SS\haea) replacements for standard and ReGen implementations. Additionally, the table includes the best fitness based on the minimum median performance, following the standard deviation of the observed value, and the iteration where the reported fitness is found, which is enclosed in square brackets. The last row displays \haea(XUG) implementation with the best results reported by Gomez \cite{GOMEZa, GOMEZb} using three different genetic operators: Single real point crossover, Uniform mutation, and Gaussian mutation (XUG). \haea(XUG) performed experiments with real encoding. 

\begin{figure}[H]
\centering
\includegraphics[width=2in]{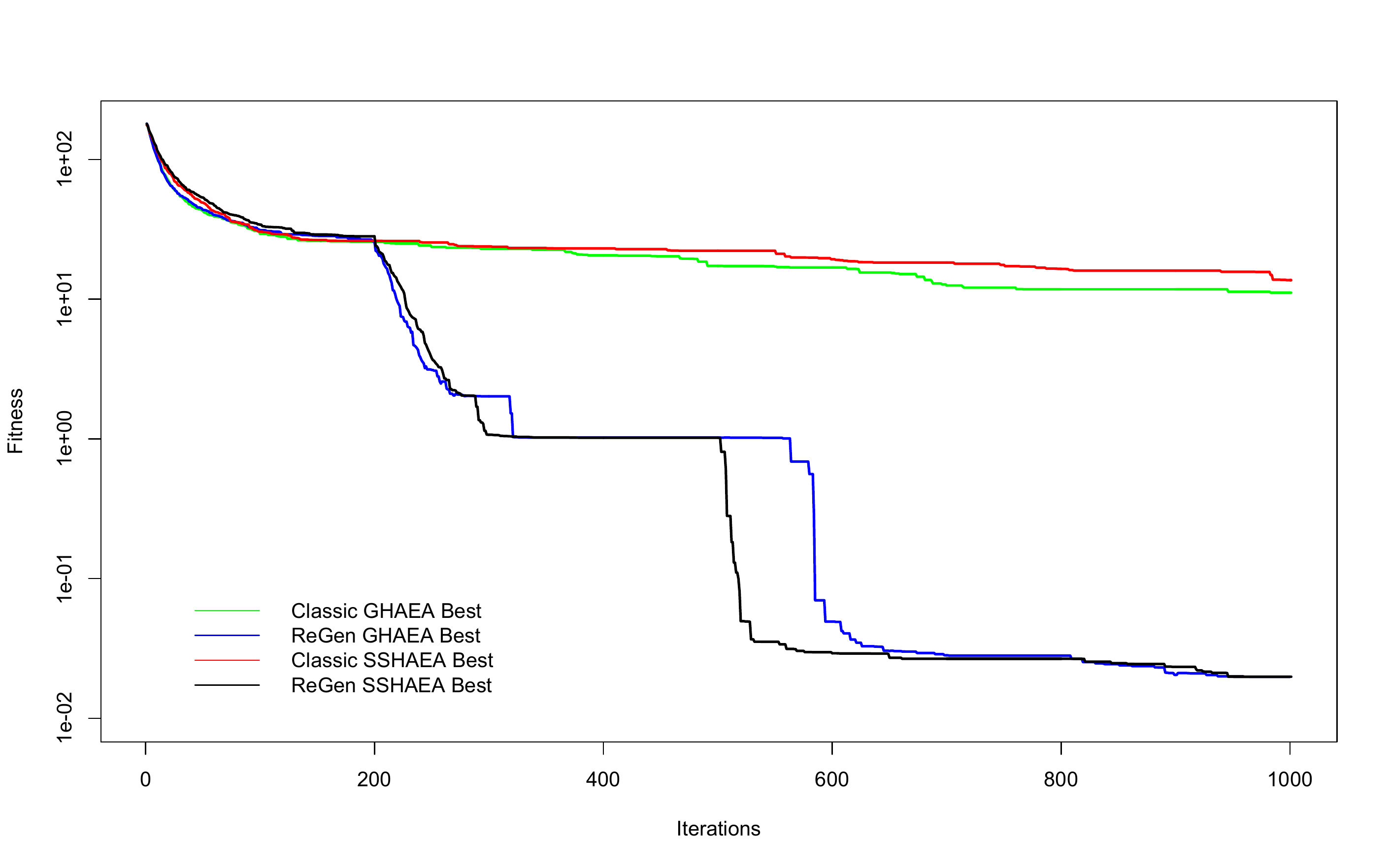}
\includegraphics[width=2in]{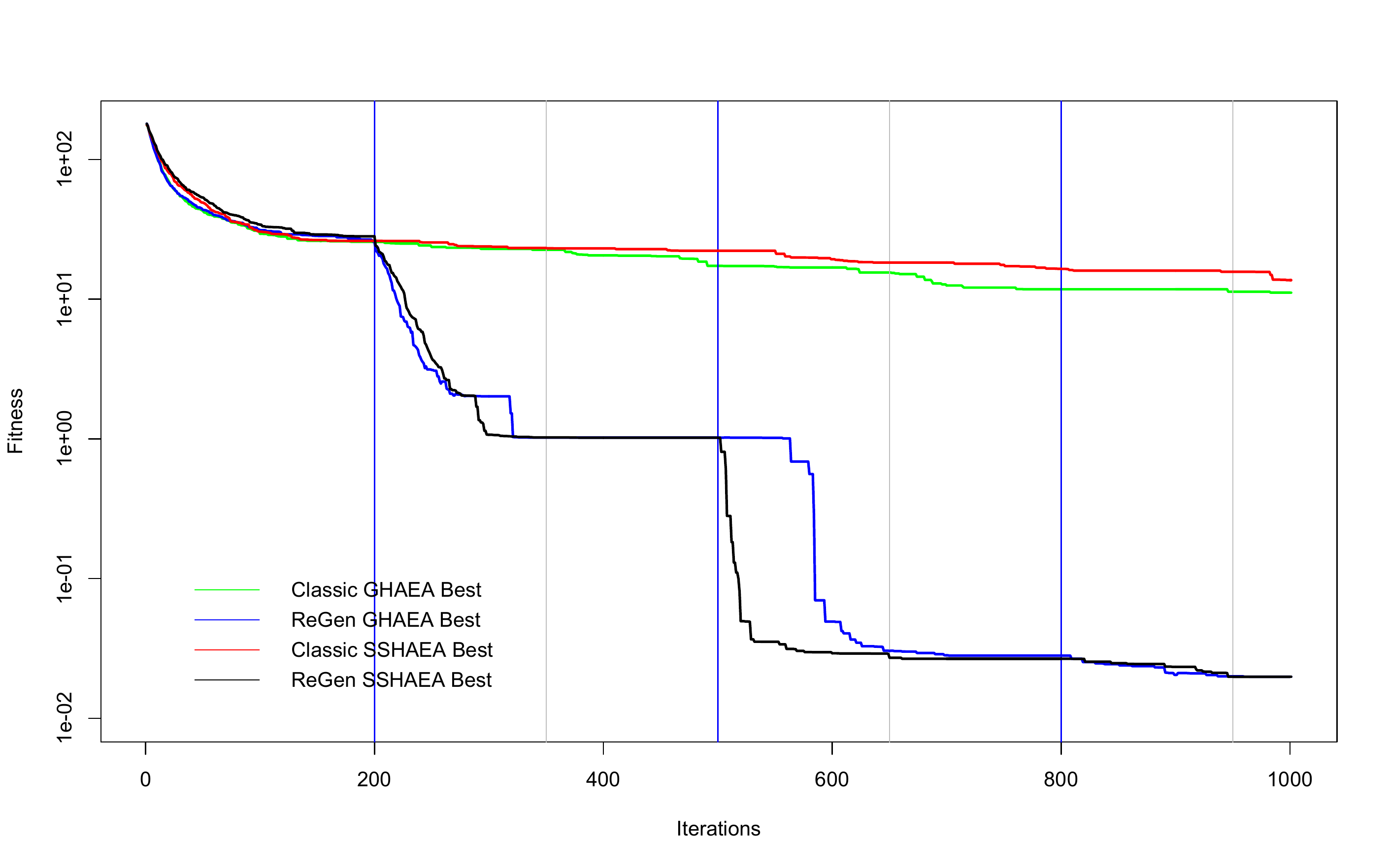}
\caption{Rastrigin. Generational replacement (G\haea) and Steady state replacement (SS\haea).}
\label{c5fig3}
\end{figure}

\begin{figure}[H]
\centering
\includegraphics[width=2in]{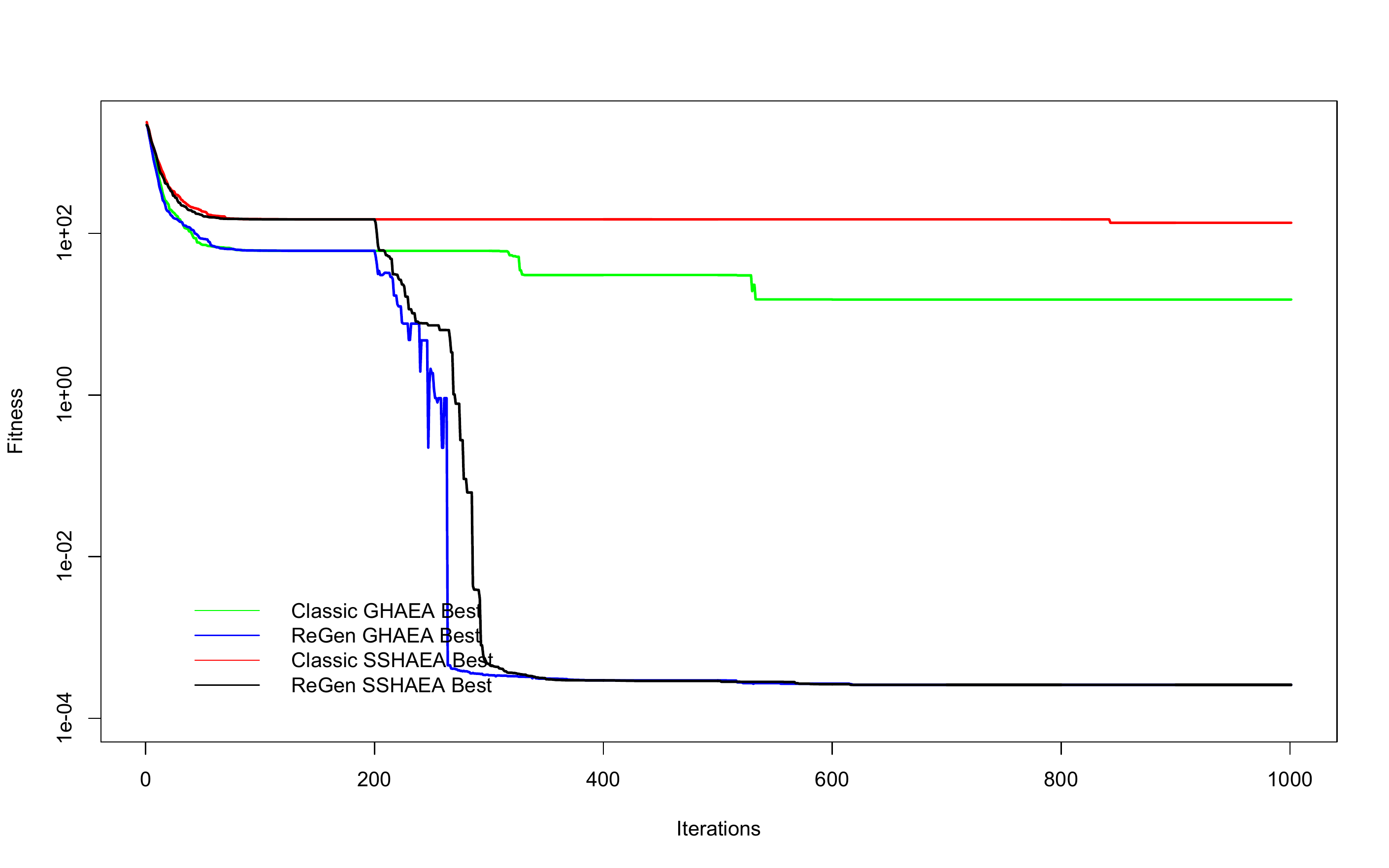}
\includegraphics[width=2in]{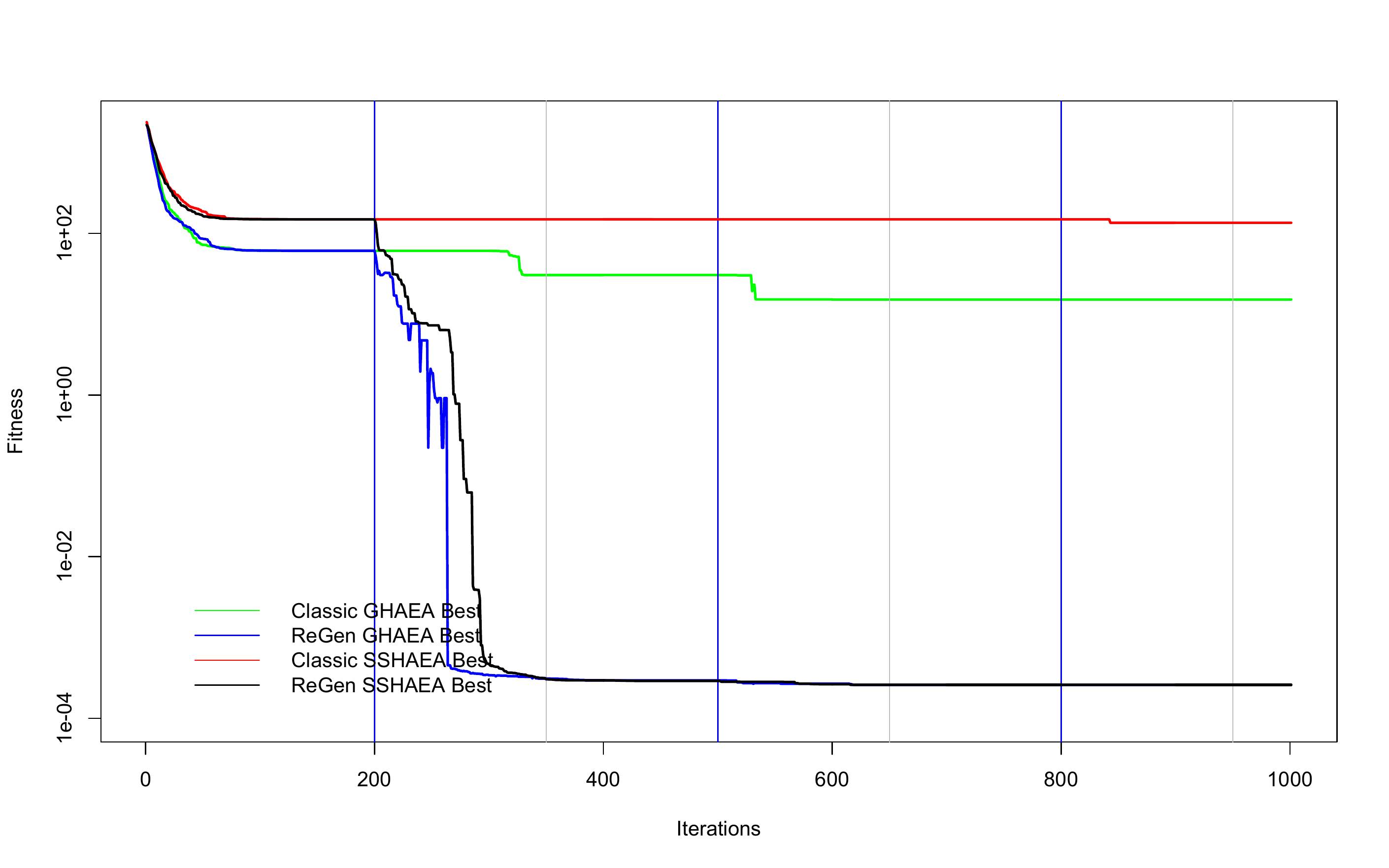}
\caption{Schwefel. Generational replacement (G\haea) and Steady state replacement (SS\haea).}
\label{c5fig4}
\end{figure}

\begin{figure}[H]
\centering
\includegraphics[width=2in]{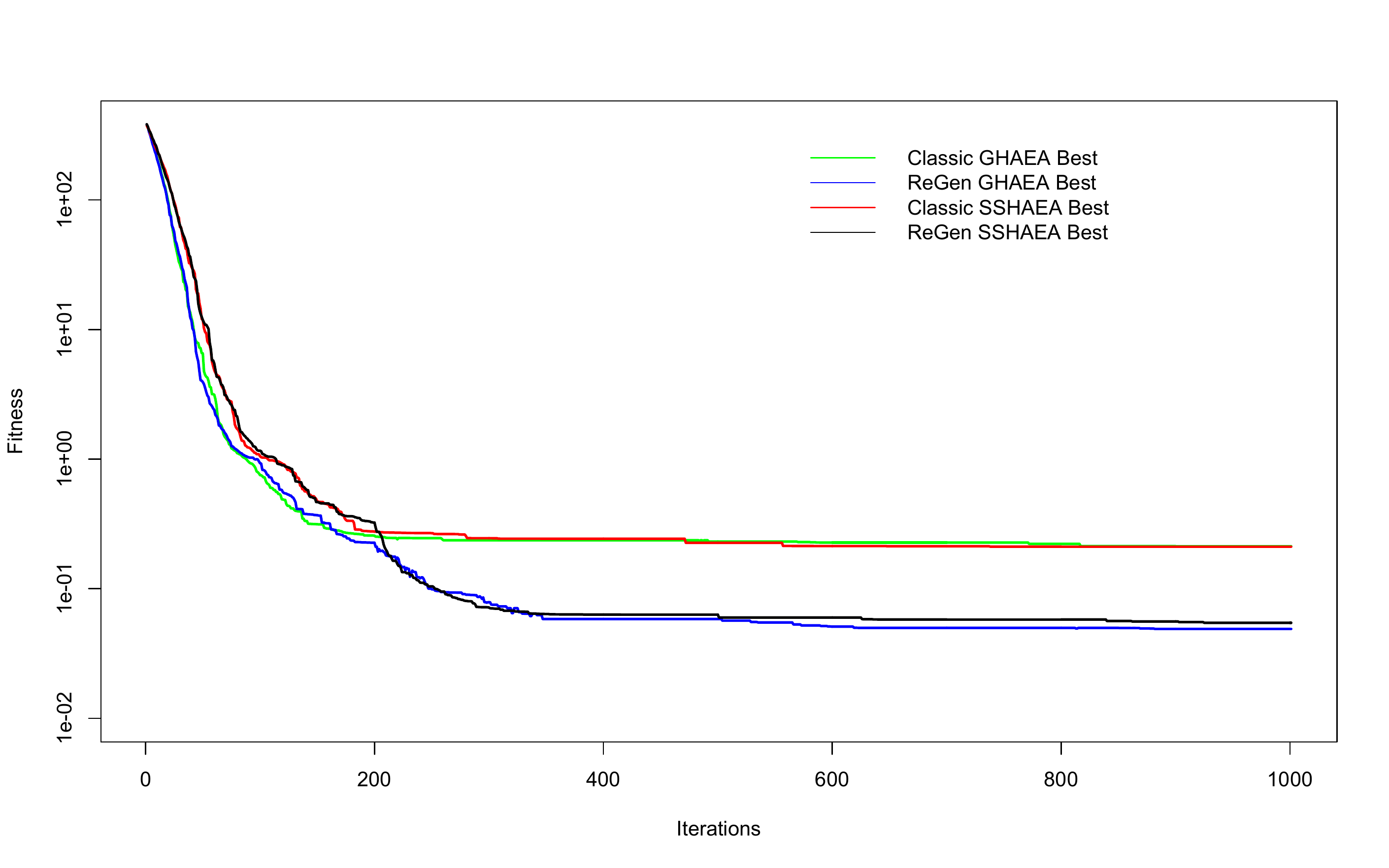}
\includegraphics[width=2in]{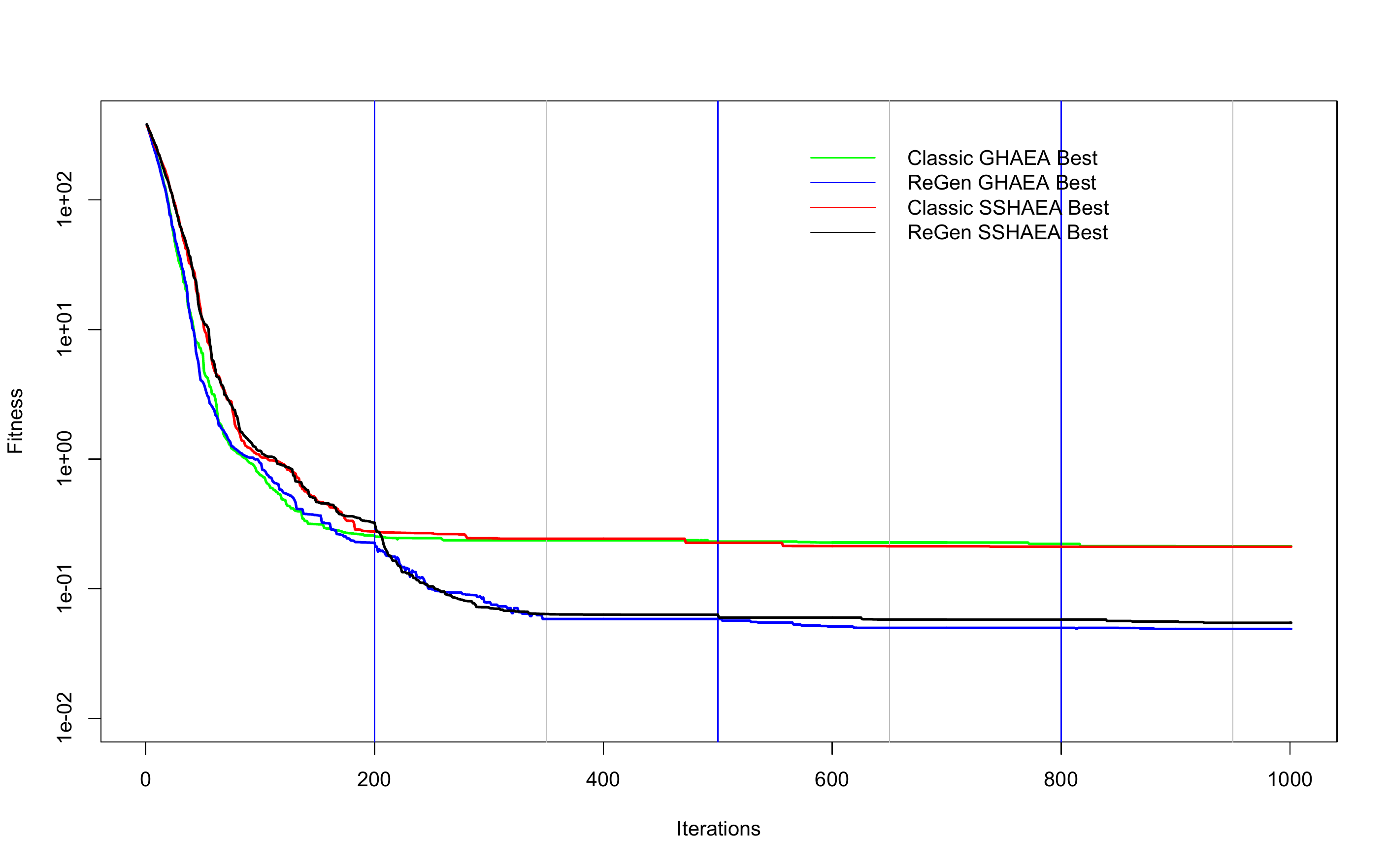}
\caption{Griewank. Generational replacement (G\haea) and Steady state replacement (SS\haea).}
\label{c5fig5}
\end{figure}

Graphs Fig.~\ref{c5fig3}, Fig.~\ref{c5fig4}, and Fig.~\ref{c5fig5} illustrate the fitness of best individuals of populations in performed experiments, reported fitnesses are based on the minimum median performance. Each figure shows the course of best individuals per technique. For \haea and ReGen \haea, two methods are applied: steady state and generational population replacements. The fitness evolution of individuals can be noted by tracking green and red lines, which depict best individuals' fitness for the standard \haea. Blue and black lines trace best individuals' fitness for ReGen \haea. Figures on the right side show defined marking periods. Vertical lines in blue depict the starting of a marking period, lines in gray delimit the end of such periods.

\begin{table}[H]
\centering
  \caption{Results of the experiments for Generational and Steady state replacements}
  \label{c5table3}
  \begin{tabular}{llll} \hline
      \textbf{EA} & \textbf{Rastrigin} & \textbf{Schwefel} & \textbf{Griewank} \\\hline
      ReGen G\haea & $0.019836 \pm 0.400 [969]$ & $0.000259 \pm 05.660 [998]$ & $0.048921 \pm 0.04 [975]$\\ 
      ReGen SS\haea & $0.019799 \pm 0.32 [1000]$ & $0.000259 \pm 24.770 [923]$ & $0.054499 \pm 0.02 [956]$\\ 
      G\haea & $11.14796 \pm 4.53 [1000]$ &  $15.24888 \pm 101.59 [601]$ & $0.212970 \pm 0.12 [818]$\\
      SS\haea & $13.68203 \pm 5.12 [1000]$ & $135.4623 \pm 115.92 [843]$ & $0.211103 \pm 0.16 [783]$\\ 
      \haea(XUG) & $0.053614\pm0.2168080$ & $0.005599\pm0.01170200$ &  $0.054955\pm0.029924$\\\hline
\end{tabular}
\end{table}

Based on tabulated results in Table~\ref{c5table3} it can be noted that ReGen \haea implementations perform better than standards \haea implementations, including results from \haea(XUG), which used real encoding for experiments. ReGen \haea is able to discover suitable candidate solutions. In Fig.~\ref{c5fig3}, Fig.~\ref{c5fig4}, and Fig.~\ref{c5fig5} is observable that marking periods applied on chromosomes at iterations $200$, $500$, and $800$ does cause a great change on the evolution of individuals. After starting the first marking period, populations improve their performance once tags are added, especially for Rastrigin and Schwefel functions. It is remarkable how in every defined marking period (delimited with vertical blue lines), individuals improve their fitness. For the Griewank function, there is a small margin of difference between the two implementation performances, even though ReGen \haea accomplishes better results. In Fig.~\ref{c5fig5} is evident that the pressure applied on chromosomes at iteration $200$ affects the evolution of individuals, the fitness improves, and keeps stable for best individuals until the evolution process finishes. ReGen \haea found a variety of good solutions during the evolution process, exposing the ability of the proposed approach to discover local minima that are not identified by standard \haea implementations.

\section{Statistical Analysis}\label{c5s4}

Three different tests are performed, One-Way ANOVA test, Pairwise Student’s t-test, and Paired Samples Wilcoxon Test (also known as Wilcoxon signed-rank test). The data set ReGen EAs Samples in Appendix \ref{appendB} is used. The samples contain four \haea implementations for each of the following functions: Deceptive Order Three, Deceptive Order Four Trap, Rastrigin, Schwefel, and Griewank. The samples refer to the best fitness of a solution found in each run, the number of executions per algorithm is $30$. Different implementations involve standard \haea and ReGen \haea with Generational (G) and Steady State (SS) population replacements.

\begin{table}[H]
\centering
\caption{Anova Single Factor: SUMMARY}
\label{c5table4}
\begin{tabular}{lllll}
\hline
\multicolumn{5}{c}{\textbf{Deceptive Order Three}} \\
Groups & Count & Sum & Average & Variance \\\hline
GHAEA & 30 & 103146 & 3438.2 & 102.993103 \\
SSHAEA & 30 & 103052 & 3435.066667 & 111.650575 \\
ReGenGHAEA & 30 & 107554 & 3585.133333 & 93.4298851 \\
ReGenSSHAEA & 30 & 107650 & 3588.333333 & 78.3678161 \\\hline
\multicolumn{5}{c}{\textbf{Deceptive Order Four Trap}} \\
Groups & Count & Sum & Average & Variance \\\hline
GHAEA & 30 & 11815 & 393.8333333 & 9.385057471 \\
SSHAEA & 30 & 11737 & 391.2333333 & 20.73678161 \\
ReGenGHAEA & 30 & 13410 & 447 & 4.75862069 \\
ReGenSSHAEA & 30 & 13390 & 446.3333333 & 3.471264368 \\\hline
\multicolumn{5}{c}{\textbf{Rastrigin}} \\
Groups & Count & Sum & Average & Variance \\\hline
GHAEA & 30 & 329.0666924 & 10.96888975 & 20.2816323 \\
SSHAEA & 30 & 403.9728574 & 13.46576191 & 26.1883055 \\
ReGenGHAEA & 30 & 4.911251327 & 0.163708378 & 0.14340697 \\
ReGenSSHAEA & 30 & 3.576815371 & 0.119227179 & 0.09299807 \\\hline
\multicolumn{5}{c}{\textbf{Schwefel}} \\
Groups & Count & Sum & Average & Variance \\\hline
GHAEA & 30 & 1344.597033 & 44.81990111 & 7258.77527 \\
SSHAEA & 30 & 4439.27726 & 147.9759087 & 13144.3958 \\
ReGenGHAEA & 30 & 30.50459322 & 1.016819774 & 31.002189 \\
ReGenSSHAEA & 30 & 218.4970214 & 7.283234047 & 527.103699 \\\hline
\multicolumn{5}{c}{\textbf{Griewank}} \\
Groups & Count & Sum & Average & Variance \\\hline
GHAEA & 30 & 6.520457141 & 0.217348571 & 0.01290201 \\
SSHAEA & 30 & 7.181644766 & 0.239388159 & 0.02090765 \\
ReGenGHAEA & 30 & 1.624738713 & 0.054157957 & 0.0015871 \\
ReGenSSHAEA & 30 & 1.61989951 & 0.05399665 & 0.000493\\\hline
\end{tabular}
\end{table}

Based on ReGen EAs Samples in Appendix \ref{appendB}, the analysis of variance is computed to know the difference between evolutionary algorithms with different implementations that include standard \haea and ReGen \haea with generational and steady state replacement strategies. Algorithms are four in total, in Table~\ref{c5table4} a summary of each function and algorithm is shown. The summary presents the number of samples per algorithm (30), the sum of fitnesses, the average fitness, and their variances. Results of the ANOVA single factor is tabulated in Table~\ref{c5table5}.

\begin{table}[H]
\centering
\caption{Anova Single Factor: ANOVA}
\label{c5table5}
\begin{tabular}{lllllll}
\hline
\multicolumn{7}{c}{\textbf{Deceptive Order Three}} \\
Source   of Variation & SS & df & MS & F & P-value & F crit \\\hline
Between Groups & 676201.16 & 3 & 225400.38 & 2333.0875 & 1.7577E-103 & 2.6828 \\
Within Groups & 11206.8 & 116 & 96.6103 &  &  &  \\
 &  &  &  &  &  &  \\
Total & 687407.96 & 119 &  &  &  &  \\\hline
\multicolumn{7}{c}{\textbf{Deceptive Order Four Trap}} \\
Source   of Variation & SS & df & MS & F & P-value & F crit \\\hline
Between Groups & 88020.6 & 3 & 29340.2 & 3060.1179 & 3.2412E-110 & 2.6828 \\
Within Groups & 1112.2 & 116 & 9.5879 &  &  &  \\
 &  &  &  &  &  &  \\
Total & 89132.8 & 119 &  &  &  &  \\\hline
\multicolumn{7}{c}{\textbf{Rastrigin}} \\
Source   of Variation & SS & df & MS & F & P-value & F crit \\\hline
Between Groups & 4468.33 & 3 & 1489.44 & 127.5582 & 1.39871E-36 & 2.6828 \\
Within Groups & 1354.48 & 116 & 11.6765 &  &  &  \\
 &  &  &  &  &  &  \\
Total & 5822.8196 & 119 &  &  &  &  \\\hline
\multicolumn{7}{c}{\textbf{Schwefel}} \\
Source   of Variation & SS & df & MS & F & P-value & F crit \\\hline
Between Groups & 415496.57 & 3 & 138498.85 & 26.4294 & 4.22517E-13 & 2.6828 \\
Within Groups & 607877.03 & 116 & 5240.3192 &  &  &  \\
 &  &  &  &  &  &  \\
Total & 1023373.61 & 119 &  &  &  &  \\\hline
\multicolumn{7}{c}{\textbf{Griewank}} \\
Source   of Variation & SS & df & MS & F & P-value & F crit \\\hline
Between Groups & 0.918607 & 3 & 0.306202 & 34.1270 & 6.92905E-16 & 2.6828 \\
Within Groups & 1.040803 & 116 & 0.008972 &  &  &  \\
 &  &  &  &  &  &  \\
Total & 1.959410 & 119 &  &  &  &  \\\hline
\end{tabular}
\end{table}


As P-values for Deceptive Order Three, Deceptive Order Four Trap, Rastrigin, Schwefel, and Griewank functions are less than the significance level $0.05$, results allow concluding that there are significant differences between the groups as shown in Table~\ref{c5table5}. In one-way ANOVA tests, significant P-values indicate that some of the group means are different, but it is not evident which pairs of groups are different. In order to interpret one-way ANOVA test' results, multiple pairwise-comparison with Student's t-test is performed to determine if the mean difference between specific pairs of the group is statistically significant. Also, paired-sample Wilcoxon tests are computed.

\begin{landscape}
\begin{figure}[H]
\centering
\includegraphics[width=4.2in]{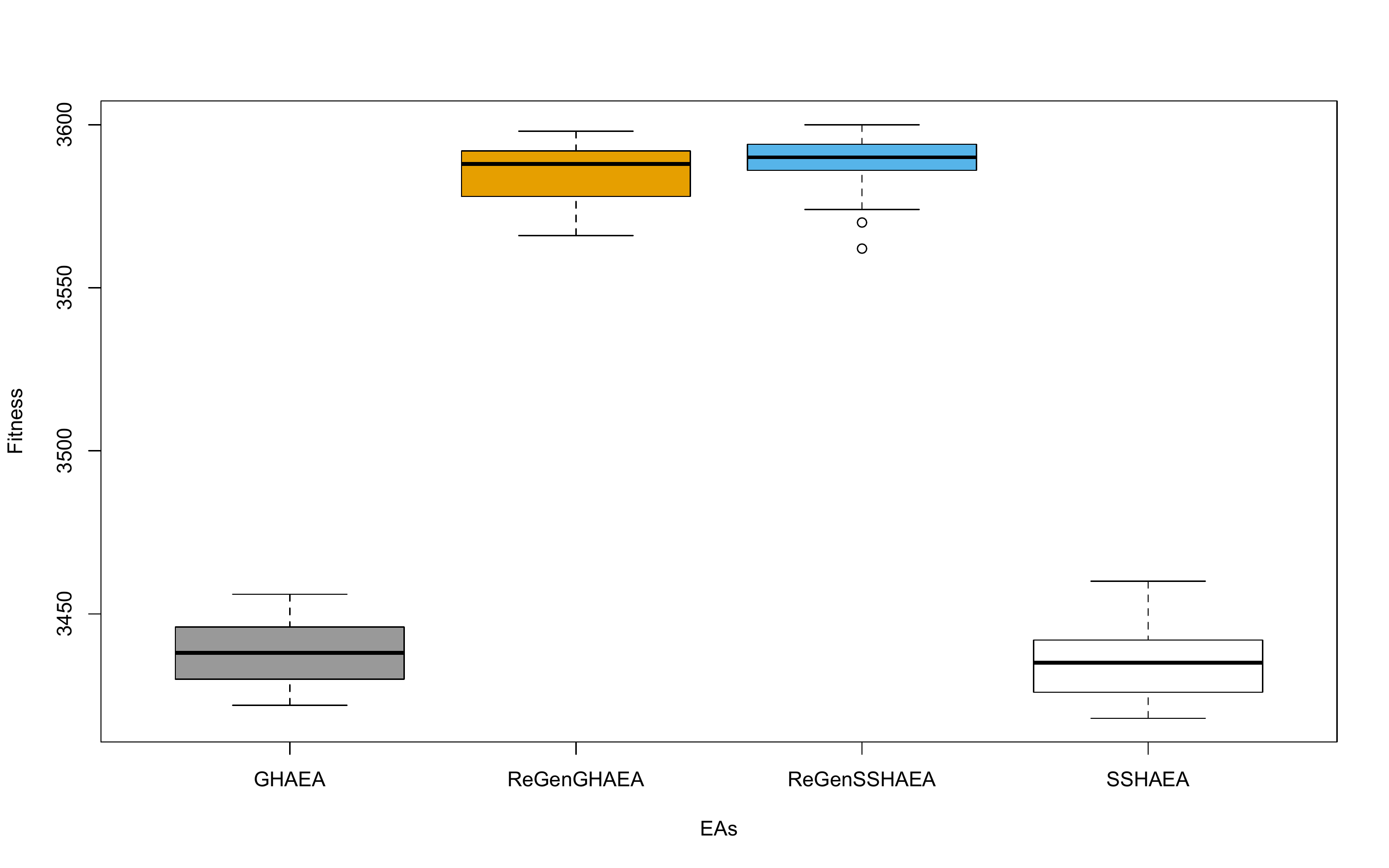}
\includegraphics[width=4.2in]{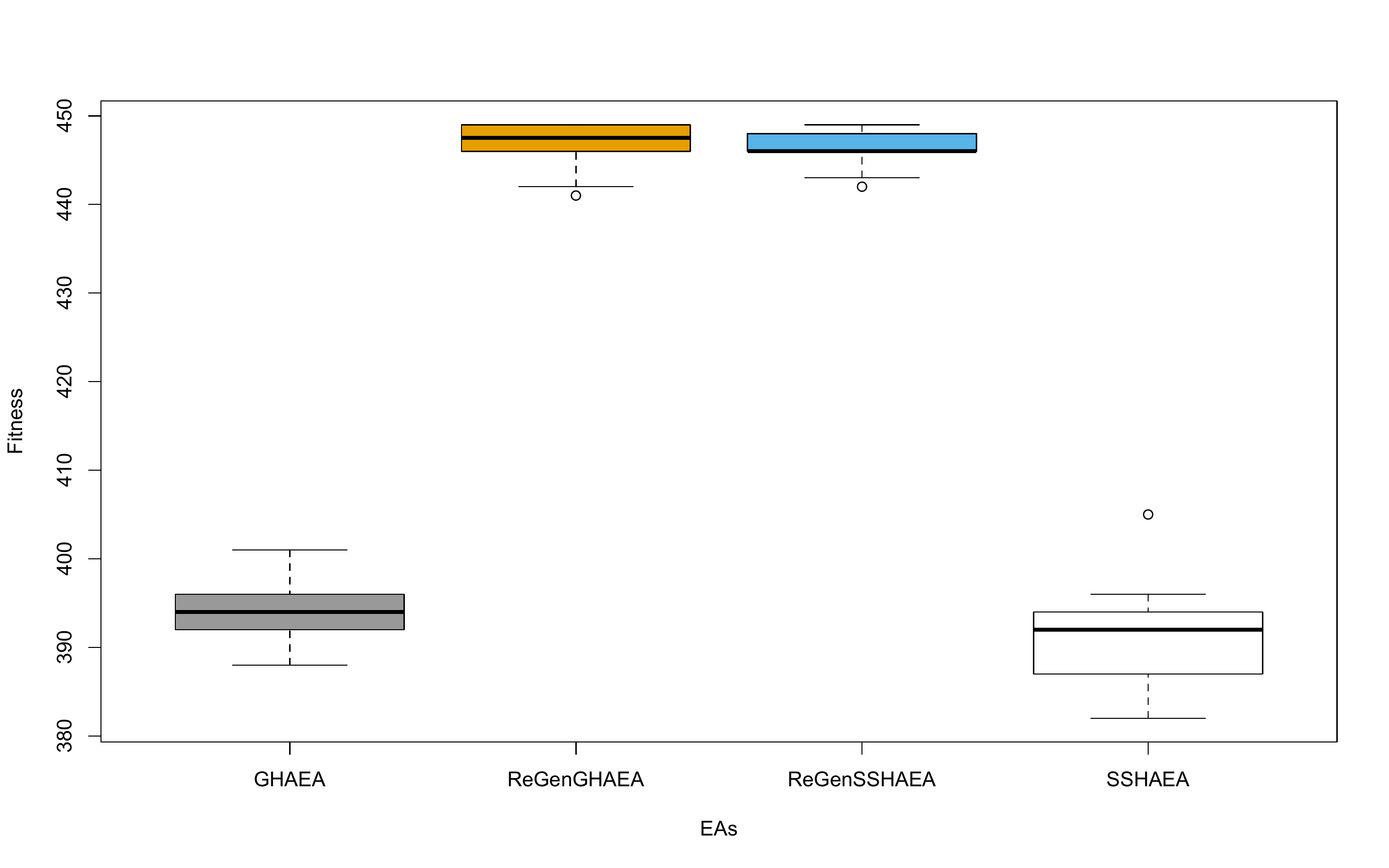}
\includegraphics[width=4.2in]{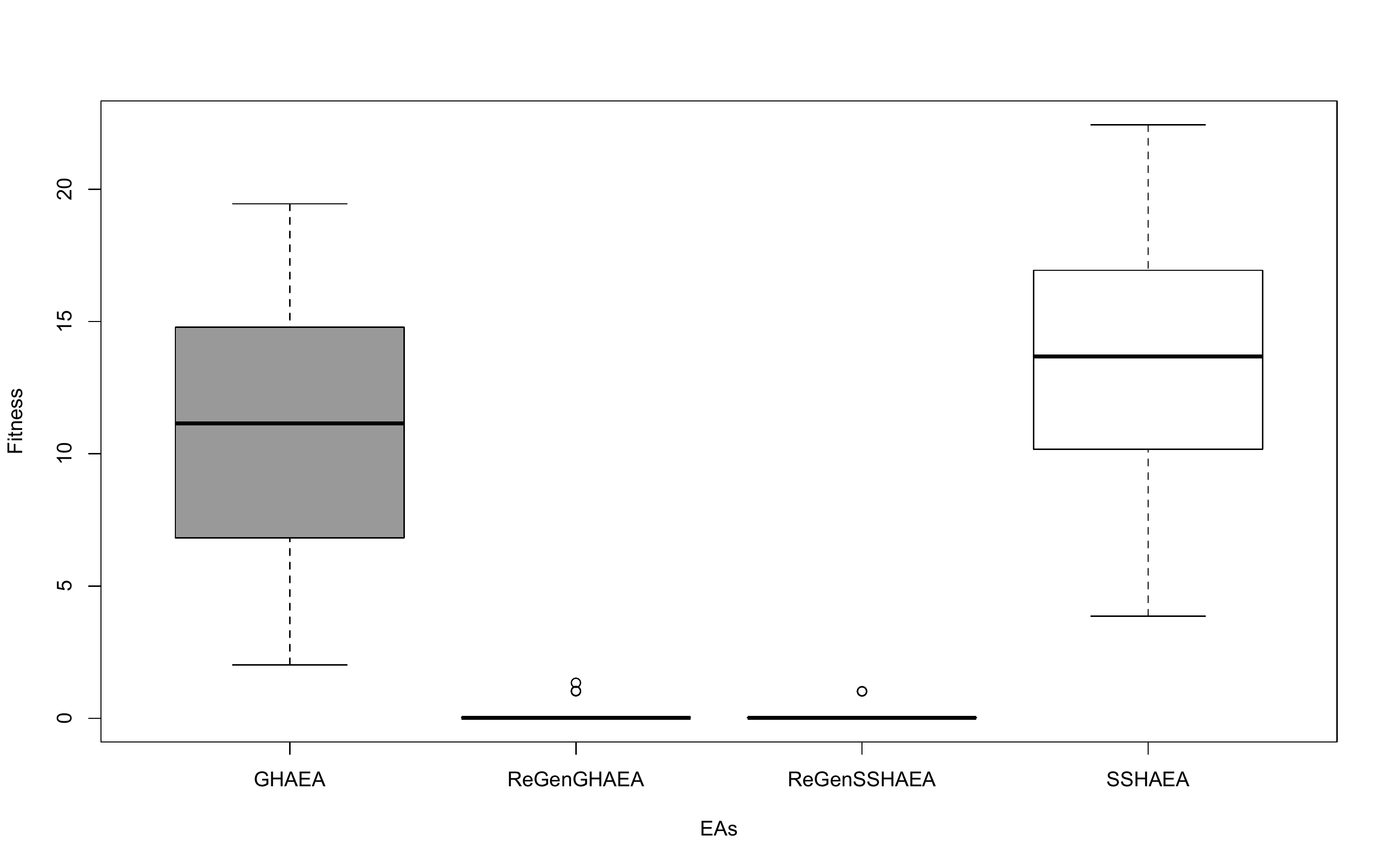}
\includegraphics[width=4.2in]{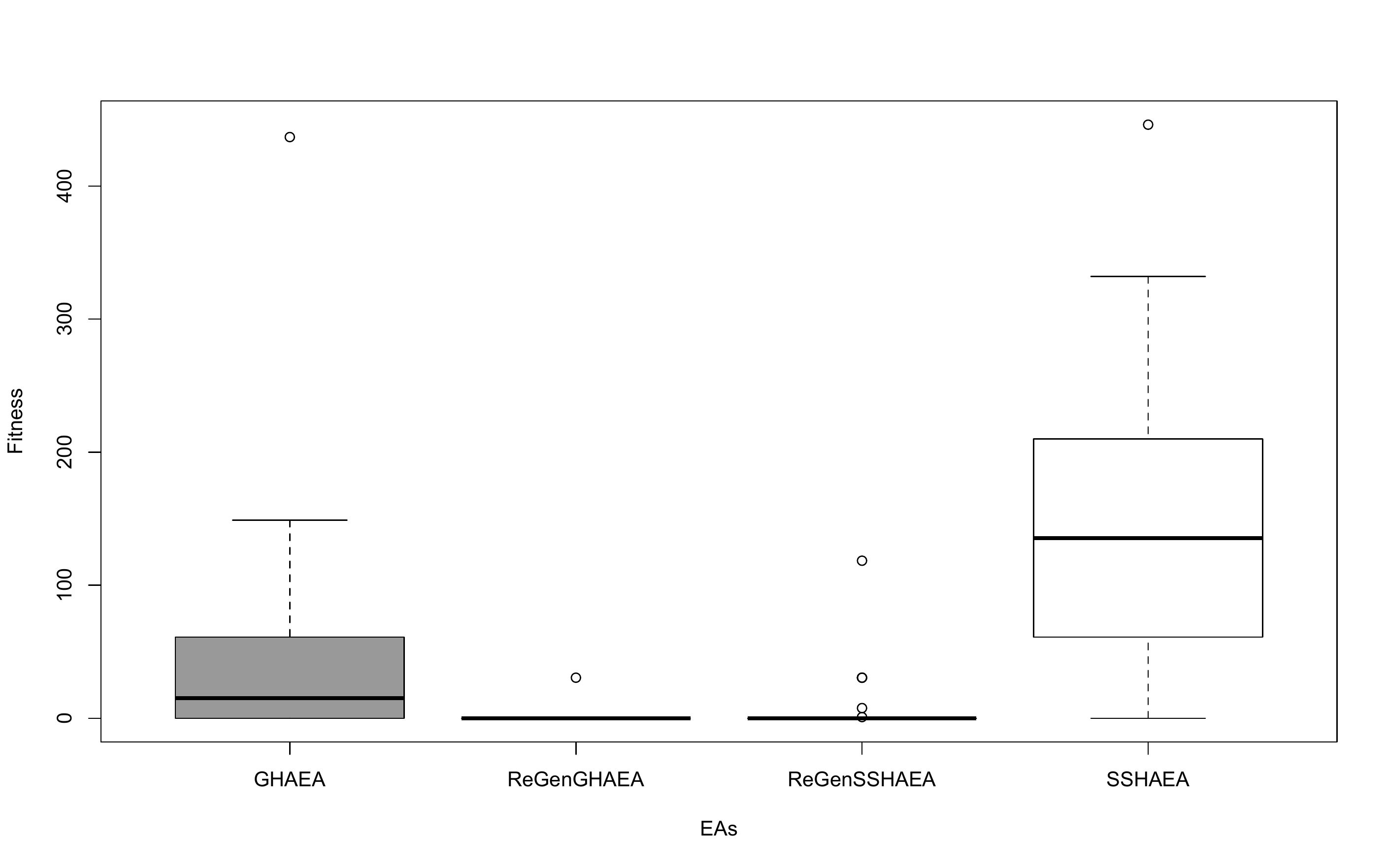}
\caption{EAs with Generational replacement (G\haea) and Steady State replacement (SS\haea). On top: Deceptive Order Three and Deceptive Order Four Trap Functions. On the bottom: Rastrigin and Schwefel functions.}
\label{c5fig6}
\end{figure}
\end{landscape}

\begin{figure}[H]
\centering
\includegraphics[width=4.2in]{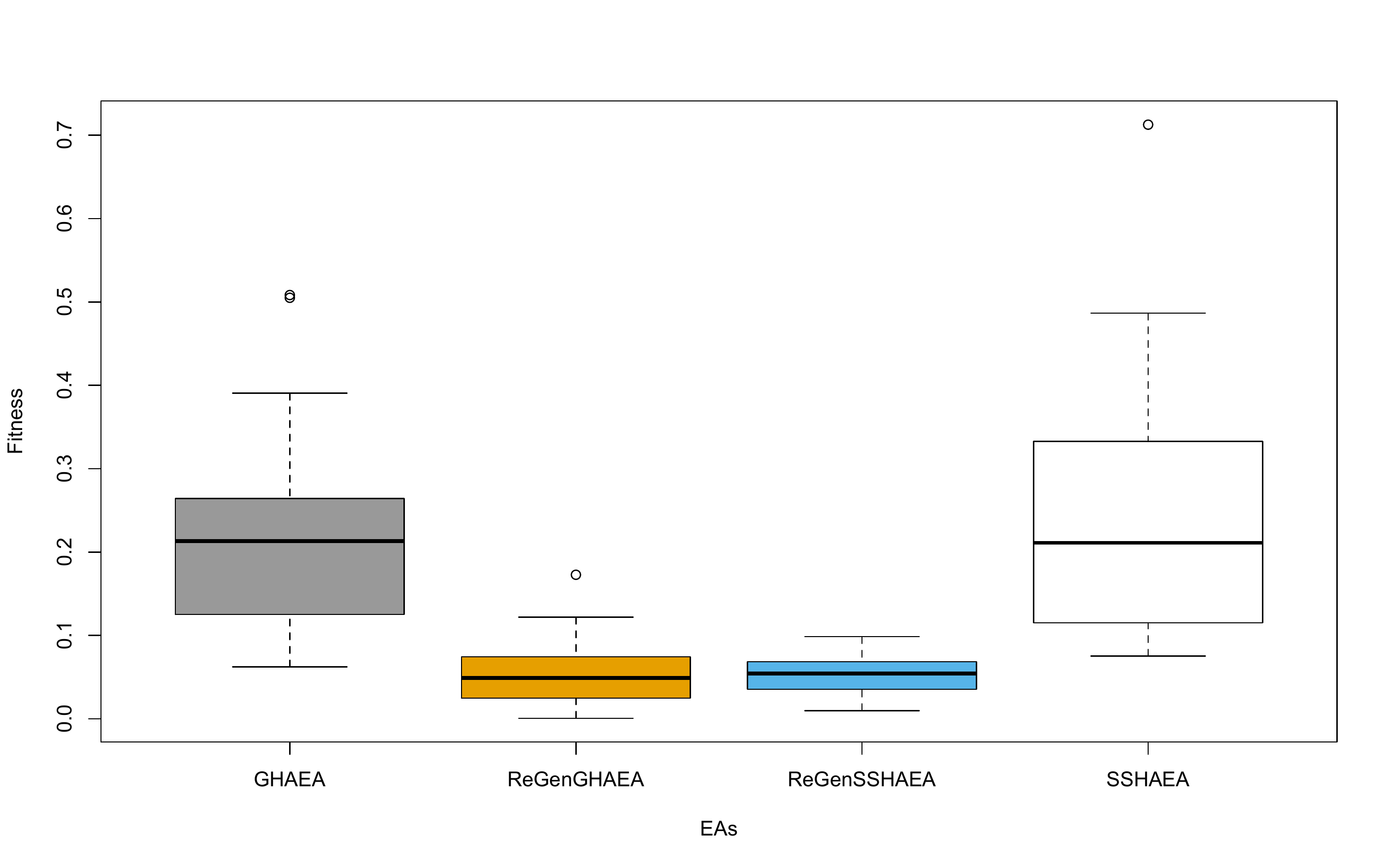}
\caption{EAs with Generational replacement (G\haea) and Steady State replacement (SS\haea). Griewank function.}
\label{c5fig7}
\end{figure}

Box plots in Fig.~\ref{c5fig6} and Fig.~\ref{c5fig7} depict the median fitness of EAs' best solutions (ReGen EAs Samples in Appendix \ref{appendB}). Four EAs are illustrated, epigenetic EAs in Orange (ReGen G\haea) and Blue (ReGen SS\haea), standard EAs in Gray (G\haea) and White (SS\haea). For Deceptive Order Three function, the median fitness for each Epigenetic EA is close to the global optimum ($3600$), while median fitnesses for classic \haea are under the local optimum ($3450$). On the other hand, Deceptive Order Four Trap median fitness surpasses $440$ for all Epigenetic implementations; in contrast, for standard \haea, the median fitness does not reach $400$. For Rastrigin function, the median fitness for each Epigenetic EA is lower than the local minima ($1.0$), while median fitnesses for standard \haea are over the local optimum ($10$). Next in order, epigenetic EAs for Schwefel achieved median fitness inferior to $0.0003$; conversely, \haea median fitnesses are greater than $0.0004$ for generational replacement and higher than 100 for steady state implementation. Finally, for Griewank function's box plots, depicted median fitnesses are below the local optimum $0.1$ for epigenetic evolutionary algorithms, while median fitness values for standard versions of \haea are above $0.2$. So, based on these data, it seems that Epigenetic \haea versions find better solutions than classic \haea implementations. However, it is needed to determine whether these findings are statistically significant.

\begin{table}[H]
\centering
\caption{Student T-tests pairwise comparisons with pooled standard deviation. Benjamini Hochberg (BH) as p-value adjustment method.}
\label{c5table6}
\begin{tabular}{llll}
\hline
\multicolumn{4}{c}{\textbf{Deceptive Order Three}} \\
EAs & GHAEA & ReGenGHAEA & ReGenSSHAEA \\\hline
ReGenGHAEA & \cellcolor[HTML]{EFEFEF}2.92E-87 & - & - \\
ReGenSSHAEA & \cellcolor[HTML]{EFEFEF}3.65E-88 & 0.2194596 & - \\
SSHAEA & 0.2194596 & \cellcolor[HTML]{EFEFEF}3.65E-88 & \cellcolor[HTML]{EFEFEF}1.02E-88 \\\hline
\multicolumn{4}{c}{\textbf{Deceptive Order Four Trap}} \\
EAs & GHAEA & ReGenGHAEA & ReGenSSHAEA \\\hline
ReGenGHAEA & \cellcolor[HTML]{EFEFEF}6.52E-94 & - & - \\
ReGenSSHAEA & \cellcolor[HTML]{EFEFEF}2.04E-93 & 0.40607501 & - \\
SSHAEA & \cellcolor[HTML]{EFEFEF}0.00180068 & \cellcolor[HTML]{EFEFEF}8.80E-96 & \cellcolor[HTML]{EFEFEF}1.72E-95 \\\hline
\multicolumn{4}{c}{\textbf{Rastrigin}} \\
EAs & GHAEA & ReGenGHAEA & ReGenSSHAEA \\\hline
ReGenGHAEA & \cellcolor[HTML]{EFEFEF}1.83E-22 & - & - \\
ReGenSSHAEA & \cellcolor[HTML]{EFEFEF}1.83E-22 & 0.959877985 & - \\
SSHAEA & \cellcolor[HTML]{EFEFEF}0.006585082 & \cellcolor[HTML]{EFEFEF}1.27E-28 & \cellcolor[HTML]{EFEFEF}1.27E-28 \\\hline
\multicolumn{4}{c}{\textbf{Schwefel}} \\
EAs & GHAEA & ReGenGHAEA & ReGenSSHAEA \\\hline
ReGenGHAEA & \cellcolor[HTML]{EFEFEF}0.03120645 & - & - \\
ReGenSSHAEA & 0.05632517 & 0.73803211 & - \\
SSHAEA & \cellcolor[HTML]{EFEFEF}4.21E-07 & \cellcolor[HTML]{EFEFEF}1.29E-11 & \cellcolor[HTML]{EFEFEF}3.65E-11 \\\hline
\multicolumn{4}{c}{\textbf{Griewank}} \\
EAs & GHAEA & ReGenGHAEA & ReGenSSHAEA \\\hline
ReGenGHAEA & \cellcolor[HTML]{EFEFEF}1.35E-09 & - & - \\
ReGenSSHAEA & \cellcolor[HTML]{EFEFEF}1.35E-09 & 0.99474898 & - \\
SSHAEA & 0.44325525 & \cellcolor[HTML]{EFEFEF}2.87E-11 & \cellcolor[HTML]{EFEFEF}2.87E-11\\\hline
\end{tabular}
\end{table}

\paragraph{\em{Multiple pairwise t-test:}}
Multiple pairwise-comparison between means of groups is performed. In the one-way ANOVA test described above, significant p-values indicate that some group means are different. In order to know which pairs of groups are different, multiple pairwise-comparison is performed for Deceptive Order Three (D3), Deceptive Order Four Trap (D4), Rastrigin (RAS), Schwefel (SCHW), and Griewank (GRIE) best solutions samples. Table~\ref{c5table6} presents Pairwise comparisons using t-tests with pooled standard deviation (SD) with their respective p-values. The test adjusts p-values with the Benjamini-Hochberg method. Pairwise comparisons show that only highlighted values in gray between two algorithms are significantly different ($p < 0.05$). Therefore, the alternative hypothesis is true.

Now, to find out any significant difference between the median fitness of individuals in the two experimental groups (standard \haea and \haea with regulated genes), the Wilcoxon test is conducted.

\paragraph{\em{Paired Samples Wilcoxon Test:}}  
For this test, algorithms are grouped per population replacement strategy. Wilcoxon signed rank test for generational EAs (G\haea and ReGen G\haea) and Wilcoxon signed rank test for steady state EAs (SS\haea and ReGen SS\haea). The test assesses standard \haea versus Epigenetic \haea implementations. 

\begin{itemize}
    \item Deceptive Order Three (D3)
    \begin{enumerate}
      \item \par Wilcoxon signed rank test with continuity correction for generational EAs uses all data-set samples from G\haea and ReGen G\haea implementations. $V = 0$, {\em P-value} is equal to $1.792453e-06$, which is less than the significance level alpha ($0.05$).
      
      \item \par Wilcoxon signed rank test with continuity correction for steady state EAs uses all data-set samples from SS\haea and ReGen SS\haea algorithms. $V = 0$, {\em P-value} is equal to $1.803748e-06$, which is less than the significance level $alpha = 0.05$.
      \end{enumerate}
      
    \item Deceptive Order Four Trap (D4)
    \begin{enumerate}
      \item \par Wilcoxon signed rank test with continuity correction for generational EAs uses all data-set samples from G\haea and ReGen G\haea implementations. $V = 0$, {\em P-value} is equal to $1.760031e-06$, which is less than the significance level alpha ($0.05$).
      
      \item \par Wilcoxon signed rank test with continuity correction for steady state EAs uses all data-set samples from SS\haea and ReGen SS\haea versions. $V = 0$, {\em P-value} is equal to $1.768926e-06$, which is less than the significance level $alpha = 0.05$.
      \end{enumerate}
      
    \item  Rastrigin (RAS)
    \begin{enumerate}
      \item \par Wilcoxon signed rank test with continuity correction for generational EAs uses all data-set samples from G\haea and ReGen G\haea implementations. $V = 465$, {\em P-value} is equal to $1.863e-09$, which is less than the significance level alpha ($0.05$).
      
      \item \par Wilcoxon signed rank test with continuity correction for steady state EAs uses all data-set samples from SS\haea and ReGen SS\haea algorithms. $V = 465$, {\em P-value} is equal to $1.863e-09$, which is less than the significance level $alpha = 0.05$.
      \end{enumerate}
      
      \item Schwefel (SCHW)
    \begin{enumerate}
      \item \par Wilcoxon signed rank test with continuity correction for generational EAs uses all data-set samples from G\haea and ReGen G\haea implementations. $V = 450$, {\em P-value} is equal to $2.552e-07$, which is less than the significance level $alpha = 0.05$.
      
      \item \par Wilcoxon signed rank test with continuity correction for steady state EAs uses all data-set samples from SS\haea and ReGen SS\haea versions. $V = 452$, {\em P-value} is equal to $1.639e-07$, which is less than the significance level alpha ($0.05$). 
      \end{enumerate}
      
    \item Griewank (GRIE)
    \begin{enumerate}
      \item \par Wilcoxon signed rank test with continuity correction for generational EAs uses all data-set samples from G\haea and ReGen G\haea implementations. $V = 462$, {\em P-value} is equal to $9.313e-09$, which is less than the significance level $alpha = 0.05$.
      
      \item \par Wilcoxon signed rank test with continuity correction for steady state EAs uses all data-set samples from SS\haea and ReGen SS\haea algorithms. $V = 465$, {\em P-value} is equal to $1.863e-09$, which is less than the significance level alpha ($0.05$). 
      \end{enumerate}
      
\end{itemize}

The above leads to conclude that median fitnesses of solutions found by standard generational Hybrid Adaptive Evolutionary Algorithms (G\haea) are significantly different from median fitnesses of solutions found by generational \haea with regulated genes (ReGen G\haea) with p-values equal to $1.792453e-06$ (D3 samples), $1.760031e-06$ (D4 samples), $1.863e-09$ (RAS samples), $2.552e-07$ (SCHW samples), and $9.313e-09$ (GRIE samples). So, the alternative hypothesis is true.

The median fitness of solutions found by classic steady state Hybrid Adaptive Evolutionary Algorithms (SS\haea) is significantly different from the median fitness of solutions found by steady state \haea with regulated genes (ReGen SS\haea) with p-values equal to $1.803748e-06$ (D3 sampling fitness), $1.768926e-06$ (D4 sampling fitness), $1.863e-09$ (RAS sampling fitness), $1.639e-07$ (SCHW sampling fitness), and $1.863e-09$ (GRIE sampling fitness). As p-values are less than the significance level $0.05$, it may be concluded that there are significant differences between the two EAs groups in each Wilcoxon Test. 

\section{Summary}\label{c5s5}
The epigenetic technique is implemented on \haea to solve both binary and real encoding problems. Results have shown that the marking process did impact the way populations evolve, and the fitness of individuals considerably improves to the optimum. It is important to point out that only two operators are used: single point Crossover and single bit Mutation. This thesis intends to avoid giving too many advantages to implemented EAs in terms of parametrization and specialized operators in order to identify the applicability of the proposed epigenetic model. The statistical analysis helps to conclude that epigenetic implementations performed better than standard versions.
    \chapter{Concluding Remarks}\label{chapter6}

\section{Conclusions}

Epigenetics has proven to be a useful field of study to extract elements for improving the framework of evolutionary algorithms. Primarily because epigenetic encompasses mechanisms that support inheritance and prolongation of experiences, so future generations have enough information to adapt to changing environments. Based on the preceding, some elements are used to bring into life the ReGen EA technique. This research abstracts epigenetics fundamental concepts and introduces them as part of standard evolutionary algorithms' elements or operations. 

Modeling epigenetic evolutionary algorithms is not easy, mainly because epigenetic involves too many elements, concepts, principles, and interactions to describe what is known about the epigenetic landscape today. The process of designing epigenetic algorithms requires well-defined abstractions to simplify the epigenetic dynamics and its computational implementation. Epigenetic strategies for EAs variations have been designed during the last decade; those strategies have reported improvement in EAs performance and reduction in the computational cost when solving specific problems. Nevertheless, almost all strategies use the same idea of switching genes off and on (gene activation mechanism), or silencing chromosome sections in response to a changing environment. This approach is correct, but epigenetic goes beyond on and off states. The ReGen EA approach focuses on developing interactions by affecting genetic codes with tags that encode epigenetic instructions.

One thing ReGen EA has in common with other strategies is the use of epigenetic mechanisms such as DNA Methylation. This research characterizes DNA Methylation along with the Histone Modification mechanism. These epigenetic mechanisms are the best characterization among all epigenetic modifications, the most studied, and offer a description that is easy to understand and represent computationally. Most epigenetic approaches have abstracted the repression and activation principles from these mechanisms; basically, they use a large genotype in order to activate advantageous genes/sections and deactivate others to only express parts of the genome that produce suitable phenotypic variations. Note that, traditional evolutionary algorithms assume a finite number of genes, and to obtain novelty, EAs require not only mutations in their chromosome but also new genes. Epigenetics satisfies these needs; epigenetics becomes a problem-solver; it optimizes the number of genes and reduces classic mutation dependence. This thesis takes advantage of that; for the ReGen EA does not exist good genes, a fixed number of genes is defined; the ReGen EA involves epigenetic tags that positively or negatively affect individuals, in this way, tags promote or prevent individuals from becoming more suitable to a specific problem.

Metaphorical representations of epigenetic elements and principles such as epigenotype, tags, marking (to add, modify, and remove tags), reading (tags interpretation), and tags inheritance help in optimizing a defined number of genes, and avoiding the use of high and varied mutation rates; the ReGen EA is capable to produce suitable solutions without enlarging individuals genome and with a fixed mutation rate of {\em 1.0/chomosome length}. Compared to other approaches, emerging interactions from the dynamic of marking and reading processes is beneficial to combine multiple schemes and build varied phenotypes; avoiding to create such large genomes and regulate them by activation and deactivation mechanisms. The epigenetic mechanisms aforementioned have been useful for the design of the markers, but it has been too complicated to define what tags encode today. Tags design involves encoding, structure, and meaning. These properties have helped in proposing tags sections and rules to be interpreted by the reader function.

In this thesis, gene regulation is accomplished by adding, modifying, and removing epigenetic tags from individuals genome. The complete regulation process produces phenotypes that, in most cases, become feasible solutions to a problem. Tags structure contains binary operations; defining operations has implied a process of trial and error. The operations do not represent any biological mechanism, this fact may be miss-interpreted; it is clear that the decision to include binary operations to build the instructions may be seen as advantageous, but it is not the case. The operations have been selected taking into account many factors, three of the most prominent are: first, designed tags are meant to only solve binary and real defined problems since it is the scope of this research; second, the idea has always been to avoid giving advantages to the marking process and follow some basic principles that biological epigenetic mechanisms offer when attaching and interpreting tags, based on this, simpler operations, that do not cause abrupt changes to the phenotype generation process are chosen; and third, chemical tags in biology contain epigenetic codes that are interpreted to maintain the dynamic of many natural processes, in this case, defined operations are considered plausible to solve binary and real coded problems. Even though other operations must be explored, it is conceivable that better operations have not been taken into consideration yet.

It is important to point out that there are other molecular units, epigenetically heritable, such epigenetic factors are not abstracted in this thesis, because any non-genetic factor is reduced to only be represented as tags. With tags-only regulation, the ReGen EA proves that epigenetic factors in general influence individuals fitness by propagating permanent tags; but does not evidence tags instability that many classical geneticists question about epigenetics. Epigenetic mechanisms influence is not only at the individuals level, but also on evolution, and is caused by the environment; marking periods represent the surrounding environment of individuals and abstract the Dutch Hunger Winter study case between the winter of 1944 and 1945 in the Netherlands. Based on this study case, the ReGen EA simulates periods where individuals' genetic codes are affected by external factors -designed tags-, during different periods or iteration ranges, not continuous, but separated. The impact is evident in conducted experiments where individuals' fitness improves to a certain extent during the starting of marking periods. The impact of such an element becomes more influential when configuring more than one period; the use of marking periods reflects an acceleration towards the global optimum. On the contrary, when a single long period is set, without interruptions, fitness variations are only seen at the beginning of that period, reaching good local optima; but then, the fitness keeps stable without variation. The above confirms that repeated periods with pauses in between allow to establish and make permanent modifications to gain more variety -at the phenotypic level- and discovery of new search areas.

The marking process through periods reveals a prominent behavior between populations; the ReGen EA finds better solutions (although the optimum is not always reached) than standard EAs. There are a better exploration and exploitation of the search space and it is observable through the trajectory of depicted curves, how individuals fitness improves, noticing a steep speedup in the first marking period, and then stabilizing curves slowly, with more moderate variations during the remaining periods as the end of the curve approaches. For Binary experiments in the first place, more pronounced jumps are seen, except by Max Ones problem, which does not show any alteration by epigenetic marks, especially because before the first marking period starts, it almost has reached the global maximum. For Rosenbrock and Griewank real defined functions, it is noticed that marking periods generates more subtle changes and take more time to reach better local optima, it moves slowly towards the global minimum. On the other side, Rastrigin and Schwefel evidence abrupt changes from the first marking period, leading to obtaining better local optima. Despite behaving differently, it is worth using markers to show prominent breaks; the use of epigenetic factors shows different behaviors in the evolution of the populations, which is not noticed in the classic EAs. The discrepancy between abrupt and gradual fitness changes may be related to functions features such as the domain, modality, space dimensionality, constraints, defined schemata, among others. This brings questions about the effectiveness of designed tags, how should be tags redefined in order to cope with such problem restrictions? what if the gap between good and poor performances is giving ideas to consider other epigenetics elements that may be missed? can this approach be assessed with other harder and broader problems?. Future work must give some closer ideas.

Regarding the definition of the marking rate, it has taken a long time to tests a variety of rates, so individuals genomes do not be over-marked and keep tags in any position, as it is reflected in biology, except creating islands or groups of marks as particularized by the methylation mechanism ({\em CpG islands}). The hardest activity has been focused on testing the entire framework through the marking and reading processes. The positive thing is the ReGen EA architecture is simple, the epigenetic elements do not add complexity in its implementation, ReGen EA follows a generic idea of an individual with a genome and a well-defined epigenome structure that is shaped during the evolution process. This approach obeys to a population-based bio-inspired technique that adapts while signals from the environment influence the epigenome; the phenotype is configured from interactions between the genotype and the environment; tags are transmitted through generations to maintaining a notion of memory between generations.

It is important to mention that, EA implementations are also statistically analyzed; the median analysis of samples allows graphically depict groups of data and explain them visually to identify the distribution of the samples' median fitnesses. For all functions, except by Max Ones function, median values outpoint the samples from EA standard versions. The analysis to find differences statistically significant between EA groups also confirms there are remarkable differences between the algorithms. This process has evidenced an improvement in epigenetic implementations performance compared to standard versions as reported in experimental results; differences between epigenetic algorithms and standard EAs fitnesses vary significantly, leading to conclude that introducing epigenetic factors to classical versions of EAs do accelerate the search process. These analyses also demonstrate it is not needed to increase or vary mutations rate -for classical mutations-; experiments have used the same rate, inversely, ReGen EA takes advantage of the recombination operator -to promote inheritance-; this operator is a powerful element to evaluate results. For many strategies mutation operator introduces diversity, this thesis does not deny it but instead embraces the idea that mutations can be used with a low rate to have a closer occurrence as seen in biology, and contemplates epigenetic assimilation -fixed changes- to influence the fitness.

Epigenetic components presented in this thesis for the Evolutionary Algorithms framework, describe a way to model Epigenetic Evolutionary Algorithms. ReGen Evolutionary Algorithms involve populations of individuals with genetic and epigenetic codes. This research mainly focuses on those experiences that individuals could acquire during their life cycle and how epigenetic mechanisms lead to learn and adapt for themselves under different conditions. So, such experiences can be inherited over time, and populations would evidence a kind of power of survival. To validate the technique applicability, only problems with real definition and binary encoding schemes were selected. Designed operations are meant to exclusively cover problems with these kinds of encoding, even though, problems with different encoding should be addressed by transforming their domain set into a binary representation; the performance and possible results of such implementations are unknown since no experiments of that kind have been conducted, but ReGen EA must allow them to be performed. The journey with this research allows concluding that epigenetics has many elements to continue improving this work and expanding it in such a way that it can be used for any problem.

\section{Future Work}

Epigenetic mechanisms offer a variety of elements to extend this work and improve the adaptation of individuals in population-based methods to identify novelties during the evolution process. From a biological point of view, some ideas are linked to the fact that there are mechanisms that make epigenetic tags keep fixed and maintained over a long time. It seems to be a process that let tags be bound without changing, just being preserved in the same location under specific environmental conditions or in certain life stages of an individual to avoid their degradation. This hypothesis supports the idea of memory consolidation and adds another dimension to describe a kind of intelligence at a molecular level.

It is intended to extend this model to cover a wider set of optimization problems, different from binary and real encoding problems. The plan is to design a mechanism to create dynamic tags during the evolution process, tags that use a generic encoding and do not depend on specific encoding problems. Problems with numbers-forms, chars, instructions, permutations, commands, expressions among others encoding schemes that can be influenced by generic-defined tags. Base on the former, designed operations need to be redefined, the reading process might be extended, any domain-specific problem must keep its encoding and not be transformed into a binary representation, as it is the current case, and also the marking process may be expanded.

Currently, the application of the marking actions by the ReGen EA is mutually exclusive; marking actions are not happening at the same time. The proceeding opens the possibility to think about another marking process enhancement so that adding, removing, and modifying actions are applied independently based on their distributed probabilities. Performed experiments changing this configuration reveal that by having the possibility of applying them simultaneously with their probability rate, they can produce more suitable individuals with scores closer to the global optimum. However, experiments are required to see individuals' behavior in problems with different encoding from binary and real defined, such as a permutation problem.

From the computational point of view, attempts are made to facilitate this model's replicability in the evolutionary algorithms community. It is expected to elaborate more tests by designing a complete benchmark to continue assessing the ReGen EA performance and improving what it integrates today.

\appendix
    \chapter{Examples of Individuals with Tags}\label{append}

\begin{table}[H]
\centering
\caption{Individual representation for Binary functions, $D=20$.}
\label{apptable1}
\begin{tabular}{lllllllllllllllllllll}
 & \cellcolor[HTML]{34CDF9}0 &  &  &  & 1 & 1 &  & 0 &  & \cellcolor[HTML]{34CDF9}0 &  &  & 1 &  & 0 &  &  & 0 & 1 &  \\
 & \cellcolor[HTML]{34CDF9}1 &  &  &  & 1 & 0 &  & 0 &  & \cellcolor[HTML]{34CDF9}1 &  &  & 0 &  & 0 &  &  & 0 & 1 &  \\
 & \cellcolor[HTML]{34CDF9}0 &  &  &  & 0 & 1 &  & 0 &  & \cellcolor[HTML]{34CDF9}0 &  &  & 1 &  & 1 &  &  & 0 & 1 &  \\
 & \cellcolor[HTML]{FFCC67}0 &  &  &  & 1 & 0 &  & 1 &  & \cellcolor[HTML]{67FD9A}1 &  &  & 0 &  & 0 &  &  & 0 & 0 &  \\
 & \cellcolor[HTML]{FFCC67}1 &  &  &  & 0 & 0 &  & 1 &  & \cellcolor[HTML]{67FD9A}1 &  &  & 0 &  & 0 &  &  & 0 & 0 &  \\
 & \cellcolor[HTML]{FFCC67}0 &  &  &  & 1 & 1 &  & 1 &  & \cellcolor[HTML]{67FD9A}0 &  &  & 1 &  & 1 &  &  & 1 & 1 &  \\
 & \cellcolor[HTML]{FFCC67}0 &  &  &  & 0 & 1 &  & 0 &  & \cellcolor[HTML]{67FD9A}0 &  &  & 1 &  & 0 &  &  & 0 & 0 &  \\
\multirow{-8}{*}{Epigenotype} & \cellcolor[HTML]{FFCC67}0 &  &  &  & 1 & 1 &  & 1 &  & \cellcolor[HTML]{67FD9A}0 &  &  & 0 &  & 1 &  &  & 0 & 0 &  \\
Genotype & \cellcolor[HTML]{C0C0C0}0 & \cellcolor[HTML]{C0C0C0}1 & \cellcolor[HTML]{C0C0C0}0 & \cellcolor[HTML]{C0C0C0}1 & \cellcolor[HTML]{C0C0C0}1 & \cellcolor[HTML]{C0C0C0}1 & \cellcolor[HTML]{C0C0C0}1 & \cellcolor[HTML]{C0C0C0}0 & 0 & \cellcolor[HTML]{C0C0C0}0 & \cellcolor[HTML]{C0C0C0}0 & \cellcolor[HTML]{C0C0C0}1 & \cellcolor[HTML]{C0C0C0}1 & \cellcolor[HTML]{C0C0C0}0 & \cellcolor[HTML]{C0C0C0}1 & \cellcolor[HTML]{C0C0C0}1 & \cellcolor[HTML]{C0C0C0}1 & \cellcolor[HTML]{C0C0C0}1 & \cellcolor[HTML]{C0C0C0}1 & \cellcolor[HTML]{C0C0C0}0 \\
BitString & \cellcolor[HTML]{EFEFEF}0 & \cellcolor[HTML]{EFEFEF}0 & \cellcolor[HTML]{EFEFEF}0 & \cellcolor[HTML]{EFEFEF}0 & \cellcolor[HTML]{EFEFEF}0 & \cellcolor[HTML]{EFEFEF}0 & \cellcolor[HTML]{EFEFEF}0 & \cellcolor[HTML]{EFEFEF}0 & \cellcolor[HTML]{EFEFEF}0 & \cellcolor[HTML]{EFEFEF}0 & \cellcolor[HTML]{EFEFEF}0 & \cellcolor[HTML]{EFEFEF}0 & \cellcolor[HTML]{EFEFEF}0 & \cellcolor[HTML]{EFEFEF}0 & \cellcolor[HTML]{EFEFEF}0 & \cellcolor[HTML]{EFEFEF}0 & \cellcolor[HTML]{EFEFEF}0 & \cellcolor[HTML]{EFEFEF}0 & \cellcolor[HTML]{EFEFEF}0 & \cellcolor[HTML]{EFEFEF}0 \\
Phenotype & \cellcolor[HTML]{EFEFEF}0 & \cellcolor[HTML]{EFEFEF}0 & \cellcolor[HTML]{EFEFEF}0 & \cellcolor[HTML]{EFEFEF}0 & \cellcolor[HTML]{EFEFEF}0 & \cellcolor[HTML]{EFEFEF}0 & \cellcolor[HTML]{EFEFEF}0 & \cellcolor[HTML]{EFEFEF}0 & \cellcolor[HTML]{EFEFEF}0 & \cellcolor[HTML]{EFEFEF}0 & \cellcolor[HTML]{EFEFEF}0 & \cellcolor[HTML]{EFEFEF}0 & \cellcolor[HTML]{EFEFEF}0 & \cellcolor[HTML]{EFEFEF}0 & \cellcolor[HTML]{EFEFEF}0 & \cellcolor[HTML]{EFEFEF}0 & \cellcolor[HTML]{EFEFEF}0 & \cellcolor[HTML]{EFEFEF}0 & \cellcolor[HTML]{EFEFEF}0 & \cellcolor[HTML]{EFEFEF}0
\end{tabular}
\end{table}

The illustrated individual in Appendix \ref{apptable1} describes a solution for the Deceptive Order Four Trap function with a dimension of $20$. The individual depicted in Appendix \ref{apptable2} describes a solution for Rastrigin function with a problem dimension of $2$.  Individuals representation shows in the first row the tags attached to specific alleles, only colored tags are read during tags decoding process ({\em epiGrowingFunction}). The second row presents the genotype code; the third row exhibits the bit string generated by the epigenetic growth function. Finally, the fourth row shows the phenotype representation of the individual. For Real defined functions, binary strings of {\em32}-bits encode real values. 

\begin{landscape}
\begin{table}
\centering
\caption{Individual representation for Real functions, $D=2$.}
\label{apptable2}
\begin{tabular}{lllllllllllllllllllllllllllllllll}
 &  &  & \cellcolor[HTML]{34CDF9}0 & 1 &  &  &  &  & 0 & 1 & 0 & 0 &  & 1 & 1 & 1 & 0 &  &  &  &  & 0 &  & 1 &  &  &  &  & \cellcolor[HTML]{34CDF9}0 &  & 1 &  \\
 &  &  & \cellcolor[HTML]{34CDF9}1 & 0 &  &  &  &  & 1 & 1 & 0 & 0 &  & 0 & 0 & 1 & 1 &  &  &  &  & 0 &  & 1 &  &  &  &  & \cellcolor[HTML]{34CDF9}0 &  & 0 &  \\
 &  &  & \cellcolor[HTML]{34CDF9}0 & 1 &  &  &  &  & 1 & 1 & 1 & 1 &  & 1 & 1 & 0 & 1 &  &  &  &  & 0 &  & 0 &  &  &  &  & \cellcolor[HTML]{34CDF9}1 &  & 1 &  \\
 &  &  & \cellcolor[HTML]{FFC702}1 & 0 &  &  &  &  & 0 & 1 & 1 & 1 &  & 1 & 1 & 0 & 1 &  &  &  &  & 0 &  & 0 &  &  &  &  & \cellcolor[HTML]{FD6864}1 &  & 0 &  \\
 &  &  & \cellcolor[HTML]{FFC702}0 & 0 &  &  &  &  & 0 & 1 & 0 & 0 &  & 0 & 0 & 0 & 1 &  &  &  &  & 0 &  & 0 &  &  &  &  & \cellcolor[HTML]{FD6864}1 &  & 0 &  \\
 &  &  & \cellcolor[HTML]{FFC702}1 & 1 &  &  &  &  & 1 & 1 & 1 & 1 &  & 1 & 1 & 1 & 1 &  &  &  &  & 0 &  & 0 &  &  &  &  & \cellcolor[HTML]{FD6864}0 &  & 1 &  \\
 &  &  & \cellcolor[HTML]{FFC702}1 & 1 &  &  &  &  & 0 & 0 & 1 & 1 &  & 1 & 0 & 1 & 1 &  &  &  &  & 1 &  & 0 &  &  &  &  & \cellcolor[HTML]{FD6864}0 &  & 0 &  \\
\multirow{-8}{*}{Epigenotype} &  &  & \cellcolor[HTML]{FFC702}0 & 1 &  &  &  &  & 1 & 0 & 0 & 0 &  & 1 & 0 & 1 & 0 &  &  &  &  & 1 &  & 1 &  &  &  &  & \cellcolor[HTML]{FD6864}1 &  & 1 &  \\
Genotype & 1 & 0 & \cellcolor[HTML]{C0C0C0}1 & \cellcolor[HTML]{C0C0C0}0 & \cellcolor[HTML]{C0C0C0}1 & \cellcolor[HTML]{C0C0C0}1 & \cellcolor[HTML]{C0C0C0}0 & \cellcolor[HTML]{C0C0C0}0 & \cellcolor[HTML]{C0C0C0}1 & \cellcolor[HTML]{C0C0C0}1 & \cellcolor[HTML]{C0C0C0}0 & \cellcolor[HTML]{C0C0C0}1 & \cellcolor[HTML]{C0C0C0}0 & \cellcolor[HTML]{C0C0C0}1 & \cellcolor[HTML]{C0C0C0}0 & \cellcolor[HTML]{C0C0C0}0 & \cellcolor[HTML]{C0C0C0}1 & \cellcolor[HTML]{C0C0C0}1 & \cellcolor[HTML]{C0C0C0}0 & \cellcolor[HTML]{C0C0C0}0 & \cellcolor[HTML]{C0C0C0}1 & \cellcolor[HTML]{C0C0C0}0 & \cellcolor[HTML]{C0C0C0}0 & \cellcolor[HTML]{C0C0C0}0 & 1 & 1 & 1 & 1 & \cellcolor[HTML]{C0C0C0}1 & \cellcolor[HTML]{C0C0C0}1 & \cellcolor[HTML]{C0C0C0}1 & \cellcolor[HTML]{C0C0C0}1 \\
BitString & \cellcolor[HTML]{EFEFEF}1 & \cellcolor[HTML]{EFEFEF}0 & \cellcolor[HTML]{EFEFEF}1 & \cellcolor[HTML]{EFEFEF}1 & \cellcolor[HTML]{EFEFEF}1 & \cellcolor[HTML]{EFEFEF}1 & \cellcolor[HTML]{EFEFEF}1 & \cellcolor[HTML]{EFEFEF}1 & \cellcolor[HTML]{EFEFEF}1 & \cellcolor[HTML]{EFEFEF}1 & \cellcolor[HTML]{EFEFEF}1 & \cellcolor[HTML]{EFEFEF}1 & \cellcolor[HTML]{EFEFEF}1 & \cellcolor[HTML]{EFEFEF}1 & \cellcolor[HTML]{EFEFEF}1 & \cellcolor[HTML]{EFEFEF}1 & \cellcolor[HTML]{EFEFEF}1 & \cellcolor[HTML]{EFEFEF}1 & \cellcolor[HTML]{EFEFEF}1 & \cellcolor[HTML]{EFEFEF}1 & \cellcolor[HTML]{EFEFEF}1 & \cellcolor[HTML]{EFEFEF}1 & \cellcolor[HTML]{EFEFEF}1 & \cellcolor[HTML]{EFEFEF}1 & \cellcolor[HTML]{EFEFEF}1 & \cellcolor[HTML]{EFEFEF}1 & \cellcolor[HTML]{EFEFEF}1 & \cellcolor[HTML]{EFEFEF}1 & \cellcolor[HTML]{EFEFEF}1 & \cellcolor[HTML]{EFEFEF}1 & \cellcolor[HTML]{EFEFEF}1 & \cellcolor[HTML]{EFEFEF}1 \\
Phenotype & \multicolumn{32}{c}{\cellcolor[HTML]{FFFC9E}2.552499999404536} \\
 &  &  &  &  &  &  &  &  &  &  &  &  &  &  &  &  &  &  &  &  &  &  &  &  &  &  &  &  &  &  &  &  \\
Epigenotype & 1 & 0 &  & 1 &  &  &  &  &  &  & 0 &  &  &  &  & 0 & 0 & 0 &  &  &  & \cellcolor[HTML]{34CDF9}0 &  & 1 &  &  & 1 &  &  &  &  & 0 \\
 & 1 & 1 &  & 1 &  &  &  &  &  &  & 0 &  &  &  &  & 0 & 0 & 1 &  &  &  & \cellcolor[HTML]{34CDF9}0 &  & 0 &  &  & 0 &  &  &  &  & 0 \\
 & 1 & 1 &  & 1 &  &  &  &  &  &  & 0 &  &  &  &  & 1 & 0 & 1 &  &  &  & \cellcolor[HTML]{34CDF9}1 &  & 0 &  &  & 0 &  &  &  &  & 0 \\
 & 1 & 0 &  & 0 &  &  &  &  &  &  & 1 &  &  &  &  & 0 & 1 & 0 &  &  &  & \cellcolor[HTML]{FD6864}1 &  & 1 &  &  & 0 &  &  &  &  & 0 \\
 & 0 & 1 &  & 0 &  &  &  &  &  &  & 1 &  &  &  &  & 0 & 0 & 1 &  &  &  & \cellcolor[HTML]{FD6864}1 &  & 0 &  &  & 1 &  &  &  &  & 1 \\
 & 1 & 0 &  & 0 &  &  &  &  &  &  & 0 &  &  &  &  & 1 & 0 & 1 &  &  &  & \cellcolor[HTML]{FD6864}0 &  & 0 &  &  & 0 &  &  &  &  & 0 \\
 & 1 & 0 &  & 1 &  &  &  &  &  &  & 1 &  &  &  &  & 1 & 0 & 1 &  &  &  & \cellcolor[HTML]{FD6864}0 &  & 1 &  &  & 1 &  &  &  &  & 1 \\
 & 0 & 0 &  & 1 &  &  &  &  &  &  & 0 &  &  &  &  & 0 & 1 & 1 &  &  &  & \cellcolor[HTML]{FD6864}1 &  & 1 &  &  & 1 &  &  &  &  & 0 \\
Genotype & \cellcolor[HTML]{C0C0C0}1 & \cellcolor[HTML]{C0C0C0}1 & \cellcolor[HTML]{C0C0C0}1 & \cellcolor[HTML]{C0C0C0}1 & \cellcolor[HTML]{C0C0C0}1 & \cellcolor[HTML]{C0C0C0}1 & \cellcolor[HTML]{C0C0C0}1 & \cellcolor[HTML]{C0C0C0}1 & \cellcolor[HTML]{C0C0C0}1 & \cellcolor[HTML]{C0C0C0}1 & \cellcolor[HTML]{C0C0C0}1 & \cellcolor[HTML]{C0C0C0}1 & \cellcolor[HTML]{C0C0C0}1 & \cellcolor[HTML]{C0C0C0}1 & \cellcolor[HTML]{C0C0C0}1 & \cellcolor[HTML]{C0C0C0}1 & \cellcolor[HTML]{C0C0C0}0 & \cellcolor[HTML]{C0C0C0}1 & \cellcolor[HTML]{C0C0C0}1 & \cellcolor[HTML]{C0C0C0}1 & \cellcolor[HTML]{C0C0C0}1 & \cellcolor[HTML]{C0C0C0}1 & \cellcolor[HTML]{C0C0C0}1 & \cellcolor[HTML]{C0C0C0}1 & \cellcolor[HTML]{C0C0C0}1 & \cellcolor[HTML]{C0C0C0}1 & \cellcolor[HTML]{C0C0C0}1 & \cellcolor[HTML]{C0C0C0}1 & \cellcolor[HTML]{C0C0C0}1 & \cellcolor[HTML]{C0C0C0}1 & \cellcolor[HTML]{C0C0C0}1 & \cellcolor[HTML]{C0C0C0}1 \\
BitString & \cellcolor[HTML]{EFEFEF}0 & \cellcolor[HTML]{EFEFEF}1 & \cellcolor[HTML]{EFEFEF}1 & \cellcolor[HTML]{EFEFEF}1 & \cellcolor[HTML]{EFEFEF}1 & \cellcolor[HTML]{EFEFEF}1 & \cellcolor[HTML]{EFEFEF}1 & \cellcolor[HTML]{EFEFEF}1 & \cellcolor[HTML]{EFEFEF}1 & \cellcolor[HTML]{EFEFEF}1 & \cellcolor[HTML]{EFEFEF}1 & \cellcolor[HTML]{EFEFEF}1 & \cellcolor[HTML]{EFEFEF}1 & \cellcolor[HTML]{EFEFEF}1 & \cellcolor[HTML]{EFEFEF}1 & \cellcolor[HTML]{EFEFEF}1 & \cellcolor[HTML]{EFEFEF}1 & \cellcolor[HTML]{EFEFEF}1 & \cellcolor[HTML]{EFEFEF}1 & \cellcolor[HTML]{EFEFEF}1 & \cellcolor[HTML]{EFEFEF}1 & \cellcolor[HTML]{EFEFEF}1 & \cellcolor[HTML]{EFEFEF}1 & \cellcolor[HTML]{EFEFEF}1 & \cellcolor[HTML]{EFEFEF}1 & \cellcolor[HTML]{EFEFEF}1 & \cellcolor[HTML]{EFEFEF}1 & \cellcolor[HTML]{EFEFEF}1 & \cellcolor[HTML]{EFEFEF}1 & \cellcolor[HTML]{EFEFEF}1 & \cellcolor[HTML]{EFEFEF}1 & \cellcolor[HTML]{EFEFEF}1 \\
Phenotype & \multicolumn{32}{c}{\cellcolor[HTML]{9AFF99}-0.005000001190928138}
\end{tabular}
\end{table}
\end{landscape}

\chapter{Standard and ReGen EAs Samples for Statistical Analysis}\label{appendB}

Standard and ReGen GAs Samples are tabulated from Appendices \ref{table1} to \ref{table8} by function. Tables present fitness samples of twenty implementations with crossover rates (X) from $0.6$ to $1.0$, generational (G), and steady state (SS) population replacements. Columns report EA implementations and rows contain the runs per algorithm, in total there are thirty runs. Algorithms are represented by numbers from one to twenty, for example, GGAX06 refers to a standard generational GA with crossover rate $0.6$ and the samples are tabulated in column name {\em1}\footnote{
GGAX06 (1),	GGAX07 (2), GGAX08 (3), GGAX09 (4), GGAX10 (5), SSGAX06 (6), SSGAX07 (7), SSGAX08 (8), SSGAX09 (9), SSGAX10 (10), ReGenGGAX06 (11), ReGenGGAX07 (12), ReGenGGAX08 (13), ReGenGGAX09 (14), ReGenGGAX10 (15), ReGenSSGAX06 (16), ReGenSSGAX07 (17), ReGenSSGAX08 (18), ReGenSSGAX09 (19), ReGenSSGAX10 (20).}.

On the other hand, \haea implementations in Appendix \ref{table9} are grouped by function, each function reports four implementations: standard generational \haea (GHAEA (1)), steady state \haea (SSHAEA (2)), ReGen generational \haea (ReGenGHAEA (3)), and ReGen steady state \haea (ReGenSSHAEA (4)). Columns report EA implementations and rows contain the runs per algorithm, in total there are thirty runs. Algorithms are represented by numbers from one to four.

\begin{landscape}
\begin{table}
\centering
\caption{Deceptive Order Three Fitness Sampling: Ten Classic GAs and Ten ReGen GAs with different crossover rates and 30 runs. Best fitness value per run.}
\label{table1}
\footnotesize
\begin{tabular}{llllllllllllllllllll}
\hline
1 & 2 & 3 & 4 & 5 & 6 & 7 & 8 & 9 & 10 & 11 & 12 & 13 & 14 & 15 & 16 & 17 & 18 & 19 & 20 \\\hline
3438 & 3426 & 3456 & 3448 & 3434 & 3422 & 3444 & 3442 & 3420 & 3438 & 3562 & 3564 & 3542 & 3546 & 3584 & 3570 & 3578 & 3578 & 3592 & 3578 \\
3436 & 3426 & 3434 & 3428 & 3448 & 3448 & 3446 & 3424 & 3428 & 3446 & 3560 & 3570 & 3578 & 3570 & 3596 & 3584 & 3584 & 3542 & 3554 & 3580 \\
3426 & 3414 & 3436 & 3440 & 3430 & 3436 & 3424 & 3418 & 3438 & 3414 & 3582 & 3578 & 3576 & 3596 & 3580 & 3564 & 3570 & 3592 & 3566 & 3588 \\
3438 & 3428 & 3436 & 3450 & 3432 & 3432 & 3438 & 3412 & 3432 & 3440 & 3570 & 3584 & 3548 & 3566 & 3548 & 3572 & 3570 & 3582 & 3582 & 3582 \\
3422 & 3420 & 3432 & 3438 & 3442 & 3422 & 3454 & 3432 & 3444 & 3444 & 3574 & 3596 & 3586 & 3556 & 3584 & 3546 & 3586 & 3586 & 3566 & 3574 \\
3454 & 3426 & 3446 & 3446 & 3442 & 3426 & 3438 & 3424 & 3426 & 3434 & 3554 & 3574 & 3584 & 3594 & 3560 & 3568 & 3530 & 3588 & 3592 & 3588 \\
3438 & 3436 & 3448 & 3416 & 3430 & 3422 & 3424 & 3438 & 3438 & 3454 & 3570 & 3584 & 3584 & 3576 & 3582 & 3584 & 3580 & 3568 & 3590 & 3592 \\
\hline
\end{tabular}
\end{table}

\begin{table}
\centering
\caption{Deceptive Order Four Fitness Sampling: Ten Classic GAs and Ten ReGen GAs with different crossover rates and 30 runs. Best fitness value per run.}
\label{table2}
\footnotesize
\begin{tabular}{llllllllllllllllllll}
\hline
1 & 2 & 3 & 4 & 5 & 6 & 7 & 8 & 9 & 10 & 11 & 12 & 13 & 14 & 15 & 16 & 17 & 18 & 19 & 20 \\\hline
390 & 389 & 389 & 389 & 387 & 394 & 389 & 387 & 386 & 390 & 440 & 445 & 442 & 444 & 448 & 441 & 437 & 445 & 447 & 437 \\
391 & 396 & 388 & 389 & 401 & 386 & 391 & 384 & 392 & 397 & 449 & 447 & 447 & 444 & 447 & 444 & 447 & 445 & 444 & 445 \\
388 & 394 & 396 & 391 & 392 & 391 & 387 & 395 & 385 & 393 & 441 & 447 & 445 & 449 & 447 & 443 & 447 & 447 & 444 & 443 \\
387 & 390 & 395 & 388 & 399 & 385 & 384 & 393 & 395 & 394 & 447 & 446 & 444 & 445 & 445 & 445 & 446 & 439 & 448 & 446 \\
394 & 390 & 396 & 395 & 387 & 388 & 392 & 396 & 393 & 388 & 447 & 446 & 443 & 445 & 444 & 444 & 443 & 440 & 449 & 443 \\
387 & 389 & 386 & 396 & 388 & 382 & 392 & 391 & 388 & 388 & 439 & 446 & 433 & 443 & 444 & 444 & 449 & 448 & 445 & 444 \\
387 & 393 & 387 & 390 & 393 & 389 & 386 & 391 & 387 & 391 & 442 & 444 & 447 & 447 & 447 & 443 & 441 & 448 & 449 & 446 \\
\hline
\end{tabular}
\end{table}

\begin{table}
\centering
\caption{Royal Road Fitness Sampling: Ten Classic GAs and Ten ReGen GAs with different crossover rates and 30 runs. Best fitness value per run.}
\label{table3}
\footnotesize
\begin{tabular}{llllllllllllllllllll}
\hline
1 & 2 & 3 & 4 & 5 & 6 & 7 & 8 & 9 & 10 & 11 & 12 & 13 & 14 & 15 & 16 & 17 & 18 & 19 & 20 \\\hline
256 & 240 & 240 & 256 & 280 & 80 & 112 & 104 & 104 & 128 & 352 & 360 & 360 & 360 & 352 & 344 & 352 & 352 & 360 & 344 \\
216 & 184 & 264 & 272 & 312 & 104 & 96 & 120 & 128 & 112 & 360 & 360 & 360 & 360 & 352 & 344 & 352 & 344 & 352 & 360 \\
208 & 248 & 240 & 264 & 256 & 120 & 96 & 128 & 128 & 96 & 360 & 360 & 360 & 360 & 360 & 344 & 344 & 352 & 352 & 360 \\
168 & 200 & 248 & 280 & 280 & 136 & 80 & 112 & 96 & 144 & 360 & 344 & 360 & 360 & 360 & 344 & 352 & 360 & 360 & 352 \\
224 & 224 & 264 & 272 & 272 & 80 & 112 & 128 & 80 & 112 & 360 & 360 & 360 & 360 & 360 & 352 & 360 & 352 & 360 & 360 \\
200 & 256 & 264 & 272 & 296 & 96 & 128 & 120 & 56 & 120 & 360 & 360 & 360 & 360 & 360 & 360 & 336 & 312 & 360 & 360 \\
232 & 216 & 264 & 288 & 272 & 80 & 96 & 88 & 120 & 128 & 360 & 360 & 360 & 360 & 360 & 352 & 360 & 352 & 360 & 344 \\
\hline
\end{tabular}
\end{table}

\begin{table}
\centering
\caption{Max Ones Fitness Sampling: Ten Classic GAs and Ten ReGen GAs with different crossover rates and 30 runs. Best fitness value per run.}
\label{table4}
\footnotesize
\begin{tabular}{llllllllllllllllllll}
\hline
1 & 2 & 3 & 4 & 5 & 6 & 7 & 8 & 9 & 10 & 11 & 12 & 13 & 14 & 15 & 16 & 17 & 18 & 19 & 20 \\\hline
360 & 360 & 360 & 360 & 360 & 360 & 360 & 360 & 360 & 360 & 360 & 360 & 360 & 360 & 360 & 360 & 360 & 360 & 360 & 360 \\
360 & 360 & 360 & 360 & 360 & 360 & 360 & 360 & 360 & 360 & 360 & 360 & 360 & 360 & 360 & 360 & 360 & 360 & 360 & 360 \\
360 & 360 & 360 & 360 & 360 & 360 & 360 & 360 & 360 & 360 & 360 & 360 & 360 & 360 & 360 & 360 & 360 & 360 & 360 & 360 \\
360 & 360 & 360 & 360 & 360 & 360 & 360 & 360 & 360 & 360 & 360 & 360 & 360 & 360 & 360 & 360 & 360 & 360 & 360 & 360 \\
360 & 360 & 360 & 360 & 360 & 360 & 360 & 360 & 360 & 360 & 360 & 360 & 360 & 360 & 360 & 360 & 360 & 360 & 360 & 360 \\
360 & 360 & 360 & 360 & 360 & 360 & 360 & 360 & 360 & 360 & 360 & 360 & 360 & 360 & 360 & 360 & 360 & 360 & 360 & 360 \\
360 & 360 & 360 & 360 & 360 & 360 & 360 & 360 & 360 & 360 & 360 & 360 & 360 & 360 & 360 & 360 & 360 & 360 & 360 & 360 \\
\hline
\end{tabular}
\end{table}

\begin{table}
\centering
\caption{Rastrigin Fitness Sampling: Ten Classic GAs and Ten ReGen GAs with different crossover rates and 30 runs. Best fitness value per run.}
\label{table5}
\scriptsize
\begin{tabular}{llllllllllllllllllll}
\hline
1 & 2 & 3 & 4 & 5 & 6 & 7 & 8 & 9 & 10 & 11 & 12 & 13 & 14 & 15 & 16 & 17 & 18 & 19 & 20 \\\hline
8.821 & 16.925 & 17.668 & 6.890 & 12.467 & 10.058 & 5.658 & 13.224 & 7.675 & 8.346 & 1.025 & 1.034 & 3.340 & 0.020 & 0.029 & 0.028 & 0.020 & 1.025 & 0.018 & 0.029 \\
4.695 & 8.487 & 6.029 & 5.685 & 7.170 & 6.469 & 8.996 & 4.190 & 8.969 & 7.050 & 2.038 & 0.025 & 0.025 & 0.031 & 1.017 & 1.026 & 0.033 & 0.034 & 1.020 & 0.030 \\
10.883 & 11.481 & 11.148 & 7.373 & 9.042 & 16.263 & 16.694 & 11.852 & 9.621 & 5.029 & 0.037 & 2.027 & 2.010 & 0.025 & 0.010 & 1.019 & 0.018 & 0.015 & 0.016 & 0.020 \\
6.638 & 12.630 & 16.947 & 9.698 & 11.349 & 8.382 & 13.939 & 13.364 & 15.860 & 5.565 & 0.036 & 0.034 & 1.022 & 3.003 & 0.020 & 1.102 & 1.038 & 0.023 & 0.019 & 0.026 \\
9.881 & 24.151 & 23.382 & 9.000 & 11.102 & 13.115 & 7.649 & 15.759 & 8.495 & 11.857 & 0.028 & 1.012 & 0.016 & 0.025 & 1.334 & 0.014 & 0.016 & 0.031 & 0.029 & 0.019 \\
9.811 & 11.291 & 10.149 & 13.611 & 12.122 & 6.996 & 14.611 & 10.682 & 10.328 & 4.372 & 2.203 & 0.020 & 0.022 & 0.037 & 0.010 & 0.033 & 0.013 & 1.028 & 0.015 & 0.021 \\
8.844 & 7.067 & 10.632 & 6.024 & 4.705 & 13.976 & 10.343 & 7.968 & 7.535 & 10.845 & 1.022 & 2.198 & 0.025 & 0.026 & 0.016 & 0.042 & 2.014 & 1.003 & 0.035 & 1.025 \\
\hline
\end{tabular}
\end{table}

\begin{table}
\centering
\caption{Rosenbrock Fitness Sampling: Ten Classic GAs and Ten ReGen GAs with different crossover rates and 30 runs. Best fitness value per run.}
\label{table6}
\scriptsize
\begin{tabular}{llllllllllllllllllll}
\hline
1 & 2 & 3 & 4 & 5 & 6 & 7 & 8 & 9 & 10 & 11 & 12 & 13 & 14 & 15 & 16 & 17 & 18 & 19 & 20 \\\hline
0.389 & 0.350 & 0.597 & 0.464 & 0.282 & 1.056 & 0.504 & 1.107 & 1.423 & 0.361 & 0.254 & 0.335 & 0.220 & 1.734 & 0.160 & 0.445 & 0.205 & 0.113 & 0.975 & 0.321 \\
2.150 & 4.640 & 0.277 & 0.385 & 2.225 & 1.048 & 0.411 & 0.471 & 1.071 & 11.332 & 0.335 & 0.226 & 0.593 & 0.283 & 0.434 & 0.643 & 0.199 & 0.235 & 0.295 & 0.486 \\
0.505 & 0.360 & 1.144 & 10.815 & 0.496 & 0.973 & 0.417 & 0.341 & 0.300 & 0.623 & 0.203 & 0.257 & 0.251 & 0.215 & 0.135 & 0.254 & 0.169 & 0.268 & 0.144 & 0.153 \\
0.356 & 0.594 & 0.404 & 0.658 & 0.332 & 11.167 & 0.485 & 0.308 & 0.445 & 0.363 & 0.332 & 0.589 & 1.376 & 0.133 & 0.083 & 0.046 & 0.282 & 0.212 & 0.150 & 0.411 \\
0.480 & 0.283 & 0.331 & 11.445 & 10.102 & 0.623 & 1.847 & 4.572 & 0.446 & 0.304 & 0.331 & 0.492 & 0.219 & 0.314 & 0.163 & 3.257 & 0.260 & 0.147 & 0.268 & 0.169 \\
0.330 & 0.413 & 10.829 & 0.343 & 0.685 & 1.058 & 0.660 & 0.408 & 0.596 & 0.276 & 0.239 & 0.219 & 0.101 & 0.251 & 0.099 & 0.151 & 0.135 & 0.318 & 0.343 & 0.211 \\
0.353 & 0.407 & 10.731 & 0.479 & 0.277 & 0.295 & 4.506 & 10.195 & 5.862 & 0.346 & 0.302 & 0.212 & 0.077 & 0.386 & 0.297 & 0.075 & 0.223 & 0.102 & 0.185 & 0.288 \\
\hline
\end{tabular}
\end{table}

\begin{table}
\centering
\caption{Schwefel Fitness Sampling: Ten Classic GAs and Ten ReGen GAs with different crossover rates and 30 runs. Best fitness value per run.}
\label{table7}
\tiny
\begin{tabular}{llllllllllllllllllll}
\hline
1 & 2 & 3 & 4 & 5 & 6 & 7 & 8 & 9 & 10 & 11 & 12 & 13 & 14 & 15 & 16 & 17 & 18 & 19 & 20 \\\hline
267.4 & 30.5 & 91.5 & 61.0 & 4E-04 & 91.5 & 508.7 & 237.6 & 131.5 & 297.9 & 3.4E-04 & 118.439 & 17.208 & 7.656 & 3.5E-04 & 3.1E-04 & 2.8E-04 & 2.3E-04 & 30.497 & 2.6E-04 \\
192.5 & 61.0 & 118.4 & 118.5 & 61.0 & 354.5 & 162.0 & 118.4 & 30.5 & 4E-04 & 7.655 & 2.8E-04 & 2.3E-04 & 30.497 & 7.655 & 30.497 & 2.4E-04 & 2.7E-04 & 2.4E-04 & 2.6E-04 \\
4E-04 & 162.0 & 4E-04 & 148.9 & 118.5 & 119.1 & 61.0 & 2E-04 & 149.0 & 30.5 & 3.4E-04 & 2.9E-04 & 30.497 & 3.0E-04 & 2.6E-04 & 17.208 & 17.208 & 3.2E-04 & 3.0E-04 & 2.9E-04 \\
30.5 & 179.4 & 30.5 & 118.5 & 118.4 & 328.4 & 148.9 & 213.7 & 4E-04 & 30.5 & 23.391 & 3.6E-04 & 2.6E-04 & 47.705 & 2.7E-04 & 17.208 & 3.2E-04 & 2.6E-04 & 2.9E-04 & 2.6E-04 \\
61.0 & 30.5 & 4E-04 & 118.4 & 122.2 & 402.6 & 61.0 & 30.5 & 236.9 & 30.5 & 2.5E-04 & 2.6E-04 & 2.6E-04 & 3.2E-04 & 2.6E-04 & 2.9E-04 & 118.681 & 7.655 & 57.327 & 2.3E-04 \\
162.0 & 148.9 & 122.2 & 61.0 & 30.5 & 212.2 & 192.5 & 119.2 & 420.0 & 4E-04 & 2.4E-04 & 3.1E-04 & 3.2E-04 & 2.6E-04 & 2.7E-04 & 2.7E-04 & 3.1E-04 & 2.7E-04 & 2.6E-04 & 2.4E-04 \\
61.0 & 118.5 & 30.5 & 30.5 & 6E-04 & 179.5 & 118.5 & 61.0 & 162.0 & 149.6 & 30.490 & 3.2E-04 & 10.106 & 3.7E-04 & 17.208 & 3.6E-04 & 2.5E-04 & 2.4E-04 & 17.208 & 3.5E-04 \\
\hline
\end{tabular}
\end{table}

\begin{table}
\centering
\caption{Griewank Fitness Sampling: Ten Classic GAs and Ten ReGen GAs with different crossover rates and 30 runs. Best fitness value per run.}
\label{table8}
\scriptsize
\begin{tabular}{llllllllllllllllllll}
\hline
1 & 2 & 3 & 4 & 5 & 6 & 7 & 8 & 9 & 10 & 11 & 12 & 13 & 14 & 15 & 16 & 17 & 18 & 19 & 20 \\\hline
0.146 & 0.237 & 0.120 & 0.152 & 0.090 & 0.077 & 0.202 & 0.075 & 0.131 & 0.334 & 0.085 & 0.030 & 0.083 & 0.268 & 0.122 & 0.058 & 0.070 & 0.084 & 0.045 & 0.176 \\
0.092 & 0.503 & 0.156 & 0.124 & 0.114 & 0.586 & 0.244 & 0.167 & 0.119 & 0.131 & 0.021 & 0.037 & 0.190 & 0.042 & 0.100 & 0.111 & 0.016 & 0.057 & 0.066 & 0.024 \\
0.201 & 0.262 & 0.113 & 0.119 & 0.041 & 0.141 & 0.269 & 0.220 & 0.389 & 0.145 & 0.084 & 0.020 & 0.117 & 0.010 & 0.036 & 0.206 & 0.045 & 0.067 & 0.083 & 0.064 \\
0.235 & 0.148 & 0.196 & 0.132 & 0.274 & 0.140 & 0.158 & 0.160 & 0.105 & 0.299 & 0.058 & 0.087 & 0.160 & 0.098 & 0.063 & 0.055 & 0.109 & 0.081 & 0.199 & 0.062 \\
0.163 & 0.409 & 0.199 & 0.370 & 0.162 & 0.261 & 0.157 & 0.118 & 0.244 & 0.310 & 0.050 & 0.085 & 0.084 & 0.080 & 0.076 & 0.134 & 0.037 & 0.253 & 0.064 & 0.025 \\
0.135 & 0.262 & 0.189 & 0.131 & 0.048 & 0.144 & 0.218 & 0.463 & 0.197 & 0.232 & 0.050 & 0.040 & 0.028 & 0.139 & 0.051 & 0.042 & 0.088 & 0.080 & 0.012 & 0.077 \\
0.132 & 0.089 & 0.220 & 0.093 & 0.165 & 0.244 & 0.111 & 0.169 & 0.160 & 0.056 & 0.244 & 0.086 & 0.157 & 0.008 & 0.076 & 0.109 & 0.058 & 0.036 & 0.044 & 0.057 \\
\hline
\end{tabular}
\end{table}

\begin{table}
\centering
\caption{Fitness Sampling: Two standard \haea and Two ReGen \haea implementations with 30 runs. Best fitness value per run.}
\label{table9}
\tiny
\begin{tabular}{llllllllllllllllllll}
\hline
\multicolumn{4}{c}{Deceptive Order Three} & \multicolumn{4}{c}{Deceptive Order Four} & \multicolumn{4}{c}{Ratrigin} & \multicolumn{4}{c}{Schwefel} & \multicolumn{4}{c}{Griewank} \\
1 & 2 & 3 & 4 & 1 & 2 & 3 & 4 & 1 & 2 & 3 & 4 & 1 & 2 & 3 & 4 & 1 & 2 & 3 & 4 \\\hline
3430 & 3440 & 3596 & 3576 & 392 & 387 & 447 & 447 & 5.8084 & 5.6855 & 0.0337 & 0.0298 & 1.489E+02 & 9.149E+01 & 2.284E-04 & 2.636E-04 & 0.1294 & 0.3327 & 0.0609 & 0.0619 \\
3456 & 3438 & 3586 & 3584 & 391 & 393 & 446 & 446 & 17.8005 & 18.9239 & 0.0147 & 0.0223 & 1.184E+02 & 2.979E+02 & 2.672E-04 & 2.459E-04 & 0.0805 & 0.1648 & 0.1170 & 0.0261 \\
3444 & 3446 & 3592 & 3594 & 390 & 387 & 446 & 443 & 5.3688 & 19.2643 & 0.0198 & 0.0160 & 9.149E+01 & 4.131E-04 & 2.870E-04 & 3.050E+01 & 0.2237 & 0.3112 & 0.0248 & 0.0559 \\
3454 & 3436 & 3590 & 3590 & 395 & 393 & 448 & 449 & 14.7865 & 10.5025 & 0.0000 & 0.0099 & 3.515E-04 & 1.496E+02 & 2.284E-04 & 2.284E-04 & 0.2458 & 0.1858 & 0.1020 & 0.0985 \\
3436 & 3450 & 3570 & 3590 & 393 & 382 & 441 & 447 & 6.6905 & 14.0264 & 0.0246 & 0.0259 & 6.099E+01 & 6.099E+01 & 2.591E-04 & 1.184E+02 & 0.0965 & 0.4865 & 0.0517 & 0.0822 \\
3430 & 3434 & 3588 & 3574 & 391 & 394 & 449 & 444 & 12.0701 & 20.2973 & 0.0239 & 0.0099 & 4.130E-04 & 4.438E-04 & 2.365E-04 & 2.591E-04 & 0.1813 & 0.0913 & 0.0228 & 0.0607 \\
\hline
\end{tabular}
\end{table}
\end{landscape}




\backmatter
\nocite{*}
    \bibliography{bibliografia}
\end{document}